\pdfoutput=1
\documentclass[11pt]{article}
\usepackage[preprint]{acl}
\usepackage{times}
\usepackage{latexsym}
\usepackage[T1]{fontenc}
\usepackage[utf8]{inputenc}
\usepackage{microtype}
\usepackage{inconsolata}
\usepackage{amsmath}
\usepackage{amssymb}
\usepackage{xspace}
\usepackage{booktabs}
\usepackage{multirow}
\usepackage{graphicx}
\usepackage{tabularx}
\usepackage{subcaption}
\usepackage{tcolorbox}

\newcommand{\paratitle}[1]{\vspace{1.5ex}\noindent\textbf{#1}}
\newcommand{\ie}{\emph{i.e.,}\xspace}

\newcommand{\OURS}{\textbf{A3PO}\xspace}
\newcommand{\std}[1]{\normalsize{$\pm$#1}}
\newcommand{\ignore}[1]{}

\title{Rethinking Sample Polarity in Reinforcement \\ Learning with Verifiable Rewards}

\author{
    \textbf{
        Xinyu Tang\textsuperscript{\rm{1}\thanks{\ \ Equal contribution.}},
        Yuliang Zhan\textsuperscript{\rm{1}\footnotemark[1]},
        Zhixun Li\textsuperscript{\rm{2}\footnotemark[1]},
        Wayne Xin Zhao\textsuperscript{\rm{1}\thanks{\ \ Corresponding author.}},
    } \\
    \textbf{
        Zhenduo Zhang\textsuperscript{\rm{3}\ },
        Zujie Wen\textsuperscript{\rm{3}\ },
        Zhiqiang Zhang\textsuperscript{\rm{3}\ }
        Jun Zhou\textsuperscript{\rm{3}\ }
    } \\
    \textsuperscript{1}Gaoling School of Artificial Intelligence, Renmin University of China \\
    \textsuperscript{2}The Chinese University of Hong Kong
    \textsuperscript{3}Ant Group \\
}

\begin{document}
\maketitle

\begin{abstract}

Large reasoning models~(LRMs) are typically trained using reinforcement learning with verifiable reward~(RLVR) to enhance their reasoning abilities.
In this paradigm, policies are updated using both positive and negative self-generated rollouts, which correspond to distinct \textbf{\textit{sample polarities}}.
In this paper, we provide a systematic investigation into how these sample polarities affect RLVR training dynamics and behaviors.
We find that positive samples sharpen existing correct reasoning patterns, while negative samples encourage exploration of new reasoning paths.
We further explore how adjusting the advantage values of positive and negative samples at both the sample level and the token level affects RLVR training.
Based on these insights, we propose an \textbf{A}daptive and \textbf{A}symmetric token-level \textbf{A}dvantage shaping method for \textbf{P}olicy \textbf{O}ptimization, namely \OURS, that more precisely allocates advantage signals to key tokens across different polarities. 
Experiments across five reasoning benchmarks demonstrate the effectiveness of our approach.

\end{abstract}

\section{Introduction}
\label{sec-introdction}

Large reasoning models~\citep{Deepseek-R1,Kimi-K2,Qwen3-8B-Base} have recently gained significant attention due to their impressive performance in mathematical, coding, and scientific reasoning tasks.
These models typically adopt the reinforcement learning with verifiable reward~(RLVR) paradigm~\citep{DAPO,CISPO}, where they generate multiple long chain-of-thought reasoning trajectories and use verifiable binary rewards to assess the correctness of the final answers.
The reward signals are then used to update the model’s policy.
Unlike supervised fine-tuning~\citep{SFT}, which imitates external teachers by memorizing correct examples, RLVR enables models to learn from their own generated rollouts, including both positive and negative samples.
Positive samples help reinforce reasoning paths that the model already handles correctly, while negative samples facilitate self-correction by learning from mistakes.
However, within the RLVR framework, the distinct roles of positive and negative samples, which are referred to as \textit{\textbf{sample polarity}}, remain underexplored.

To explore this question, prior studies have attempted to analyze the respective contributions of positive and negative samples in RLVR.
For instance, \citet{PSRNSR} decomposes RLVR into two learning paradigms: positive and negative sample reinforcement. 
Their findings show that training solely with negative samples consistently improves Pass@k metrics.
However, their experiments were constrained to a simple math training dataset and a small set of models, which limits the generalizability of their conclusions.
Subsequent studies observe an asymmetry between positive and negative samples and propose methods to improve importance sampling~\citep{ASPO}, advantage shaping~\citep{PSRNSR}, and clipping mechanisms~\citep{STEER,BAPO}.
Nevertheless, a thorough analysis of how positive and negative samples influence RLVR training dynamics remains incomplete.

In this paper, we systematically analyze the roles of positive and negative samples in RLVR by applying PSR and NSR to three different base LLMs.
We find that positive samples sharpen the model's existing correct reasoning paths, reduce entropy, and result in shorter outputs. 
In contrast, negative samples promote the discovery of new reasoning patterns, increase entropy, and encourage longer responses.
However, using only one sample polarity impairs reasoning performance and boundary, demonstrating both types are important for RLVR.

We further investigate how modulating the influence of positive and negative samples at different granularities affects RLVR training.
At the sample level, assigning higher weights to positive samples accelerates reward improvement but narrows exploration diversity, whereas emphasizing negative samples encourages broader exploration at the expense of slower reward progress.
To examine the training process in finer granularity, we perform token-level advantage shaping to determine which specific tokens in positive and negative samples contribute more to the training dynamics. 
Our results indicate that weighting tokens based on their entropy and probability has distinct effects for each polarity.
Building on these findings, we propose an \textbf{\underline{A}}daptive and \textbf{\underline{A}}symmetric token-level \textbf{\underline{A}}dvantage shaping method for \textbf{\underline{P}}olicy \textbf{\underline{O}}ptimization, namely \OURS. 
This approach dynamically adjusts the advantages of high-probability tokens in negative samples and low-probability tokens in positive samples, enabling finer-grained advantage allocation. 
Experiments across three LLMs and five reasoning benchmarks validate the effectiveness of \OURS.

Our contributions are summarized as follows:


$\bullet$ We conduct a comprehensive analysis of sample polarity in RLVR. We identify distinct training dynamics between them and observe that both sample polarities are crucial for RLVR.

$\bullet$ We investigate how varying the influence of positive and negative samples at different granularities affects RLVR training through sample-level and token-level advantage shaping.

$\bullet$ We propose an adaptive and asymmetric token-level advantage shaping method, which enables finer-grained advantage allocation and leads to more effective and stable RLVR training.
\section{Related Work}
\label{sec-related_work}

\subsection{Reinforcement Learning with Verifiable Rewards}

Reinforcement learning with verifiable rewards (RLVR) effectively improves the reasoning ability of large language models.
Under this paradigm, an LLM acts as a policy model that generates multiple long chain-of-thought reasoning paths.
The model is optimized using binary outcome-based rewards, which removes the need for a learned reward model.
As a representative algorithm, Group Relative Policy Optimization (GRPO)~\citep{Deepseek-R1} computes advantages directly from groups of rollouts, avoiding reliance on a learned value network and enabling scalable reasoning through zero-RL.
Following GRPO, subsequent studies have refined the algorithm by introducing enhanced techniques for advantage estimation~\citep{PRIME,VAPO}, loss aggregation~\citep{GMPO,GSPO}, importance sampling~\citep{CISPO}, and sampling strategies~\citep{Treepo,SPO}.

\subsection{Sample Polarity in RLVR}

In RLVR, both positive and negative samples are important for policy optimization.
Positive samples reinforce correct reasoning paths, while negative samples allow models to learn from their mistakes.
Prior work has examined the distinct effects of these two sample types~\citep{PSRNSR}.
They propose methods that treat them differently, including importance sampling~\citep{ASPO}, advantage reweighting~\citep{PSRNSR,STEER}, and clipping mechanisms~\citep{BAPO}.
In this paper, we provide a more thorough investigation of how different sample polarities affect training dynamics and analyze their respective contributions to RLVR.
In addition, we conduct a finer-grained analysis of different sample polarities via polarity-level and token-level advantage shaping.
\section{Rethinking the Role of Positive and Negative Samples in RLVR}
\label{sec-pn-sample-reinforcement}

In this section, we analyze how positive and negative samples affect RLVR training across different base LLMs.
Specifically, we study reinforcement using only positive and negative samples and compare their training dynamics and model behaviors.

\subsection{Experimental Setup}

We conduct experiments with three different types of LLMs: a math-enhanced LLM Qwen2.5-7B-Math~\citep{Qwen2.5-Math-7B}, a general pretrained LLM Qwen3-8B-Base~\citep{Qwen3-8B-Base}, and a distilled LLM after supervised fine-tuning DeepSeek-R1-Distill-Qwen-7B~\citep{Deepseek-R1}.
Following ~\citet{PSRNSR}, we perform reinforcement separately using only positive and negative samples for each LLM, and include DAPO, which utilizes both types of samples, for comparison.
More details on positive and negative sample reinforcement are included in Appendix~\ref{appsec:psrnsr}, and experimental setups are provided in Appendix~\ref{app:exp-setup}.

\begin{figure*}[t]
    \centering
    \begin{subfigure}[b]{0.32\linewidth}
        \centering
        \includegraphics[width=\linewidth]{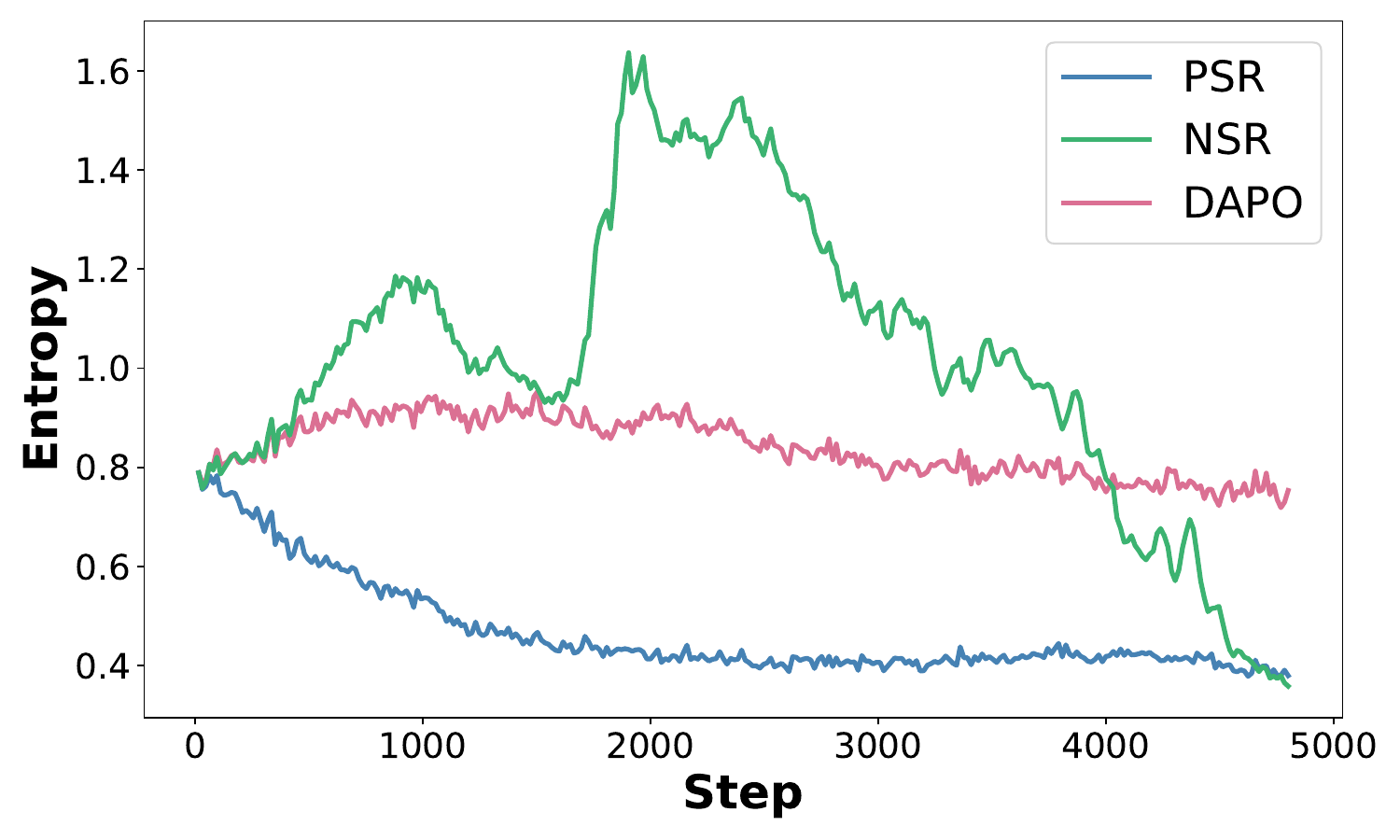}
        \caption{Entropy}
        \label{subfig:ds-entropy}
    \end{subfigure}
    \begin{subfigure}[b]{0.32\linewidth}
        \centering
        \includegraphics[width=\linewidth]{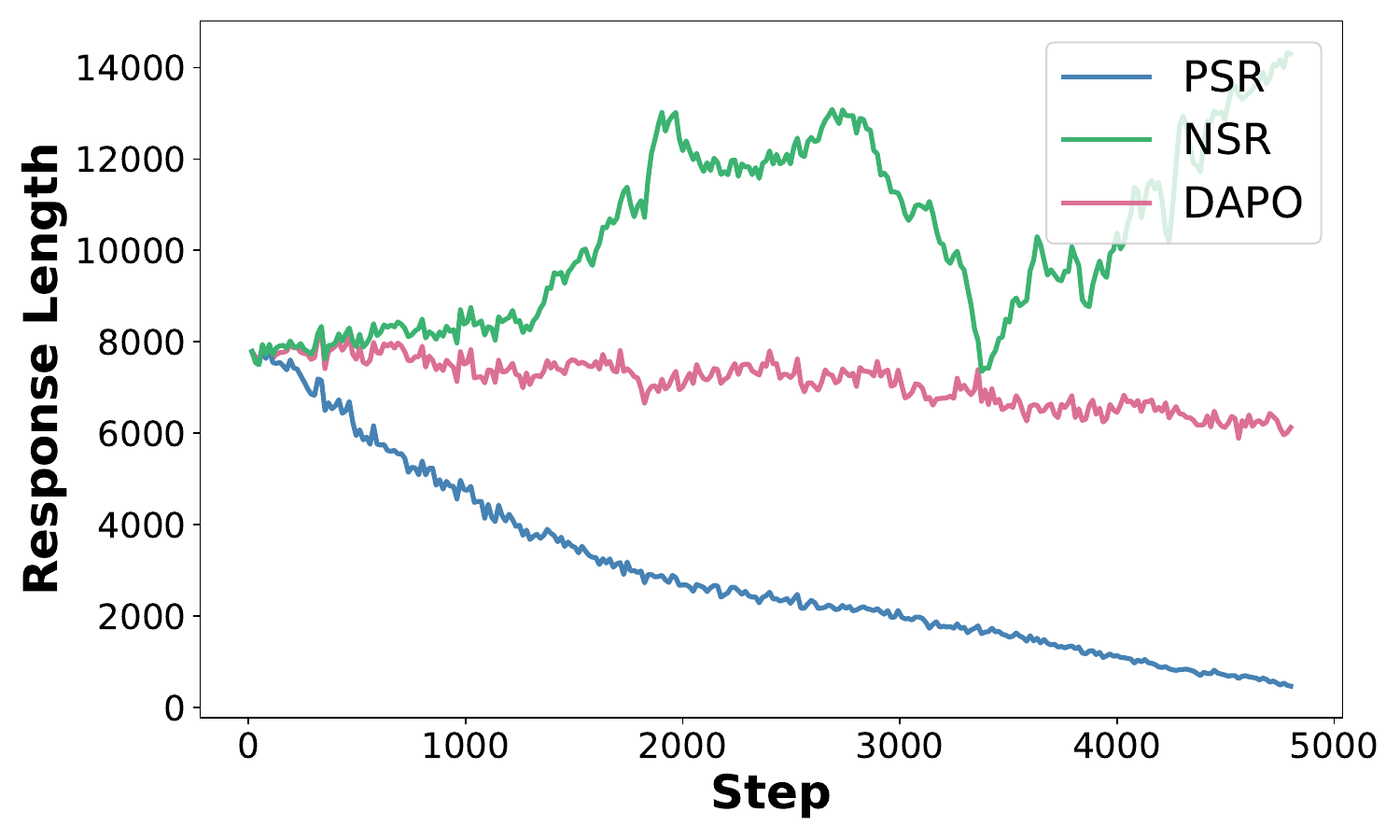}
        \caption{Response Length}
        \label{subfig:ds-length}
    \end{subfigure}
    \begin{subfigure}[b]{0.32\linewidth}
        \centering
        \includegraphics[width=\linewidth]{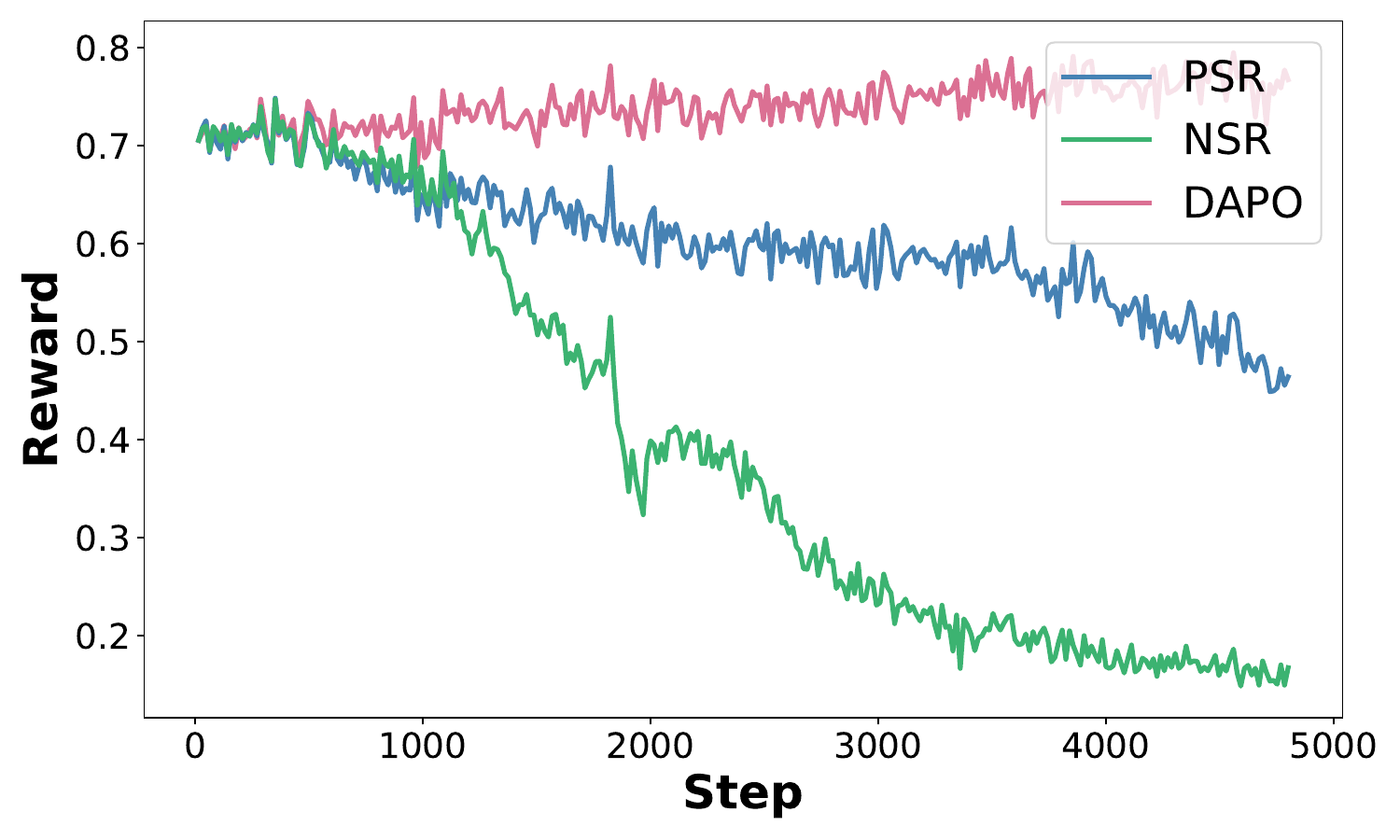}
        \caption{Reward}
        \label{subfig:ds-reward}
    \end{subfigure}
    \begin{subfigure}[b]{0.32\linewidth}
        \centering
        \includegraphics[width=\linewidth]{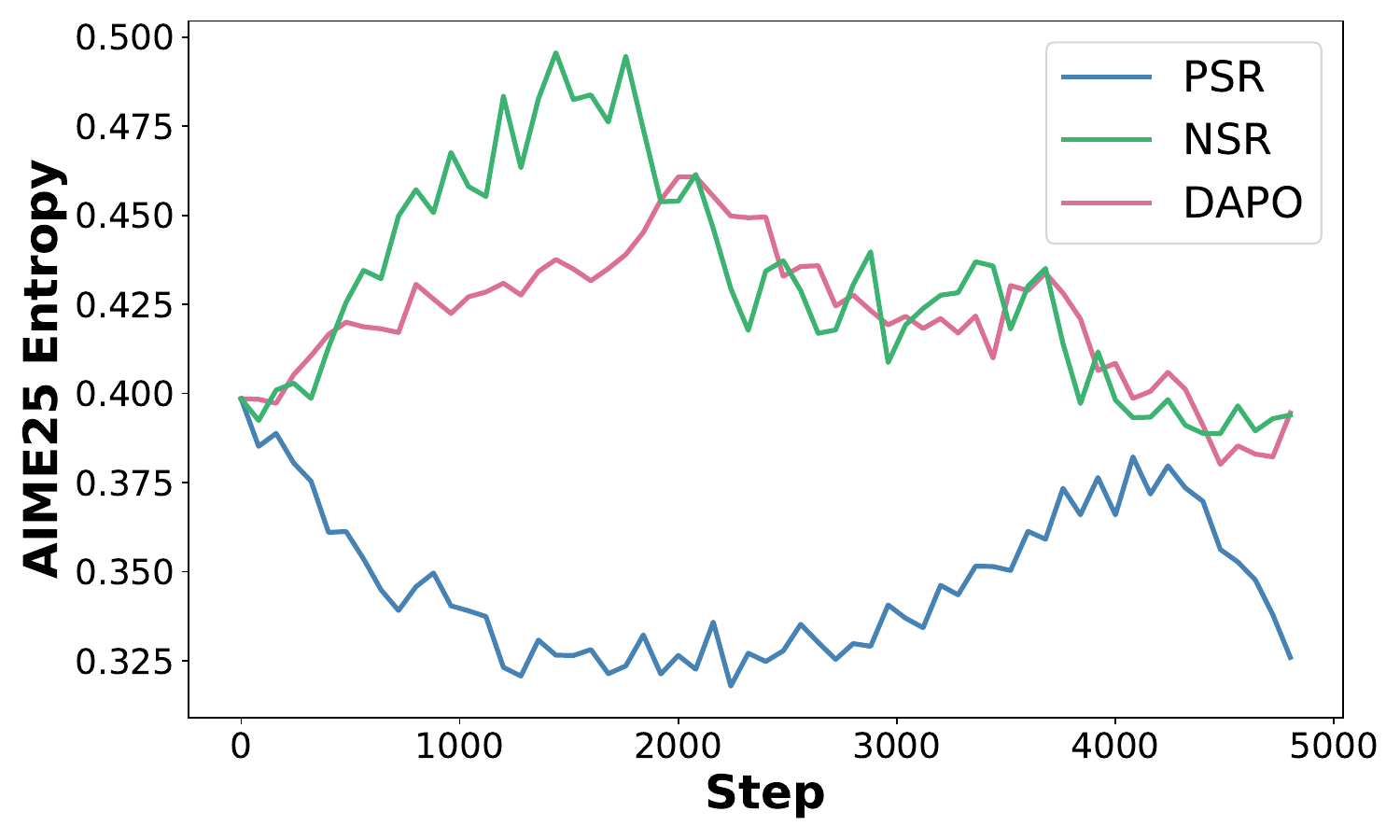}
        \caption{AIME25 Entropy}
        \label{subfig:ds-aime25-entropy}
    \end{subfigure}
    \begin{subfigure}[b]{0.32\linewidth}
        \centering
        \includegraphics[width=\linewidth]{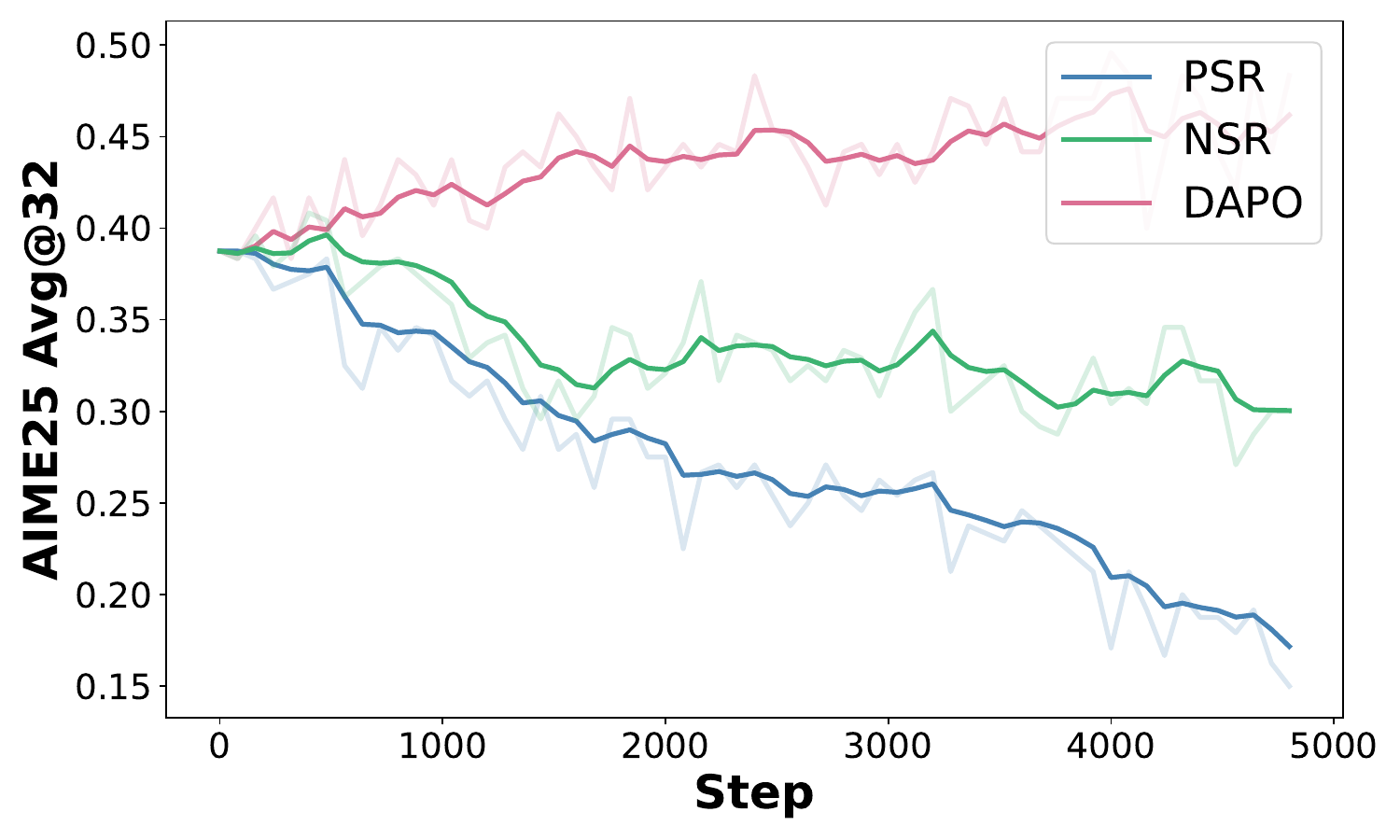}
        \caption{AIME25 Avg@32}
        \label{subfig:ds-aime25-avg}
    \end{subfigure}
    \begin{subfigure}[b]{0.32\linewidth}
        \centering
        \includegraphics[width=\linewidth]{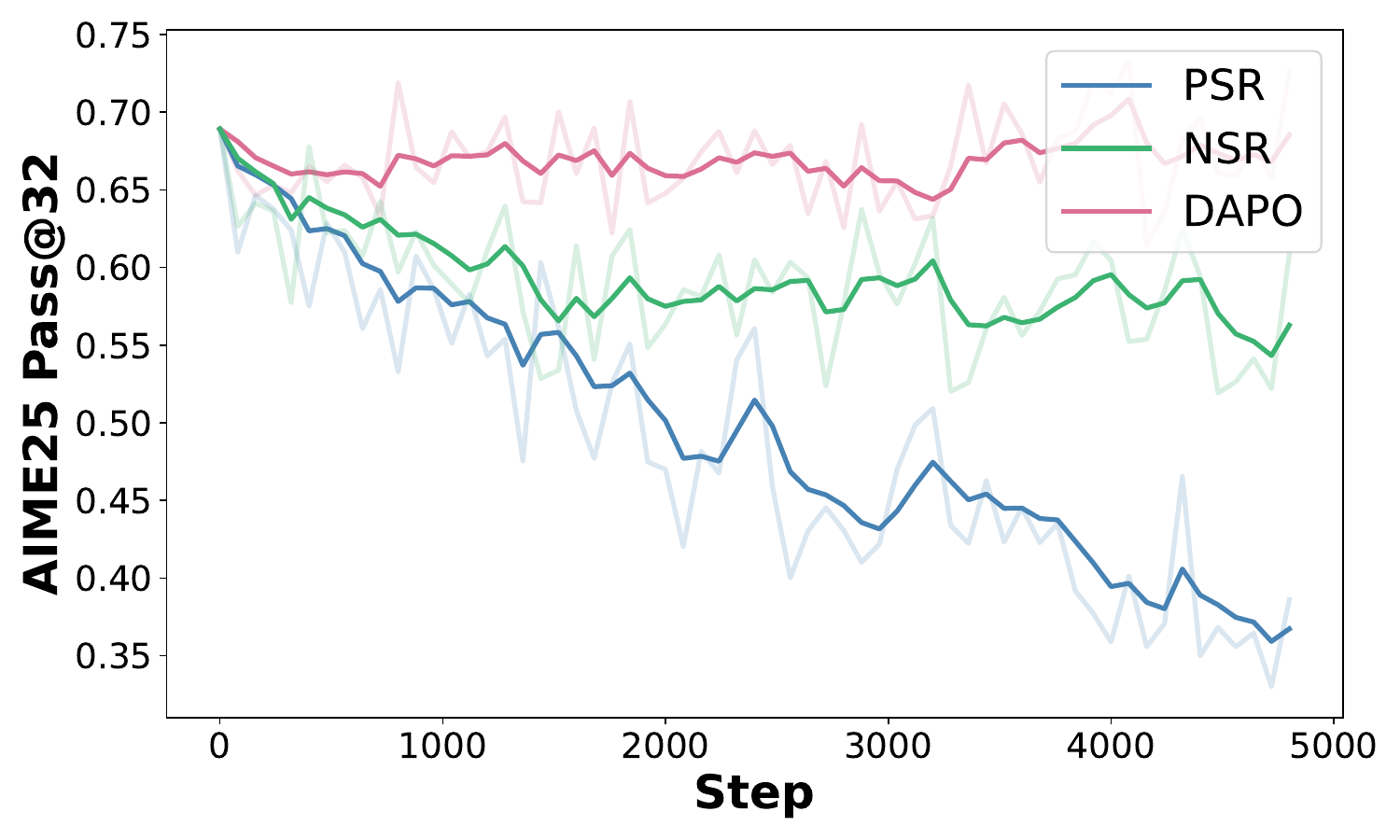}
        \caption{AIME25 Pass@32}
        \label{subfig:ds-aime25-pass}
    \end{subfigure}
    \caption{RLVR training dynamics under three training paradigms on Deepseek-R1-Distilled-Qwen-7B.}
\label{fig:ds-qwen-7b-training_dynamic-main}
\end{figure*}

\subsection{Different Training Dynamics of Positive and Negative Sample Reinforcement}

\paratitle{Positive samples reduce entropy, negative samples maintain it.}
As shown in Figure~\ref{subfig:ds-entropy} and~\ref{subfig:ds-aime25-entropy}, reinforcement with only positive samples leads to a rapid decline in model entropy, while reinforcement with only negative samples helps maintain higher entropy levels on both training and validation data.
This occurs because positive reinforcement amplifies the logits of tokens that appear in correct solutions, making the model more confident in a narrow set of high-probability predictions.
In contrast, negative reinforcement reduces the logits of tokens present in incorrect solutions and indirectly boosts alternatives, 
thus preserving greater exploration diversity and higher entropy.

\paratitle{Positive samples produce shorter responses, negative samples yield longer ones.}
Figure~\ref{subfig:ds-length} shows that models trained with positive samples alone generate increasingly shorter responses, whereas those trained with negative samples produce longer outputs.
This is because positive reinforcement rewards the most efficient path to correct answers, implicitly penalizing extra reasoning steps. 
On the other hand, negative reinforcement suppresses incorrect tokens without encouraging brevity, allowing models to explore longer reasoning chains.

\paratitle{Using only one sample polarity harms reasoning abilities and boundaries.}
As shown by the training reward in Figure~\ref{subfig:ds-reward}) and validation performance in Figure~\ref{subfig:ds-aime25-avg} and~\ref{subfig:ds-aime25-pass}), training with only positive or negative samples damages the model's reasoning ability and boundary, with further degradation over training.
It is worth noting that although \citet{PSRNSR} suggests that negative-only reinforcement can improve reasoning boundaries, we find that it only maintains Pass@32 performance comparable to DAPO on Qwen2.5-7B-MATH, indicating that such a conclusion is limited to certain models.
This further confirms that both positive and negative samples are essential in RLVR.

\paratitle{Negative samples are key to preserving generalization.}
As illustrated in Figure~\ref{subfig:ds-reward} and Figure~\ref{subfig:ds-aime25-avg}, reward on the training set declines faster or grows slower with negative sample reinforcement compared to positive sample reinforcement. 
However, models trained with negative samples achieve better performance on the validation set.
This suggests that negative samples are crucial for maintaining the model's generalization ability in RLVR.

\begin{figure*}[t]
    \centering
    \begin{subfigure}[b]{0.32\linewidth}
        \centering
        \includegraphics[width=\linewidth]{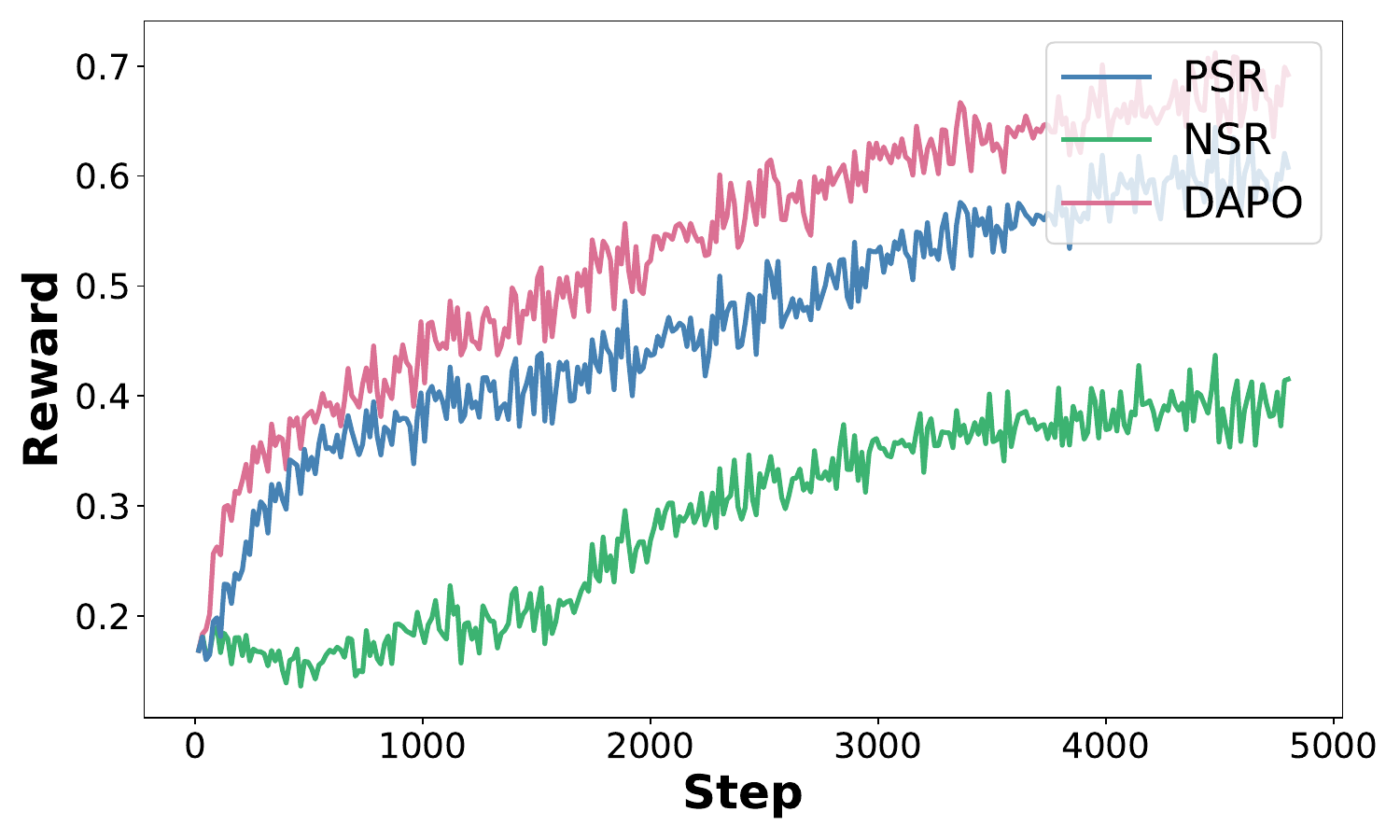}
        \caption{Qwen2.5-7B-Math}
        \label{subfig:qwen2.5-7b-math-reward}
    \end{subfigure}
    \begin{subfigure}[b]{0.32\linewidth}
        \centering
        \includegraphics[width=\linewidth]{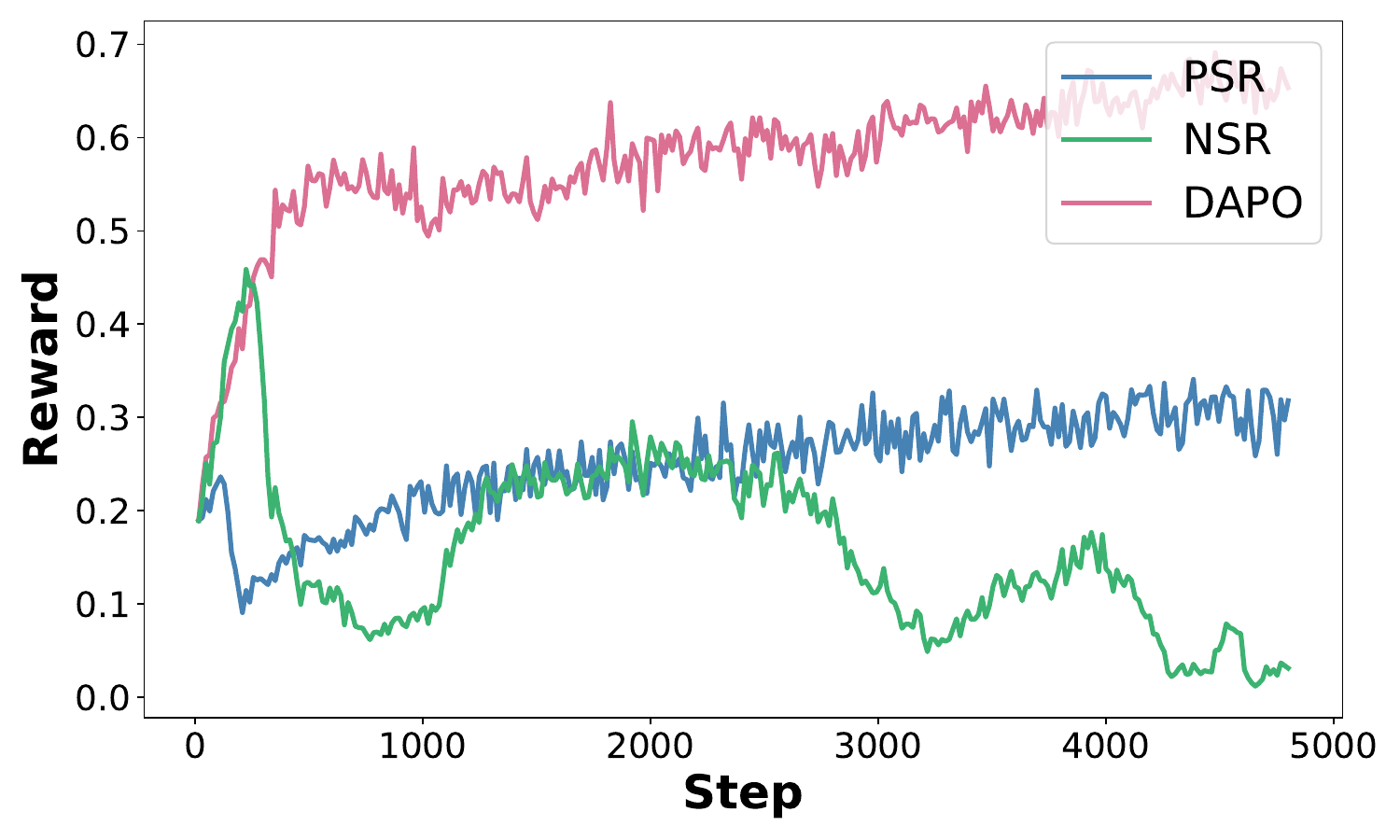}
        \caption{Qwen3-8B-Base}
        \label{subfig:qwen3-8b-base-reward}
    \end{subfigure}
    \begin{subfigure}[b]{0.32\linewidth}
        \centering
        \includegraphics[width=\linewidth]{figures/psr_nsr/ds-qwen-7b/Reward.pdf}
        \caption{Deepseek-R1-Distilled-Qwen-7B}
        \label{subfig:ds-7b-qwen-reward}
    \end{subfigure}
    \caption{RLVR training reward across different training paradigms and base LLMs.}
\label{fig:training-dynamics}
\end{figure*}

\subsection{Different Training Dynamics across Base LLMs}

Figure~\ref{fig:training-dynamics} illustrates how reward changes during RLVR training across various base LLMs.
For Qwen2.5-7B-Math, using only positive or negative samples can improve reward, but is less effective than using both together.
Here, both types of samples jointly accelerate RL training and lead to better final performance.
In contrast, for Deepseek-R1-Distilled-Qwen-7B, training with only one polarity damages reasoning ability.
Only when both positive and negative samples are combined does reward improve consistently.

When training Qwen3-8B-Base with only positive samples, the reward initially drops sharply and then recovers in later stages.
We observe that the model exhibits reward hacking, where it learns to guess answers directly rather than perform step-by-step reasoning.
On the other hand, using only negative samples results in reward fluctuation without steady progress.
This occurs because negative sample reinforcement continuously shifts probability away from high-probability tokens to others, which increases the likelihood of generating irrelevant tokens and ultimately leads to mojibake output.
A case study on Qwen3-8B-Base is provided in Appendix~\ref{app:case_study}.
Detailed analyses of accuracy changes on validation samples are presented in Appendix~\ref{app:diff_train_dynamics}.



\begin{figure}[t]
    \centering
    \begin{subfigure}[b]{0.48\linewidth}
        \centering
        \includegraphics[width=\linewidth]{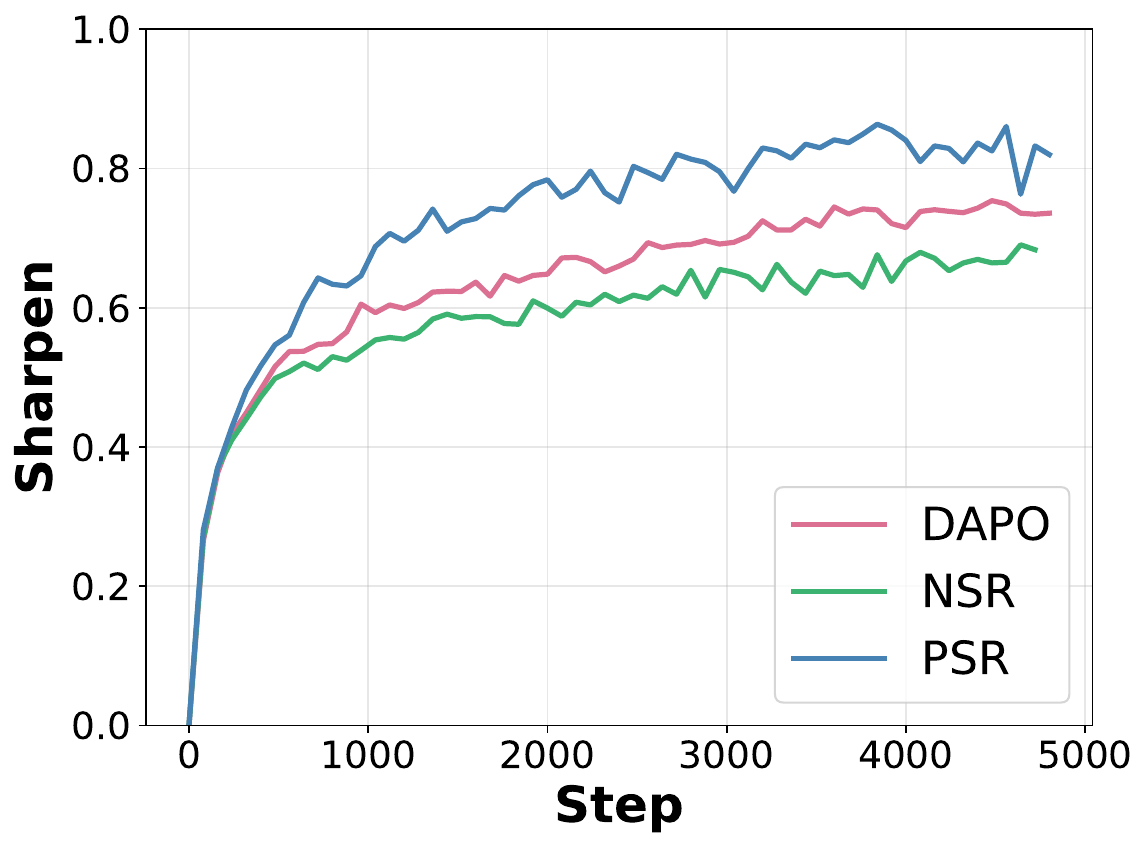}
        \caption{Sharpen}
        \label{subfig:ds-sharpen-main}
    \end{subfigure}
    \begin{subfigure}[b]{0.48\linewidth}
        \centering
        \includegraphics[width=\linewidth]{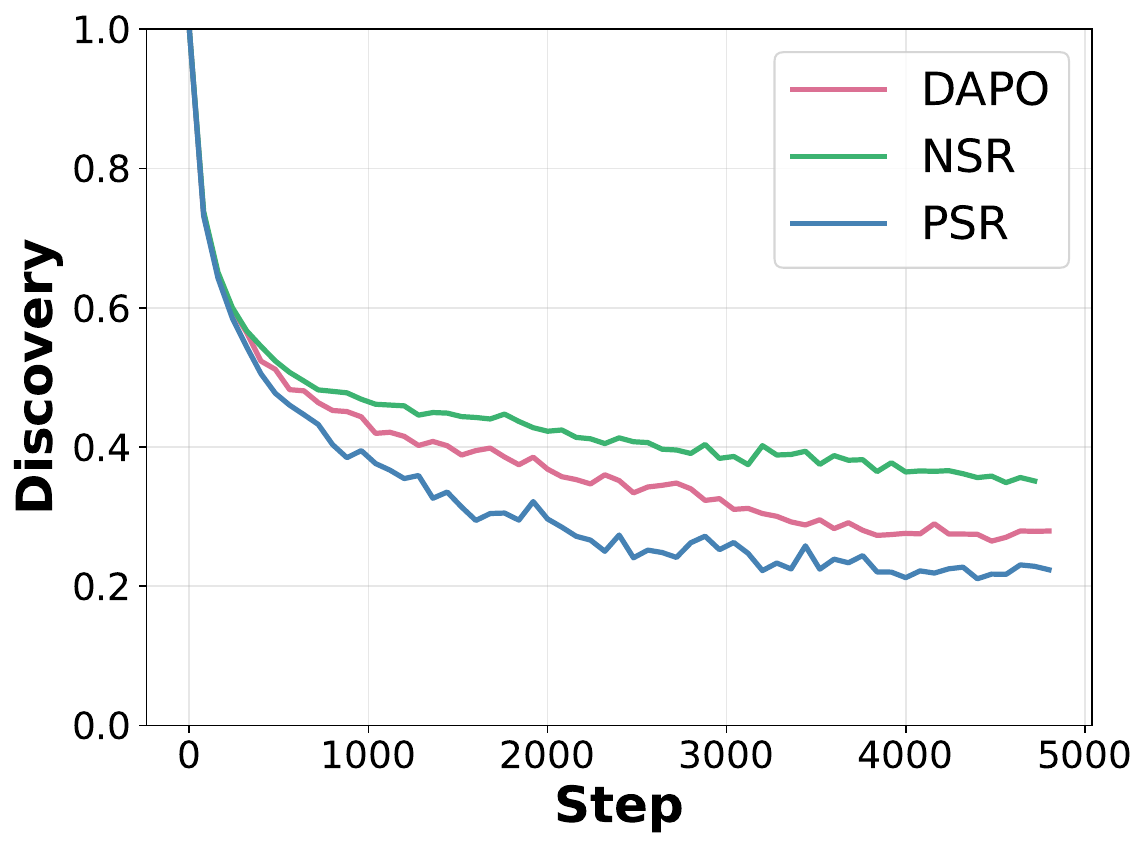}
        \caption{Discovery}
        \label{subfig:ds-discovery-main}
    \end{subfigure}
    \caption{Training behaviors of different paradigms.}
\label{fig:sharpen-discovery-main}
\end{figure}

\subsection{Positive Samples Encourage Shapren, Negative Samples Help Discovery}

There are two prevailing views on RLVR: sharpening and discovery, which appear to be in direct opposition~\citep{RL-survey}. 
The sharpening view~\citep{limit-of-RLVR} posits that RLVR does not create genuinely new patterns, but instead refines and reweights correct responses already available in the base model. 
In contrast, the discovery view~\citep{ProRL} suggests that RLVR can uncover new reasoning patterns not acquired during pre-training and not generated through repeated sampling.
To investigate how sample polarities contribute to each perspective, we examine the model's generated rollouts from an n-gram perspective. 
Here, we define two metrics:

$\bullet$ \textbf{Sharpening}: The proportion of n-grams in the current rollout that have appeared in previously correct rollouts, which measures how much the model reinforces existing correct patterns.

$\bullet$ \textbf{Discovery}: The proportion of n-grams in the current rollout that have never appeared before, reflecting the model's exploration of new paths.

The results are shown in Figure~\ref{fig:sharpen-discovery-main}. 
We observe that as training proceeds, the model increasingly reinforces previously correct reasoning processes while reducing the frequency of exploration.
For sharpening, the ranking is PSR > DAPO > NSR. 
For discovery, the ranking is NSR > DAPO > PSR. 
These findings indicate that positive samples help models exploit and strengthen previously correct trajectories, while negative samples facilitate exploration of unseen reasoning paths.

\section{Impacts of Advantage Shaping with Different Sample Polarities at Varying Granularities on RLVR Training}
\label{sec-as}

Our previous analyses show that training with only one sample polarity impairs performance, confirming the importance of both positive and negative samples in RLVR. 
In this section, we further investigate how adjusting the influence of each polarity at different granularities affects RLVR training.

\begin{figure*}
    \centering
    \includegraphics[width=\linewidth]{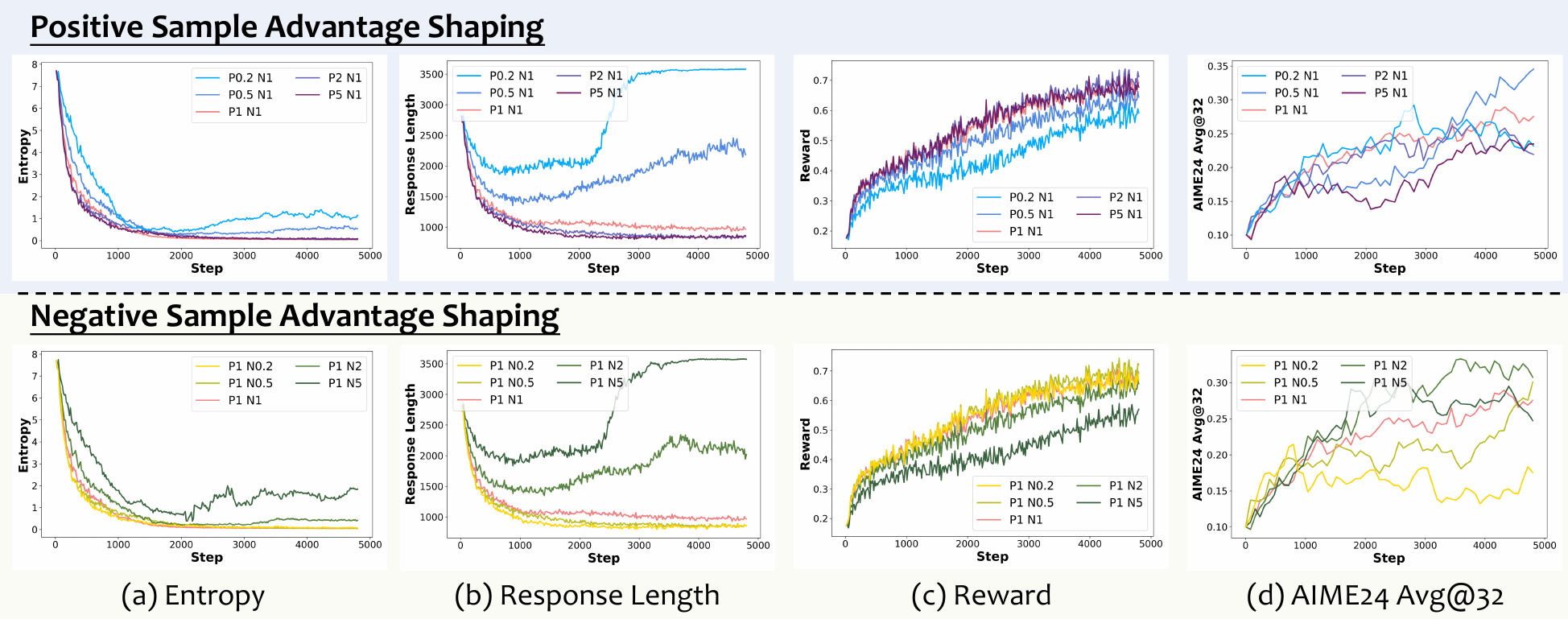}
    \caption{Polarity-level advantage shaping results on Qwen2.5-7B-Math. Each label is formatted as ``PXNY'', where ``X'' and ``Y'' represent the advantage scaling factors for positive and negative samples. For example, ``P1N5'' denotes positive sample weight $\times$1 and negative sample weight $\times$5.}
    \label{fig:polarity-level-as}
\end{figure*}

\subsection{Polarity-level Advantage Shaping}
\label{subsec-pas}

In this part, we conduct polarity-level advantage shaping using Qwen2.5-7B-Math.
Specifically, we scale the advantage values of one sample type by factors of 0.2, 0.5, 2, and 5, while keeping the other type fixed.
More details on polarity-level advantage shaping are provided in Appendix~\ref{appsec:pas}.
Standard RLVR training (1$\times$ for both) is included for comparison.
The results are presented in Figure~\ref{fig:polarity-level-as}.

\paratitle{Higher positive advantage speeds up reward improvement but limits exploration diversity.}
Increasing the advantage values of positive samples accelerates reward growth on the training set, as the model learns more quickly from correct examples.
However, this also makes the model more confident and focused on reinforcing existing successful patterns, thereby limiting exploration diversity. 
Consequently, the model produces responses with lower entropy and shorter lengths.

\paratitle{Higher negative advantage encourages exploration but slows reward improvement.}
Conversely, assigning higher advantages to negative samples encourages the model to avoid mistakes and explore alternatives.
This leads to higher entropy and longer responses, as the model tests various reasoning paths. 
While this maintains diversity, it slows reward improvement on the training set because the model spends more time exploring rather than directly learning from previous successes.

\paratitle{The relative ratio between positive and negative advantage values determines training dynamics.}
Our results show that training dynamics depend primarily on the relative ratio between positive and negative advantage values, not their absolute values.
For example, settings P2N1 and P1N0.5 have the same relative ratio and exhibit similar training trends.
Besides, we also find that excessively high positive advantage causes overfitting to familiar patterns and limits exploration, while overly high negative advantage makes the model overly cautious and slows learning.
Among all settings, a positive-to-negative advantage ratio of 0.5 achieves the best performance on the validation set.
This balanced ratio enables effective learning from both positive and negative samples, which maintains exploration and ensures steady reward improvement.

\begin{figure*}
    \centering
    \includegraphics[width=\linewidth]{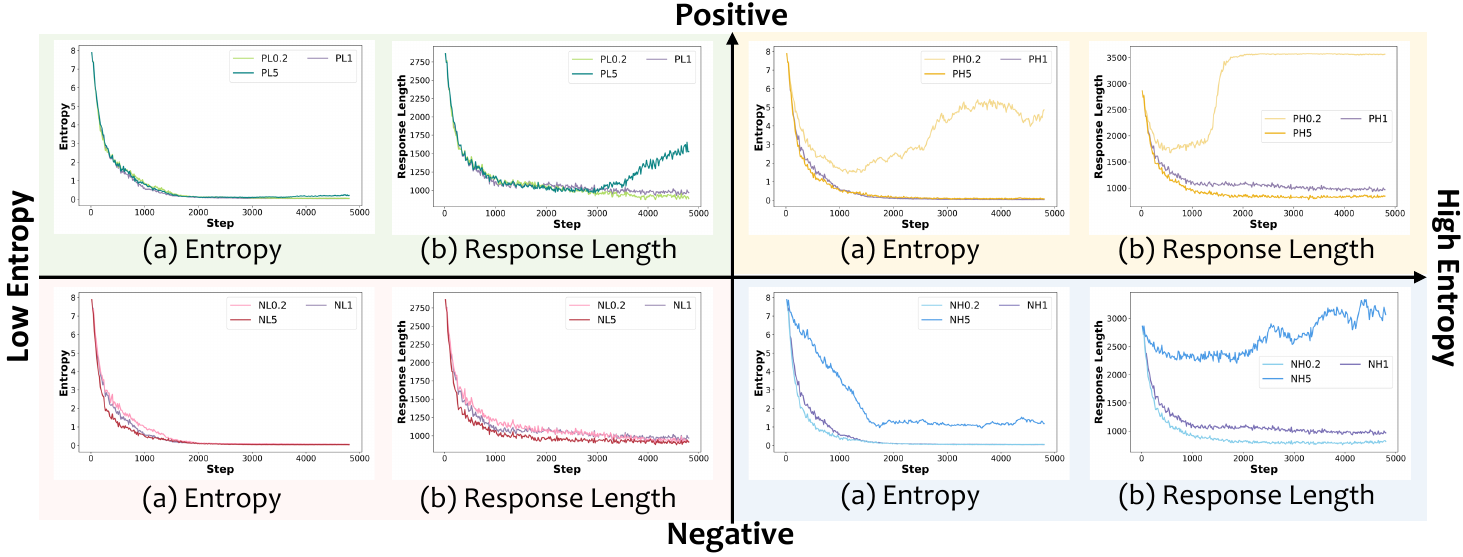}
    \caption{Token-level entropy-based advantage shaping. The x-axis indicates the entropy of shaped tokens (right: high entropy ``H''; left: low entropy ``L''). The y-axis shows shaped token polarity (top: positive ``P''; bottom: negative ``N''). Each label follows the format [Polarity][Entropy][Scaling Factor], where the first letter denotes token polarity, the second indicates entropy level, and the numeric value specifies the scaling factor applied to the advantage of those tokens. In the figure, lines with darker colors correspond to amplifying the advantage values of these tokens, while lighter colors indicate reducing their advantage values.}
    \label{fig:token-entropy-as}
\end{figure*}

\subsection{Token-level Advantage Shaping}
\label{subsec-tas}

To better understand which specific tokens in positive and negative samples contribute more to RLVR training dynamics, we perform token-level advantage shaping.
Specifically, we adjust the advantages assigned to tokens with different entropy and probability distributions, and observe the changes in RLVR training dynamics.
Following prior work~\citep{20-80}, we amplify the advantages of tokens in the top and bottom 20\% based on either entropy or probability using scaling factors of 0.2 and 5, respectively.
More details on token-level advantage shaping are provided in Appendix~\ref{appsec:tas}.
This scaling value amplifies the training dynamics, as larger scaling factors produce more pronounced effects.
Additionally, we also explore different token ratio settings in Appendix~\ref{app:diff-ratio} and find that varying the proportion of weighted tokens does not change the overall training trends.

\paratitle{Entropy-based token-level advantage shaping.}
The experimental results of entropy-based token-level advantage shaping are presented in Figure~\ref{fig:token-entropy-as}.

$\bullet$ Reinforcing \textbf{positive samples with high-entropy tokens} accelerates entropy reduction, as these tokens often represent critical decision points where the model explores multiple reasoning paths.

$\bullet$ Reinforcing \textbf{positive samples with low-entropy tokens} has little effect on training dynamics, since these tokens typically reflect familiar reasoning patterns where the model is already highly confident.

$\bullet$ Reinforcing \textbf{negative samples with high-probability tokens} slows the decrease in entropy, as the model remains uncertain about alternatives even when the current path is incorrect, which helps preserve exploration capacity.

$\bullet$ Reinforcing \textbf{negative samples with low-entropy tokens} speeds up entropy reduction, enabling the model to quickly identify and suppress confident but incorrect reasoning patterns, thereby increasing the confidence of models.

\begin{figure*}
    \centering
    \includegraphics[width=\linewidth]{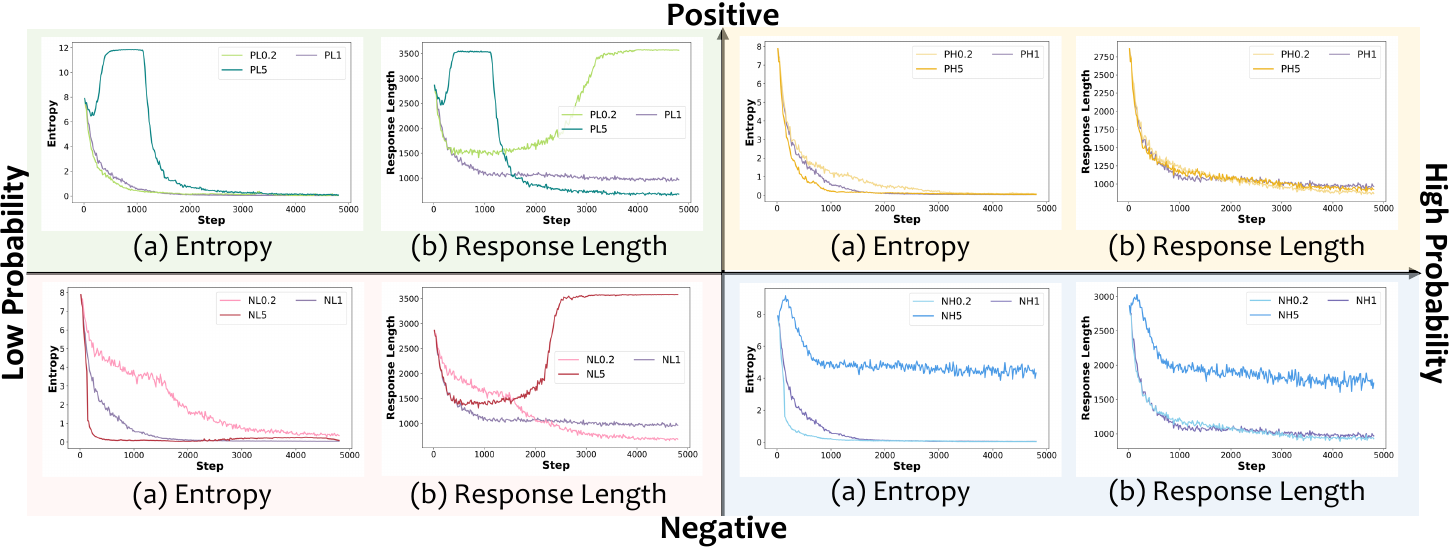}
    \caption{Token-level probability-based advantage shaping. The x-axis indicates the probability of shaped tokens (right: high probability ``H''; left: low probability ``L''). The y-axis shows shaped token polarity (top: positive ``P''; bottom: negative ``N''). Each label follows the format [Polarity][Probability][Scaling Factor], where the first letter denotes shaped token polarity, the second indicates their probability level, and the number is the scaling factor applied to the advantage for these tokens. In the figure, lines with darker colors correspond to amplifying the advantage values of these tokens, while lighter colors indicate reducing their advantage values.}
    \label{fig:token-prob-as}
\end{figure*}

\paratitle{Probability-based token-level advantage shaping.}
The results of Probability-based token-level advantage shaping are illustrated in Figure~\ref{fig:token-prob-as}.

$\bullet$ Reinforcing \textbf{high-probability positive tokens} accelerates entropy reduction, as these tokens represent correct reasoning paths the model has already mastered, which sharpens the policy distribution around these established patterns.

$\bullet$ Reinforcing \textbf{low-probability positive tokens} leads to entropy increase, because encouraging these low-confidence correct alternatives widens the policy distribution and promotes exploration.

$\bullet$ Reinforcing \textbf{high-probability negative tokens} raises entropy. This is because penalizing confidently wrong predictions reduces the model's certainty in high-probability outcomes, encouraging it to reconsider and diversify its predictions.

$\bullet$ Reinforcing \textbf{low-probability negative tokens} reduces entropy, as further suppressing already unlikely incorrect paths reinforces avoidance of these tokens and narrows the policy distribution.
\section{Adaptive and Asymmetric Advantage Shaping for Policy Optimization}
\label{sec-method}

After analyzing how sample polarity influences RLVR training dynamics, we further explore how to leverage this property to enhance the reasoning capabilities of LLMs.
To this end, we propose an adaptive and asymmetric token-level advantage shaping method to achieve stable and effective RLVR optimization.
In this section, we first introduce the method, then describe the experiment setup, and finally present the results.

\begin{table*}[t]
\centering
\Large
\caption{Performance comparison of different methods on various reasoning benchmarks. We highlight the best performance across different RLVR methods. Numbers marked with * indicate that the improvement is statistically significant compared with baselines (t-test with p-value < 0.05).}
\label{tab:performance_comparison}
\resizebox{\textwidth}{!}{
\begin{tabular}{c|l|ccccc|c}
\toprule
\textbf{Model} & \textbf{Method} & \textbf{AIME24} & \textbf{AIME25} & \textbf{MATH500} & \textbf{GPQA} & \textbf{LiveCodeBench} & \textbf{Average} \\
\midrule
\multirow{6}{*}{\begin{tabular}[c]{@{}c@{}}\textbf{Qwen2.5-}\\\textbf{7B-Math}\end{tabular}}
& GRPO & 26.4\std{1.0} & 19.3\std{0.5} & 83.8\std{0.7} & 33.7\std{0.7} & 11.6\std{1.1} & 35.0\std{0.8} \\
& DAPO & 27.6\std{1.0} & 21.4\std{1.2} & 85.4\std{0.6} & 34.6\std{0.8} & 12.4\std{0.5} & 36.3\std{0.8} \\
& DAPO w/ Fork Tokens & 28.6\std{0.7} & 22.5\std{0.9} & 86.8\std{0.5} & 36.5\std{0.7} & 14.3\std{1.0} & 37.7\std{0.8} \\
& W-REINFORCE & 28.3\std{0.9} & 21.4\std{0.7} & 87.3\std{1.0} & 36.2\std{1.1} & 13.8\std{0.5} & 37.4\std{0.8} \\
& Lp-Reg & 29.2\std{1.1} & 22.2\std{1.1} & 87.1\std{0.8} & 36.9\std{0.6} & 13.8\std{1.3} & 37.8\std{1.0} \\
& \OURS & \textbf{31.5}*\std{\textbf{0.8}} & \textbf{24.8}*\std{\textbf{0.6}} & \textbf{90.4}*\std{\textbf{0.6}} & \textbf{39.1}*\std{\textbf{1.2}} & \textbf{16.4}*\std{\textbf{1.0}} & \textbf{40.4}*\std{\textbf{0.8}} \\
\midrule
\multirow{6}{*}{\begin{tabular}[c]{@{}c@{}}\textbf{Qwen3-}\\\textbf{8B-Base}\end{tabular}} 
& GRPO & 32.4\std{0.9} & 23.1\std{0.5} & 82.3\std{0.7} & 45.3\std{0.6} & 29.4\std{0.9} & 42.5\std{0.7} \\
& DAPO & 34.2\std{0.9} & 26.1\std{1.0} & 84.5\std{0.6} & 45.8\std{0.8} & 29.7\std{0.5} & 44.1\std{0.8} \\
& DAPO w/ Fork Tokens & 35.4\std{0.6} & 25.7\std{0.8} & 86.2\std{0.5} & 47.2\std{0.6} & 31.2\std{0.9} & 45.1\std{0.7} \\
& W-REINFORCE & 35.3\std{0.8} & 26.3\std{0.6} & 86.9\std{0.9} & 47.4\std{1.0} & 30.4\std{0.5} & 45.3\std{0.8} \\
& Lp-Reg & 35.9\std{1.0} & 25.8\std{0.9} & 87.4\std{0.7} & 47.8\std{0.6} & 30.9\std{1.1} & 45.6\std{0.9} \\
& \OURS & \textbf{37.8}*\std{\textbf{0.7}} & \textbf{30.4}*\std{\textbf{0.6}} & \textbf{91.3}*\std{\textbf{0.6}} & \textbf{50.2}*\std{\textbf{1.0}} & \textbf{33.8}*\std{\textbf{0.9}} & \textbf{48.7}*\std{\textbf{0.8}} \\
\midrule
\multirow{6}{*}{\begin{tabular}[c]{@{}c@{}}\textbf{Deepseek-R1-}\\\textbf{Distill-Qwen-7B}\end{tabular}} 
& GRPO & 59.4\std{0.5} & 49.2\std{0.5} & 95.2\std{0.5} & 48.4\std{0.3} & 42.5\std{0.7} & 58.9\std{0.5} \\
& DAPO & 60.8\std{0.7} & 50.8\std{0.6} & 95.5\std{0.3} & 50.2\std{0.5} & 43.2\std{0.4} & 60.1\std{0.5} \\
& DAPO w/ Fork Tokens & 61.2\std{0.3} & 51.6\std{0.8} & 95.8\std{0.8} & 50.4\std{0.2} & 44.1\std{0.7} & 60.6\std{0.6} \\
& W-REINFORCE & 61.6\std{0.6} & 51.3\std{0.4} & 95.7\std{0.4} & 51.4\std{0.6} & 44.6\std{0.5} & 60.9\std{0.5} \\
& Lp-Reg & 61.9\std{0.6} & 52.0\std{0.6} & 96.2\std{0.3} & 51.2\std{0.7} & 44.7\std{0.8} & 61.2\std{0.6} \\
& \OURS & \textbf{65.2}*\std{\textbf{0.8}} & \textbf{54.1}*\std{\textbf{0.6}} & \textbf{96.9}*\std{\textbf{0.2}} & \textbf{53.8}*\std{\textbf{0.2}} & \textbf{47.2}*\std{\textbf{0.3}} & \textbf{63.4}*\std{\textbf{0.4}} \\
\bottomrule
\end{tabular}
}
\end{table*}

\subsection{Method}

In our previous analysis, we identified two token types that play important roles in the early stages of RLVR training: low-probability tokens from positive samples and high-probability tokens from negative samples.
These tokens help maintain higher entropy, which encourages continued exploration and prevents premature convergence.
Based on this finding, we propose an adaptive and asymmetric token-level advantage shaping method that dynamically adjusts the weighting of different token categories during training.
Our approach assigns larger advantage values to the above token types early in training to actively encourage exploration.
However, keeping such asymmetric weighting for too long can cause training-inference engine mismatch and performance collapse (See Appendix~\ref{app:prob_diff}).
Therefore, we gradually reduce these weights in a controlled manner as training progresses, allowing the optimization to smoothly transition to a standard training regime.
Our method builds on the DAPO~\citep{DAPO} framework with a modified objective function:
\begin{equation}
\begin{aligned}
& \mathcal{J}_\text{\OURS}(\theta) = \mathbb{E}_{q \sim \mathcal{D} , o\sim \pi_{\theta_\text{old}}(\cdot\mid q)}
\Bigg \{ \sum_{t=1}^{|o|} \min \Big[ 
r_{t} \hat{A}_{t},  \\ 
& \text{clip} ( r_{t}, 1 - \varepsilon_{\text{low}}, 1 + \varepsilon_{\text{high}} ) \hat{A}_{t} \Big] \Bigg \},
\label{eq:oursloss}
\end{aligned}
\end{equation}
where $r_{t}=\frac{\pi_{\theta}(o_{t} \mid q, o_{<t})}{\pi_{\theta_{\text{old}}}(o_{t} \mid q,o_{<t})}$ denotes the ratio between the current and old policies, $\hat{A_i}$ is our shaped advantage, and $\varepsilon_{\text{low}}$ and $\varepsilon_{\text{high}}$ are clipping bounds that constrain policy updates.
The asymmetric and adaptive advantage shaping is defined as:
\begin{equation}
\begin{aligned}
\hat{A}_{t} = 
\begin{cases}
A_t \cdot \max (\rho^{+} - \alpha^{+} s, 1)  & A_t > 0, p_t \le \tau_o^{+} \\
A_t \cdot \max (\rho^{-} - \alpha^{-} s, 1)  & A_t < 0, p_t \ge \tau_o^{-}  \\
A_t & \text{else}.
\end{cases}
\label{eq:oursadv}
\end{aligned}
\end{equation}
Here, $A_t$ is the normalized accuracy across groups, $\tau_o^{+}$ is the threshold for the lowest-probability token in a positive rollout, and $\tau_o^{-}$ is the threshold for the highest-probability token in a negative rollout.
$\rho^+$ and $\rho^-$ denote the initial advantage scaling factors, $\alpha^+$ and $\alpha^-$ control their decay coefficients for positive and negative samples, respectively.

\subsection{Experimental Setup}

We run our experiments on three LLMs (\ie Qwen2.5-7B-Math~\citep{Qwen2.5-Math-7B}), Qwen3-8B-Base~\citep{Qwen3-8B-Base}, and Deepseek-R1-Distill-Qwen-7B~\citep{Deepseek-R1}).
For comparison, we include GRPO~\citep{Deepseek-R1}, DAPO~\citep{DAPO}, polarity-level advantage shaping method (\ie W-REINFORCE~\citep{PSRNSR}), and token-level advantage shaping methods (\ie w/ Fork Tokens~\citep{20-80} and Lp-Reg\citep{Lp-Reg}) as baselines.
Detailed descriptions of the baselines are provided in Appendix~\ref{app:baselines}.
To evaluate the reasoning ability of the methods, we test them on three mathematical (\ie AIME24, AIME25, and MATH500~\citep{Math500}) and two other reasoning benchmarks (\ie GPQA~\citep{GPQA} and LiveCodeBench~\citep{LiveCodeBench}). 
Detailed experimental setups are presented in Appendix~\ref{app:exp-setup}.

\subsection{Main Results}

\begin{figure*}[t]
    \centering
    \begin{subfigure}[b]{0.24\linewidth}
        \centering
        \includegraphics[width=\linewidth]{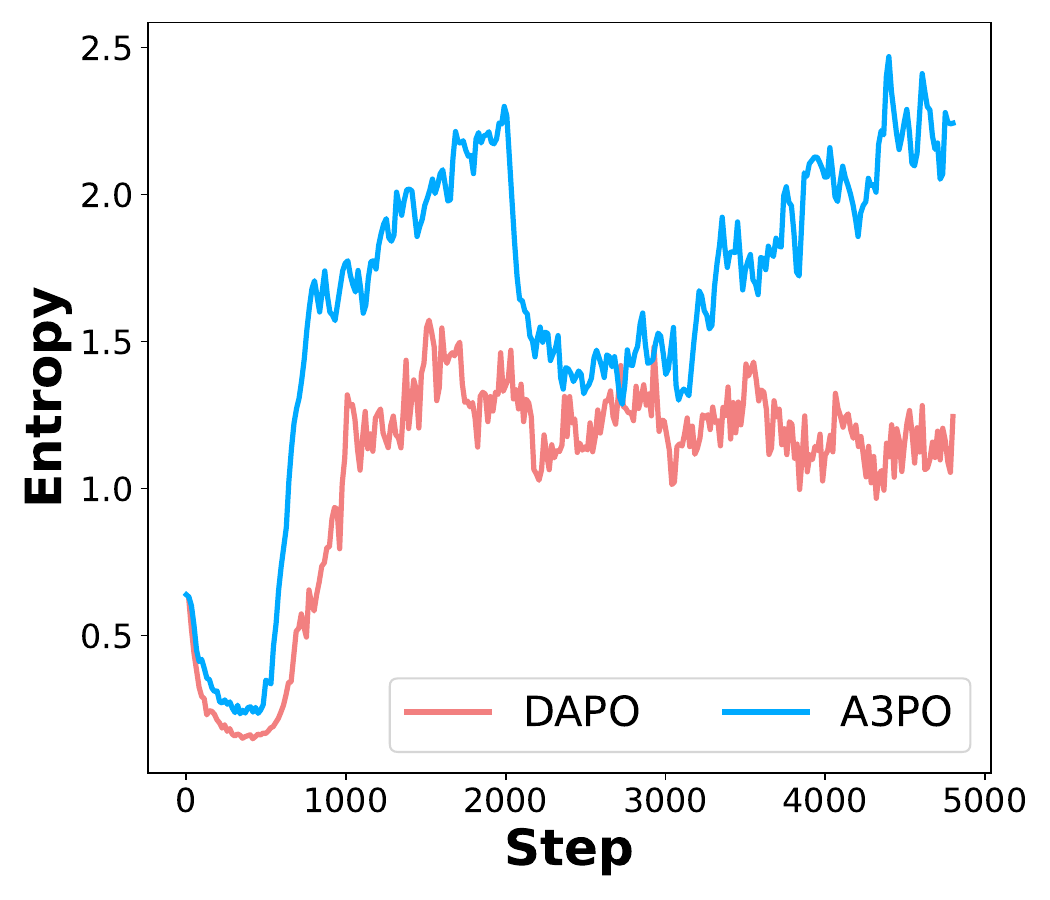}
        \caption{Entropy}
    \end{subfigure}
    \begin{subfigure}[b]{0.24\linewidth}
        \centering
        \includegraphics[width=\linewidth]{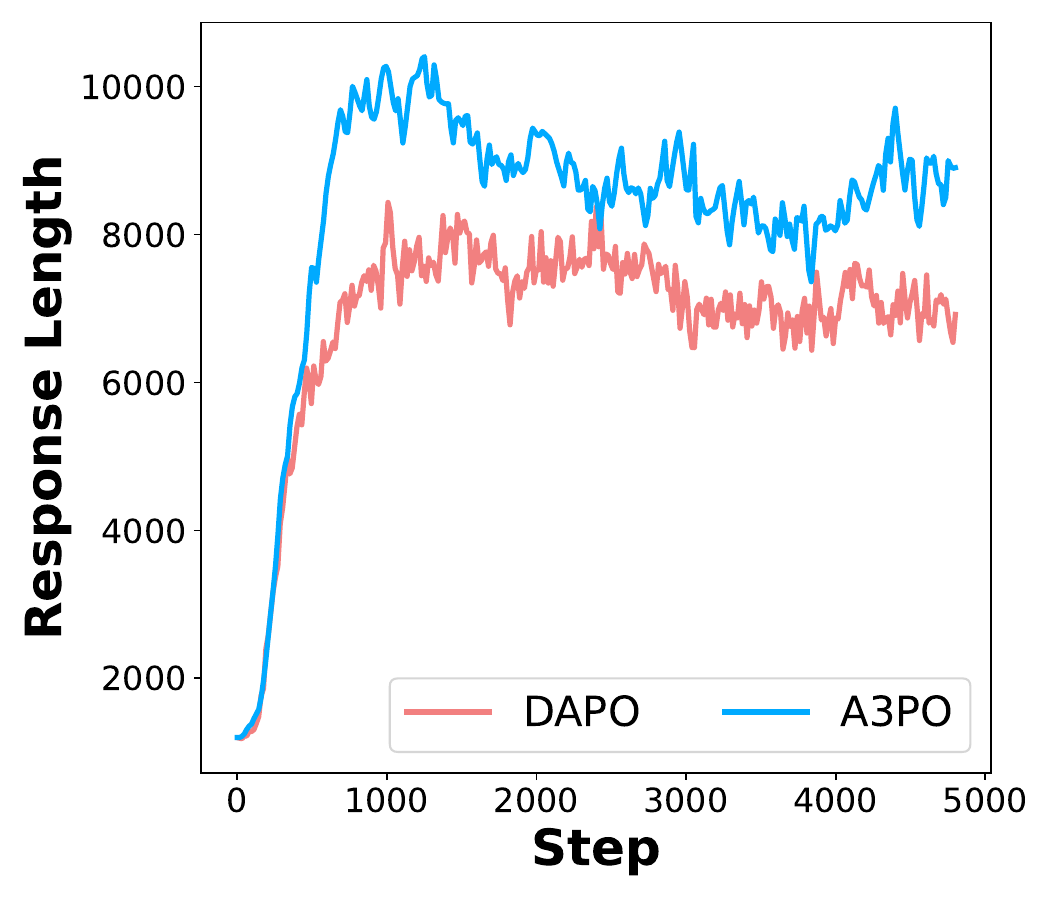}
        \caption{Response Length}
    \end{subfigure}
    \begin{subfigure}[b]{0.24\linewidth}
        \centering
        \includegraphics[width=\linewidth]{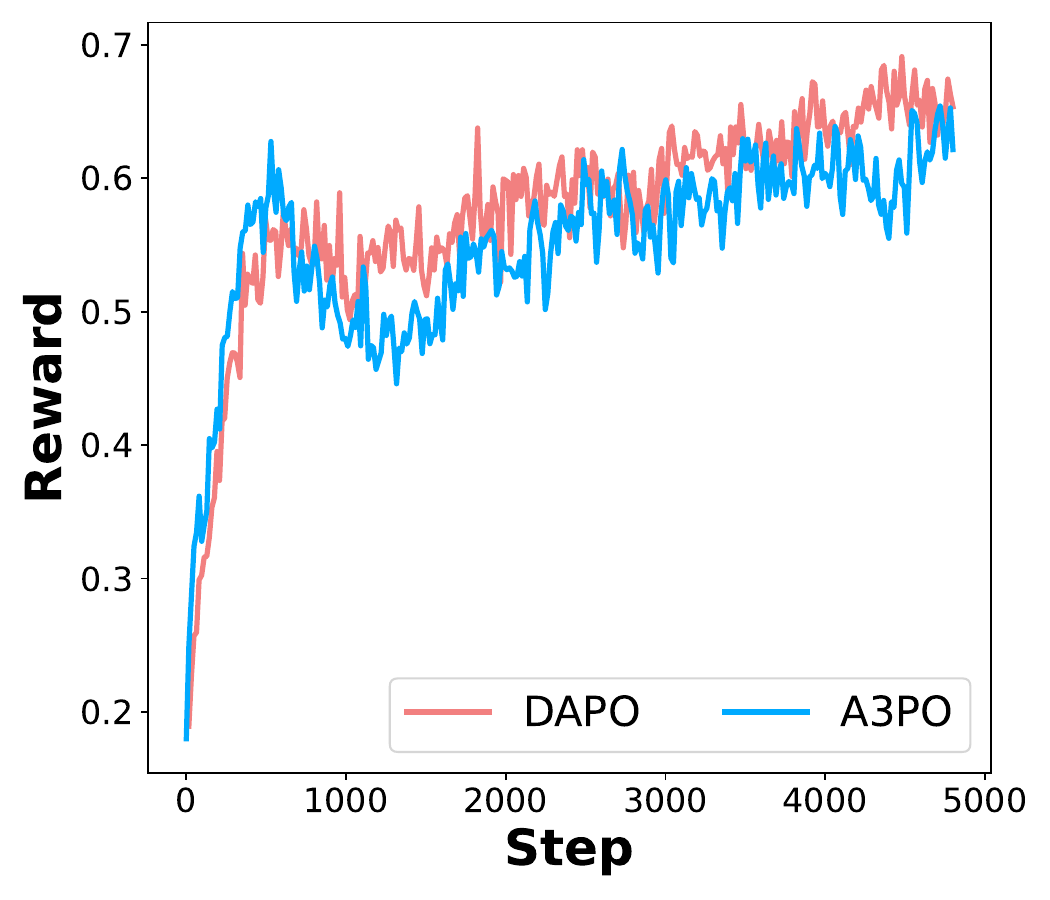}
        \caption{Reward}
    \end{subfigure}
    \begin{subfigure}[b]{0.24\linewidth}
        \centering
        \includegraphics[width=\linewidth]{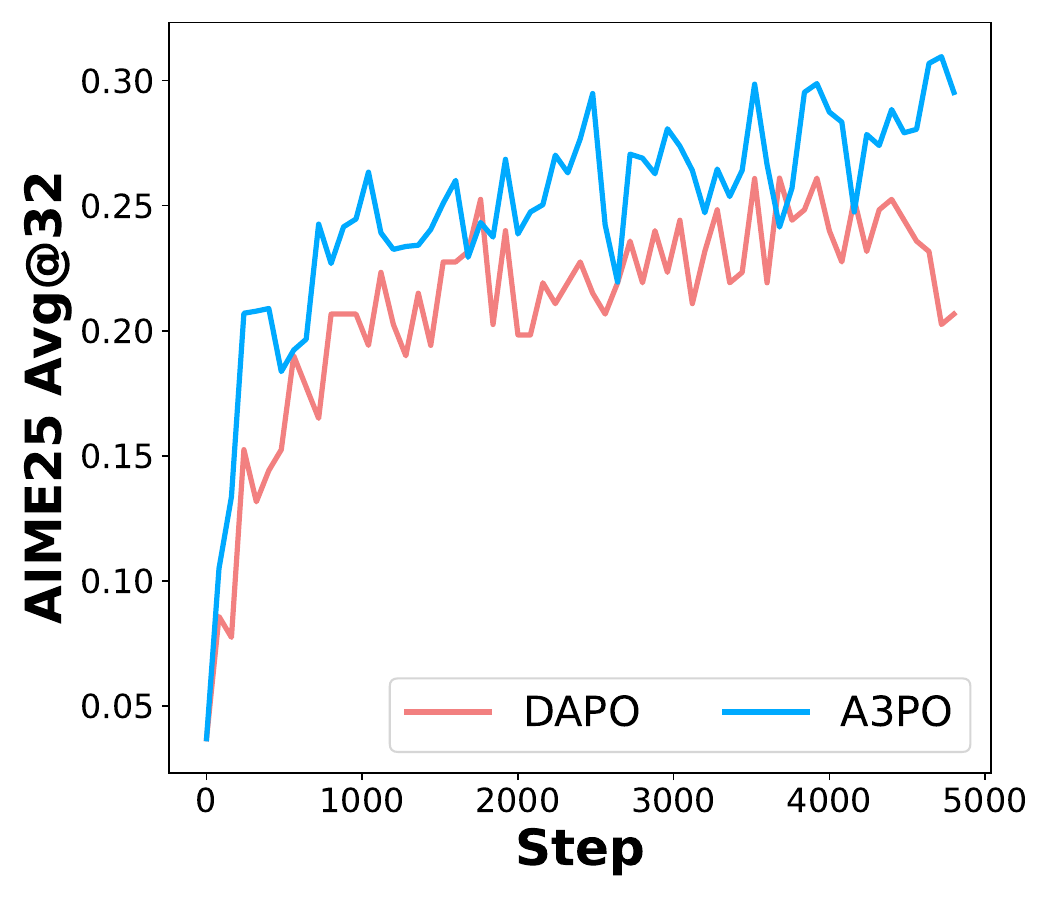}
        \caption{Accuracy}
    \end{subfigure}
    \caption{RLVR training dynamics of DAPO and \OURS on Qwen3-8B-Base.}
\label{fig:our-training-dynamics}
\end{figure*}

Figure~\ref{fig:training-dynamics} compares the training dynamics of DAPO and \OURS.
We observe that \OURS maintains higher entropy and longer responses throughout training, suggesting that the model preserves a richer probability distribution and avoids premature convergence to a narrow output mode.
Although \OURS shows a slightly slower growth in training reward compared to DAPO, it achieves higher validation accuracy, with the performance gap widening as training progresses.
These results suggest that the policy learned by \OURS generalizes better, allowing the model to acquire more general reasoning capabilities rather than merely memorizing patterns in the training data.

The main results are presented in Table~\ref{tab:performance_comparison}.
We observe that DAPO further improves performance over GRPO, which can be attributed to its clip higher mechanism, as it retains low-probability positive tokens and helps the model learn novel reasoning paths.
The sentence-level advantage shaping method further boosts performance by assigning higher advantage values to negative samples.
Additionally, DAPO w/Fork Tokens and Lp-Reg yield gains by assigning higher weights to high-entropy tokens and regularizing low-probability tokens, respectively.
However, these methods do not account for the opposing effects that high-entropy and low-probability tokens can have in positive versus negative samples on RLVR training dynamics.
Treating all tokens uniformly may partially counteract their respective contributions.
To address this question, we propose an adaptive and asymmetric token-level advantage shaping method, which dynamically adjusts the advantage values of high-probability tokens in negative samples and low-probability tokens in positive samples.
This finer-grained allocation of advantages enables more stable and effective RLVR training, ultimately achieving the best performance.
More detailed analyses of \OURS are presented in Appendix~\ref{app:detailed_analysis}.
\section{Conclusion}
\label{sec-conclusion}

In this paper, we systematically analyzed the roles of positive and negative samples in RLVR, demonstrating their distinct contributions to training dynamics.
Our findings showed that positive samples sharpen correct reasoning patterns, while negative samples promote exploration, and both are essential for RLVR training. 
Based on these findings, we proposed an adaptive and asymmetric token-level advantage shaping method that allowed more precise allocation of advantages and led to stable and improved RLVR training.
Experiments across multiple models and benchmarks validate the effectiveness of our approach.
\section{Limitations}
\label{sec-limitations}

In this paper, we provide a comprehensive analysis of Reinforcement Learning with Verifiable Rewards~(RLVR) from the perspectives of sample polarity, revealing the different roles of positive and negative samples during RLVR training.
One limitation of this work is that our experiments are conducted only on the text-based reasoning tasks.
In future work, we plan to extend our analysis and methods to other model families, including vision–language models.
Additionally, due to constraints in computational resources and budget, we have not evaluated our analysis and approach in agent-based scenarios, such as search or code agents.


\bibliography{newbib}

@article{Deepseek-R1,
  author       = {DeepSeek{-}AI and
                  Daya Guo and
                  Dejian Yang and
                  Haowei Zhang and
                  Junxiao Song and
                  Ruoyu Zhang and
                  Runxin Xu and
                  Qihao Zhu and
                  Shirong Ma and
                  Peiyi Wang and
                  Xiao Bi and
                  Xiaokang Zhang and
                  Xingkai Yu and
                  Yu Wu and
                  Z. F. Wu and
                  Zhibin Gou and
                  Zhihong Shao and
                  Zhuoshu Li and
                  Ziyi Gao and
                  Aixin Liu and
                  Bing Xue and
                  Bingxuan Wang and
                  Bochao Wu and
                  Bei Feng and
                  Chengda Lu and
                  Chenggang Zhao and
                  Chengqi Deng and
                  Chenyu Zhang and
                  Chong Ruan and
                  Damai Dai and
                  Deli Chen and
                  Dongjie Ji and
                  Erhang Li and
                  Fangyun Lin and
                  Fucong Dai and
                  Fuli Luo and
                  Guangbo Hao and
                  Guanting Chen and
                  Guowei Li and
                  H. Zhang and
                  Han Bao and
                  Hanwei Xu and
                  Haocheng Wang and
                  Honghui Ding and
                  Huajian Xin and
                  Huazuo Gao and
                  Hui Qu and
                  Hui Li and
                  Jianzhong Guo and
                  Jiashi Li and
                  Jiawei Wang and
                  Jingchang Chen and
                  Jingyang Yuan and
                  Junjie Qiu and
                  Junlong Li and
                  J. L. Cai and
                  Jiaqi Ni and
                  Jian Liang and
                  Jin Chen and
                  Kai Dong and
                  Kai Hu and
                  Kaige Gao and
                  Kang Guan and
                  Kexin Huang and
                  Kuai Yu and
                  Lean Wang and
                  Lecong Zhang and
                  Liang Zhao and
                  Litong Wang and
                  Liyue Zhang and
                  Lei Xu and
                  Leyi Xia and
                  Mingchuan Zhang and
                  Minghua Zhang and
                  Minghui Tang and
                  Meng Li and
                  Miaojun Wang and
                  Mingming Li and
                  Ning Tian and
                  Panpan Huang and
                  Peng Zhang and
                  Qiancheng Wang and
                  Qinyu Chen and
                  Qiushi Du and
                  Ruiqi Ge and
                  Ruisong Zhang and
                  Ruizhe Pan and
                  Runji Wang and
                  R. J. Chen and
                  R. L. Jin and
                  Ruyi Chen and
                  Shanghao Lu and
                  Shangyan Zhou and
                  Shanhuang Chen and
                  Shengfeng Ye and
                  Shiyu Wang and
                  Shuiping Yu and
                  Shunfeng Zhou and
                  Shuting Pan and
                  S. S. Li},
  title        = {DeepSeek-R1: Incentivizing Reasoning Capability in LLMs via Reinforcement
                  Learning},
  journal      = {CoRR},
  volume       = {abs/2501.12948},
  year         = {2025}
}

@article{Kimi-K2,
  author       = {Yifan Bai and
                  Yiping Bao and
                  Guanduo Chen and
                  Jiahao Chen and
                  Ningxin Chen and
                  Ruijue Chen and
                  Yanru Chen and
                  Yuankun Chen and
                  Yutian Chen and
                  Zhuofu Chen and
                  Jialei Cui and
                  Hao Ding and
                  Mengnan Dong and
                  Angang Du and
                  Chenzhuang Du and
                  Dikang Du and
                  Yulun Du and
                  Yu Fan and
                  Yichen Feng and
                  Kelin Fu and
                  Bofei Gao and
                  Hongcheng Gao and
                  Peizhong Gao and
                  Tong Gao and
                  Xinran Gu and
                  Longyu Guan and
                  Haiqing Guo and
                  Jianhang Guo and
                  Hao Hu and
                  Xiaoru Hao and
                  Tianhong He and
                  Weiran He and
                  Wenyang He and
                  Chao Hong and
                  Yangyang Hu and
                  Zhenxing Hu and
                  Weixiao Huang and
                  Zhiqi Huang and
                  Zihao Huang and
                  Tao Jiang and
                  Zhejun Jiang and
                  Xinyi Jin and
                  Yongsheng Kang and
                  Guokun Lai and
                  Cheng Li and
                  Fang Li and
                  Haoyang Li and
                  Ming Li and
                  Wentao Li and
                  Yanhao Li and
                  Yiwei Li and
                  Zhaowei Li and
                  Zheming Li and
                  Hongzhan Lin and
                  Xiaohan Lin and
                  Zongyu Lin and
                  Chengyin Liu and
                  Chenyu Liu and
                  Hongzhang Liu and
                  Jingyuan Liu and
                  Junqi Liu and
                  Liang Liu and
                  Shaowei Liu and
                  T. Y. Liu and
                  Tianwei Liu and
                  Weizhou Liu and
                  Yangyang Liu and
                  Yibo Liu and
                  Yiping Liu and
                  Yue Liu and
                  Zhengying Liu and
                  Enzhe Lu and
                  Lijun Lu and
                  Shengling Ma and
                  Xinyu Ma and
                  Yingwei Ma and
                  Shaoguang Mao and
                  Jie Mei and
                  Xin Men and
                  Yibo Miao and
                  Siyuan Pan and
                  Yebo Peng and
                  Ruoyu Qin and
                  Bowen Qu and
                  Zeyu Shang and
                  Lidong Shi and
                  Shengyuan Shi and
                  Feifan Song and
                  Jianlin Su and
                  Zhengyuan Su and
                  Xinjie Sun and
                  Flood Sung and
                  Heyi Tang and
                  Jiawen Tao and
                  Qifeng Teng and
                  Chensi Wang and
                  Dinglu Wang and
                  Feng Wang and
                  Haiming Wang},
  title        = {Kimi {K2:} Open Agentic Intelligence},
  journal      = {CoRR},
  volume       = {abs/2507.20534},
  year         = {2025}
}

@article{SFT,
  author       = {Shengyu Zhang and
                  Linfeng Dong and
                  Xiaoya Li and
                  Sen Zhang and
                  Xiaofei Sun and
                  Shuhe Wang and
                  Jiwei Li and
                  Runyi Hu and
                  Tianwei Zhang and
                  Fei Wu and
                  Guoyin Wang},
  title        = {Instruction Tuning for Large Language Models: {A} Survey},
  journal      = {CoRR},
  volume       = {abs/2308.10792},
  year         = {2023}
}

@article{VAPO,
  author       = {Yu Yue and
                  Yufeng Yuan and
                  Qiying Yu and
                  Xiaochen Zuo and
                  Ruofei Zhu and
                  Wenyuan Xu and
                  Jiaze Chen and
                  Cheng{-}Xiang Wang and
                  Tiantian Fan and
                  Zhengyin Du and
                  Xiangpeng Wei and
                  Xiangyu Yu and
                  Gaohong Liu and
                  Juncai Liu and
                  Lingjun Liu and
                  Haibin Lin and
                  Zhiqi Lin and
                  Bole Ma and
                  Chi Zhang and
                  Mofan Zhang and
                  Wang Zhang and
                  Hang Zhu and
                  Ru Zhang and
                  Xin Liu and
                  Mingxuan Wang and
                  Yonghui Wu and
                  Lin Yan},
  title        = {{VAPO:} Efficient and Reliable Reinforcement Learning for Advanced
                  Reasoning Tasks},
  journal      = {CoRR},
  volume       = {abs/2504.05118},
  year         = {2025}
}

@article{PRIME,
  author       = {Ganqu Cui and
                  Lifan Yuan and
                  Zefan Wang and
                  Hanbin Wang and
                  Wendi Li and
                  Bingxiang He and
                  Yuchen Fan and
                  Tianyu Yu and
                  Qixin Xu and
                  Weize Chen and
                  Jiarui Yuan and
                  Huayu Chen and
                  Kaiyan Zhang and
                  Xingtai Lv and
                  Shuo Wang and
                  Yuan Yao and
                  Xu Han and
                  Hao Peng and
                  Yu Cheng and
                  Zhiyuan Liu and
                  Maosong Sun and
                  Bowen Zhou and
                  Ning Ding},
  title        = {Process Reinforcement through Implicit Rewards},
  journal      = {CoRR},
  volume       = {abs/2502.01456},
  year         = {2025}
}

@article{CISPO,
  author       = {Aili Chen and
                  Aonian Li and
                  Bangwei Gong and
                  Binyang Jiang and
                  Bo Fei and
                  Bo Yang and
                  Boji Shan and
                  Changqing Yu and
                  Chao Wang and
                  Cheng Zhu and
                  Chengjun Xiao and
                  Chengyu Du and
                  Chi Zhang and
                  Chu Qiao and
                  Chunhao Zhang and
                  Chunhui Du and
                  Congchao Guo and
                  Da Chen and
                  Deming Ding and
                  Dianjun Sun and
                  Dong Li and
                  Enwei Jiao and
                  Haigang Zhou and
                  Haimo Zhang and
                  Han Ding and
                  Haohai Sun and
                  Haoyu Feng and
                  Huaiguang Cai and
                  Haichao Zhu and
                  Jian Sun and
                  Jiaqi Zhuang and
                  Jiaren Cai and
                  Jiayuan Song and
                  Jin Zhu and
                  Jingyang Li and
                  Jinhao Tian and
                  Jinli Liu and
                  Junhao Xu and
                  Junjie Yan and
                  Junteng Liu and
                  Junxian He and
                  Kaiyi Feng and
                  Ke Yang and
                  Kecheng Xiao and
                  Le Han and
                  Leyang Wang and
                  Lianfei Yu and
                  Liheng Feng and
                  Lin Li and
                  Lin Zheng and
                  Linge Du and
                  Lingyu Yang and
                  Lunbin Zeng and
                  Minghui Yu and
                  Mingliang Tao and
                  Mingyuan Chi and
                  Mozhi Zhang and
                  Mujie Lin and
                  Nan Hu and
                  Nongyu Di and
                  Peng Gao and
                  Pengfei Li and
                  Pengyu Zhao and
                  Qibing Ren and
                  Qidi Xu and
                  Qile Li and
                  Qin Wang and
                  Rong Tian and
                  Ruitao Leng and
                  Shaoxiang Chen and
                  Shaoyu Chen and
                  Shengmin Shi and
                  Shitong Weng and
                  Shuchang Guan and
                  Shuqi Yu and
                  Sichen Li and
                  Songquan Zhu and
                  Tengfei Li and
                  Tianchi Cai and
                  Tianrun Liang and
                  Weiyu Cheng and
                  Weize Kong and
                  Wenkai Li and
                  Xiancai Chen and
                  Xiangjun Song and
                  Xiao Luo and
                  Xiao Su and
                  Xiaobo Li and
                  Xiaodong Han and
                  Xinzhu Hou and
                  Xuan Lu and
                  Xun Zou and
                  Xuyang Shen and
                  Yan Gong and
                  Yan Ma and
                  Yang Wang and
                  Yiqi Shi and
                  Yiran Zhong and
                  Yonghong Duan},
  title        = {MiniMax-M1: Scaling Test-Time Compute Efficiently with Lightning Attention},
  journal      = {CoRR},
  volume       = {abs/2506.13585},
  year         = {2025}
}

@article{GMPO,
  author       = {Yuzhong Zhao and
                  Yue Liu and
                  Junpeng Liu and
                  Jingye Chen and
                  Xun Wu and
                  Yaru Hao and
                  Tengchao Lv and
                  Shaohan Huang and
                  Lei Cui and
                  Qixiang Ye and
                  Fang Wan and
                  Furu Wei},
  title        = {Geometric-Mean Policy Optimization},
  journal      = {CoRR},
  volume       = {abs/2507.20673},
  year         = {2025}
}

@article{DAPO,
  author       = {Qiying Yu and
                  Zheng Zhang and
                  Ruofei Zhu and
                  Yufeng Yuan and
                  Xiaochen Zuo and
                  Yu Yue and
                  Tiantian Fan and
                  Gaohong Liu and
                  Lingjun Liu and
                  Xin Liu and
                  Haibin Lin and
                  Zhiqi Lin and
                  Bole Ma and
                  Guangming Sheng and
                  Yuxuan Tong and
                  Chi Zhang and
                  Mofan Zhang and
                  Wang Zhang and
                  Hang Zhu and
                  Jinhua Zhu and
                  Jiaze Chen and
                  Jiangjie Chen and
                  Chengyi Wang and
                  Hongli Yu and
                  Weinan Dai and
                  Yuxuan Song and
                  Xiangpeng Wei and
                  Hao Zhou and
                  Jingjing Liu and
                  Wei{-}Ying Ma and
                  Ya{-}Qin Zhang and
                  Lin Yan and
                  Mu Qiao and
                  Yonghui Wu and
                  Mingxuan Wang},
  title        = {{DAPO:} An Open-Source {LLM} Reinforcement Learning System at Scale},
  journal      = {CoRR},
  volume       = {abs/2503.14476},
  year         = {2025}
}

@article{GSPO,
  author       = {Chujie Zheng and
                  Shixuan Liu and
                  Mingze Li and
                  Xiong{-}Hui Chen and
                  Bowen Yu and
                  Chang Gao and
                  Kai Dang and
                  Yuqiong Liu and
                  Rui Men and
                  An Yang and
                  Jingren Zhou and
                  Junyang Lin},
  title        = {Group Sequence Policy Optimization},
  journal      = {CoRR},
  volume       = {abs/2507.18071},
  year         = {2025}
}

@article{SPO,
  author       = {Yiran Guo and
                  Lijie Xu and
                  Jie Liu and
                  Dan Ye and
                  Shuang Qiu},
  title        = {Segment Policy Optimization: Effective Segment-Level Credit Assignment
                  in {RL} for Large Language Models},
  journal      = {CoRR},
  volume       = {abs/2505.23564},
  year         = {2025}
}

@article{Treepo,
  author       = {Yizhi Li and
                  Qingshui Gu and
                  Zhoufutu Wen and
                  Ziniu Li and
                  Tianshun Xing and
                  Shuyue Guo and
                  Tianyu Zheng and
                  Xin Zhou and
                  Xingwei Qu and
                  Wangchunshu Zhou and
                  Zheng Zhang and
                  Wei Shen and
                  Qian Liu and
                  Chenghua Lin and
                  Jian Yang and
                  Ge Zhang and
                  Wenhao Huang},
  title        = {TreePO: Bridging the Gap of Policy Optimization and Efficacy and Inference
                  Efficiency with Heuristic Tree-based Modeling},
  journal      = {CoRR},
  volume       = {abs/2508.17445},
  year         = {2025}
}

@article{PSRNSR,
  author       = {Xinyu Zhu and
                  Mengzhou Xia and
                  Zhepei Wei and
                  Wei{-}Lin Chen and
                  Danqi Chen and
                  Yu Meng},
  title        = {The Surprising Effectiveness of Negative Reinforcement in {LLM} Reasoning},
  journal      = {CoRR},
  volume       = {abs/2506.01347},
  year         = {2025}
}

@misc{ASPO,
      title={ASPO: Asymmetric Importance Sampling Policy Optimization}, 
      author={Jiakang Wang and Runze Liu and Lei Lin and Wenping Hu and Xiu Li and Fuzheng Zhang and Guorui Zhou and Kun Gai},
      year={2025},
      eprint={2510.06062},
      archivePrefix={arXiv},
      primaryClass={cs.CL},
      url={https://arxiv.org/abs/2510.06062}, 
}

@misc{STEER,
      title={Rethinking Entropy Interventions in RLVR: An Entropy Change Perspective}, 
      author={Zhezheng Hao and Hong Wang and Haoyang Liu and Jian Luo and Jiarui Yu and Hande Dong and Qiang Lin and Can Wang and Jiawei Chen},
      year={2025},
      eprint={2510.10150},
      archivePrefix={arXiv},
      primaryClass={cs.LG},
      url={https://arxiv.org/abs/2510.10150}, 
}

@misc{BAPO,
      title={BAPO: Stabilizing Off-Policy Reinforcement Learning for LLMs via Balanced Policy Optimization with Adaptive Clipping}, 
      author={Zhiheng Xi and Xin Guo and Yang Nan and Enyu Zhou and Junrui Shen and Wenxiang Chen and Jiaqi Liu and Jixuan Huang and Zhihao Zhang and Honglin Guo and Xun Deng and Zhikai Lei and Miao Zheng and Guoteng Wang and Shuo Zhang and Peng Sun and Rui Zheng and Hang Yan and Tao Gui and Qi Zhang and Xuanjing Huang},
      year={2025},
      eprint={2510.18927},
      archivePrefix={arXiv},
      primaryClass={cs.LG},
      url={https://arxiv.org/abs/2510.18927}, 
}

@article{Qwen2.5-Math-7B,
  author       = {An Yang and
                  Beichen Zhang and
                  Binyuan Hui and
                  Bofei Gao and
                  Bowen Yu and
                  Chengpeng Li and
                  Dayiheng Liu and
                  Jianhong Tu and
                  Jingren Zhou and
                  Junyang Lin and
                  Keming Lu and
                  Mingfeng Xue and
                  Runji Lin and
                  Tianyu Liu and
                  Xingzhang Ren and
                  Zhenru Zhang},
  title        = {Qwen2.5-Math Technical Report: Toward Mathematical Expert Model via
                  Self-Improvement},
  journal      = {CoRR},
  volume       = {abs/2409.12122},
  year         = {2024}
}

@article{Qwen3-8B-Base,
  author       = {An Yang and
                  Anfeng Li and
                  Baosong Yang and
                  Beichen Zhang and
                  Binyuan Hui and
                  Bo Zheng and
                  Bowen Yu and
                  Chang Gao and
                  Chengen Huang and
                  Chenxu Lv and
                  Chujie Zheng and
                  Dayiheng Liu and
                  Fan Zhou and
                  Fei Huang and
                  Feng Hu and
                  Hao Ge and
                  Haoran Wei and
                  Huan Lin and
                  Jialong Tang and
                  Jian Yang and
                  Jianhong Tu and
                  Jianwei Zhang and
                  Jian Yang and
                  Jiaxi Yang and
                  Jingren Zhou and
                  Junyang Lin and
                  Kai Dang and
                  Keqin Bao and
                  Kexin Yang and
                  Le Yu and
                  Lianghao Deng and
                  Mei Li and
                  Mingfeng Xue and
                  Mingze Li and
                  Pei Zhang and
                  Peng Wang and
                  Qin Zhu and
                  Rui Men and
                  Ruize Gao and
                  Shixuan Liu and
                  Shuang Luo and
                  Tianhao Li and
                  Tianyi Tang and
                  Wenbiao Yin and
                  Xingzhang Ren and
                  Xinyu Wang and
                  Xinyu Zhang and
                  Xuancheng Ren and
                  Yang Fan and
                  Yang Su and
                  Yichang Zhang and
                  Yinger Zhang and
                  Yu Wan and
                  Yuqiong Liu and
                  Zekun Wang and
                  Zeyu Cui and
                  Zhenru Zhang and
                  Zhipeng Zhou and
                  Zihan Qiu},
  title        = {Qwen3 Technical Report},
  journal      = {CoRR},
  volume       = {abs/2505.09388},
  year         = {2025}
}

@article{RL-survey,
  author       = {Kaiyan Zhang and
                  Yuxin Zuo and
                  Bingxiang He and
                  Youbang Sun and
                  Runze Liu and
                  Che Jiang and
                  Yuchen Fan and
                  Kai Tian and
                  Guoli Jia and
                  Pengfei Li and
                  Yu Fu and
                  Xingtai Lv and
                  Yuchen Zhang and
                  Sihang Zeng and
                  Shang Qu and
                  Haozhan Li and
                  Shijie Wang and
                  Yuru Wang and
                  Xinwei Long and
                  Fangfu Liu and
                  Xiang Xu and
                  Jiaze Ma and
                  Xuekai Zhu and
                  Ermo Hua and
                  Yihao Liu and
                  Zonglin Li and
                  Huayu Chen and
                  Xiaoye Qu and
                  Yafu Li and
                  Weize Chen and
                  Zhenzhao Yuan and
                  Junqi Gao and
                  Dong Li and
                  Zhiyuan Ma and
                  Ganqu Cui and
                  Zhiyuan Liu and
                  Biqing Qi and
                  Ning Ding and
                  Bowen Zhou},
  title        = {A Survey of Reinforcement Learning for Large Reasoning Models},
  journal      = {CoRR},
  volume       = {abs/2509.08827},
  year         = {2025}
}

@article{limit-of-RLVR,
  author       = {Yang Yue and
                  Zhiqi Chen and
                  Rui Lu and
                  Andrew Zhao and
                  Zhaokai Wang and
                  Yang Yue and
                  Shiji Song and
                  Gao Huang},
  title        = {Does Reinforcement Learning Really Incentivize Reasoning Capacity
                  in LLMs Beyond the Base Model?},
  journal      = {CoRR},
  volume       = {abs/2504.13837},
  year         = {2025}
}

@article{ProRL,
  author       = {Mingjie Liu and
                  Shizhe Diao and
                  Ximing Lu and
                  Jian Hu and
                  Xin Dong and
                  Yejin Choi and
                  Jan Kautz and
                  Yi Dong},
  title        = {ProRL: Prolonged Reinforcement Learning Expands Reasoning Boundaries
                  in Large Language Models},
  journal      = {CoRR},
  volume       = {abs/2505.24864},
  year         = {2025}
}

@article{20-80,
  author       = {Shenzhi Wang and
                  Le Yu and
                  Chang Gao and
                  Chujie Zheng and
                  Shixuan Liu and
                  Rui Lu and
                  Kai Dang and
                  Xionghui Chen and
                  Jianxin Yang and
                  Zhenru Zhang and
                  Yuqiong Liu and
                  An Yang and
                  Andrew Zhao and
                  Yang Yue and
                  Shiji Song and
                  Bowen Yu and
                  Gao Huang and
                  Junyang Lin},
  title        = {Beyond the 80/20 Rule: High-Entropy Minority Tokens Drive Effective
                  Reinforcement Learning for {LLM} Reasoning},
  journal      = {CoRR},
  volume       = {abs/2506.01939},
  year         = {2025}
}

@article{Lp-Reg,
  author       = {Guanhua Huang and
                  Tingqiang Xu and
                  Mingze Wang and
                  Qi Yi and
                  Xue Gong and
                  Siheng Li and
                  Ruibin Xiong and
                  Kejiao Li and
                  Yuhao Jiang and
                  Bo Zhou},
  title        = {Low-probability Tokens Sustain Exploration in Reinforcement Learning
                  with Verifiable Reward},
  journal      = {CoRR},
  volume       = {abs/2510.03222},
  year         = {2025}
}

@inproceedings{Verl,
  author       = {Guangming Sheng and
                  Chi Zhang and
                  Zilingfeng Ye and
                  Xibin Wu and
                  Wang Zhang and
                  Ru Zhang and
                  Yanghua Peng and
                  Haibin Lin and
                  Chuan Wu},
  title        = {HybridFlow: {A} Flexible and Efficient {RLHF} Framework},
  booktitle    = {EuroSys},
  pages        = {1279--1297},
  publisher    = {{ACM}},
  year         = {2025}
}

@inproceedings{vllm,
  author       = {Woosuk Kwon and
                  Zhuohan Li and
                  Siyuan Zhuang and
                  Ying Sheng and
                  Lianmin Zheng and
                  Cody Hao Yu and
                  Joseph Gonzalez and
                  Hao Zhang and
                  Ion Stoica},
  title        = {Efficient Memory Management for Large Language Model Serving with
                  PagedAttention},
  booktitle    = {{SOSP}},
  pages        = {611--626},
  publisher    = {{ACM}},
  year         = {2023}
}

@inproceedings{AdamW,
  author       = {Ilya Loshchilov and
                  Frank Hutter},
  title        = {Decoupled Weight Decay Regularization},
  booktitle    = {{ICLR} (Poster)},
  publisher    = {OpenReview.net},
  year         = {2019}
}

@article{FSDP,
  author       = {Yanli Zhao and
                  Andrew Gu and
                  Rohan Varma and
                  Liang Luo and
                  Chien{-}Chin Huang and
                  Min Xu and
                  Less Wright and
                  Hamid Shojanazeri and
                  Myle Ott and
                  Sam Shleifer and
                  Alban Desmaison and
                  Can Balioglu and
                  Pritam Damania and
                  Bernard Nguyen and
                  Geeta Chauhan and
                  Yuchen Hao and
                  Ajit Mathews and
                  Shen Li},
  title        = {PyTorch {FSDP:} Experiences on Scaling Fully Sharded Data Parallel},
  journal      = {Proc. {VLDB} Endow.},
  volume       = {16},
  number       = {12},
  pages        = {3848--3860},
  year         = {2023}
}

@inproceedings{Math500,
  author       = {Dan Hendrycks and
                  Collin Burns and
                  Saurav Kadavath and
                  Akul Arora and
                  Steven Basart and
                  Eric Tang and
                  Dawn Song and
                  Jacob Steinhardt},
  title        = {Measuring Mathematical Problem Solving With the {MATH} Dataset},
  booktitle    = {NeurIPS Datasets and Benchmarks},
  year         = {2021}
}

@article{GPQA,
  author       = {David Rein and
                  Betty Li Hou and
                  Asa Cooper Stickland and
                  Jackson Petty and
                  Richard Yuanzhe Pang and
                  Julien Dirani and
                  Julian Michael and
                  Samuel R. Bowman},
  title        = {{GPQA:} {A} Graduate-Level Google-Proof Q{\&}A Benchmark},
  journal      = {CoRR},
  volume       = {abs/2311.12022},
  year         = {2023}
}

@inproceedings{LiveCodeBench,
  author       = {Naman Jain and
                  King Han and
                  Alex Gu and
                  Wen{-}Ding Li and
                  Fanjia Yan and
                  Tianjun Zhang and
                  Sida Wang and
                  Armando Solar{-}Lezama and
                  Koushik Sen and
                  Ion Stoica},
  title        = {LiveCodeBench: Holistic and Contamination Free Evaluation of Large
                  Language Models for Code},
  booktitle    = {{ICLR}},
  publisher    = {OpenReview.net},
  year         = {2025}
}

\appendix

\section{Detailed Experimental Setup}
\label{app:exp-setup}

\paratitle{Models.}
We conduct experiments on three models: Qwen2.5-Math-7B~\citep{Qwen2.5-Math-7B}, Qwen3-8B-Base~\citep{Qwen3-8B-Base}, and DeepSeek-R1-Distill-Qwen-7B~\citep{Deepseek-R1}.
For DeepSeek-R1-Distill-Qwen-7B and Qwen3-8B-Base, we set a context length of 16384 tokens.
For Qwen2.5-Math-7B, we use its maximum supported length of 4,096 tokens.

\paratitle{Training.}
Our implementation is based on the Verl~\citep{verl} pipeline, with rollouts performed using vLLM~\citep{vllm}.
Models are trained on 16×H200 GPUs.
We use the DAPO-Math dataset~\citep{DAPO} for training.
During rollouts, we set the temperature to 1 and sample 8 responses per prompt.
Training follows an off-policy RL setup with a batch size of 512 and a mini-batch size of 32. 
Similar to prior work~\citet{VAPO}, we remove both the KL divergence loss and the entropy loss.
All models are trained for 300 steps, optimized with the AdamW~\citep{AdamW} optimizer using a constant learning rate of 1e-6. 
The actor module is trained efficiently with Fully Sharded Data Parallel~(FSDP)~\citep{FSDP}.
The chat template used is: ``User: \textbackslash n [question] \textbackslash n Please reason step by step, and put your final answer within \textbackslash\textbackslash boxed\{\}. \textbackslash n \textbackslash n Assistant:''.
For hyperparameter settings, we apply adaptive advantage shaping to the lowest 20\% probability tokens in positive samples and the highest 20\% probability tokens in negative samples.
The initial advantage scaling factors $\rho^+$ and $\rho^-$ are set to 2, and the decay coefficients $\alpha^+$ and $\alpha^-$ are set to 0.005.

\paratitle{Evaluation.}
We evaluate model performance on three mathematical reasoning benchmarks (\ie AIME24, AIME25, and Math500~\citep{Math500}) and two additional reasoning benchmarks (\ie GPQA~\citep{GPQA} and LiveCodeBench~\citep{LiveCodeBench}).
Models are evaluated every 5 training steps, and we report results from the checkpoint that achieves the highest average performance across five benchmarks.
All evaluations are performed in a zero-shot setting.
Following ~\citet{Deepseek-R1}, we set the temperature to 0.6 and top‑k to 0.95 during inference.
To ensure stable measurements, each test set is evaluated 32 times, and we report the average accuracy.
\section{Detailed Descriptions of Methods}
\label{app:detailed_methods}

In this section, we provide detailed descriptions of several methods used in the main text, including positive and negative sample reinforcement in Section~\ref{sec-pn-sample-reinforcement} and the polarity-level and token-level advantage shaping method in Section~\ref{sec-as}.

\subsection{Positive and Negative Sample Reinforcement}
\label{appsec:psrnsr}

In Section~\ref{sec-pn-sample-reinforcement}, we follow previous work~\citep{PSRNSR} to decompose the RLVR objective into two different learning paradigms: learning from correct rollouts and learning from incorrect rollouts. 
This decomposition allows us to examine how positive and negative responses affect training dynamics.
The RLVR objective can be expressed as the sum of two sub-objectives:
\begin{equation}
\begin{aligned}
\mathcal{L}_{\text{RLVR}}(\theta) = \mathcal{L}_{\text{PSR}}(\theta) + \mathcal{L}_{\text{NSR}}(\theta),
\end{aligned}
\end{equation}
where the two sub-objectives correspond to each learning paradigm:
\begin{align}
\label{eq:psr_obj}
\mathcal{L}_{\text{PSR}}(\theta)&= - \mathbb{E}_{{x}\sim \mathcal{D}}\left[\sum_{{y}: r({x}, {y})=1}\pi_{\theta}({y}|{x})\right], \\
\label{eq:nsr_obj}
\mathcal{L}_{\text{NSR}}(\theta) &= -\mathbb{E}_{{x}\sim \mathcal{D}}\left[\sum_{{y}: r({x}, {y})=0}-\pi_{\theta}({y}|{x})\right].
\end{align}
We refer to these two learning paradigms as \textbf{\textit{positive sample reinforcement}} and \textbf{\textit{negative sample reinforcement}}. 
Positive sample reinforcement resembles supervised fine-tuning, increasing the likelihood of correct responses.
In contrast, negative sample reinforcement acts like likelihood minimization, reducing the probability of incorrect responses.

\subsection{Polarity-level Advantage Shaping}
\label{appsec:pas}

To explore how adjusting the influence of positive and negative samples affects RLVR training, we introduce a polarity-level advantage shaping method in Section~\ref{subsec-pas}.
This approach assigns different weights to the advantage values derived from positive and negative samples, allowing us to control their relative contributions during policy optimization.
Formally, the objective is defined as:
\begin{equation}
\begin{aligned}
\mathcal{L}_{\text{Polarity-AS}}(\theta) = \beta_\text{P} \cdot \mathcal{L}_{\text{PSR}}(\theta) + \beta_\text{N} \cdot \mathcal{L}_{\text{NSR}}(\theta)
\end{aligned}
\end{equation}
Here, $\beta_\text{P}$ and $\beta_\text{N}$ are scaling factors that control the advantage values for positive and negative samples, respectively. 
By adjusting these scaling factors, we can study how emphasizing or de-emphasizing each sample polarity impacts RLVR training dynamics.

\subsection{Token-level Advantage Shaping}
\label{appsec:tas}

To further examine the contribution of specific tokens to RLVR training, we introduce a token-level advantage shaping approach in Section~\ref{subsec-tas}.
This method allows us to reweight the advantage assigned to selected tokens and observe how such adjustments affect overall training dynamics.
The modified policy optimization objective can be expressed as follows:
\begin{equation}
\begin{aligned}
& \mathcal{J}_\text{Token-AS}(\theta) = \mathbb{E}_{q \sim \mathcal{D} , o\sim \pi_{\theta_\text{old}}(\cdot\mid q)}
\Bigg \{ \sum_{t=1}^{|o|} \min \Big[ 
r_{t} \hat{A}_{t},  \\ 
& \text{clip} ( r_{t}, 1 - \varepsilon_{\text{low}}, 1 + \varepsilon_{\text{high}} ) \hat{A}_{t} \Big] \Bigg \},
\end{aligned}
\end{equation}
where $r_{t}=\frac{\pi_{\theta}(o_{t} \mid q, o_{<t})}{\pi_{\theta_{\text{old}}}(o_{t} \mid q,o_{<t})}$ denotes the probability ratio between the current and old policies, $\varepsilon_{\text{low}}$ and $\varepsilon_{\text{high}}$ are clipping thresholds that constrain policy updates, and $\hat{A_i}$ represents the shaped advantage.
The shaped advantage $\hat{A_i}$ is defined based on whether a token is selected for reweighting:
\begin{equation}
\begin{aligned}
\hat{A}_{t} = 
\begin{cases}
A_t \cdot \beta_{\text{T}}  & \text{if selected}  \\
A_t & \text{else}.
\end{cases}
\end{aligned}
\end{equation}
where ${A_i}$ is the original advantage computed from group rollouts, and $\beta_{\text{T}}$ is a token-level scaling factor applied to the selected token.
Tokens are selected based on specific criteria, such as entropy and probability, enabling us to study how reweighting distinct token categories influences training dynamics.
\section{Detailed Description of Baselines}
\label{app:baselines}

In this part, we provide detailed descriptions of the baseline methods used for comparison in our experiments.
Specifically, we compare our method with GRPO~\citep{Deepseek-R1}, DAPO~\citep{DAPO}, the polarity-level advantage shaping method (\ie W-REINFORCE~\citep{PSRNSR}), and the token-level advantage shaping method (\ie w/ Fork Tokens~\citep{20-80} and Lp-Reg\citep{Lp-Reg}).

$\bullet$ \textbf{\underline{GRPO}}~\citep{Deepseek-R1} is a reinforcement learning algorithm that improves LLM reasoning without training a separate value model. 
For each question, it samples multiple outputs from the current policy and optimizes the policy using a group-relative advantage, making it scalable for long chain-of-thought reasoning tasks.

$\bullet$ \textbf{\underline{DAPO}}~\citep{DAPO} is an enhanced RL method that introduces several improvements for LLM training.
It prevents entropy collapse by using a higher clipping threshold to encourage exploration, applies dynamic sampling to filter prompts with zero variance, adopts token-level policy gradient loss to handle varying response lengths, and removes the KL divergence term for RL training.

$\bullet$ \textbf{\underline{W-REINFORCE}}~\citep{PSRNSR} is a polarity-level advantage shaping method, which assigns higher weights to self-generated negative rollouts, enabling effective RLVR training.

$\bullet$ \textbf{\underline{DAPO w/Fork Tokens}}~\citep{20-80} is a token-level advantage shaping method that focuses policy gradient updates on high-entropy ``forking tokens''.
By masking gradients for the 80\% lowest-entropy tokens and updating only the top 20\% high-entropy tokens, it improves the reasoning performance of LLMs.

$\bullet$ \textbf{\underline{Lp-Reg}}~\citep{Lp-Reg} is a token-level method designed to mitigate exploration collapse. 
It maintains useful low-probability tokens through regularization while filtering out noisy tokens, thereby sustaining exploration throughout RLVR training.
\section{Different Training Dynamics of Base LLMs}
\label{app:diff_train_dynamics}

In this part, we analyze the training dynamics of three RLVR training paradigms (\ie positive sample reinforcement~(PSR), negative sample reinforcement~(NSR), and DAPO) across three base LLMs (\ie Qwen2.5-7B-Math, Qwen3-8B-Base, and Deepseek-R1-Distilled-Qwen-7B).
Specifically, we monitor accuracy changes on all validation samples from AIME24 and AIME25 during training and categorize them into five patterns:

$\bullet$ \textbf{Sharpen}: Accuracy improves by more than $k\%$, indicating that training strengthens the model's ability to solve the problem.

$\bullet$ \textbf{Degradation}: Accuracy drops by more than $k\%$, meaning training reduces reasoning ability

$\bullet$ \textbf{Fluctuation}: Accuracy fluctuates within $k\%$ of the original value, showing training has little effect.

$\bullet$ \textbf{Mastery}: Accuracy remains above $1-k\%$, meaning the model consistently solves the problem correctly.

$\bullet$ \textbf{Struggle}: Accuracy stays below $k\%$, meaning the problem remains too difficult for the model.

\begin{figure*}
    \centering
    \includegraphics[width=\linewidth]{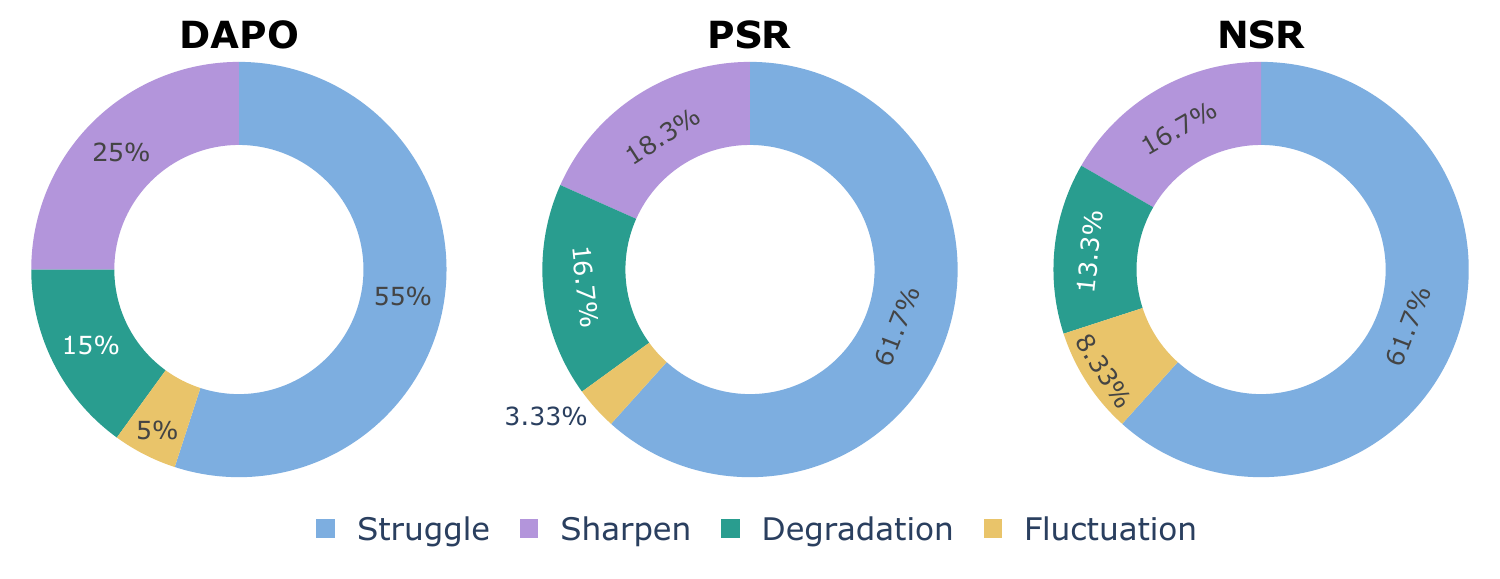}
    \caption{Training dynamics of sample accuracy changes in the validation set on Qwen2.5-7B-Math.}
    \label{fig:training-behavior-qwen7b}
\end{figure*}

\begin{figure*}
    \centering
    \includegraphics[width=\linewidth]{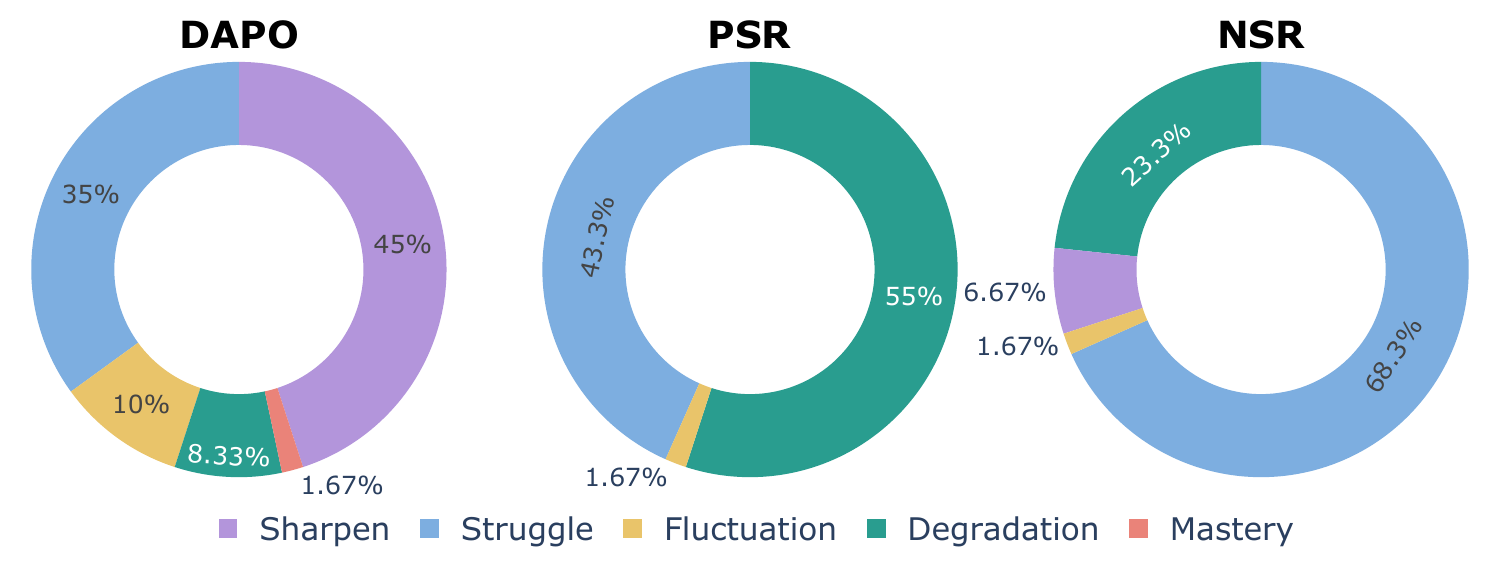}
    \caption{Training dynamics of sample accuracy changes in the validation set on Qwen3-8B-Base.}
    \label{fig:training-behavior-qwen8b}
\end{figure*}

\begin{figure*}
    \centering
    \includegraphics[width=\linewidth]{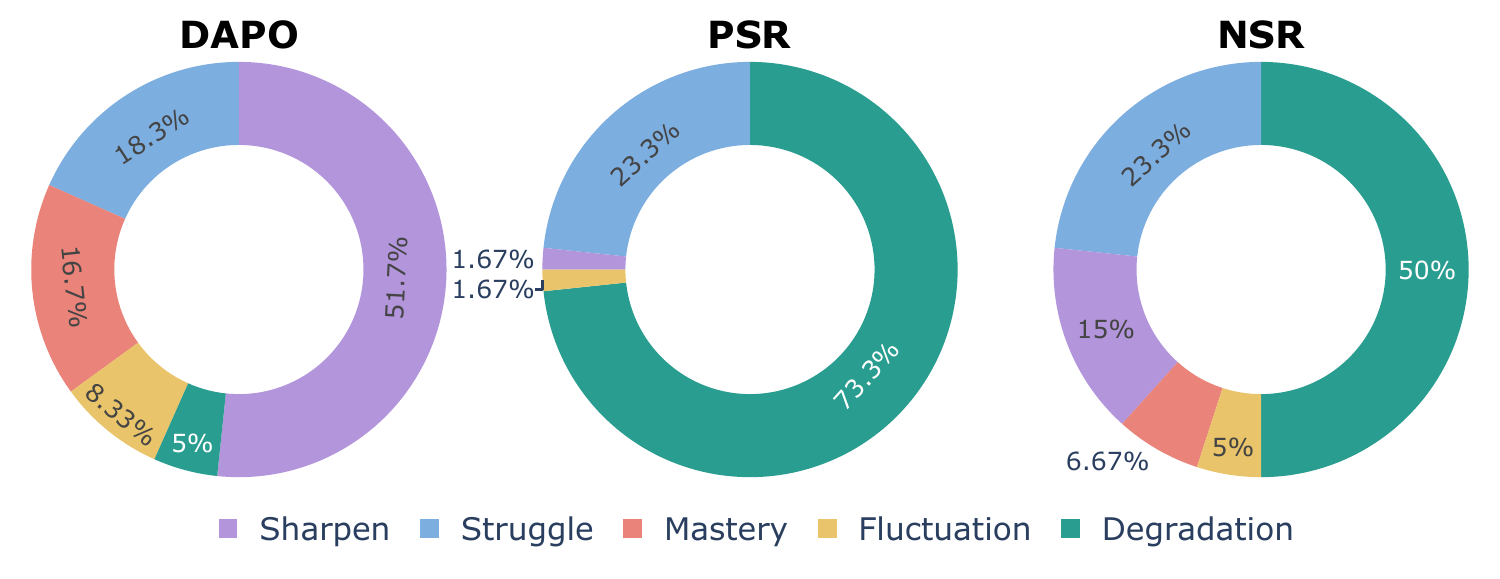}
    \caption{Training dynamics of sample accuracy changes in the validation set on DeepSeek-R1-Distilled-Qwen-7B.}
    \label{fig:training-behavior-ds7b}
\end{figure*}

We set $k$ to 10 in our analyses.
The results for Qwen2.5-7B-Math, Qwen3-8B-Base, and DeepSeek-R1-Distilled-Qwen-7B are shown in Figures~\ref{fig:training-behavior-qwen7b}, \ref{fig:training-behavior-qwen8b}, and \ref{fig:training-behavior-ds7b}, respectively.

These results reveal distinct patterns of validation accuracy changes across different base LLMs during training.
For Qwen2.5-7B-Math, which has been extensively exposed to reasoning data during pretraining, both PSR and NSR produce more sharpened samples than degraded ones.
This shows that either polarity alone can improve performance, and combining them yields a further complementary boost.
In contrast, Qwen3-8B-Base exhibits reward hacking when using only positive samples, causing degradation in most samples.
Negative sample reinforcement leads to garbled outputs, leaving the majority of samples in the struggle phase. 
Only when both polarities are combined does RLVR training become effective and improve accuracy.
For the distilled model DeepSeek-R1-Distilled-Qwen-7B, relying on a single sample polarity leads to significant degradation, and both polarities are needed together to achieve further performance improvement.

\section{Case Study}
\label{app:case_study}

\begin{figure}
    \centering
    \includegraphics[width=\linewidth]{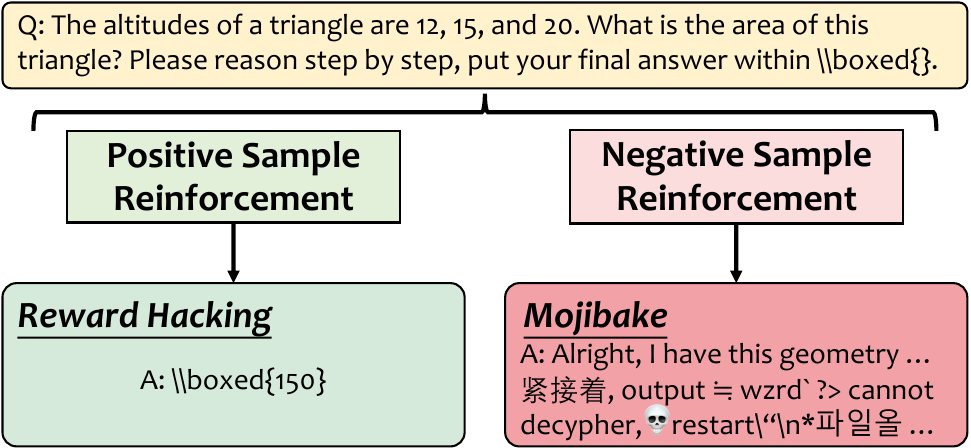}
    \caption{Case study of positive and negative sample reinforcement on Qwen3-8B-Base.}
    \label{fig:psr-nsr-case}
\end{figure}

In this part, we present a case study examining the distinct behaviors of positive sample reinforcement and negative sample reinforcement on Qwen3-8B-Base.
The results are shown in Figure~\ref{fig:psr-nsr-case}.
We observe that \textbf{continuous positive sample reinforcement} leads the model to strengthen its existing correct reasoning paths. 
Over time, this causes the model to progressively shorten its responses. 
Eventually, this results in reward hacking, where the model outputs only the final answer without step-by-step reasoning.
In contrast, \textbf{continuous negative sample reinforcement} encourages the model to repeatedly learn from its own mistakes, which drives it to explore alternative reasoning paths more broadly. 
As a result, the model ventures into low-probability regions of the output space, which sometimes leads to garbled or nonsensical outputs.
\section{Different Weighted Ratios of Token-level Advantage Shaping}
\label{app:diff-ratio}

In this part, we investigate how the proportion of tokens selected for token-level advantage shaping affects RLVR training dynamics.
Following prior work~\citep{20-80}, our main experiments adopt a shaping ratio of 20\%, meaning that advantages are reweighted for 20\% of the tokens in each response.
To assess the sensitivity of this choice, we conduct additional experiments with ratios of 5\%, 10\%, and 50\%, while keeping the scaling factor for low-probability positive tokens fixed at $0.2\times$.

\begin{figure*}[t]
    \centering
    \begin{subfigure}[b]{0.32\linewidth}
        \centering
        \includegraphics[width=\linewidth]{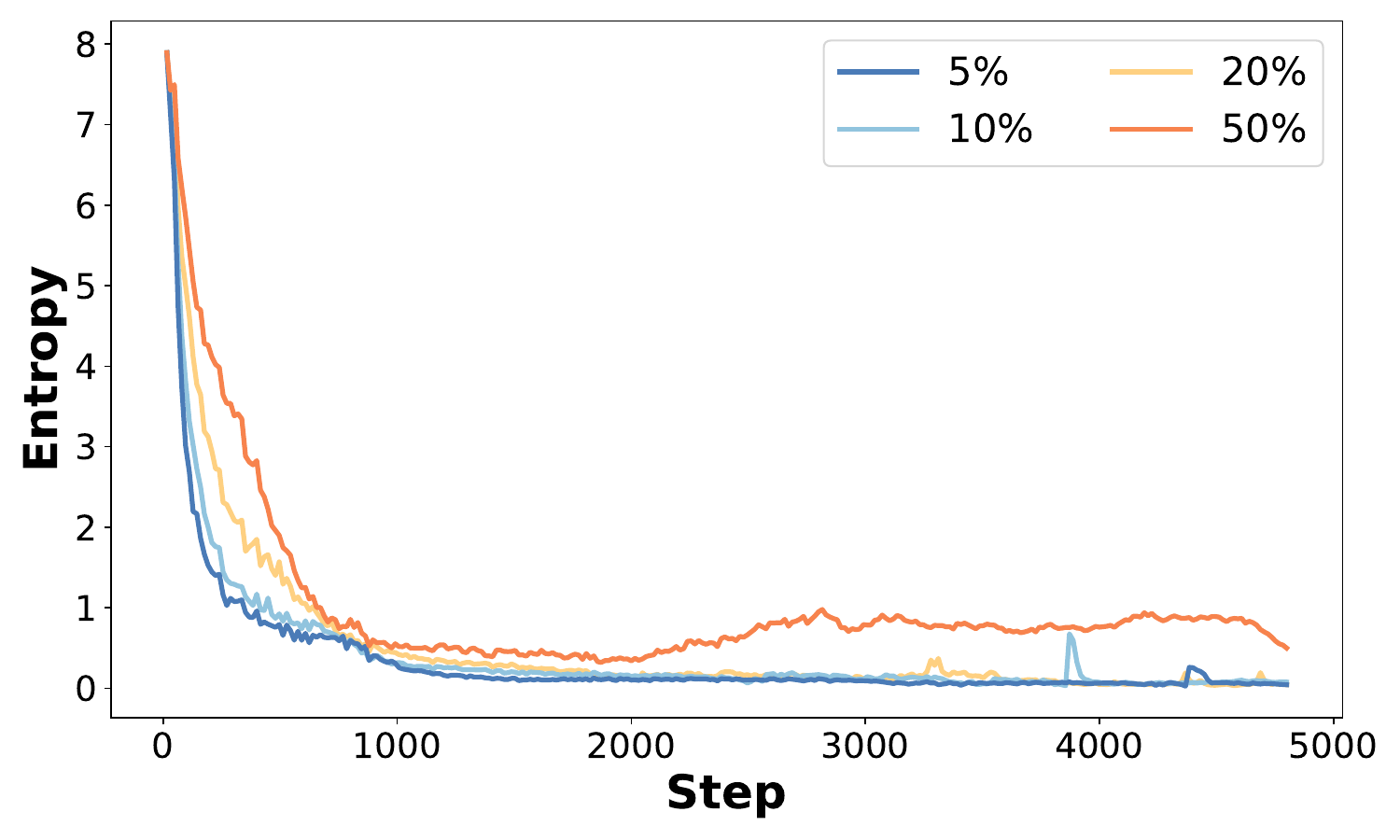}
        \caption{Entropy}
    \end{subfigure}
    \begin{subfigure}[b]{0.32\linewidth}
        \centering
        \includegraphics[width=\linewidth]{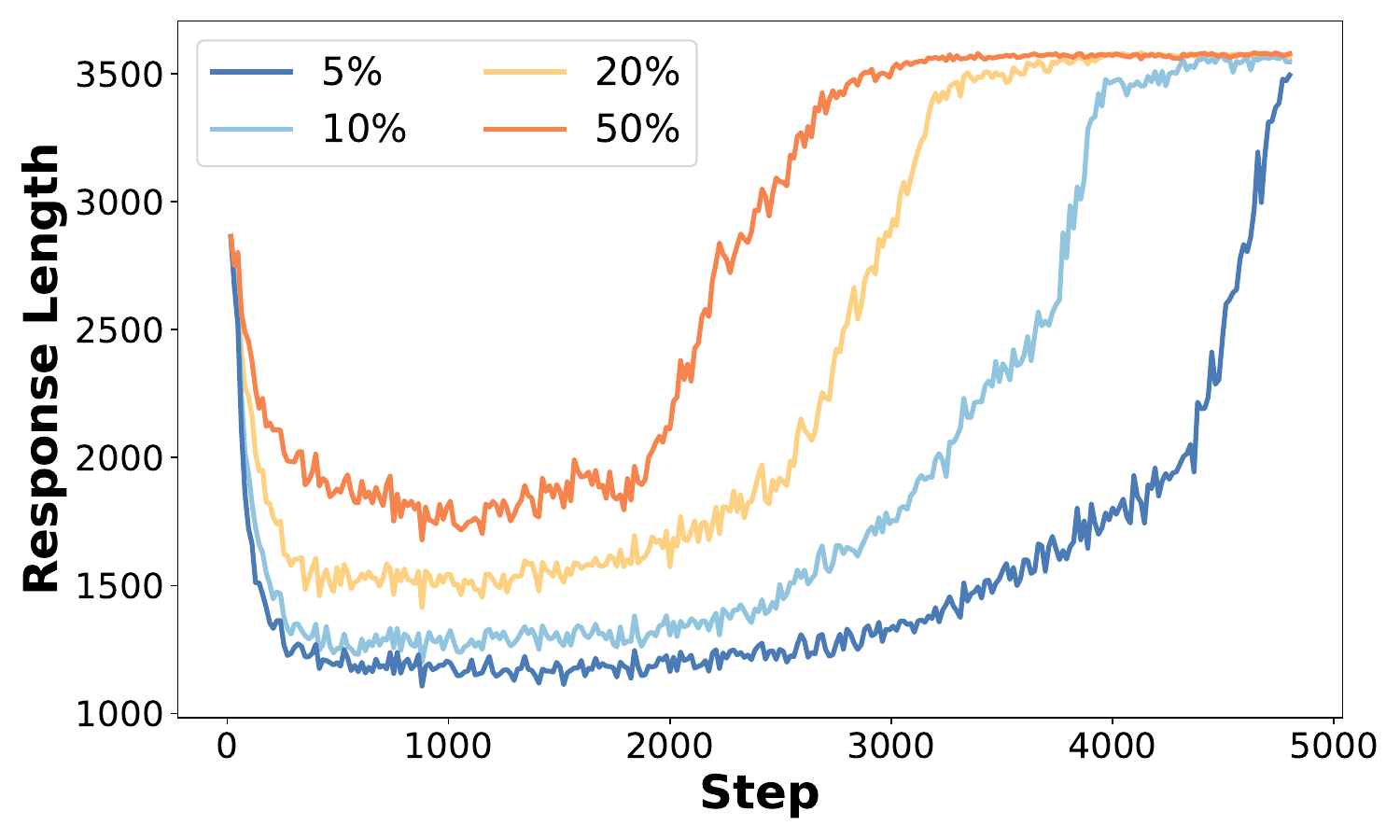}
        \caption{Response Length}
    \end{subfigure}
    \begin{subfigure}[b]{0.32\linewidth}
        \centering
        \includegraphics[width=\linewidth]{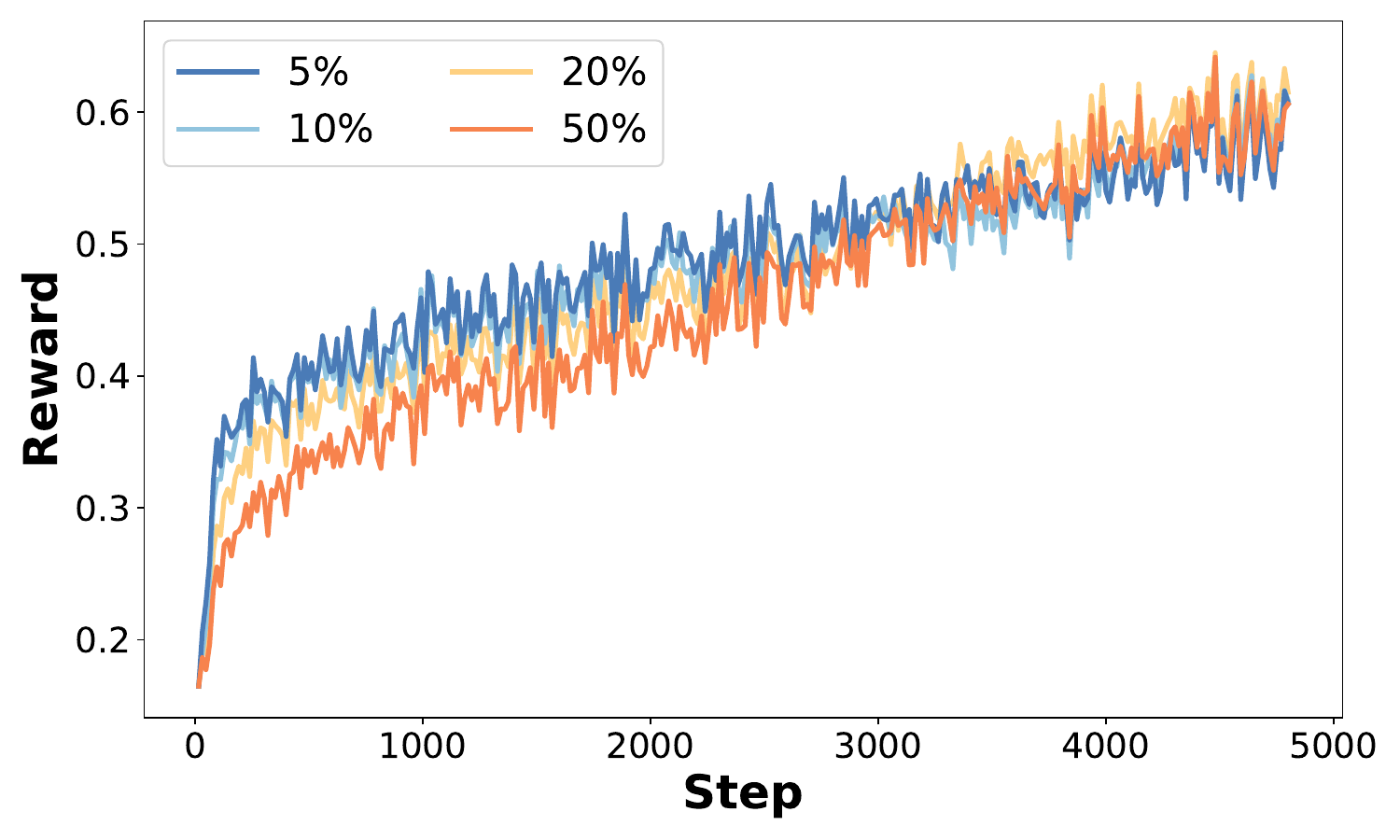}
        \caption{Reward}
    \end{subfigure}
    \caption{Impact of different ratios of advantage-shaped tokens when low-probability positive tokens are weighted at $0.2\times$ of their original values on Qwen2.5-7B-Math.}
    \label{fig:PL0.2_ratio}
\end{figure*}

Figure~\ref{fig:PL0.2_ratio} presents the results on Qwen2.5-7B-Math.
We observe that the shaping ratio mainly affects the magnitude and speed of training dynamics but does not change the overall learning trend.
Specifically, in this setting, smaller ratios lead to faster entropy reduction and a smoother transition in response length (from an initial decrease to a later increase). 
Similarly, reward improvements occur more quickly in early training but slow down later when using smaller ratios.
These results show that adjusting the proportion of advantage shaping tokens does not alter the fundamental training dynamics.
Instead, it acts as a factor that modulates the rate of policy updates.

\section{Negative Samples Amplify the Training-Inference Mismatch}
\label{app:prob_diff}

Training-inference mismatch is a critical issue in RLVR training, where token probabilities in training and inference engines exhibit significant discrepancies, potentially leading to training collapse.
In this section, we investigate which sample types contribute to this mismatch.

\begin{figure}[t]
    \centering
    \begin{subfigure}[b]{0.48\linewidth}
        \centering
        \includegraphics[width=\linewidth]{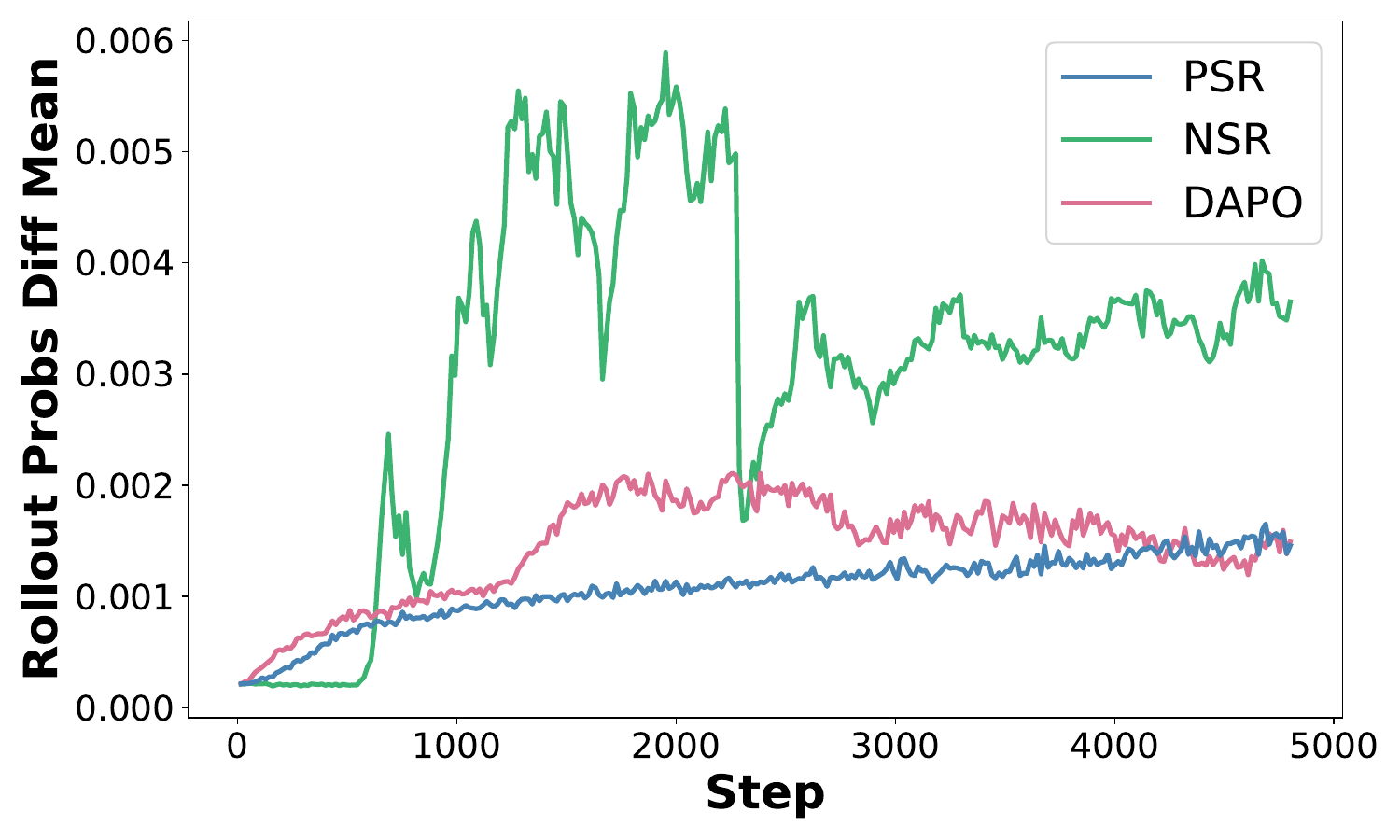}
        \caption{Positive and negative sample reinforcement}
        \label{subfig:prob_diff_nsr}
    \end{subfigure}
    \begin{subfigure}[b]{0.48\linewidth}
        \centering
        \includegraphics[width=\linewidth]{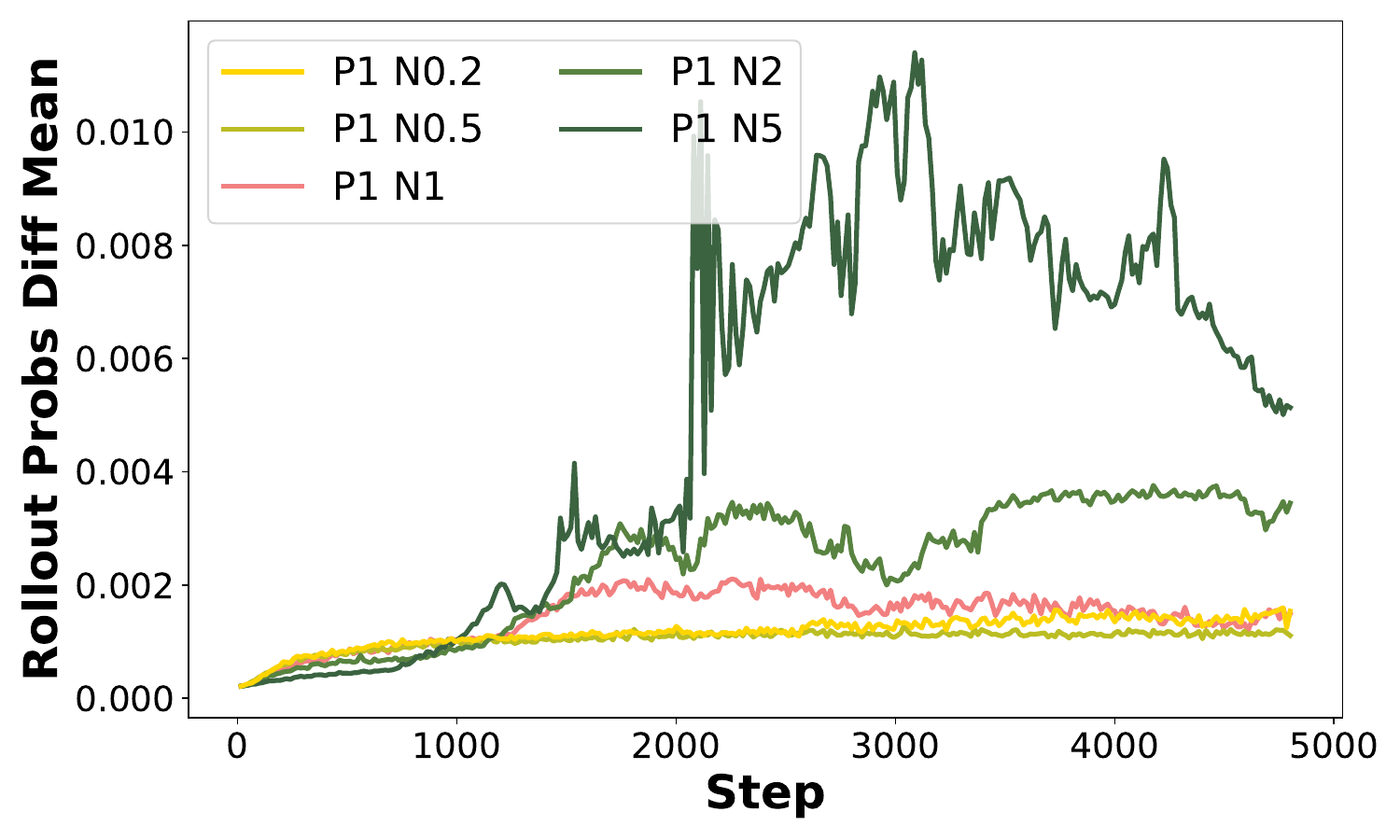}
        \caption{Polarity-level advantage shaping}
        \label{subfig:prob_diff_sample-level}
    \end{subfigure}
    \caption{Difference in token probabilities between training and inference engines.}
\label{fig:rollout_prob_diff}
\end{figure}

Figure~\ref{subfig:prob_diff_nsr} shows the difference in token probabilities between training and inference engines for three training paradigms (\ie PSR, NSR, and DAPO).
Our results reveal that utilizing negative samples widens the probability gap between training and inference engines.
Furthermore, we study the effect of polarity-level advantage shaping on negative samples.
As shown in Figure~\ref{subfig:prob_diff_sample-level}, assigning higher advantages to negative samples further increases the training–inference probability difference.
This phenomenon suggests that although weighting negative samples can raise model entropy and encourage exploration, consistently giving them higher weights enlarges the mismatch and may lead to instability.

Building on this observation, we adopt an adaptive advantage shaping strategy: we increase the weight of high-probability negative samples in early training to promote exploration, then gradually reduce it until it aligns with the weight of positive samples, thereby ensuring stable training.

\section{Detailed Analysis of \OURS}
\label{app:detailed_analysis}

In this section, we present a detailed analysis of our proposed method, \OURS.

\begin{table}
\centering
\Large
\caption{Ablation study on three math benchmarks.}
\resizebox{\linewidth}{!}{

\begin{tabular}{l | c c c | c}
\toprule
\cmidrule{2-4}
\textbf{Dataset} & \textbf{AIME 24} & \textbf{AIME 25} & \textbf{MATH500} & \textbf{Average} \\
\midrule
\OURS & \textbf{37.8} & \textbf{30.4} & \textbf{91.3} & \textbf{53.2} \\
\midrule
w/o PL adv & 36.5 & 27.8 & 88.1 & 50.8 \\
w/o NH adv & 35.8 & 29.1 & 87.4 & 50.8 \\
w/o both   & 34.2 & 26.1 & 84.5 & 48.3 \\
\midrule
w/o adaptive & 37.5 & 27.1 & 90.9 & 51.8 \\
\bottomrule
\end{tabular}
}
\label{tab:ablation}
\end{table}

\subsection{Ablation Study}

To evaluate the effectiveness of each component in our method, we conduct ablation studies on three math benchmarks using Qwen3-8B-Base.
As shown in Table~\ref{tab:ablation}, removing any component leads to performance degradation, confirming that all components are essential.
We observe that shaping advantages for both positive low-probability tokens and negative high-probability tokens improve performance, as both help maintain entropy and encourage exploration.
In addition, the adaptive strategy ensures stable RLVR training.
For instance, without this strategy, training instability due to training–inference mismatch limits further improvements on AIME25.

\subsection{Different Scales of LLMs and Training Datasets}

\begin{figure}[t]
    \centering
    \begin{subfigure}[b]{0.48\linewidth}
        \centering
        \includegraphics[width=\linewidth]{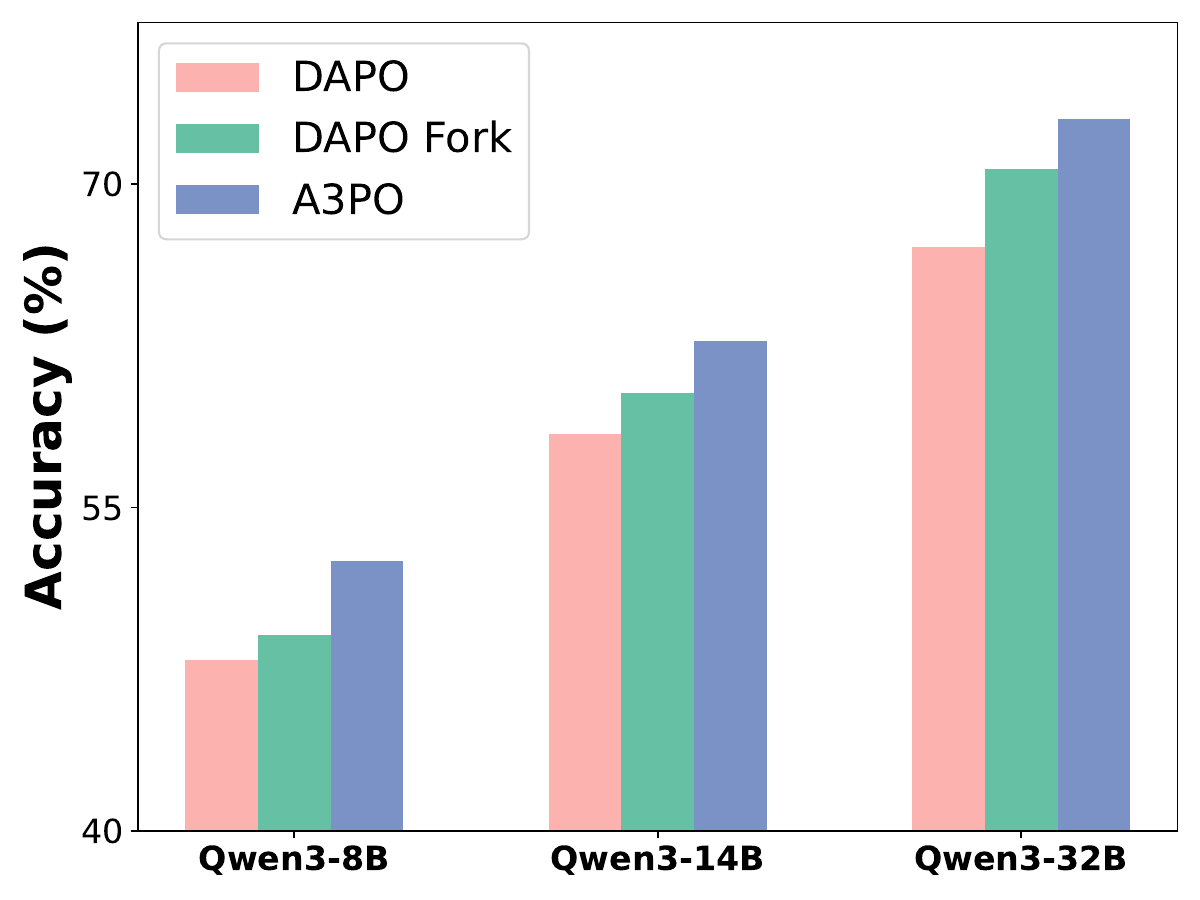}
        \caption{Different scales of LLMs}
        \label{subfig:llm}
    \end{subfigure}
    \begin{subfigure}[b]{0.48\linewidth}
        \centering
        \includegraphics[width=\linewidth]{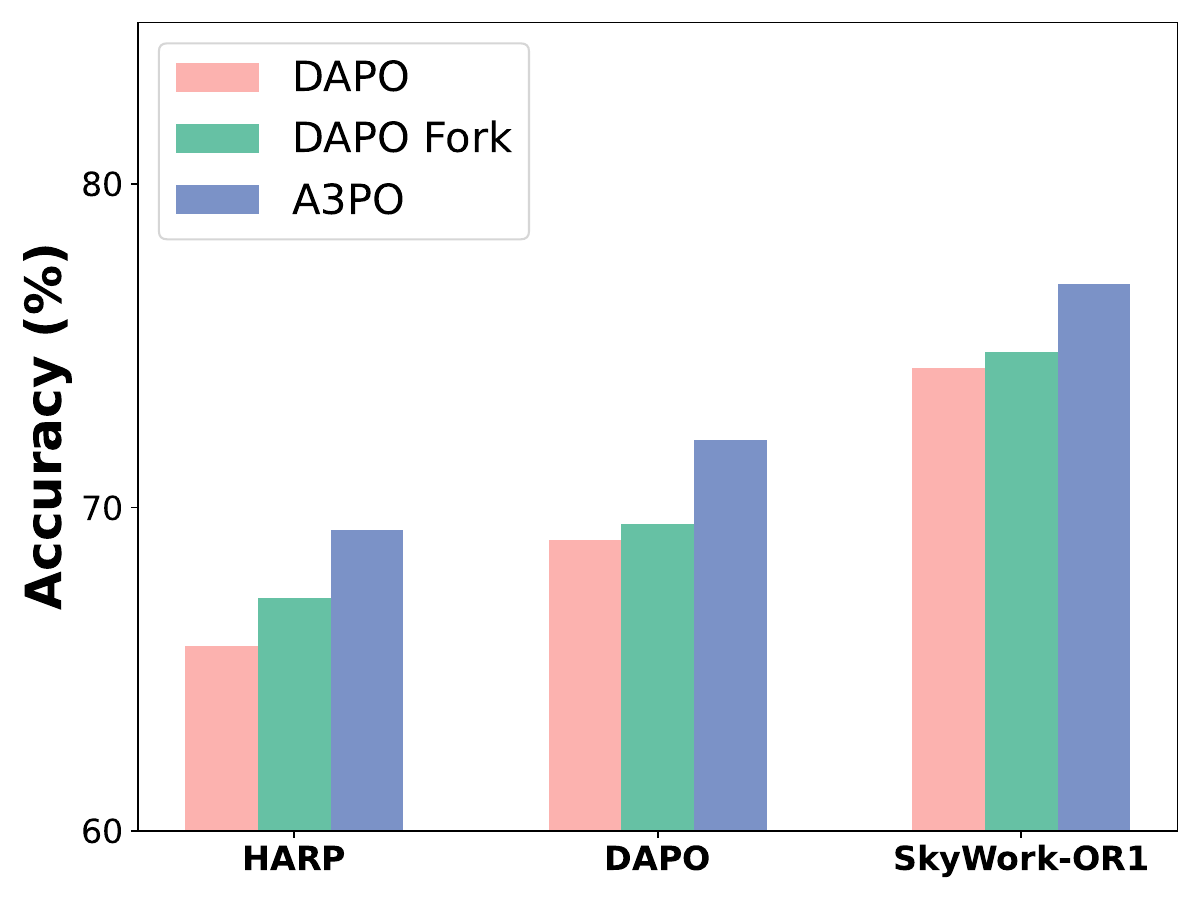}
        \caption{Different training datasets}
        \label{subfig:dataset}
    \end{subfigure}
    \caption{Different LLMs and training datasets}
\label{fig:llm-dataset}
\end{figure}

To assess the robustness and effectiveness of our proposed method, we conduct experiments using different scales of LLMs and datasets on Deepseek-R1-Distilled-Qwen-7B.
The results are shown in Figure~\ref{fig:llm-dataset}.
We find that our proposed method consistently achieves the best performance across both varying model scales and different training datasets, demonstrating its effectiveness and generalizability.

\subsection{Hyperparameter Analysis}

\begin{figure*}[t]
    \centering
    \begin{subfigure}[b]{0.32\linewidth}
        \centering
        \includegraphics[width=\linewidth]{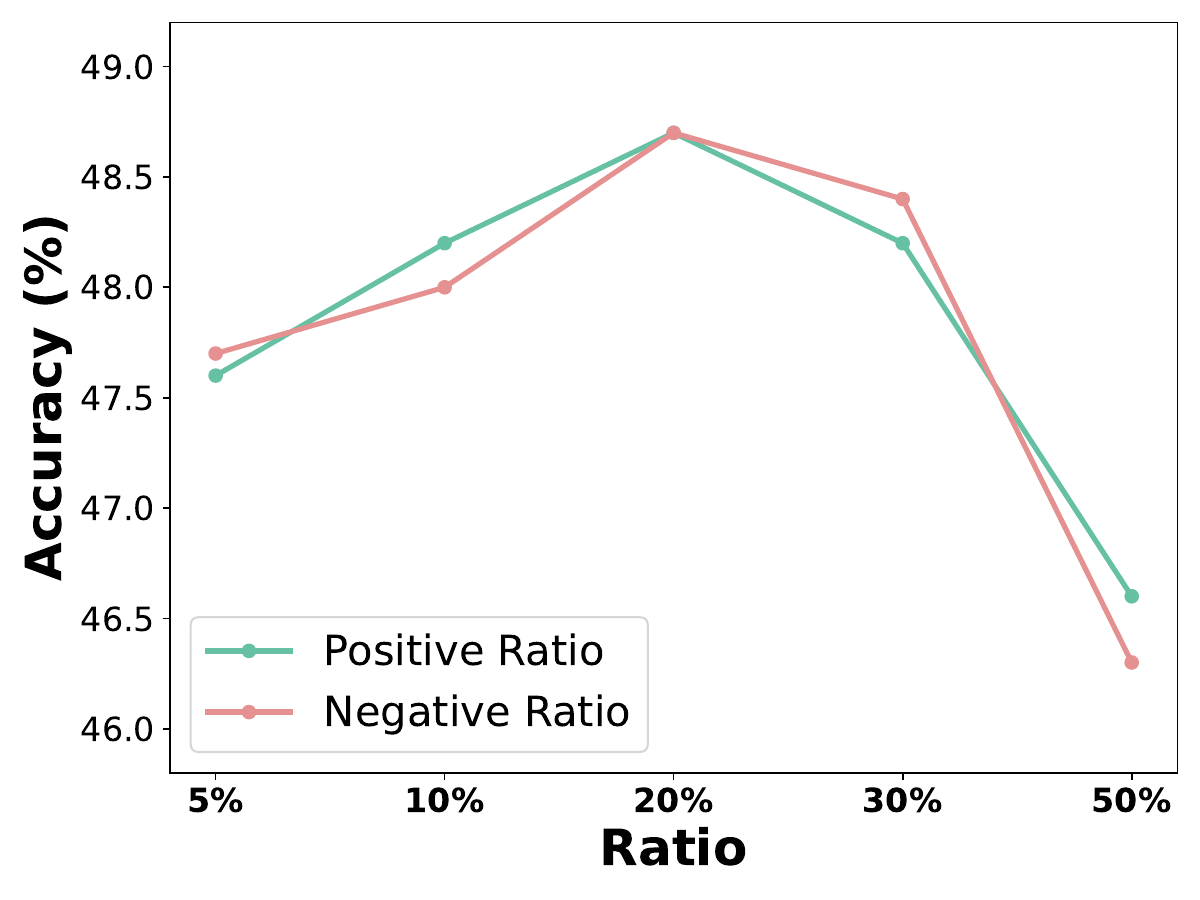}
        \label{subfig:hyper-ratio}
        \caption{Token-shaped ratios}
    \end{subfigure}
    \begin{subfigure}[b]{0.32\linewidth}
        \centering
        \includegraphics[width=\linewidth]{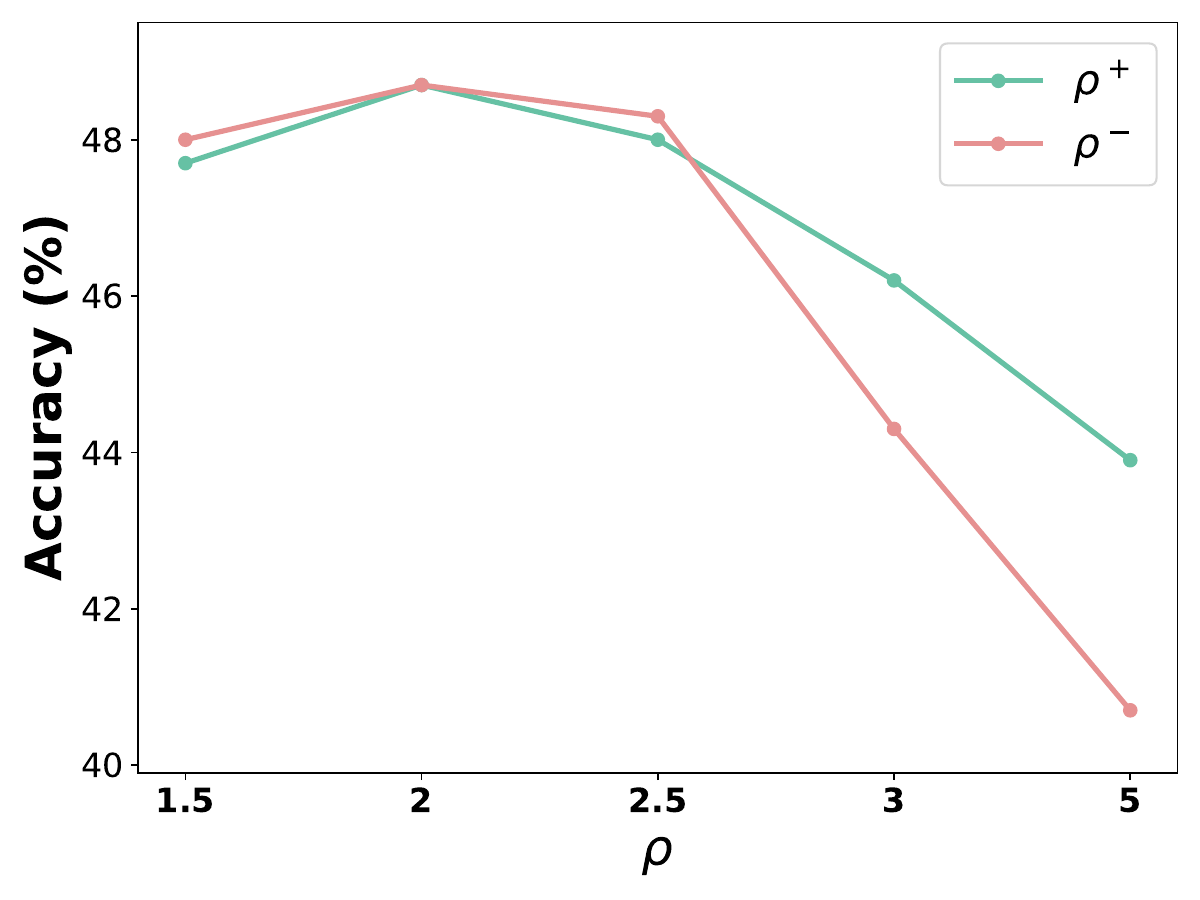}
        \label{subfig:hyper-rho}
        \caption{Decay coefficients}
    \end{subfigure}
    \begin{subfigure}[b]{0.32\linewidth}
        \centering
        \includegraphics[width=\linewidth]{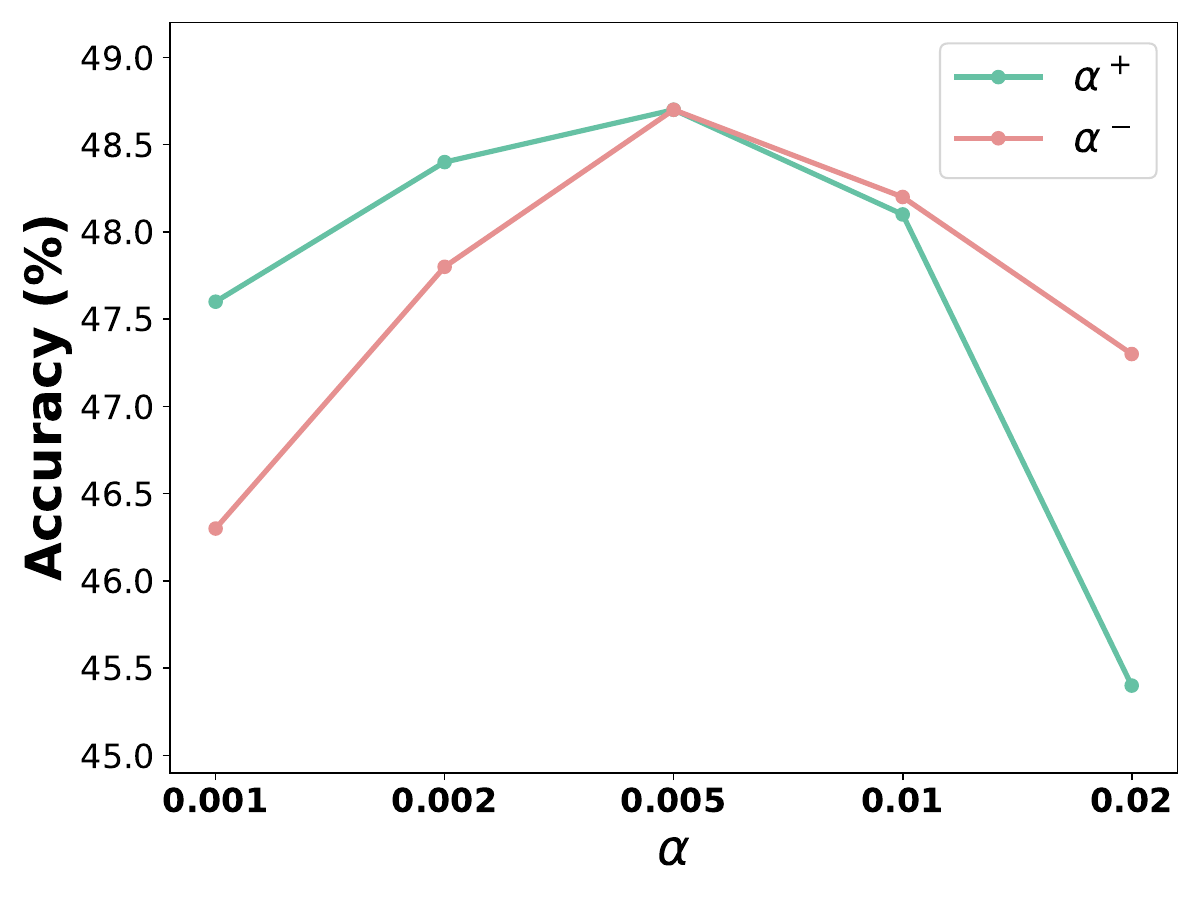}
        \label{subfig:hyper-alpha}
        \caption{Initial scaling factors}
    \end{subfigure}
    \caption{Hyperparameter Anlaysis.}
    \label{fig:hyperpara}
\end{figure*}

In this part, we analyze three key hyperparameters of \OURS: the token-shaped ratios, the initial scaling factors $\rho$, and the decay coefficients $\alpha$. 
The results are shown in Figure~\ref{fig:hyperpara}.
Our experimental results indicate that the model achieves optimal performance with the token-shaped ratios of 20\%.
If the ratio is too low, the model does not explore the token space sufficiently. 
If it is too high, performance drops because many less relevant tokens receive advantage shaping.
Next, we examine the initial scaling factors $\rho$. 
Setting appropriate values is important for stable training.
If the hyperparameter is too high, the training–inference mismatch increases. 
If it is too low, the model does not explore the solution space effectively.
Therefore, we set $\rho$ to 2 in our main experiments.
Finally, we find that the decay coefficients $\alpha$ of 0.005 yield the best performance. 
If $\alpha$ is too small, the gap between training and inference grows. 
If $\alpha$ is too large, exploration becomes insufficient and the model may converge to suboptimal solutions.
\section{Detailed Results}
\label{app:detailed-results}

\begin{figure*}[t]
    \centering
    \begin{subfigure}[b]{0.32\linewidth}
        \centering
        \includegraphics[width=\linewidth]{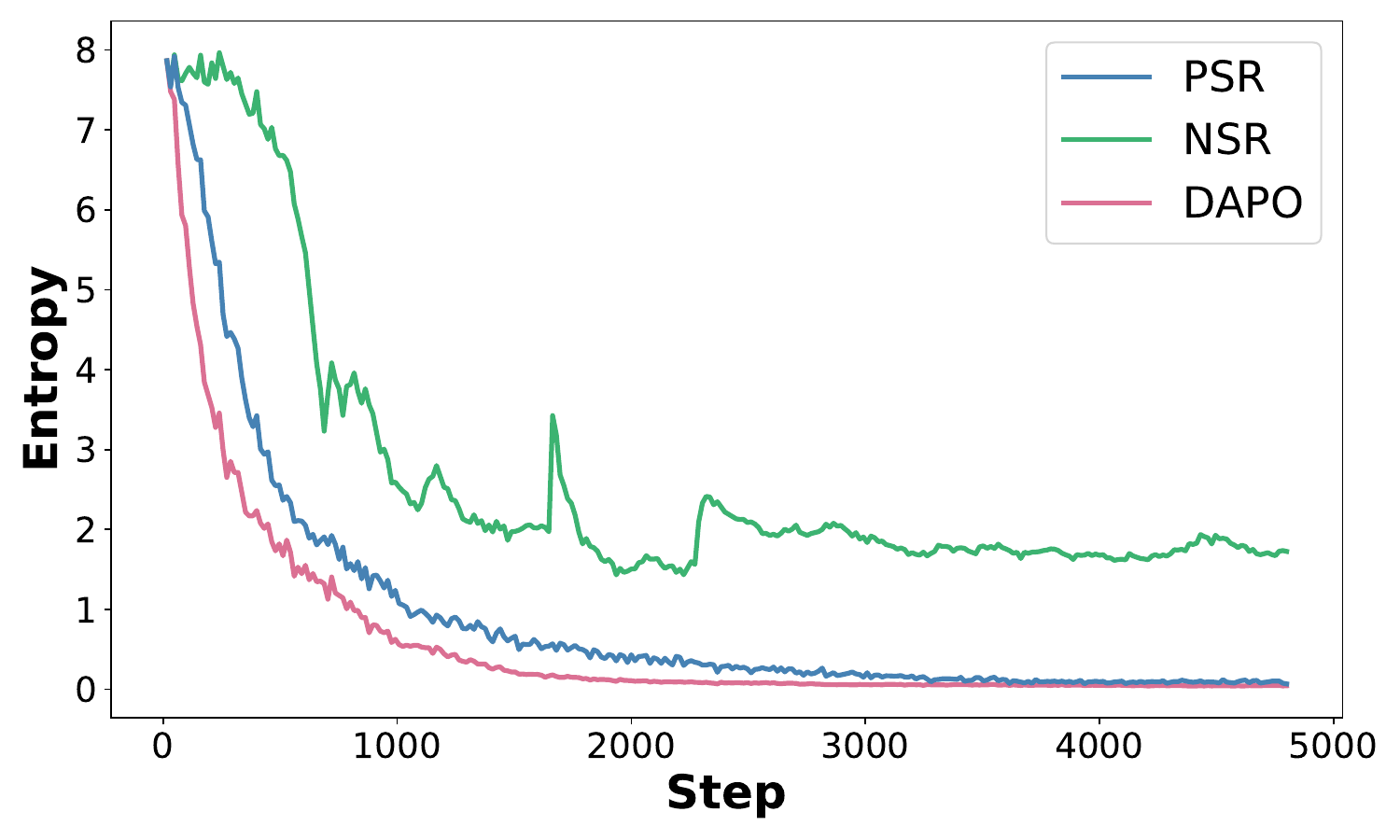}
        \caption{Entropy}
    \end{subfigure}
    \begin{subfigure}[b]{0.32\linewidth}
        \centering
        \includegraphics[width=\linewidth]{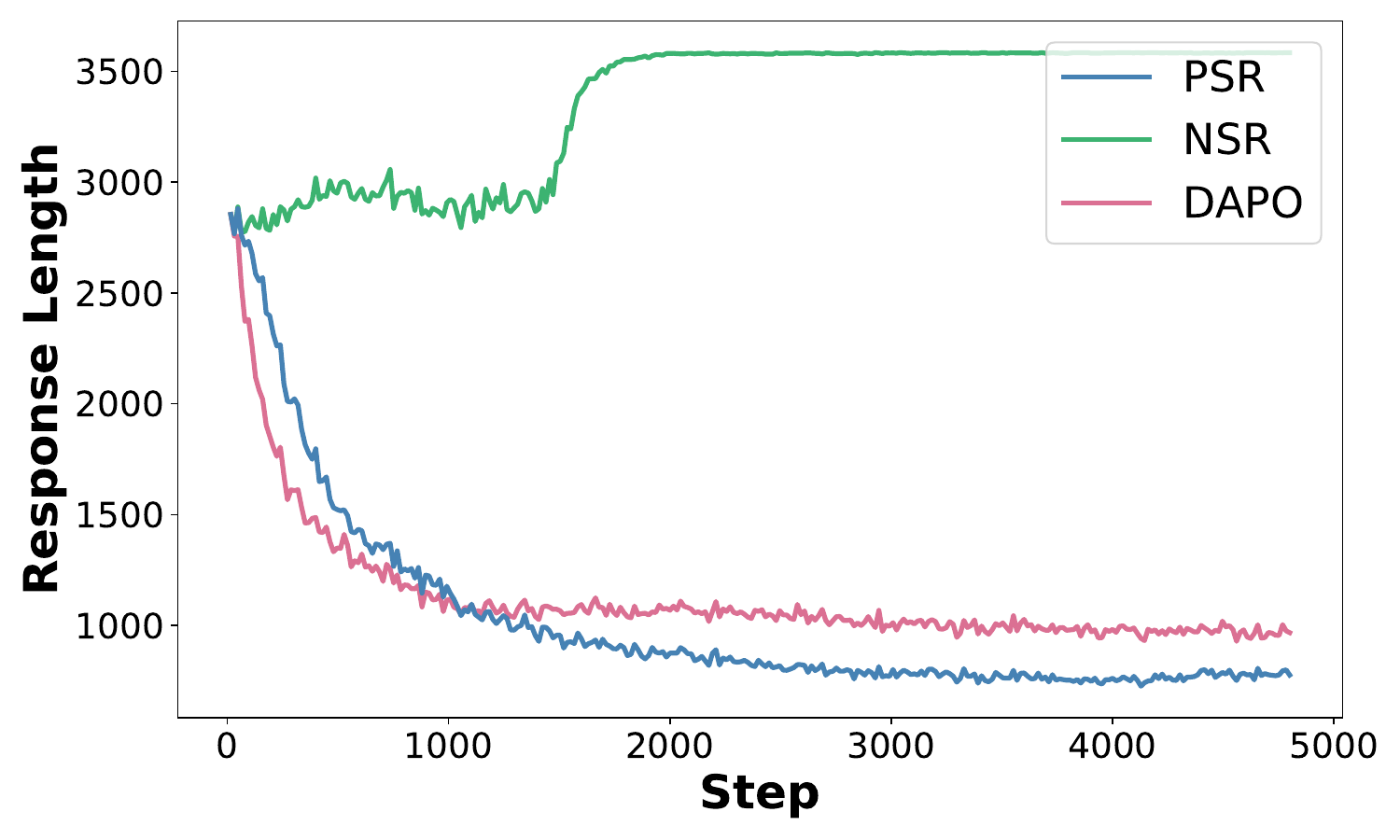}
        \caption{Length}
    \end{subfigure}
    \begin{subfigure}[b]{0.32\linewidth}
        \centering
        \includegraphics[width=\linewidth]{figures/psr_nsr/qwen2.5-7b-math/Reward.pdf}
        \caption{Reward}
    \end{subfigure}
    \begin{subfigure}[b]{0.32\linewidth}
        \centering
        \includegraphics[width=\linewidth]{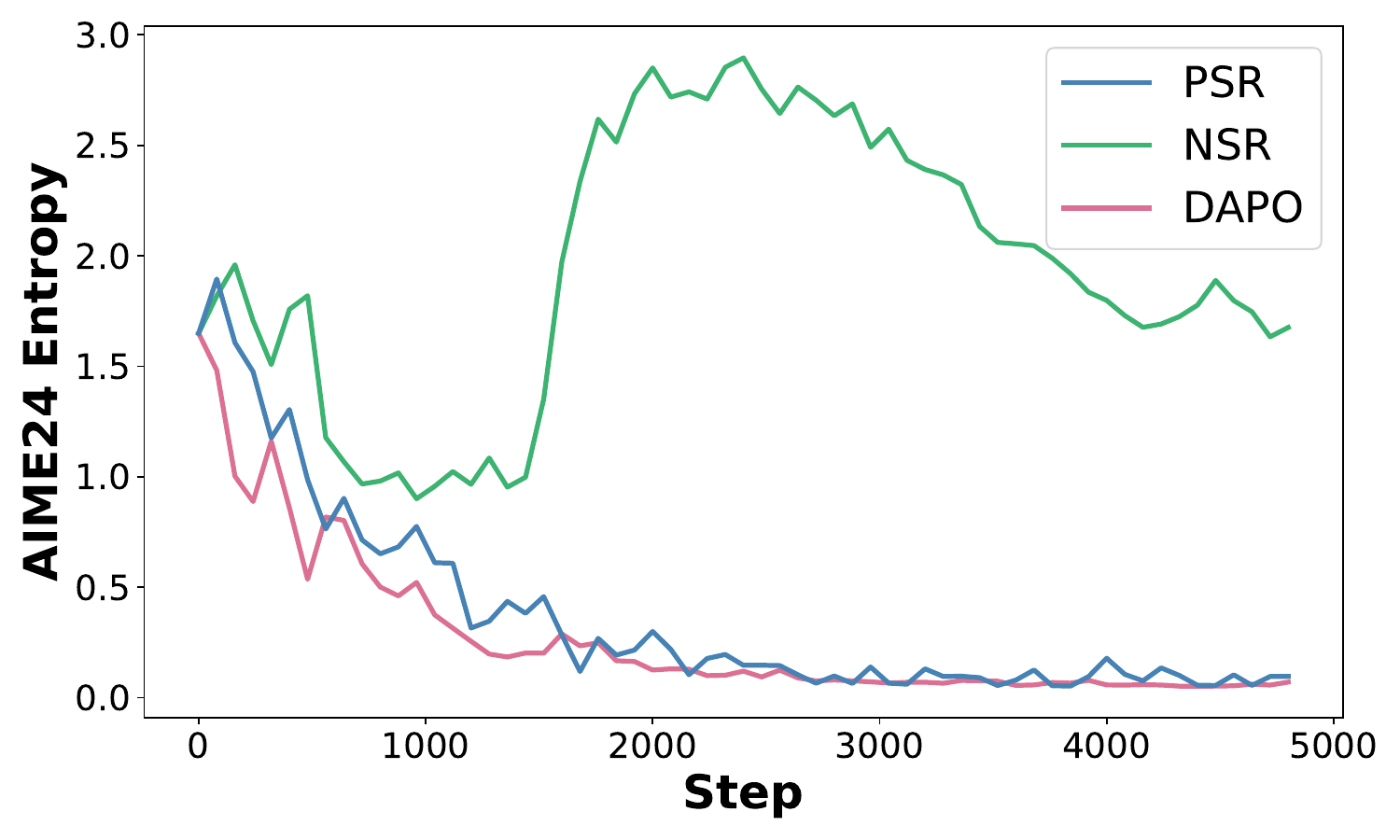}
        \caption{AIME24 Entropy}
    \end{subfigure}
    \begin{subfigure}[b]{0.32\linewidth}
        \centering
        \includegraphics[width=\linewidth]{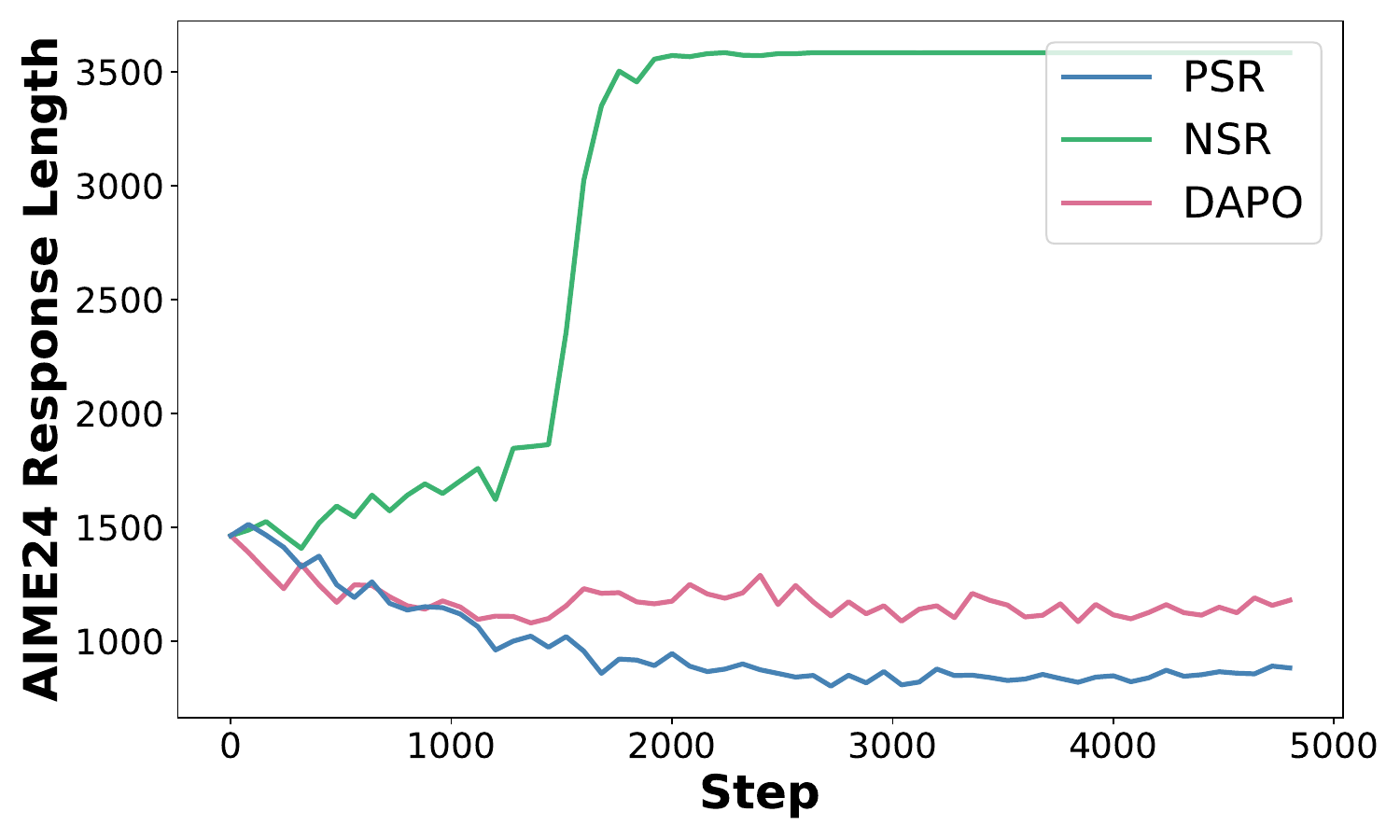}
        \caption{AIME24 Length}
    \end{subfigure}
    \begin{subfigure}[b]{0.32\linewidth}
        \centering
        \includegraphics[width=\linewidth]{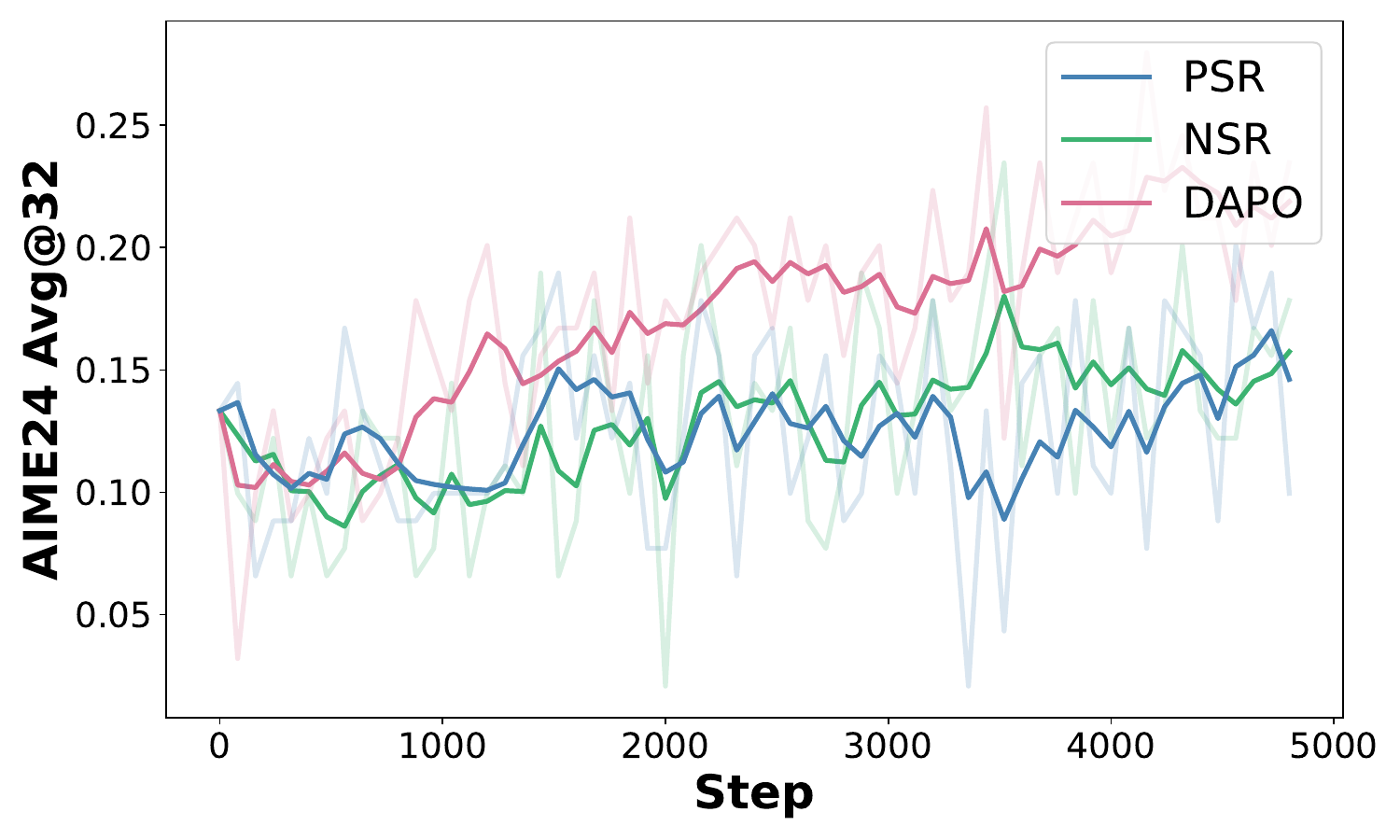}
        \caption{AIME24 Avg@32}
    \end{subfigure}
    \begin{subfigure}[b]{0.32\linewidth}
        \centering
        \includegraphics[width=\linewidth]{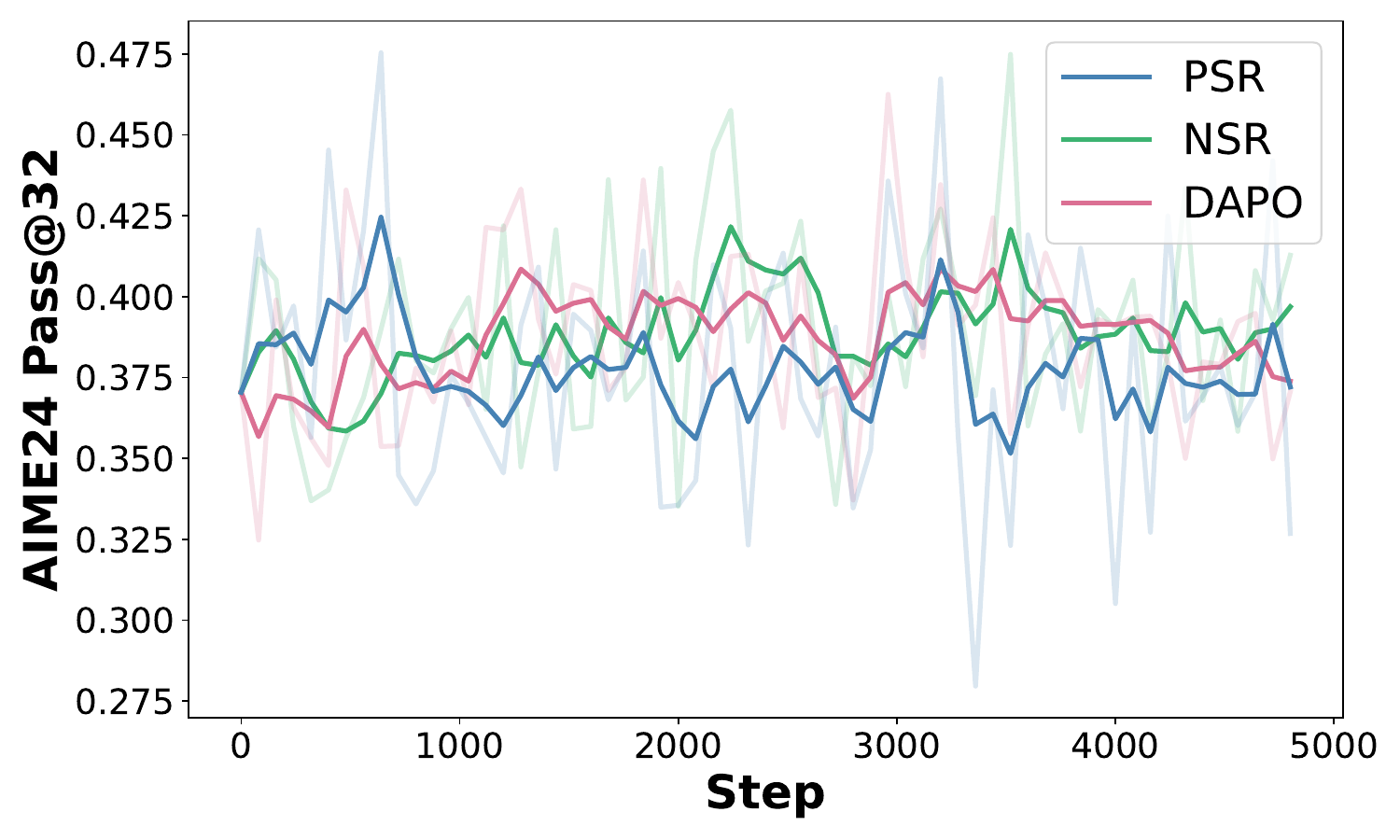}
        \caption{AIME24 Pass@32}
    \end{subfigure}
    \begin{subfigure}[b]{0.32\linewidth}
        \centering
        \includegraphics[width=\linewidth]{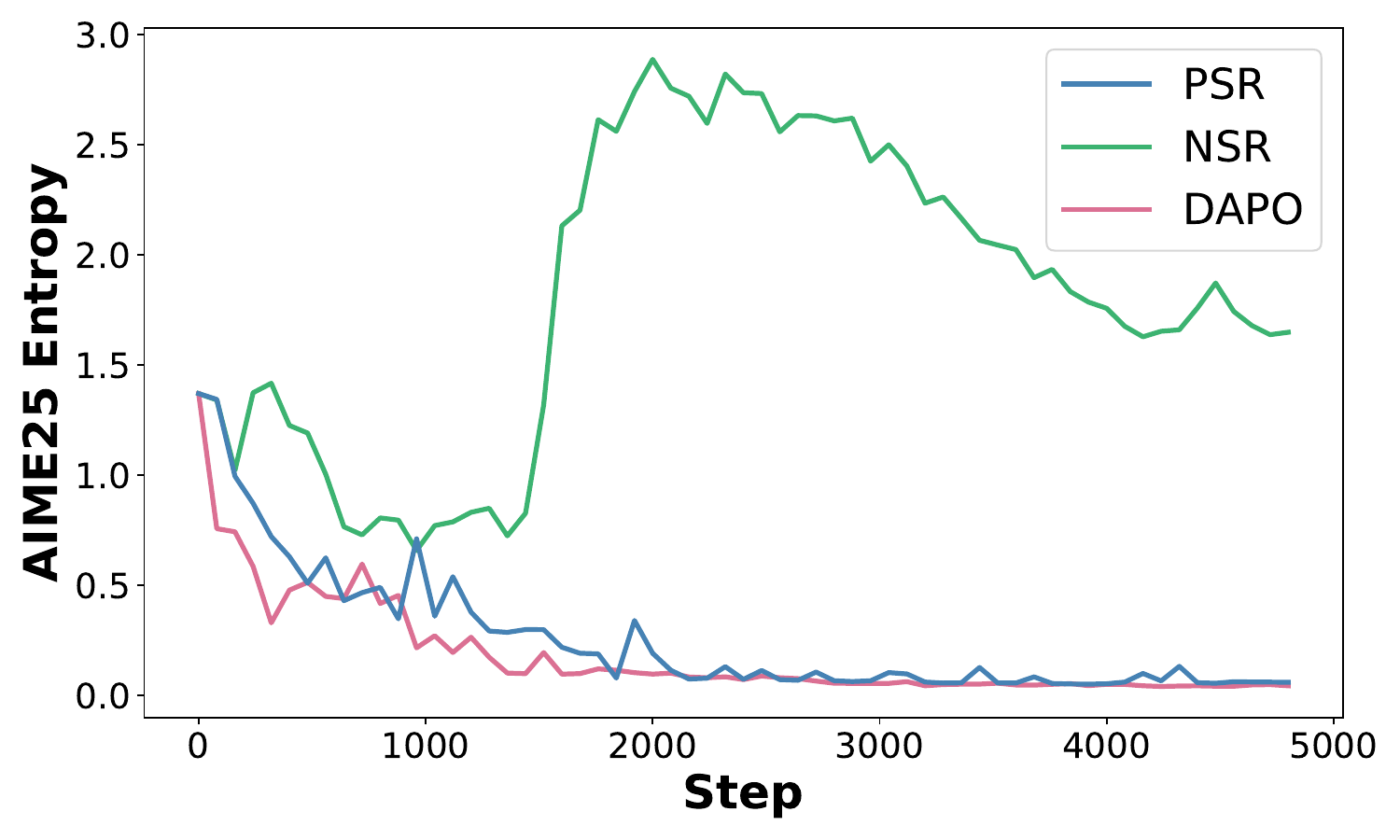}
        \caption{AIME25 Entropy}
    \end{subfigure}
    \begin{subfigure}[b]{0.32\linewidth}
        \centering
        \includegraphics[width=\linewidth]{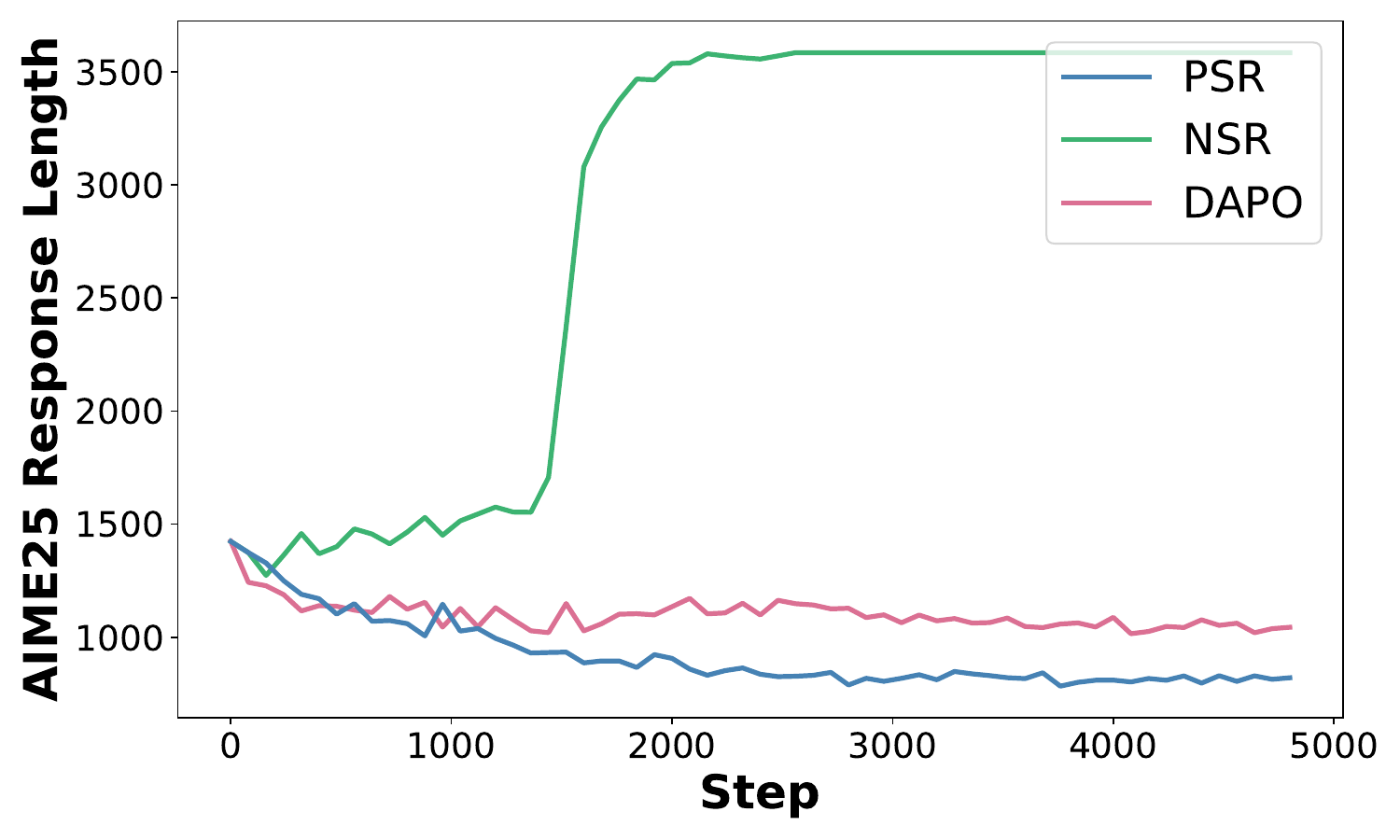}
        \caption{AIME25 Length}
    \end{subfigure}
    \begin{subfigure}[b]{0.32\linewidth}
        \centering
        \includegraphics[width=\linewidth]{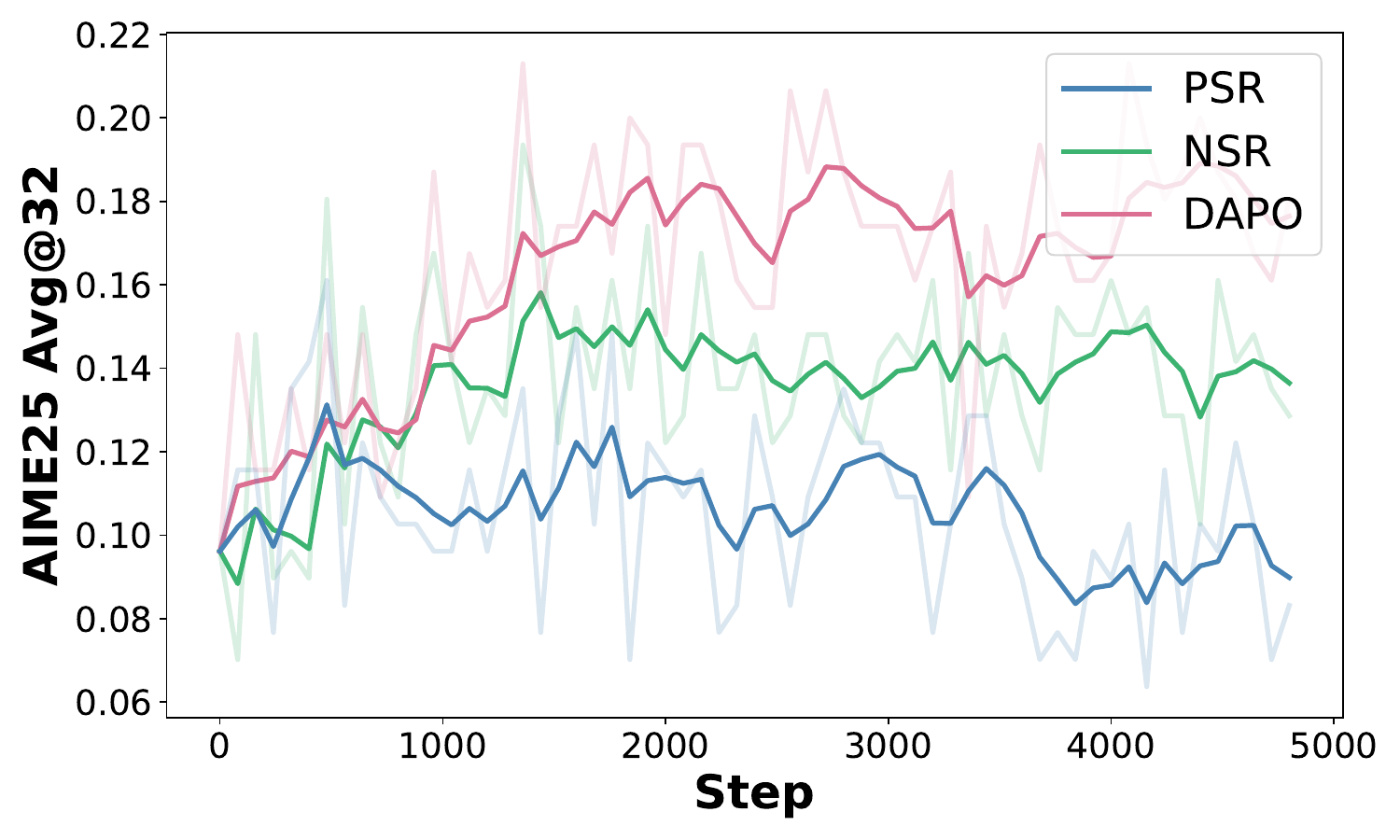}
        \caption{AIME25 Avg@32}
    \end{subfigure}
    \begin{subfigure}[b]{0.32\linewidth}
        \centering
        \includegraphics[width=\linewidth]{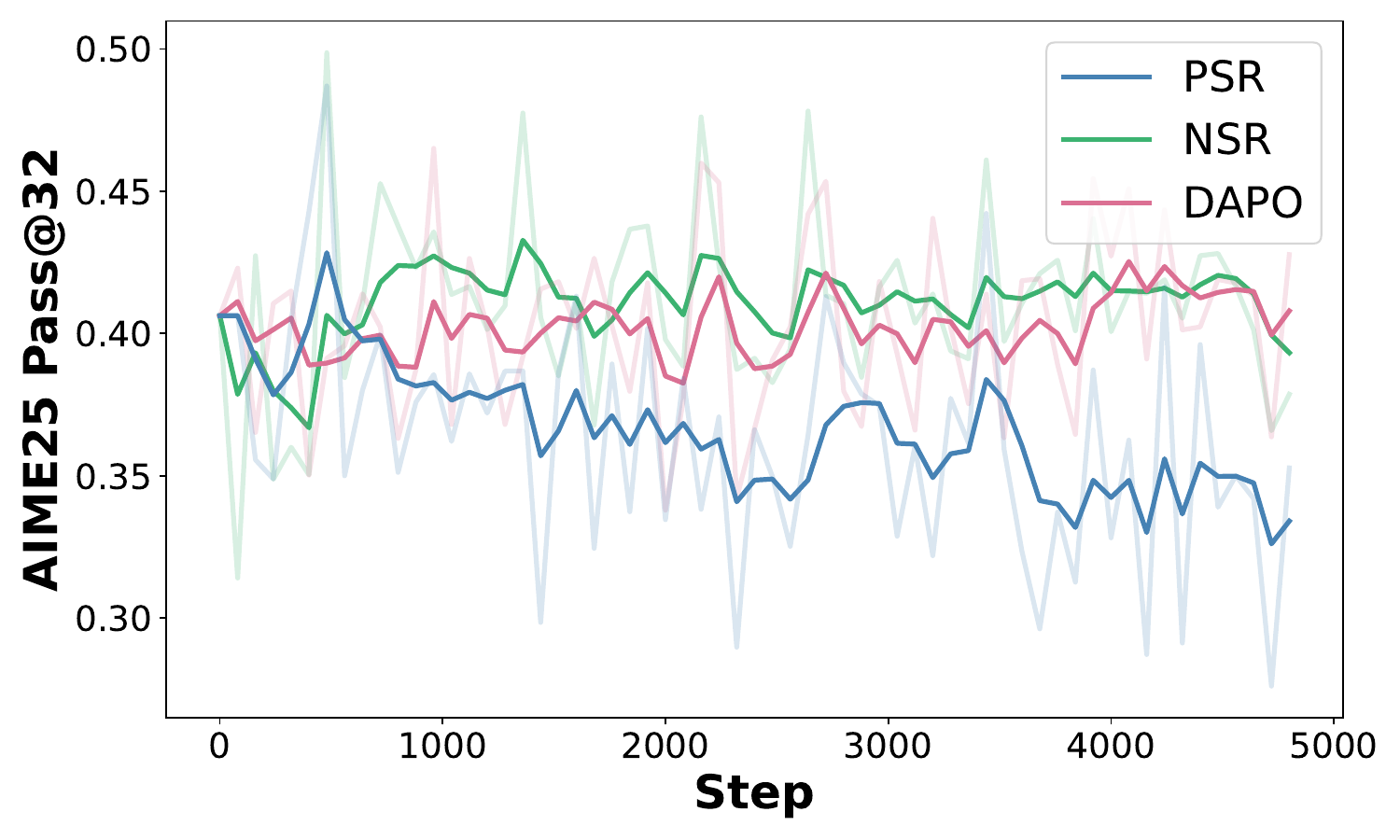}
        \caption{AIME25 Pass@32}
    \end{subfigure}
    \caption{RLVR training dynamics on Qwen2.5-7B-Math.}
\label{fig:qwen2.5-7b-math-training_dynamic}
\end{figure*}

\begin{figure*}[t]
    \centering
    \begin{subfigure}[b]{0.32\linewidth}
        \centering
        \includegraphics[width=\linewidth]{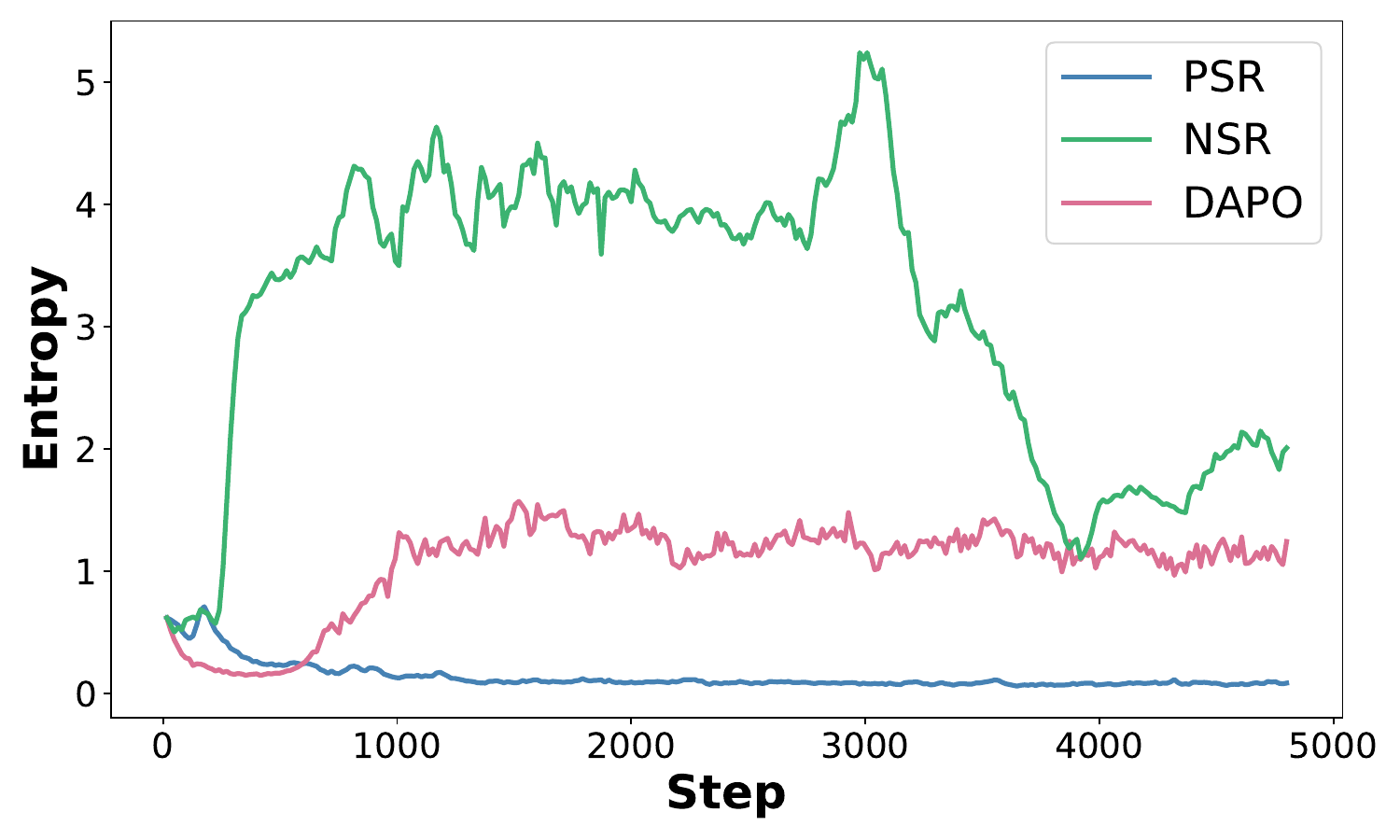}
        \caption{Entropy}
    \end{subfigure}
    \begin{subfigure}[b]{0.32\linewidth}
        \centering
        \includegraphics[width=\linewidth]{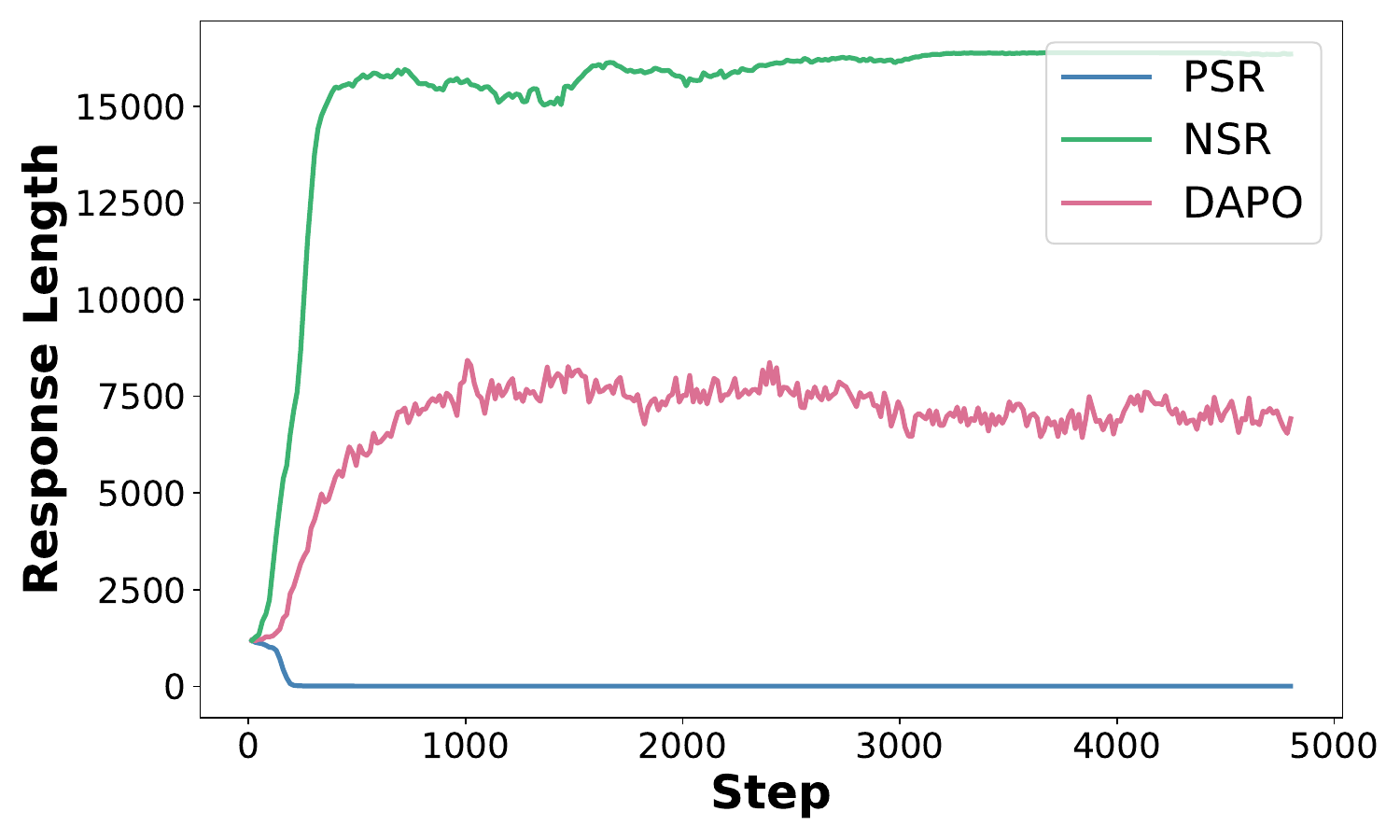}
        \caption{Length}
    \end{subfigure}
    \begin{subfigure}[b]{0.32\linewidth}
        \centering
        \includegraphics[width=\linewidth]{figures/psr_nsr/qwen3-8b-base/Reward.pdf}
        \caption{Reward}
    \end{subfigure}
    \begin{subfigure}[b]{0.32\linewidth}
        \centering
        \includegraphics[width=\linewidth]{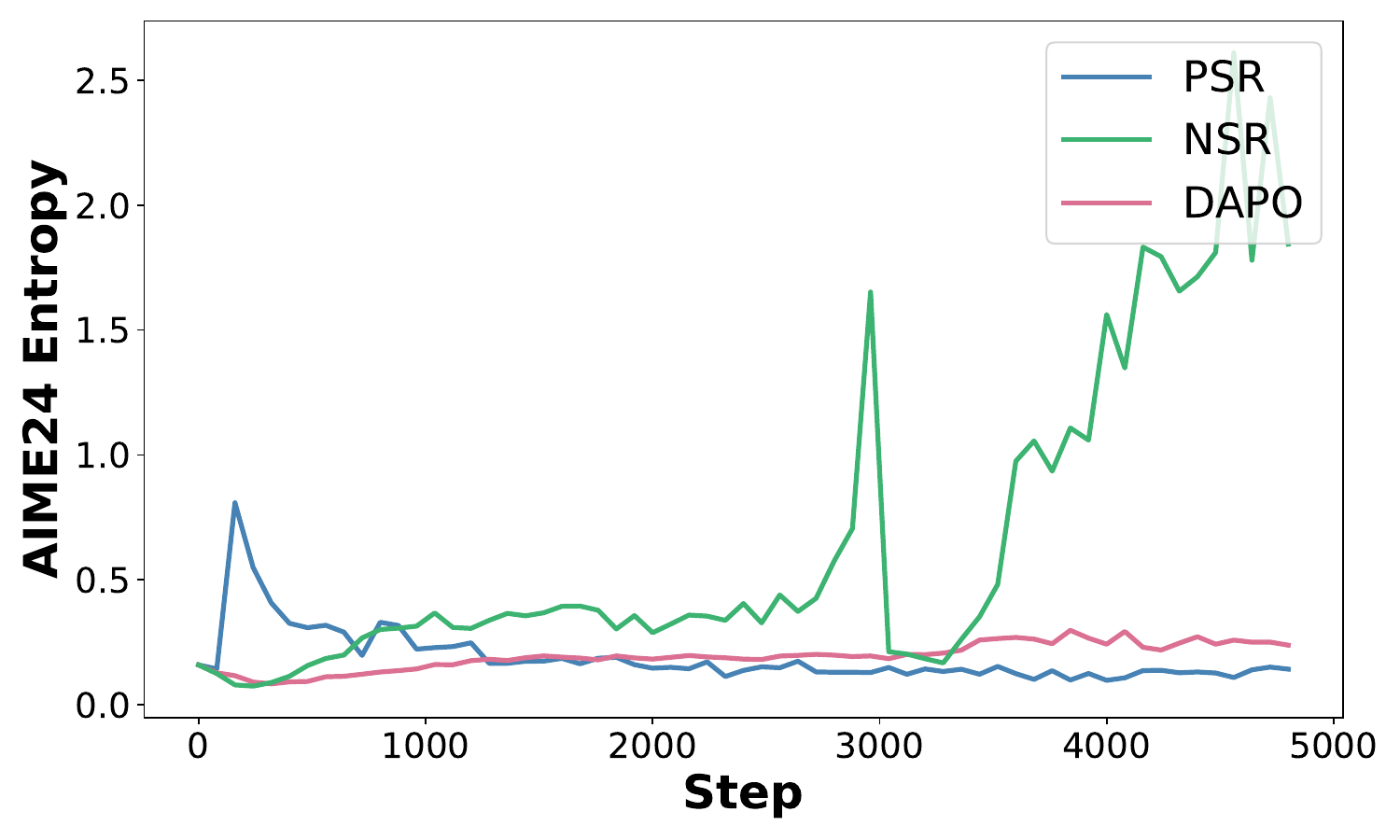}
        \caption{AIME24 Entropy}
    \end{subfigure}
    \begin{subfigure}[b]{0.32\linewidth}
        \centering
        \includegraphics[width=\linewidth]{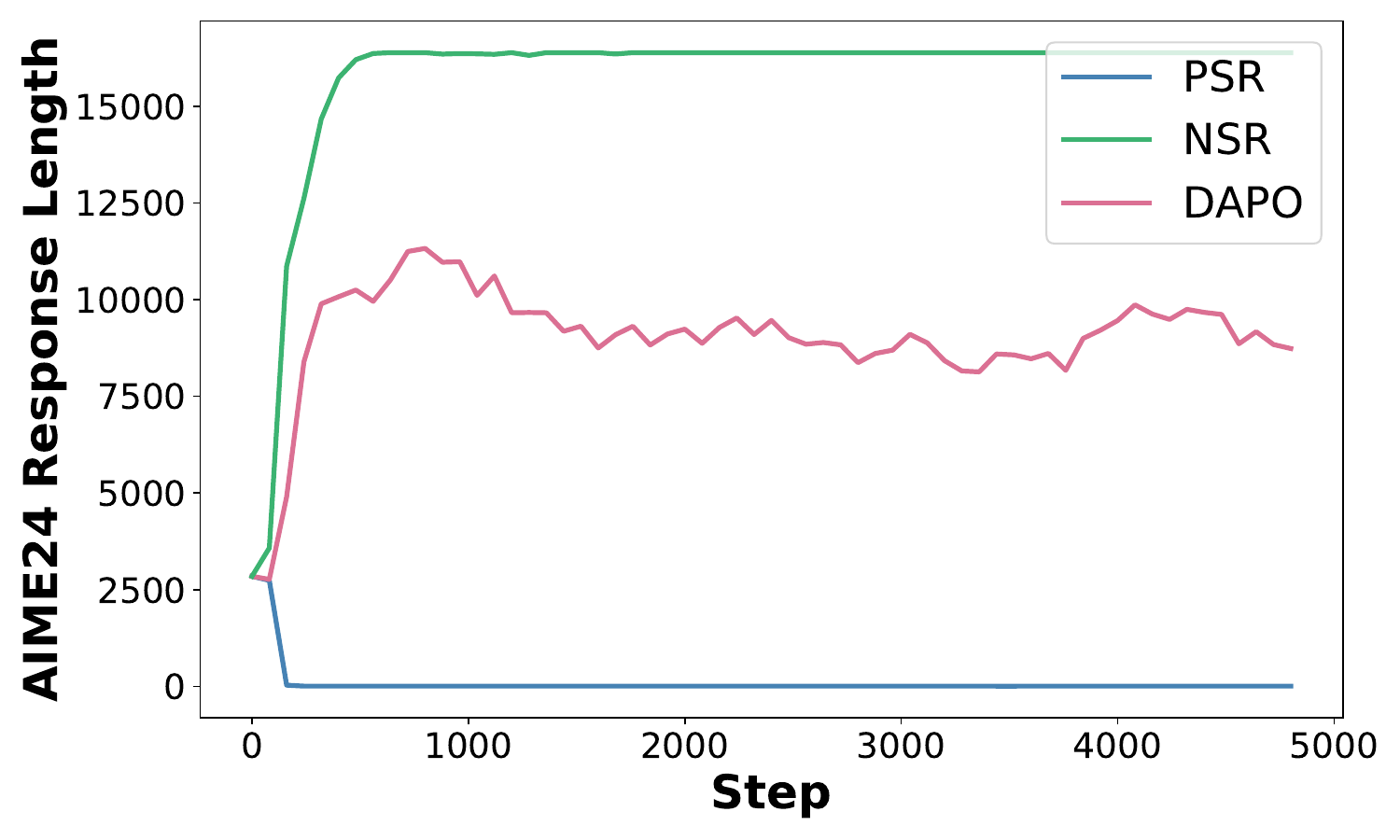}
        \caption{AIME24 Length}
    \end{subfigure}
    \begin{subfigure}[b]{0.32\linewidth}
        \centering
        \includegraphics[width=\linewidth]{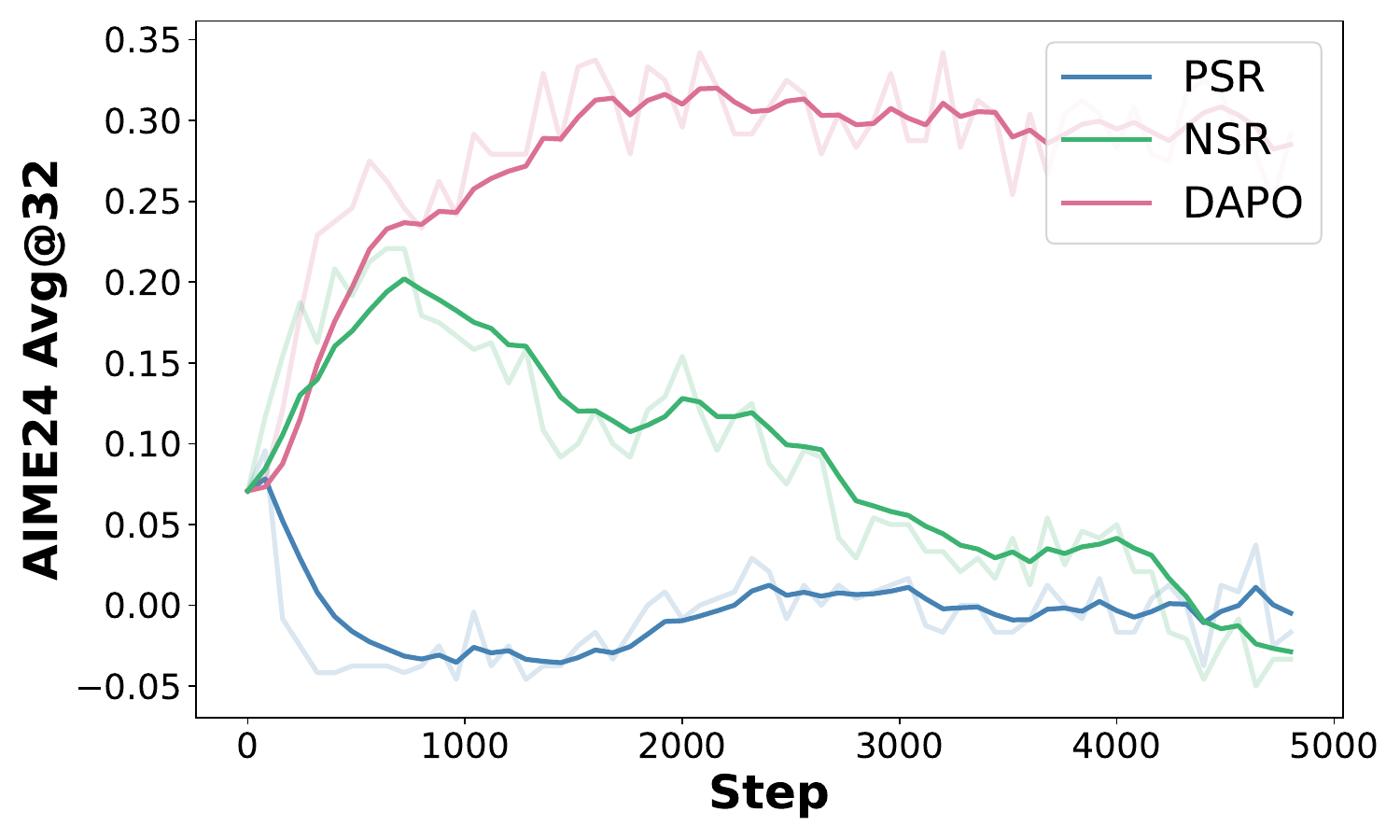}
        \caption{AIME24 Avg@32}
    \end{subfigure}
    \begin{subfigure}[b]{0.32\linewidth}
        \centering
        \includegraphics[width=\linewidth]{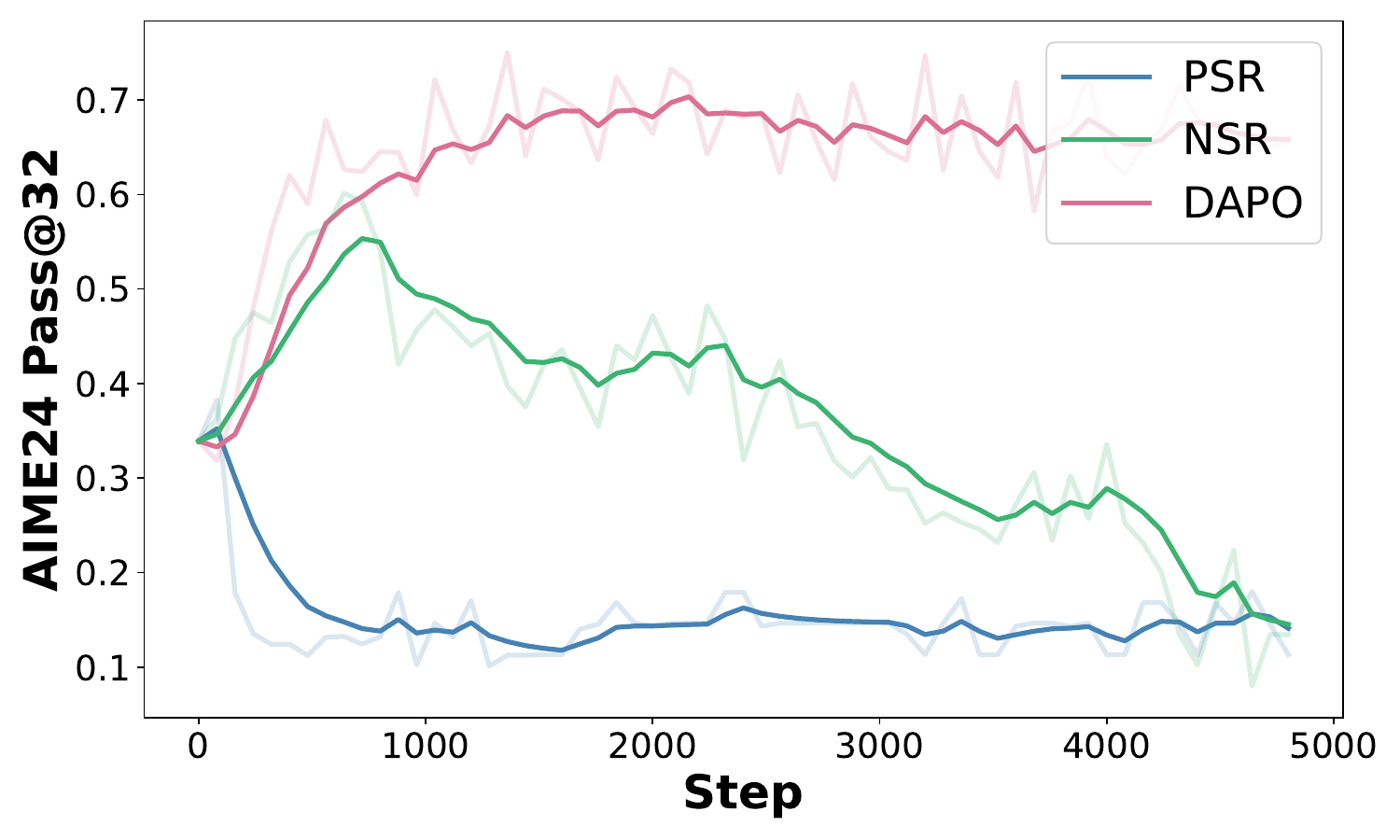}
        \caption{AIME24 Pass@32}
    \end{subfigure}
    \begin{subfigure}[b]{0.32\linewidth}
        \centering
        \includegraphics[width=\linewidth]{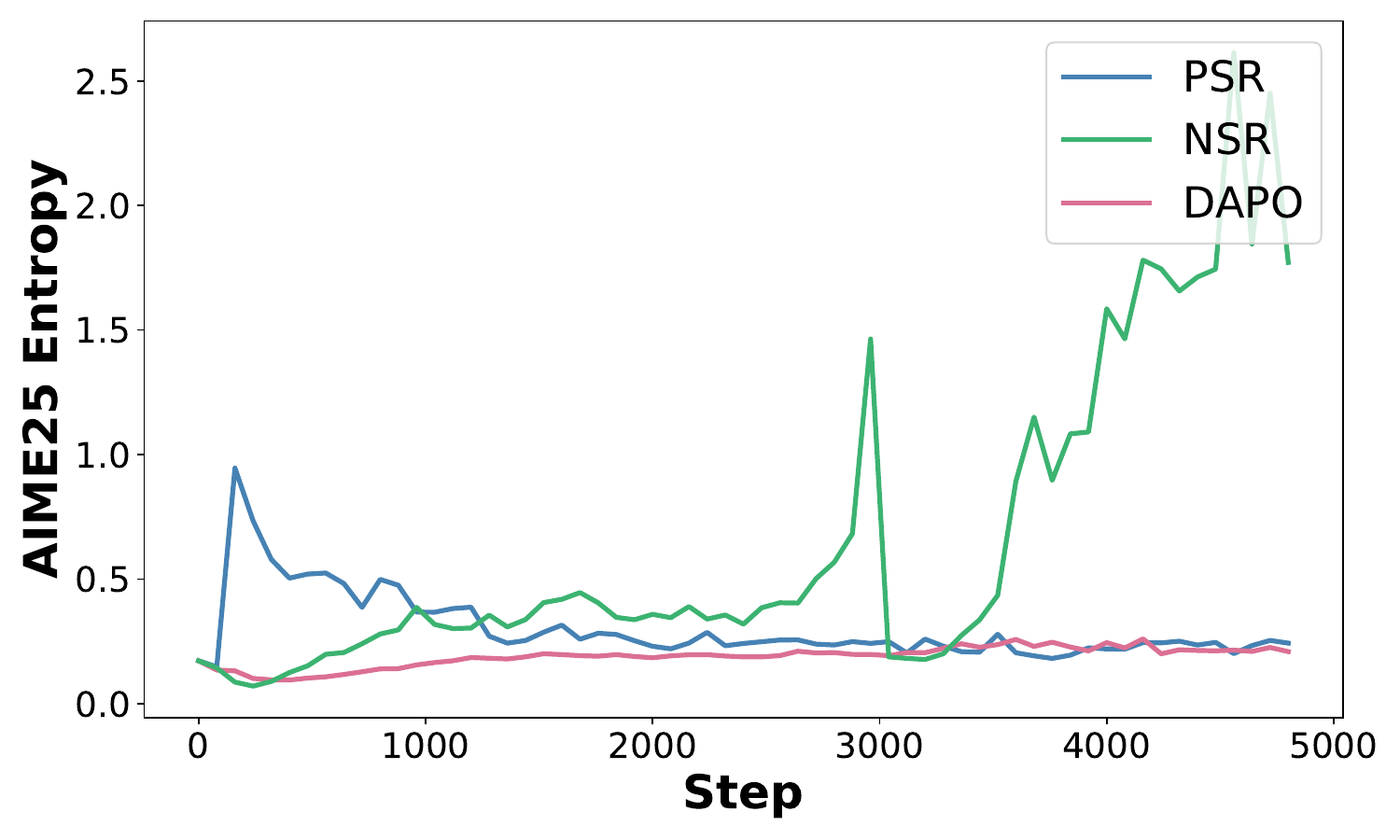}
        \caption{AIME25 Entropy}
    \end{subfigure}
    \begin{subfigure}[b]{0.32\linewidth}
        \centering
        \includegraphics[width=\linewidth]{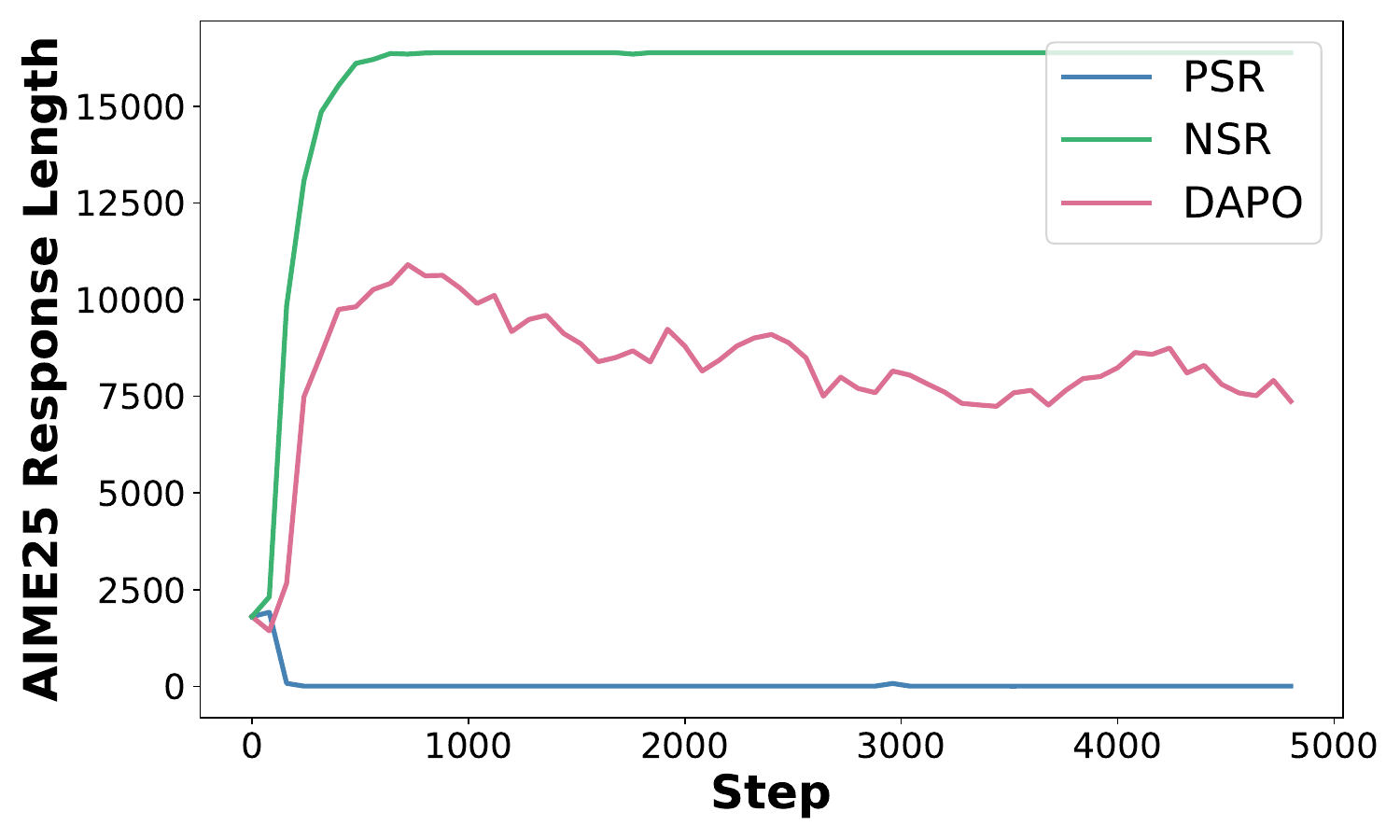}
        \caption{AIME25 Length}
    \end{subfigure}
    \begin{subfigure}[b]{0.32\linewidth}
        \centering
        \includegraphics[width=\linewidth]{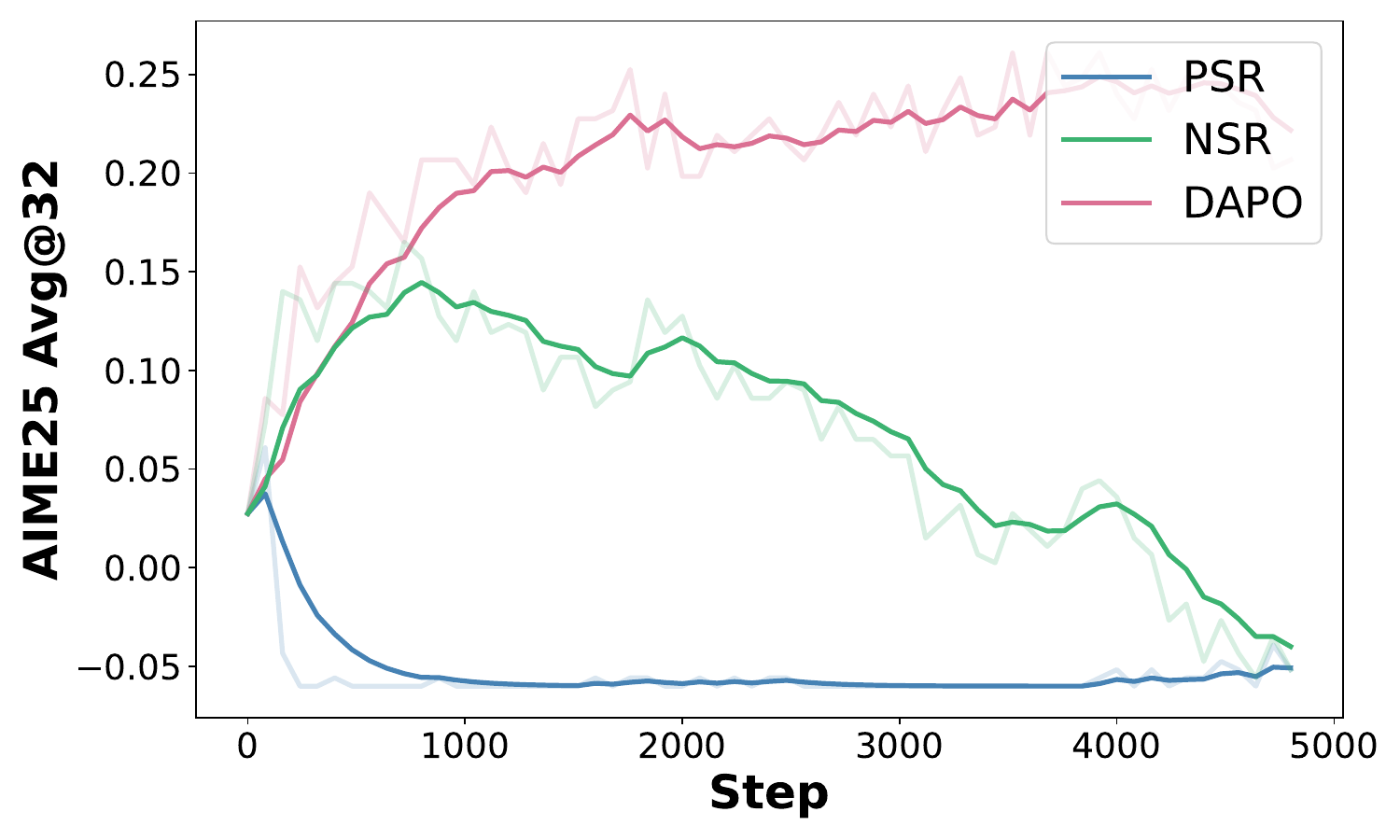}
        \caption{AIME25 Avg@32}
    \end{subfigure}
    \begin{subfigure}[b]{0.32\linewidth}
        \centering
        \includegraphics[width=\linewidth]{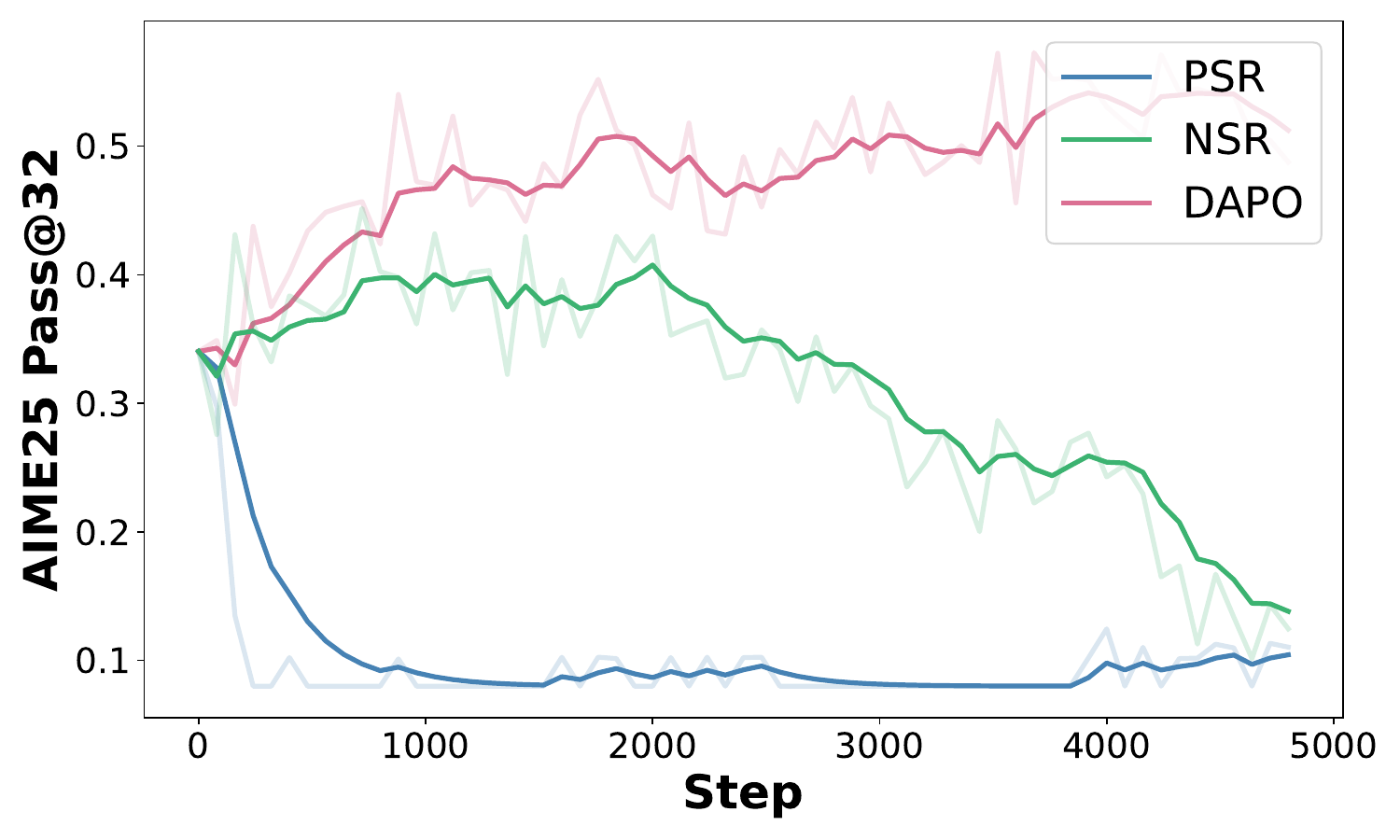}
        \caption{AIME25 Pass@32}
    \end{subfigure}
    \caption{RLVR training dynamics on Qwen3-8B-Base.}
\label{fig:qwen3-8b-base-training_dynamic}
\end{figure*}

\begin{figure*}[t]
    \centering
    \begin{subfigure}[b]{0.32\linewidth}
        \centering
        \includegraphics[width=\linewidth]{figures/psr_nsr/ds-qwen-7b/Entropy.pdf}
        \caption{Entropy}
    \end{subfigure}
    \begin{subfigure}[b]{0.32\linewidth}
        \centering
        \includegraphics[width=\linewidth]{figures/psr_nsr/ds-qwen-7b/Response_Length.pdf}
        \caption{Length}
    \end{subfigure}
    \begin{subfigure}[b]{0.32\linewidth}
        \centering
        \includegraphics[width=\linewidth]{figures/psr_nsr/ds-qwen-7b/Reward.pdf}
        \caption{Reward}
    \end{subfigure}
    \begin{subfigure}[b]{0.32\linewidth}
        \centering
        \includegraphics[width=\linewidth]{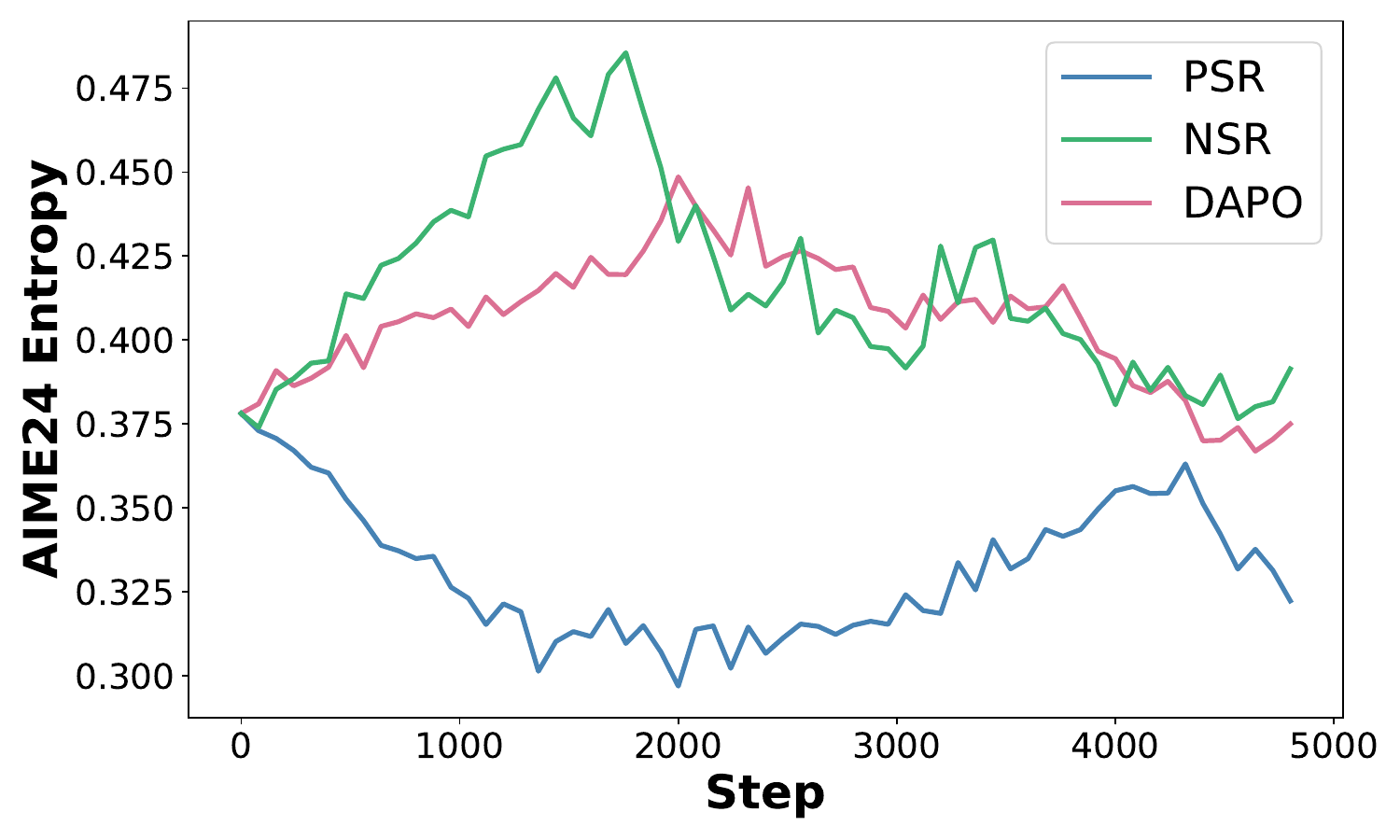}
        \caption{AIME24 Entropy}
    \end{subfigure}
    \begin{subfigure}[b]{0.32\linewidth}
        \centering
        \includegraphics[width=\linewidth]{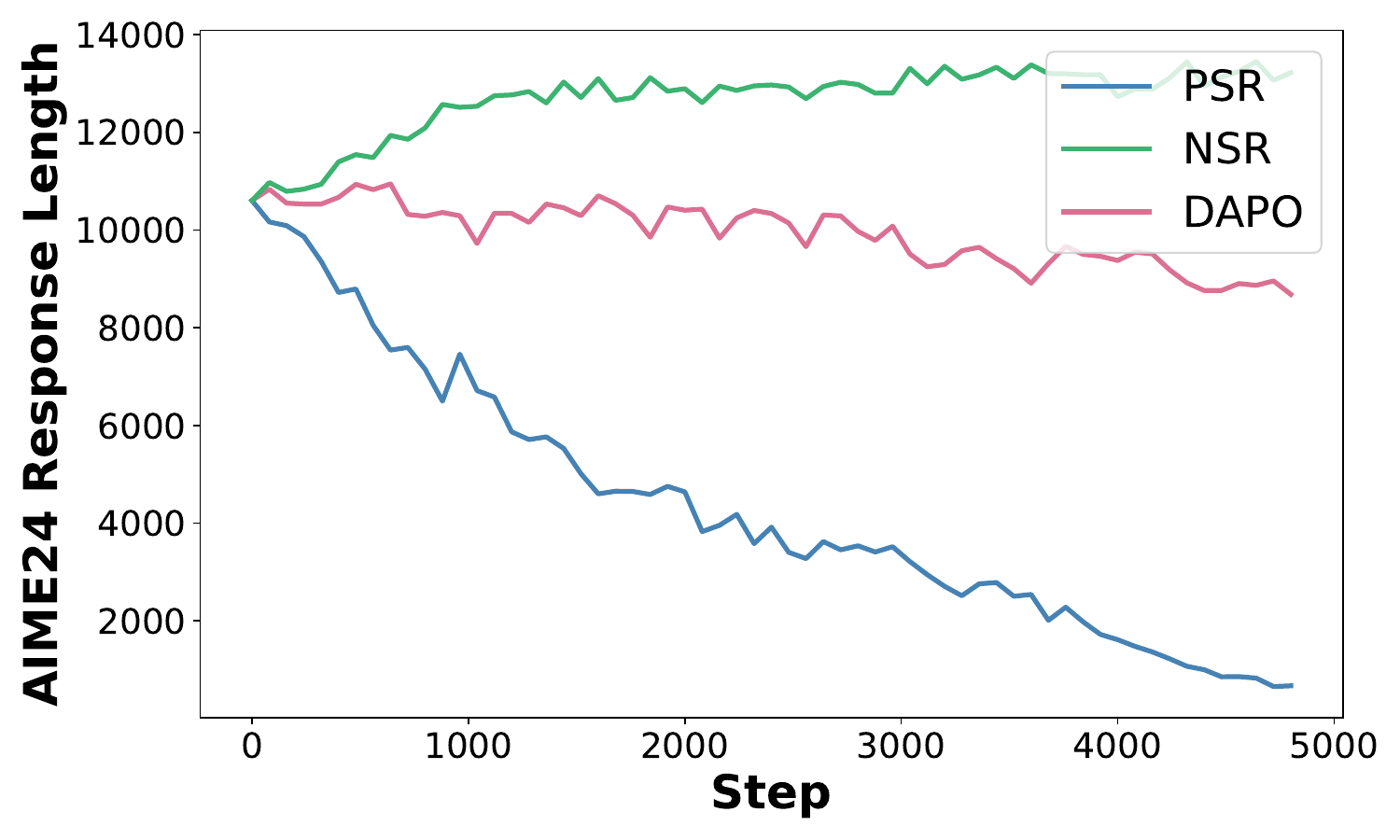}
        \caption{AIME24 Length}
    \end{subfigure}
    \begin{subfigure}[b]{0.32\linewidth}
        \centering
        \includegraphics[width=\linewidth]{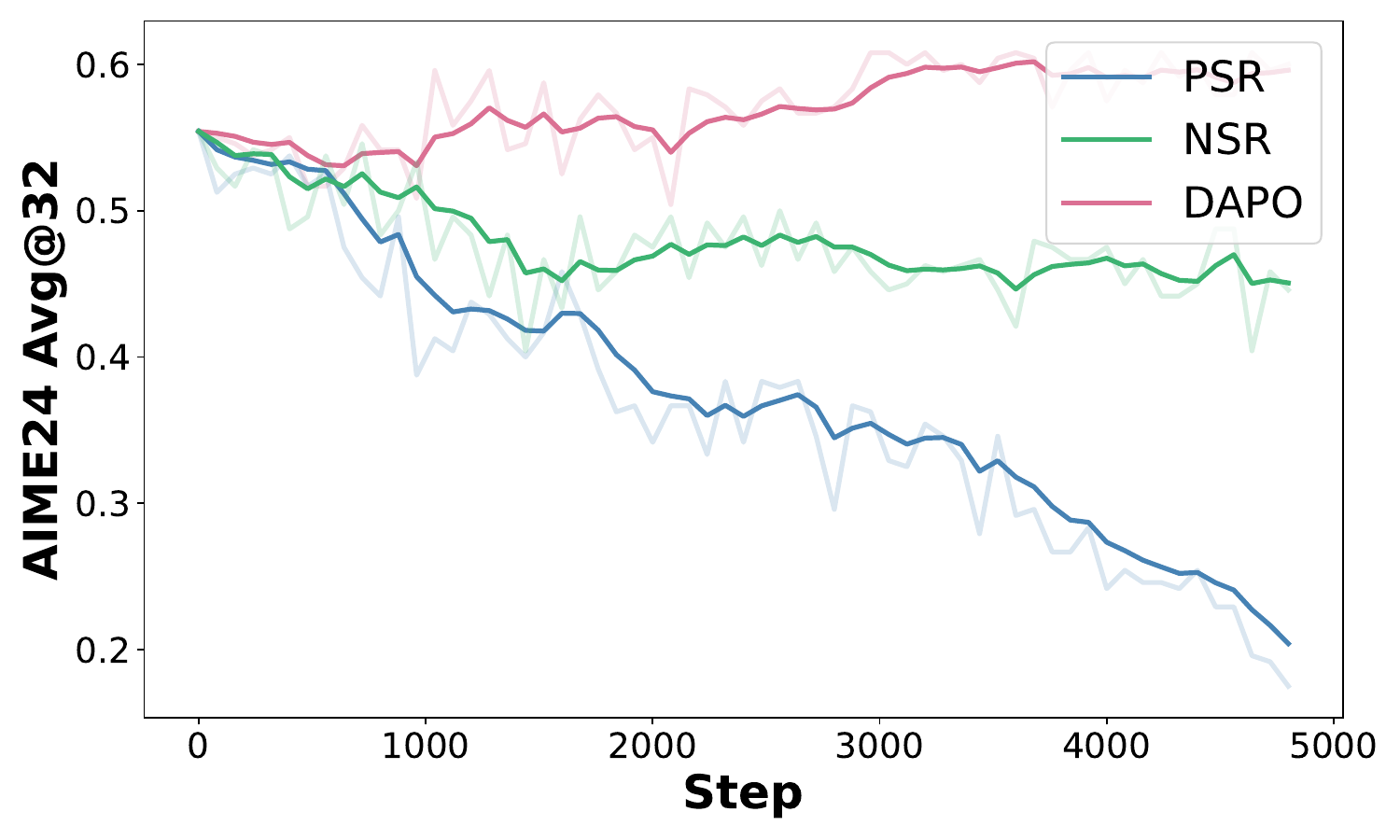}
        \caption{AIME24 Avg@32}
    \end{subfigure}
    \begin{subfigure}[b]{0.32\linewidth}
        \centering
        \includegraphics[width=\linewidth]{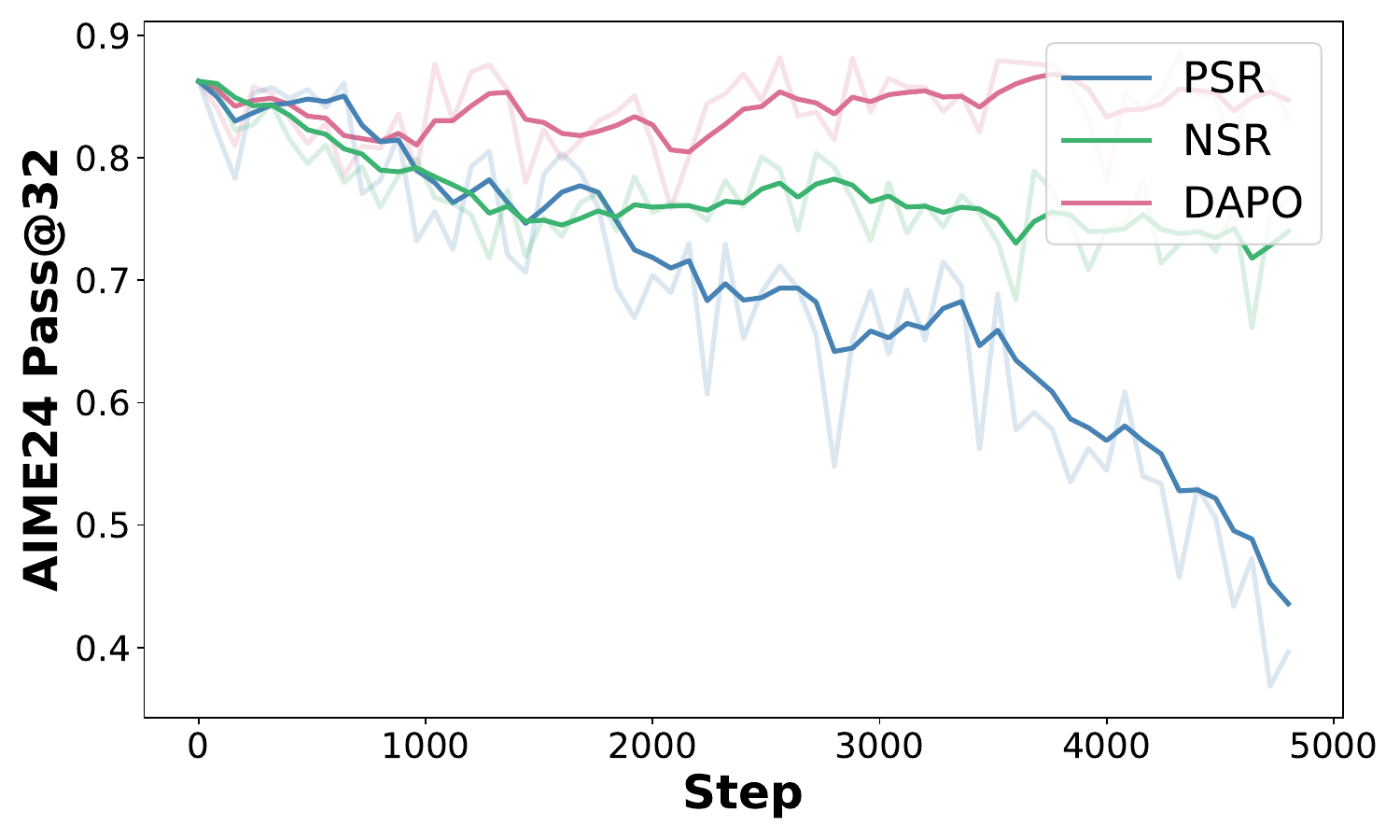}
        \caption{AIME24 Pass@32}
    \end{subfigure}
    \begin{subfigure}[b]{0.32\linewidth}
        \centering
        \includegraphics[width=\linewidth]{figures/psr_nsr/ds-qwen-7b/AIME25_Entropy.pdf}
        \caption{AIME25 Entropy}
    \end{subfigure}
    \begin{subfigure}[b]{0.32\linewidth}
        \centering
        \includegraphics[width=\linewidth]{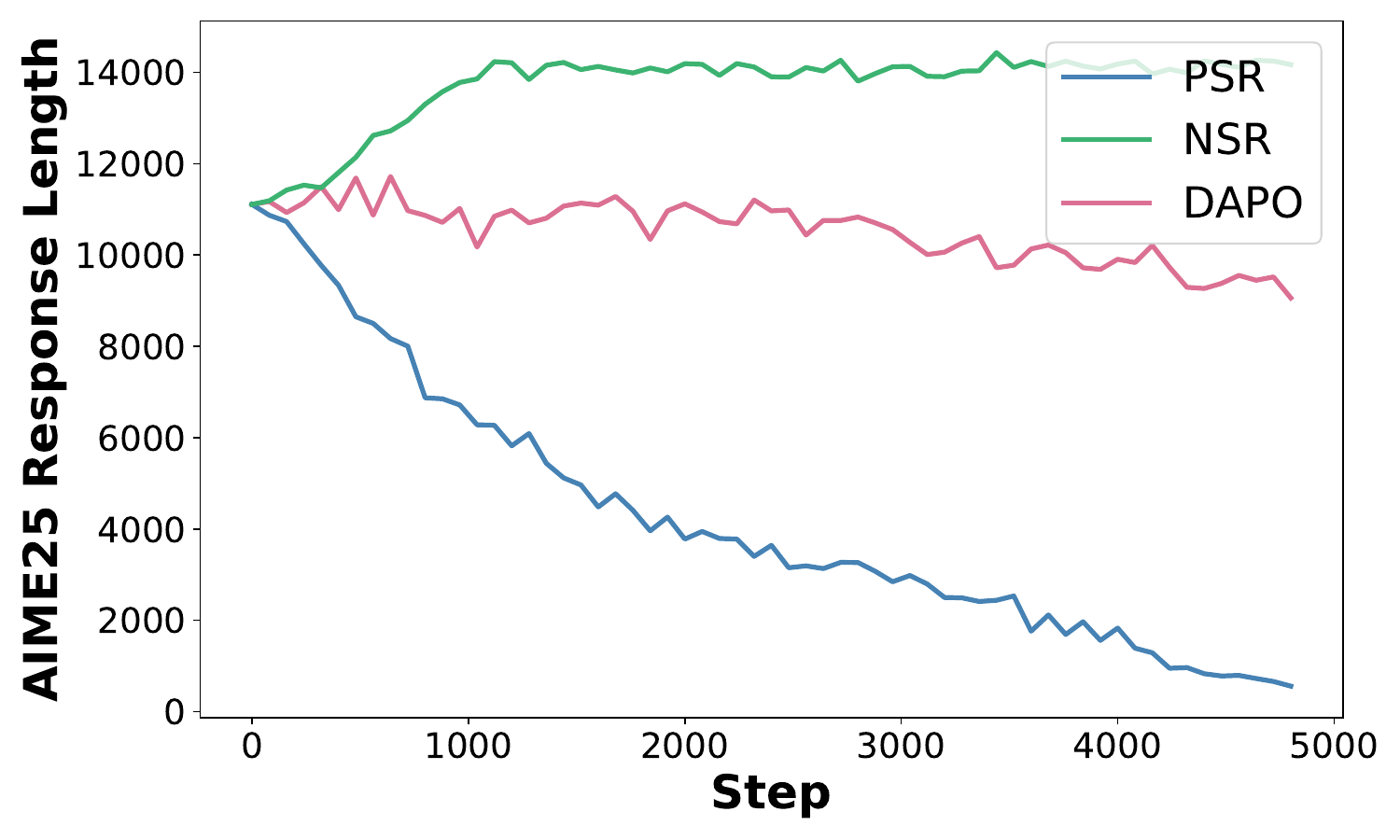}
        \caption{AIME25 Length}
    \end{subfigure}
    \begin{subfigure}[b]{0.32\linewidth}
        \centering
        \includegraphics[width=\linewidth]{figures/psr_nsr/ds-qwen-7b/AIME25_Avg32.pdf}
        \caption{AIME25 Avg@32}
    \end{subfigure}
    \begin{subfigure}[b]{0.32\linewidth}
        \centering
        \includegraphics[width=\linewidth]{figures/psr_nsr/ds-qwen-7b/AIME25_Pass32.pdf}
        \caption{AIME25 Pass@32}
    \end{subfigure}
    \caption{RLVR training dynamics on DeepSeek-R1-Distill-Qwen-7B.}
\label{fig:ds-qwen-7b-training_dynamic}
\end{figure*}

\begin{figure*}[t]
    \centering
    \begin{subfigure}[b]{0.24\linewidth}
        \centering
        \includegraphics[width=\linewidth]{figures/sample-level-exp-dis/ds-qwen-7b/sharpen_3.pdf}
        \caption{Ds Sharpen}
        \label{subfig:ds-sharpen3}
    \end{subfigure}
    \begin{subfigure}[b]{0.24\linewidth}
        \centering
        \includegraphics[width=\linewidth]{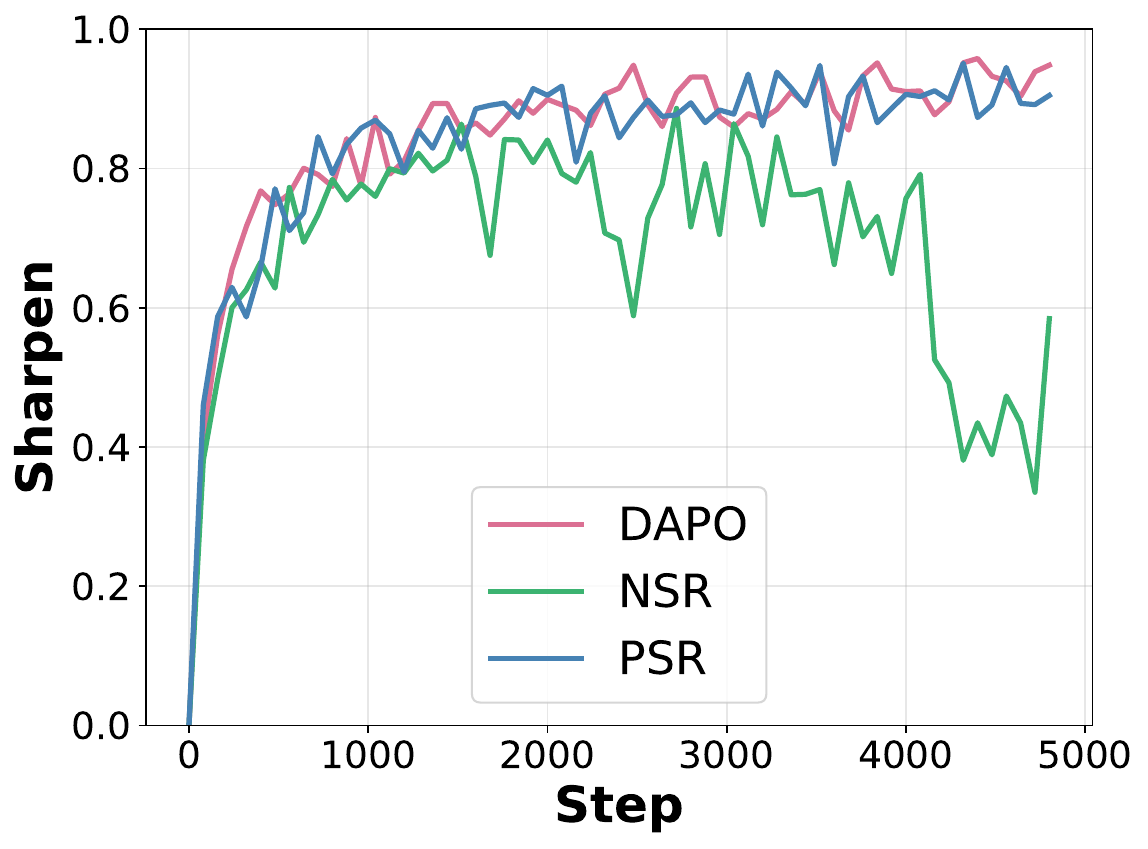}
        \caption{Qwen Math Sharpen}
        \label{subfig:qwen-math-sharpen3}
    \end{subfigure}
    \begin{subfigure}[b]{0.24\linewidth}
        \centering
        \includegraphics[width=\linewidth]{figures/sample-level-exp-dis/ds-qwen-7b/discovery_3.pdf}
        \caption{Ds Discovery}
        \label{subfig:ds-discovery3}
    \end{subfigure}
    \begin{subfigure}[b]{0.24\linewidth}
        \centering
        \includegraphics[width=\linewidth]{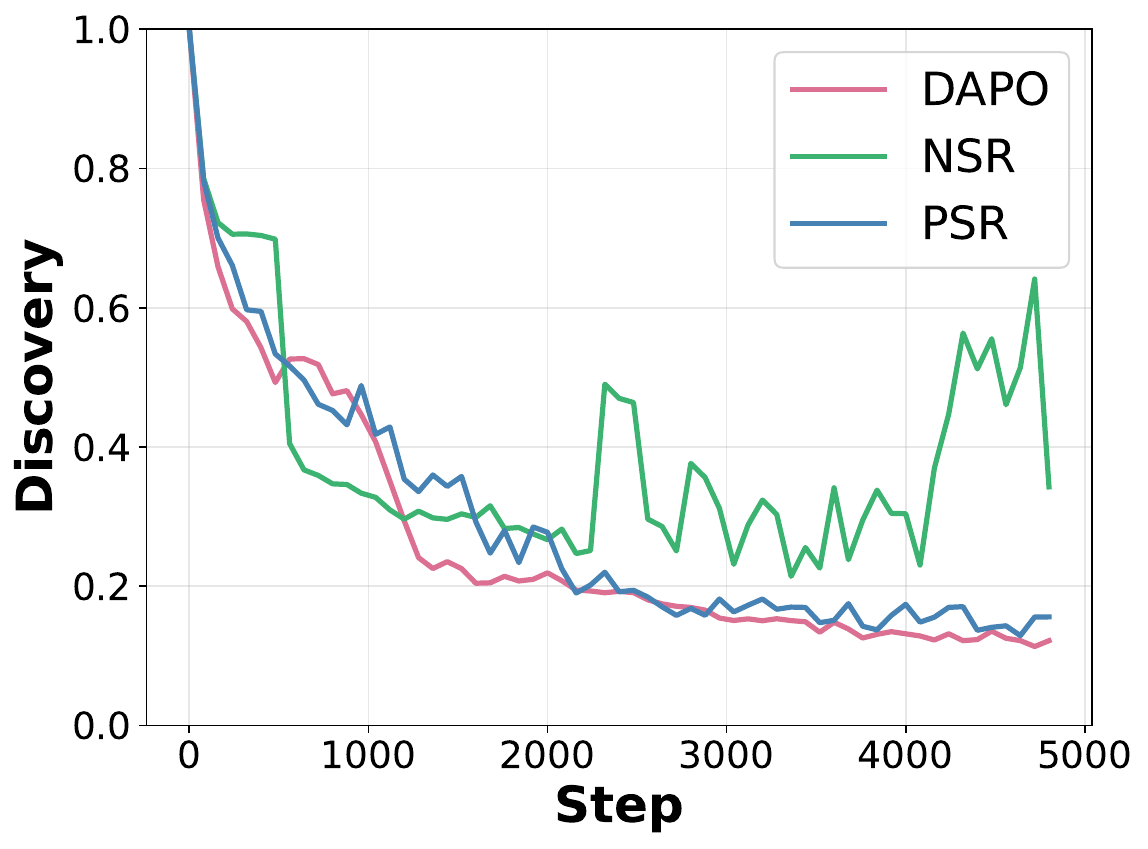}
        \caption{Qwen Math Discovery}
        \label{subfig:qwen-discovery3}
    \end{subfigure}
    \caption{Training behaviors of different RLVR training when n\_gram is 3.}
\label{fig:sharpen-discovery3}
\end{figure*}

\begin{figure*}[t]
    \centering
    \begin{subfigure}[b]{0.24\linewidth}
        \centering
        \includegraphics[width=\linewidth]{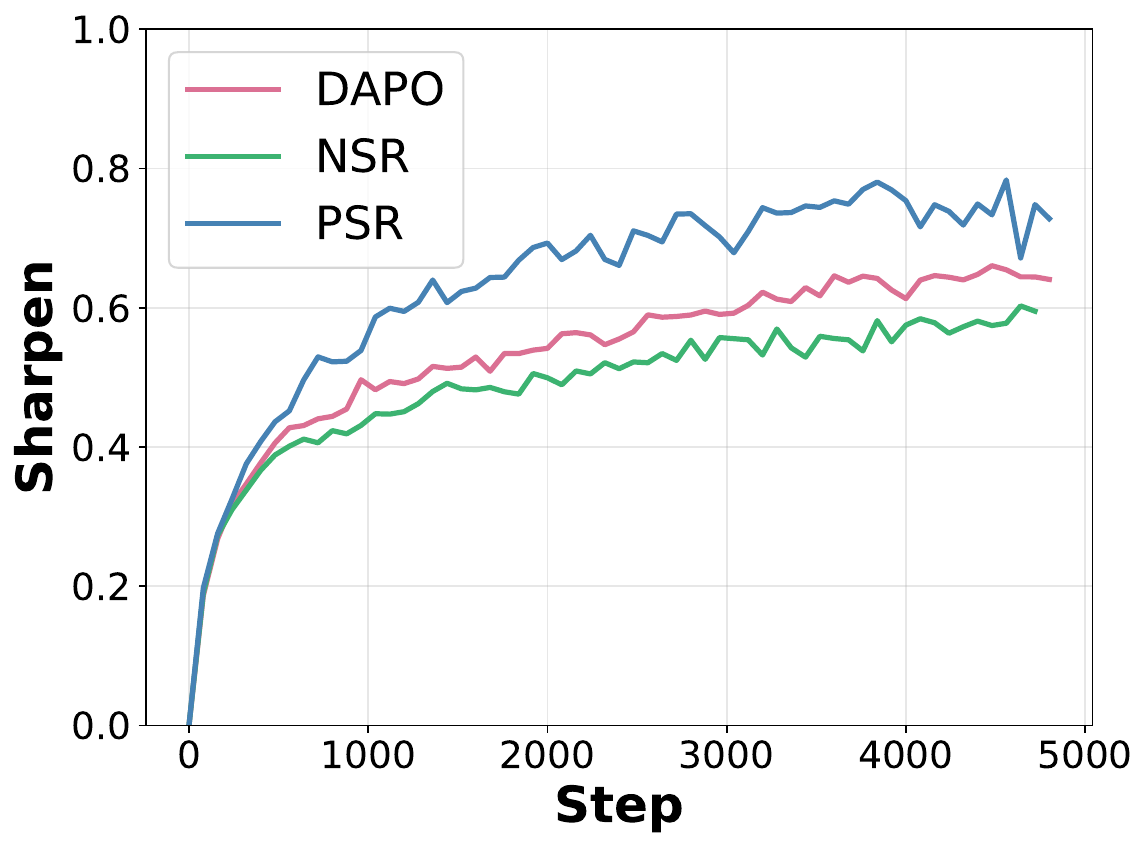}
        \caption{Ds Sharpen}
        \label{subfig:ds-sharpen4}
    \end{subfigure}
    \begin{subfigure}[b]{0.24\linewidth}
        \centering
        \includegraphics[width=\linewidth]{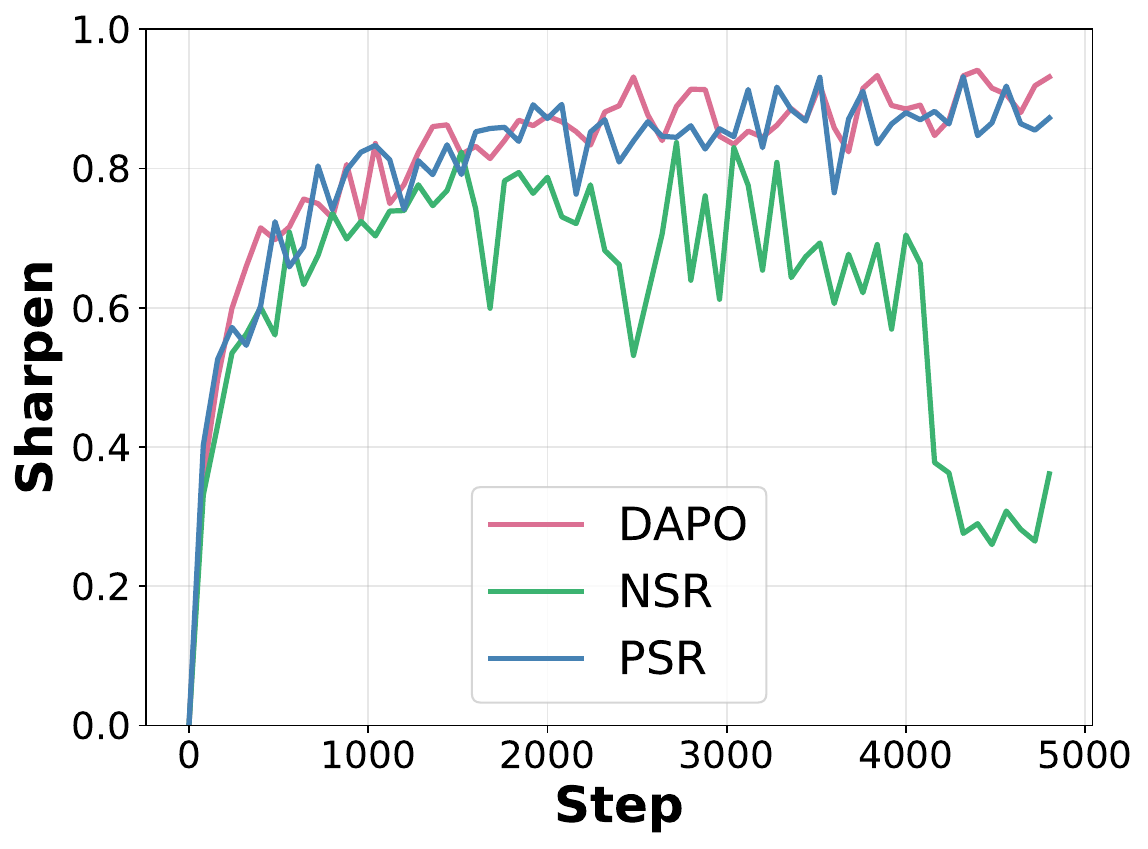}
        \caption{Qwen Math Sharpen}
        \label{subfig:qwen-sharpen4}
    \end{subfigure}
    \begin{subfigure}[b]{0.24\linewidth}
        \centering
        \includegraphics[width=\linewidth]{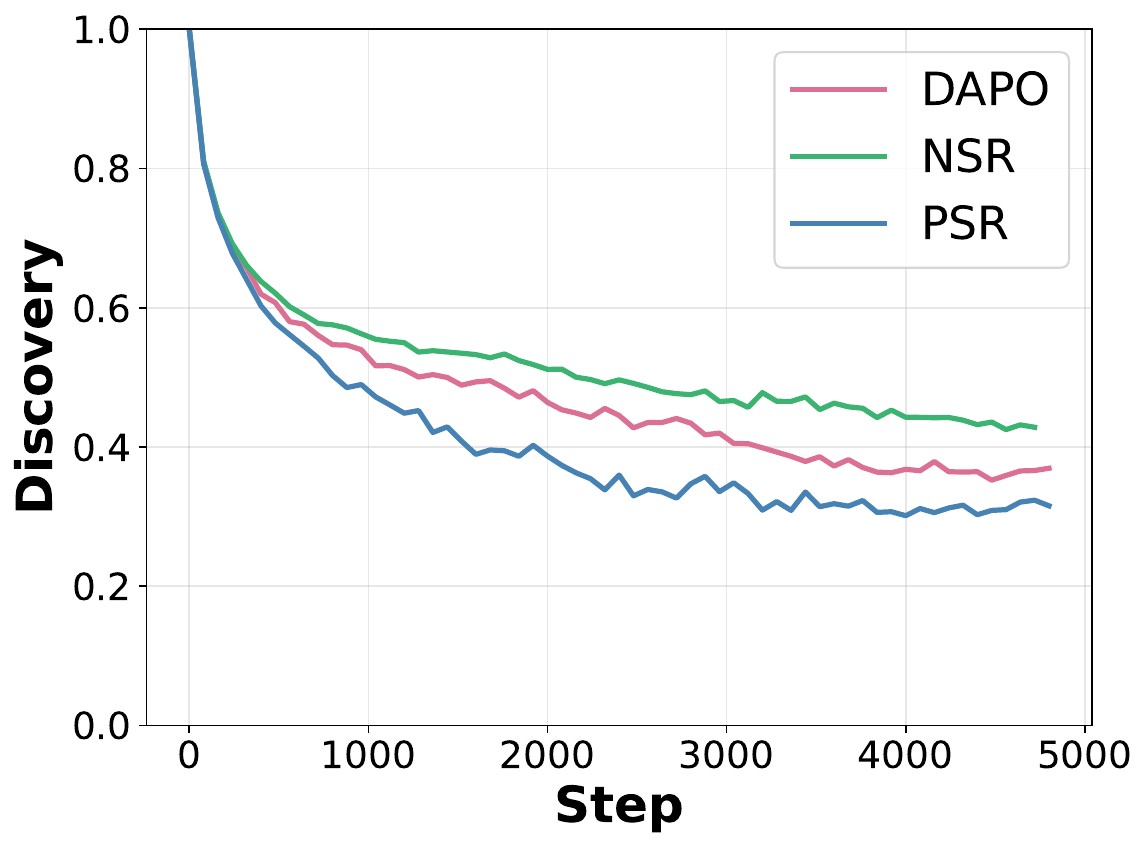}
        \caption{Ds Discovery}
        \label{subfig:ds-discovery4}
    \end{subfigure}
    \begin{subfigure}[b]{0.24\linewidth}
        \centering
        \includegraphics[width=\linewidth]{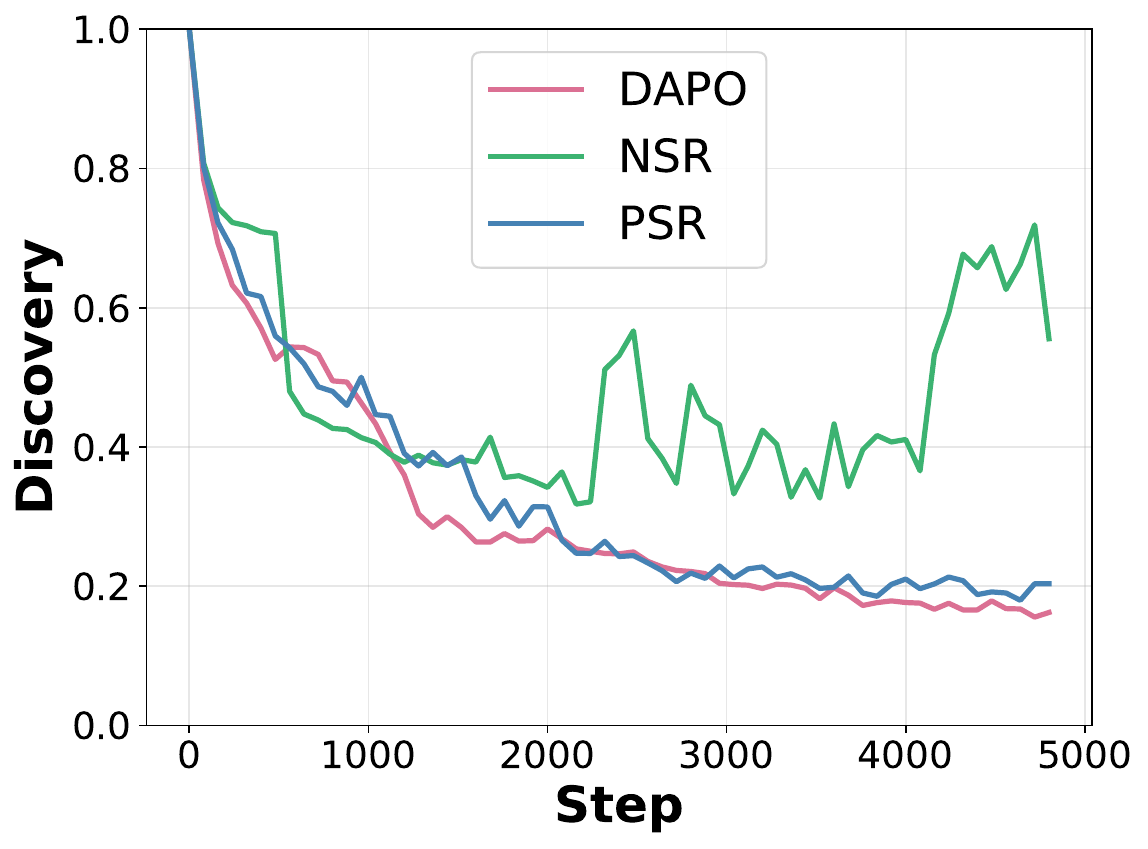}
        \caption{Qwen Math Discovery}
        \label{subfig:qwen-discovery4}
    \end{subfigure}
    \caption{Training behaviors of different RLVR training when n\_gram is 4.}
\label{fig:sharpen-discovery4}
\end{figure*}

\begin{figure*}[t]
    \centering
    \begin{subfigure}[b]{0.32\linewidth}
        \centering
        \includegraphics[width=\linewidth]{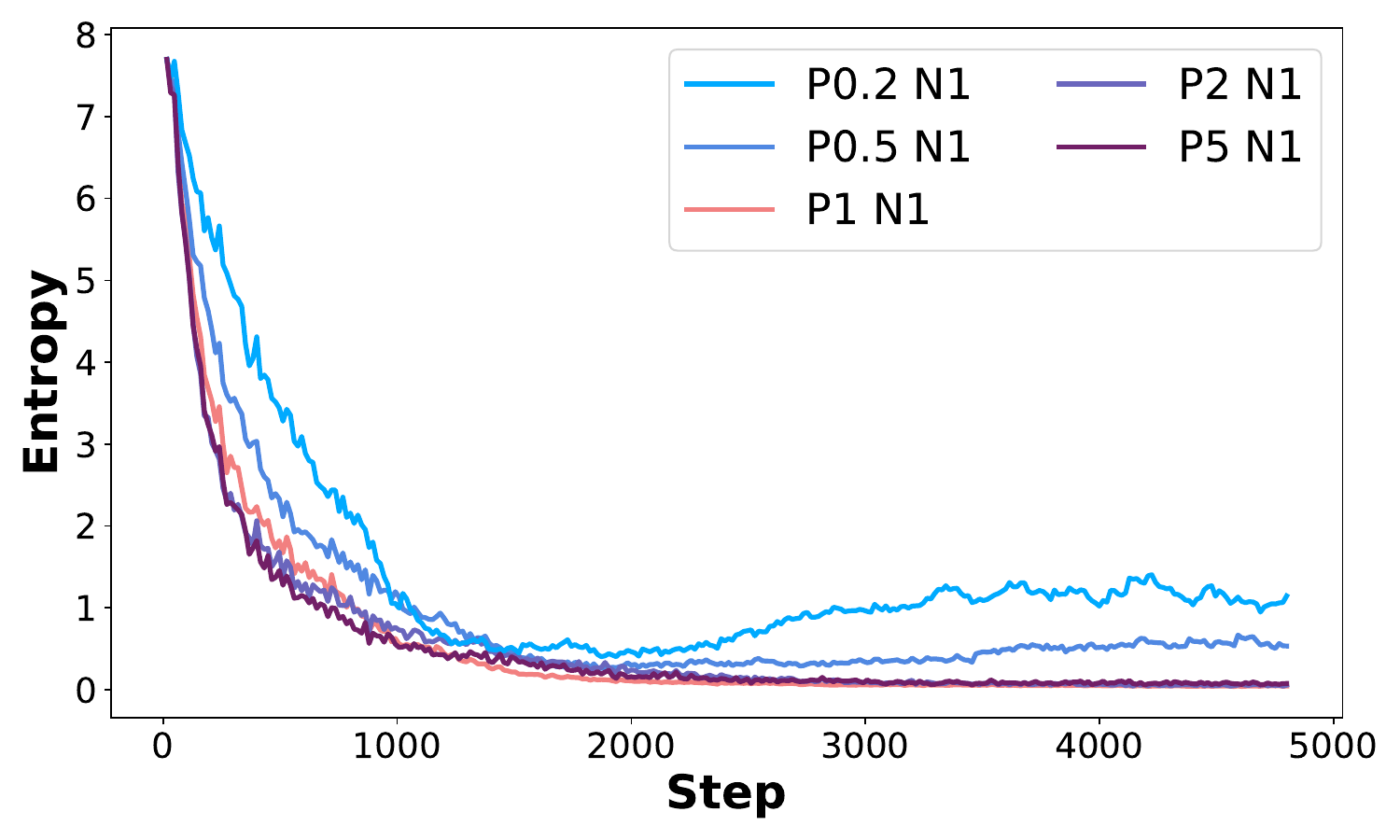}
        \caption{Entropy}
    \end{subfigure}
    \begin{subfigure}[b]{0.32\linewidth}
        \centering
        \includegraphics[width=\linewidth]{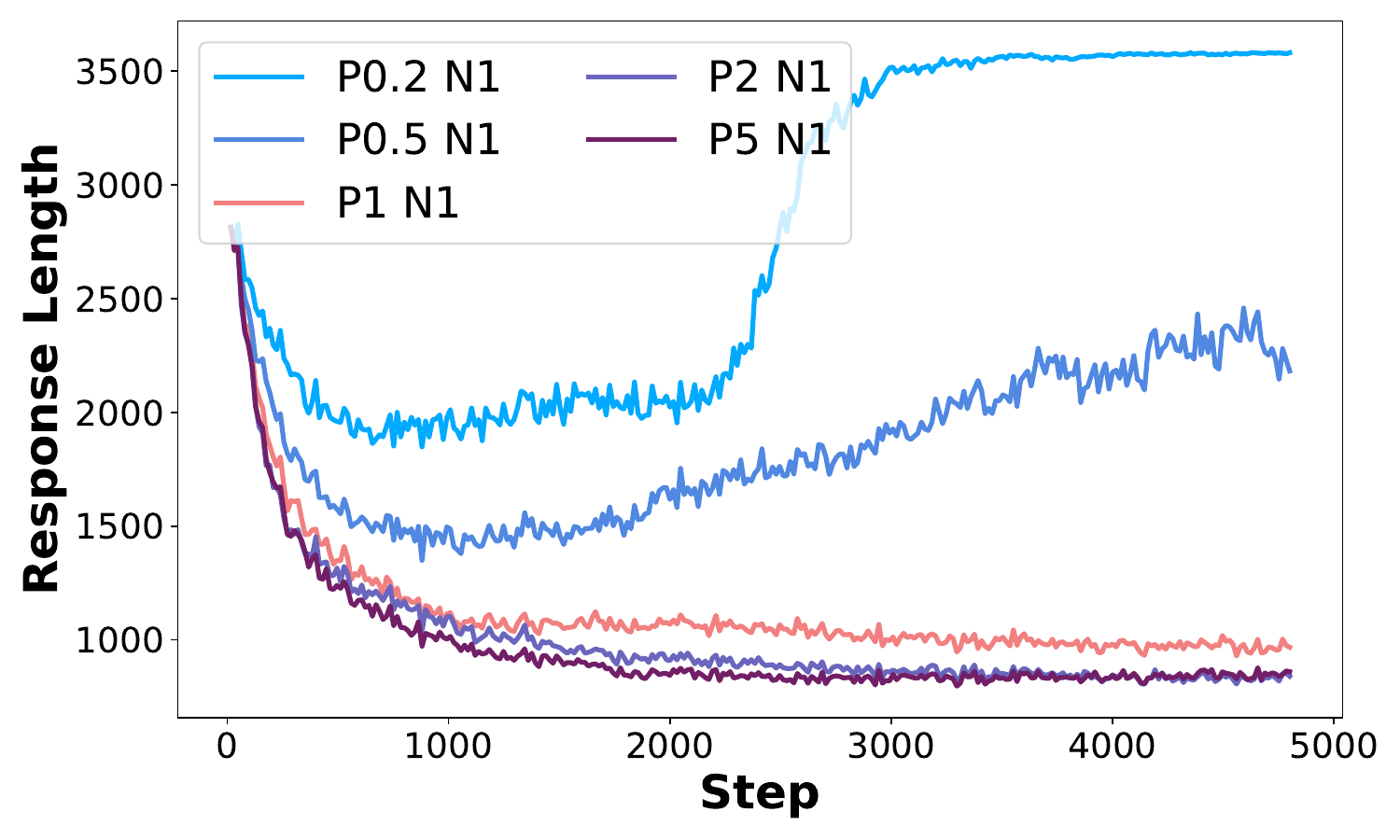}
        \caption{Length}
    \end{subfigure}
    \begin{subfigure}[b]{0.32\linewidth}
        \centering
        \includegraphics[width=\linewidth]{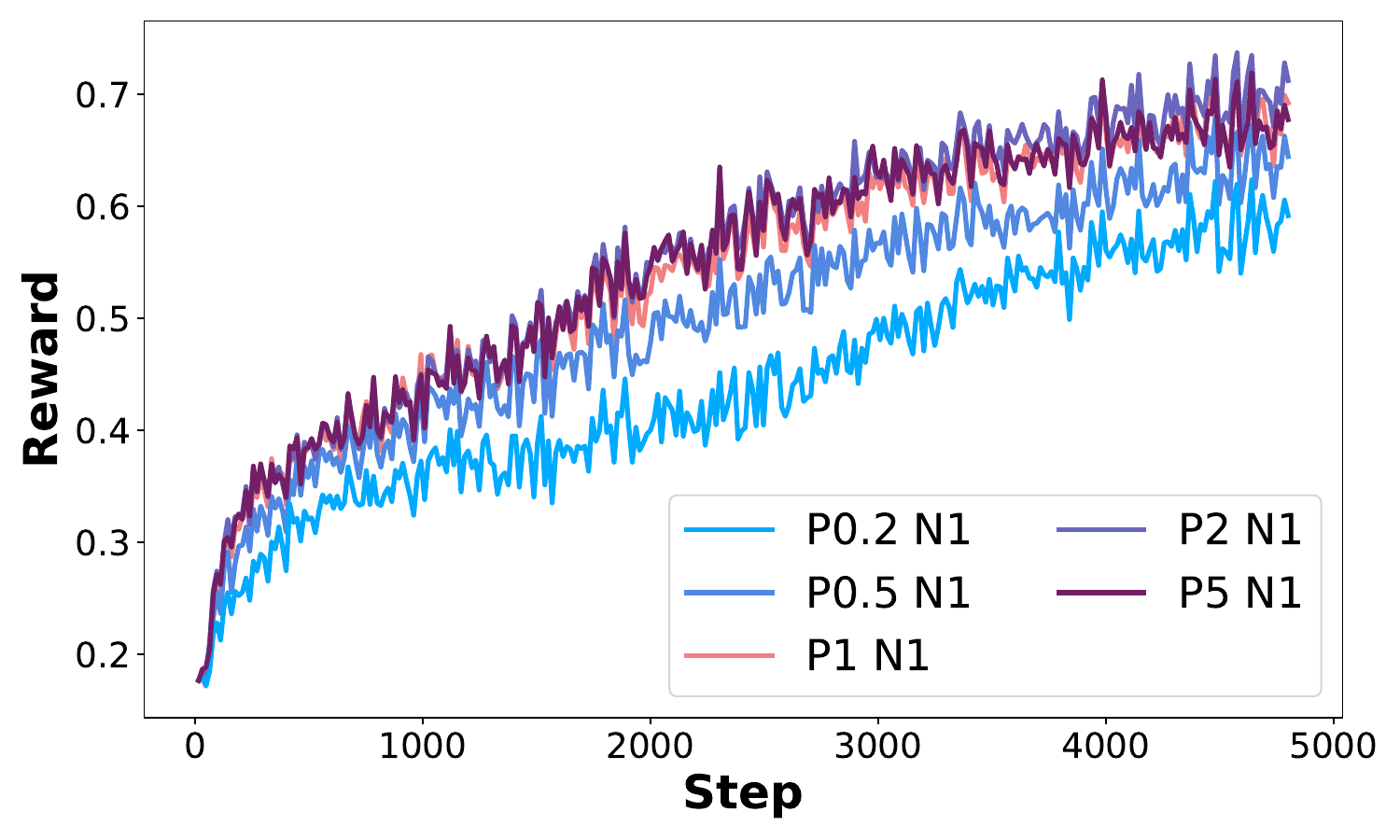}
        \caption{Reward}
    \end{subfigure}
    \begin{subfigure}[b]{0.32\linewidth}
        \centering
        \includegraphics[width=\linewidth]{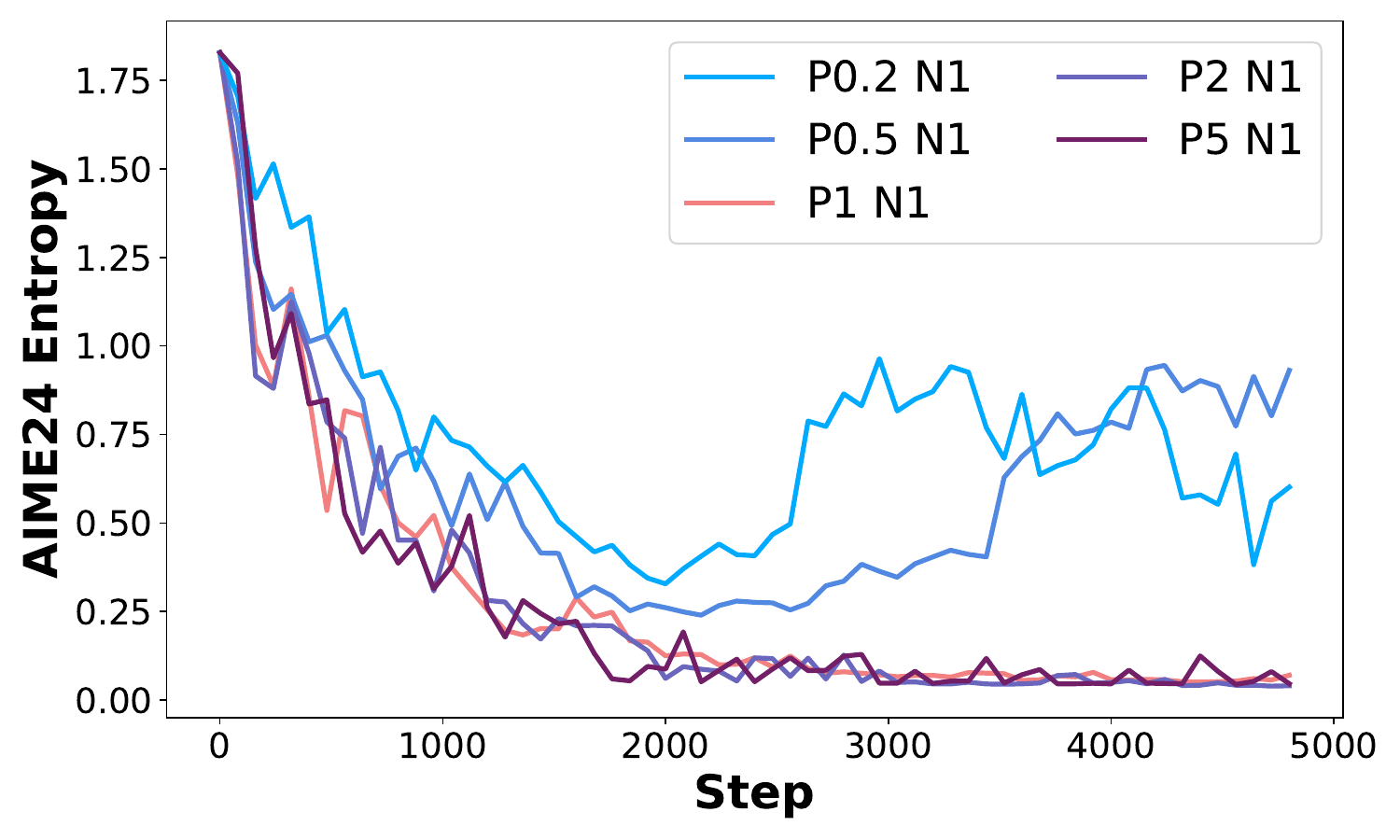}
        \caption{AIME24 Entropy}
    \end{subfigure}
    \begin{subfigure}[b]{0.32\linewidth}
        \centering
        \includegraphics[width=\linewidth]{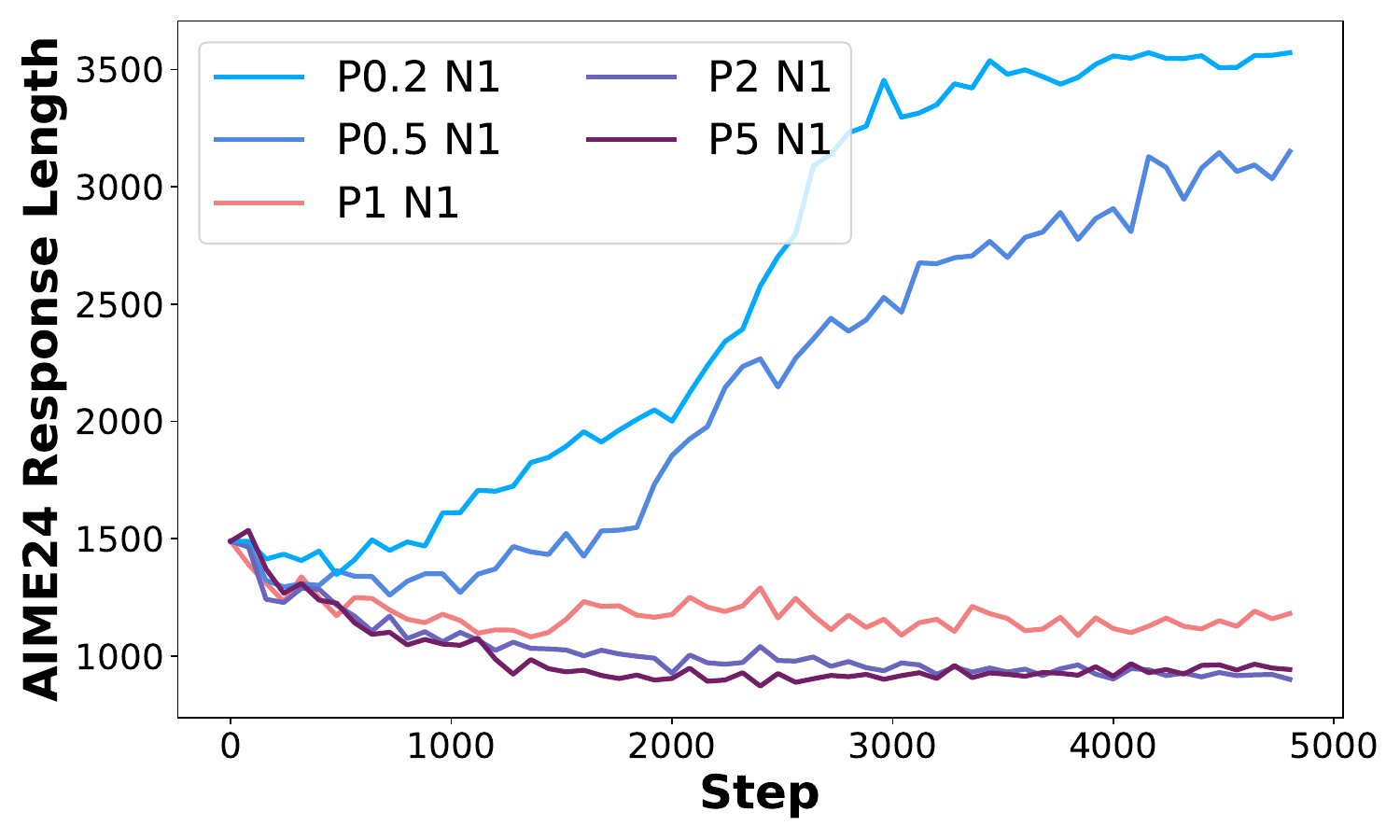}
        \caption{AIME24 Length}
    \end{subfigure}
    \begin{subfigure}[b]{0.32\linewidth}
        \centering
        \includegraphics[width=\linewidth]{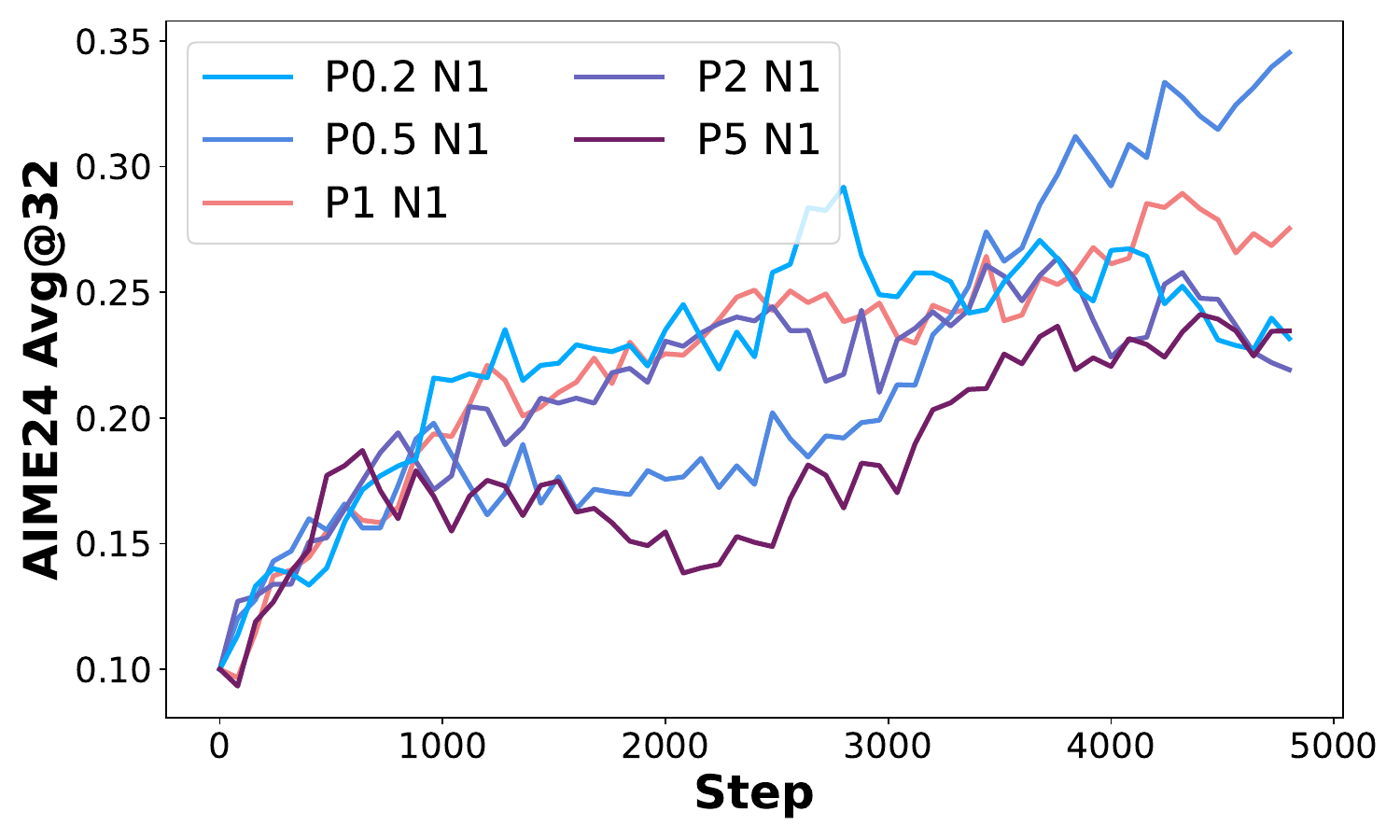}
        \caption{AIME24 Avg@32}
    \end{subfigure}
    \begin{subfigure}[b]{0.32\linewidth}
        \centering
        \includegraphics[width=\linewidth]{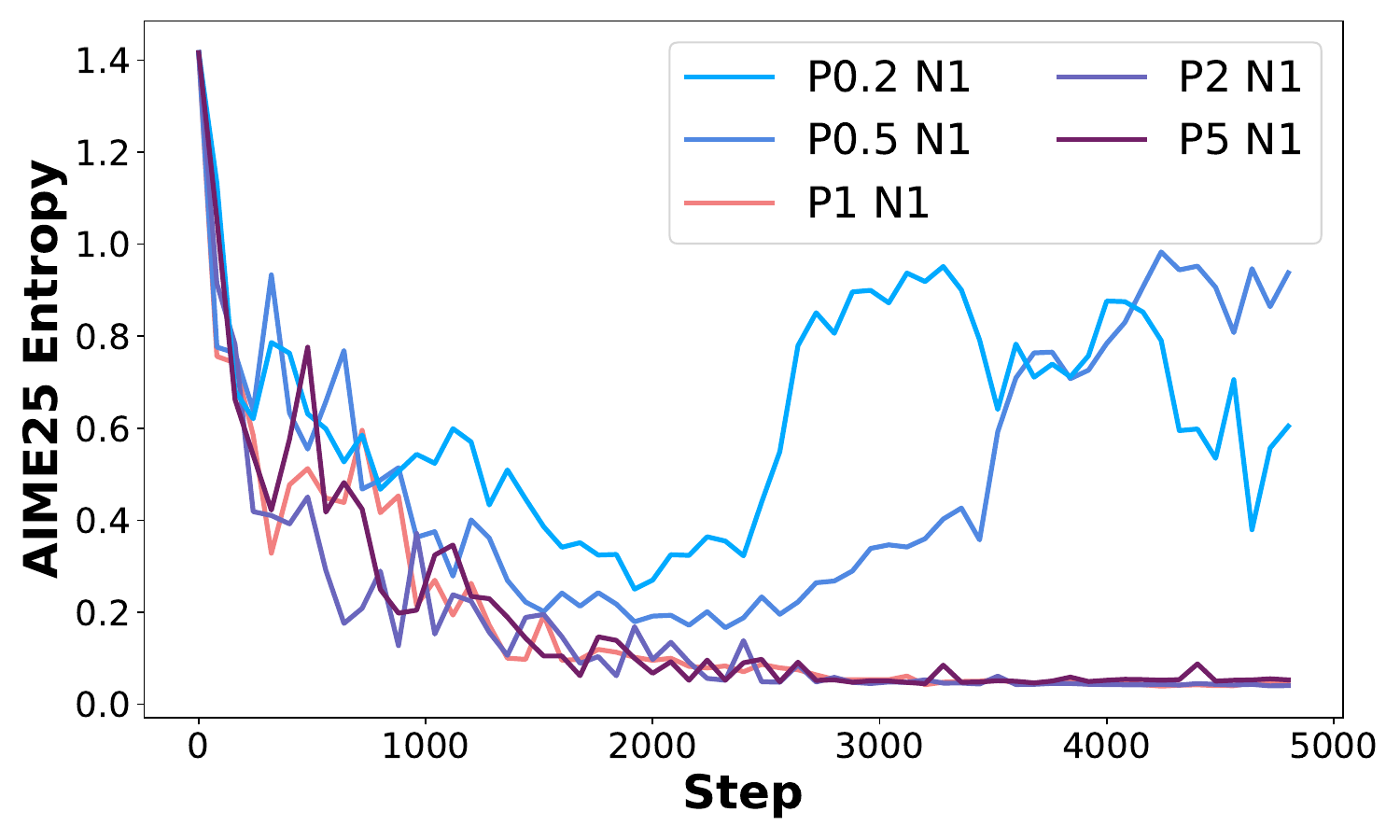}
        \caption{AIME25 Entropy}
    \end{subfigure}
    \begin{subfigure}[b]{0.32\linewidth}
        \centering
        \includegraphics[width=\linewidth]{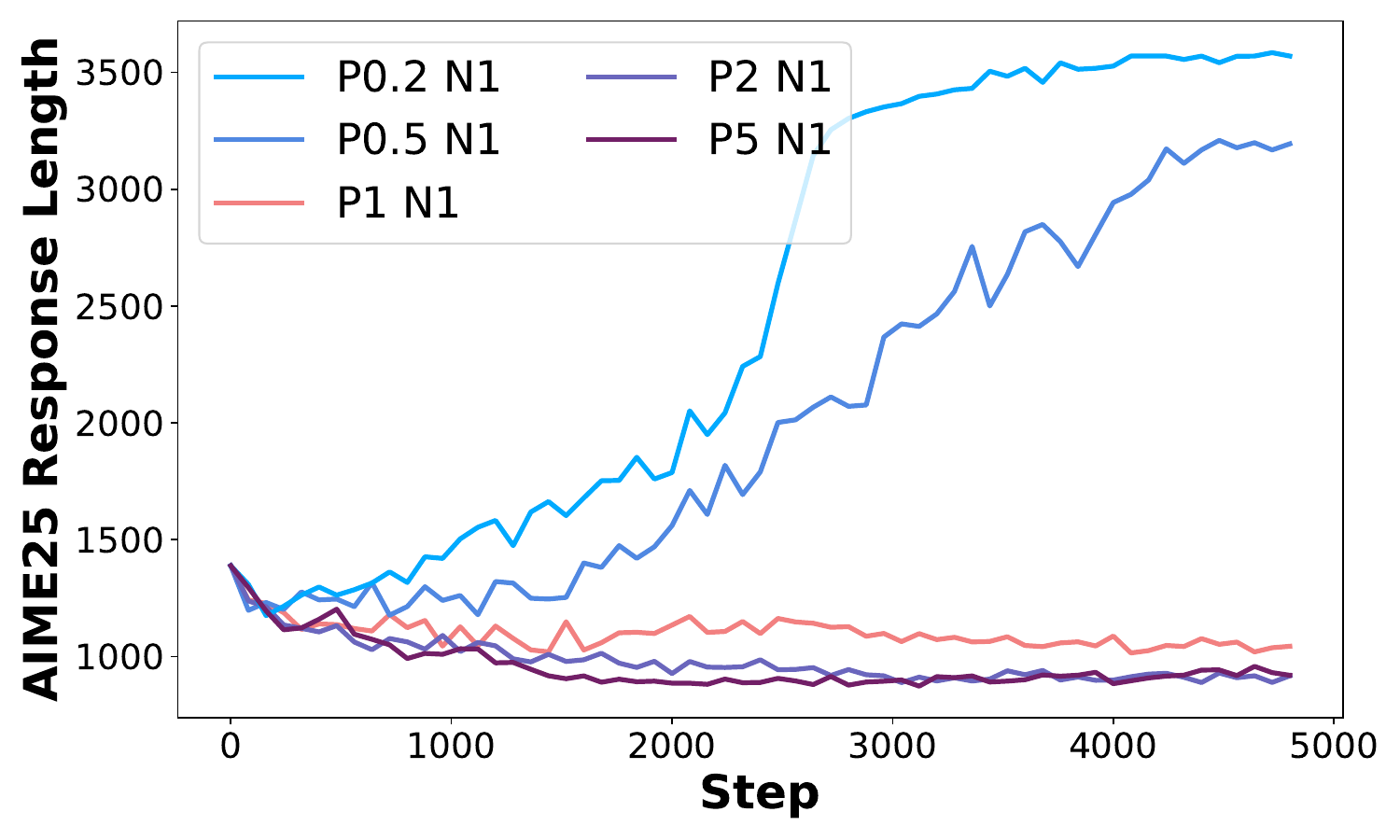}
        \caption{AIME25 Length}
    \end{subfigure}
    \begin{subfigure}[b]{0.32\linewidth}
        \centering
        \includegraphics[width=\linewidth]{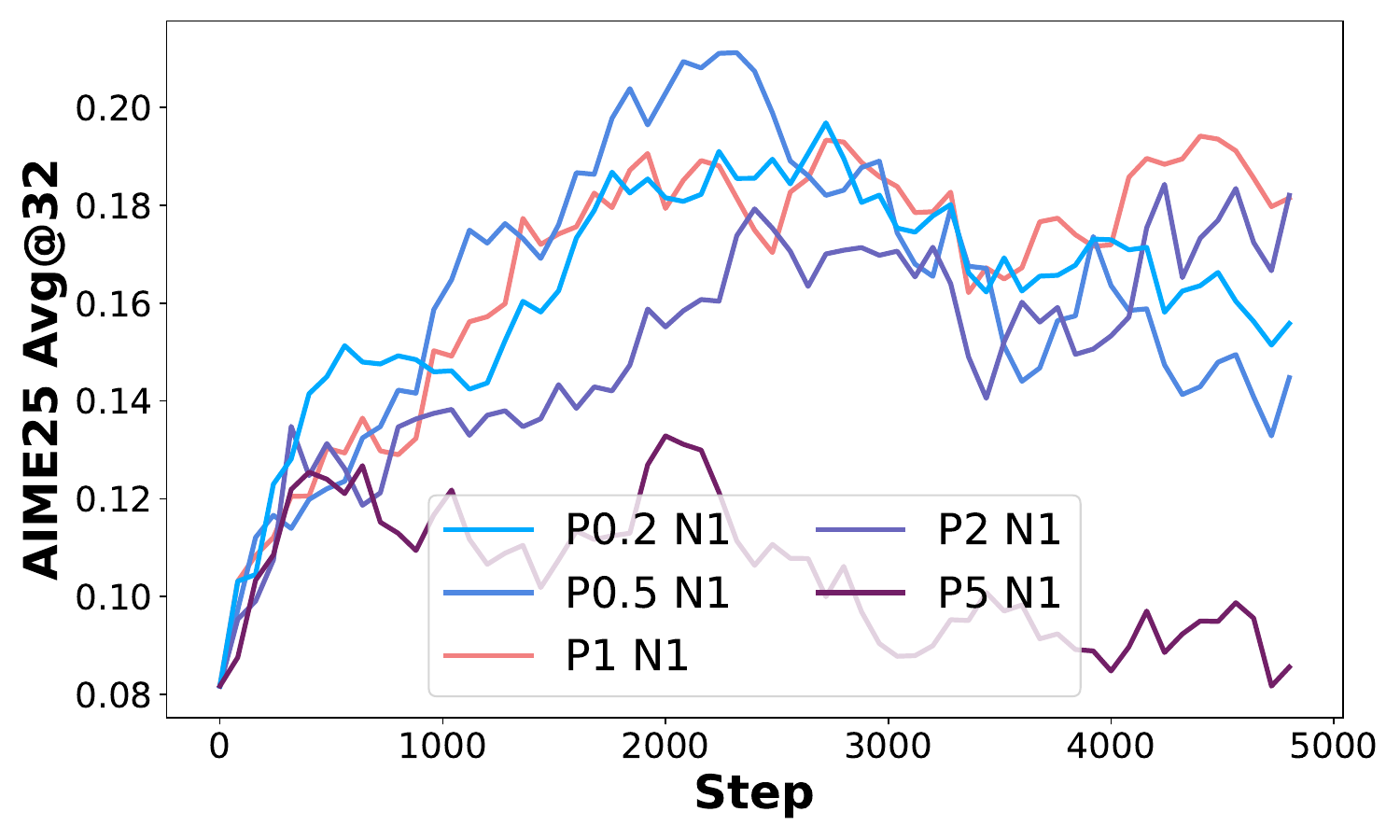}
        \caption{AIME25 Avg@32}
    \end{subfigure}
    \caption{Different training dynamics of polarity-level positive sample advantage shaping on Qwen2.5-7B-Math.}
\label{fig:polarity-pas-training_dynamic}
\end{figure*}

\begin{figure*}[t]
    \centering
    \begin{subfigure}[b]{0.32\linewidth}
        \centering
        \includegraphics[width=\linewidth]{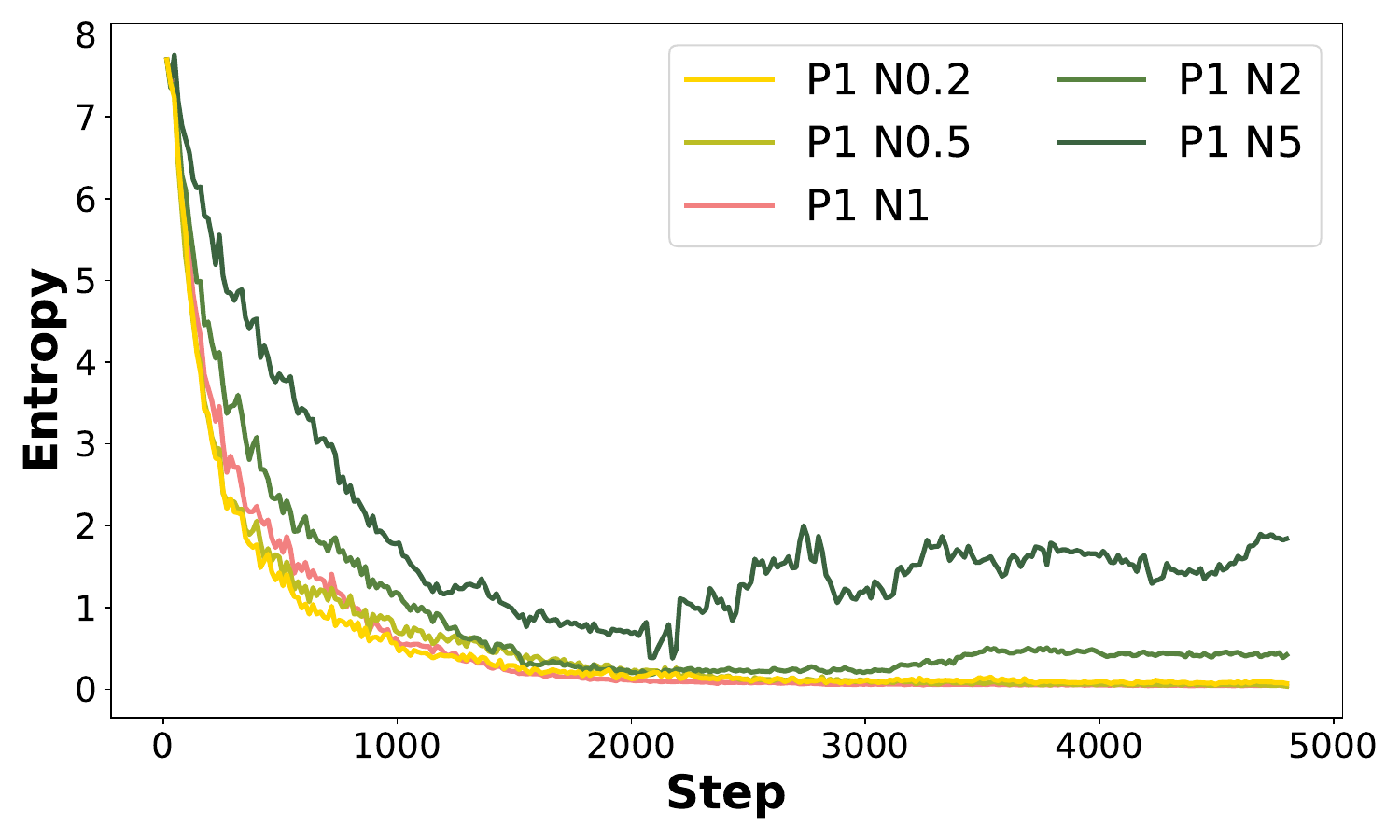}
        \caption{Entropy}
    \end{subfigure}
    \begin{subfigure}[b]{0.32\linewidth}
        \centering
        \includegraphics[width=\linewidth]{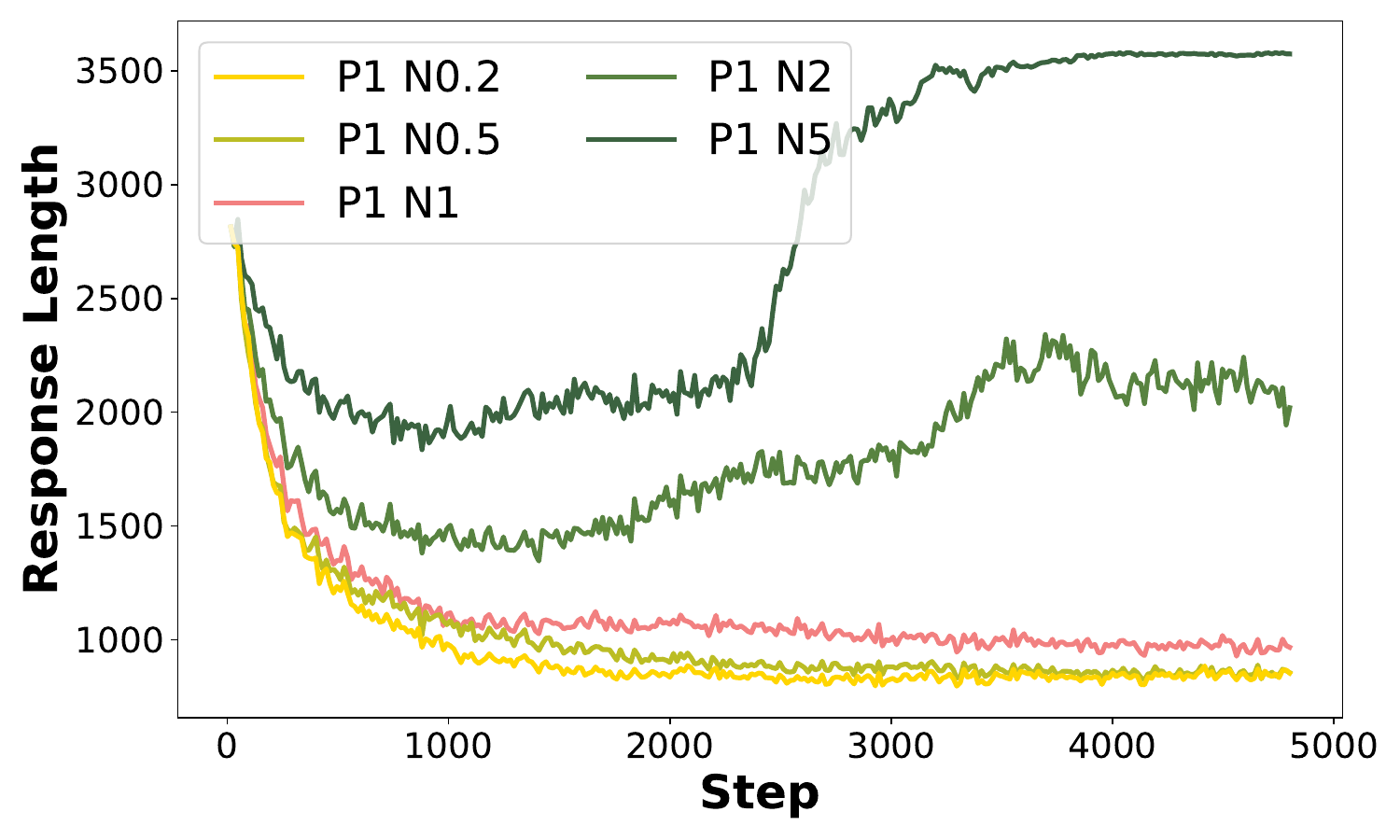}
        \caption{Length}
    \end{subfigure}
    \begin{subfigure}[b]{0.32\linewidth}
        \centering
        \includegraphics[width=\linewidth]{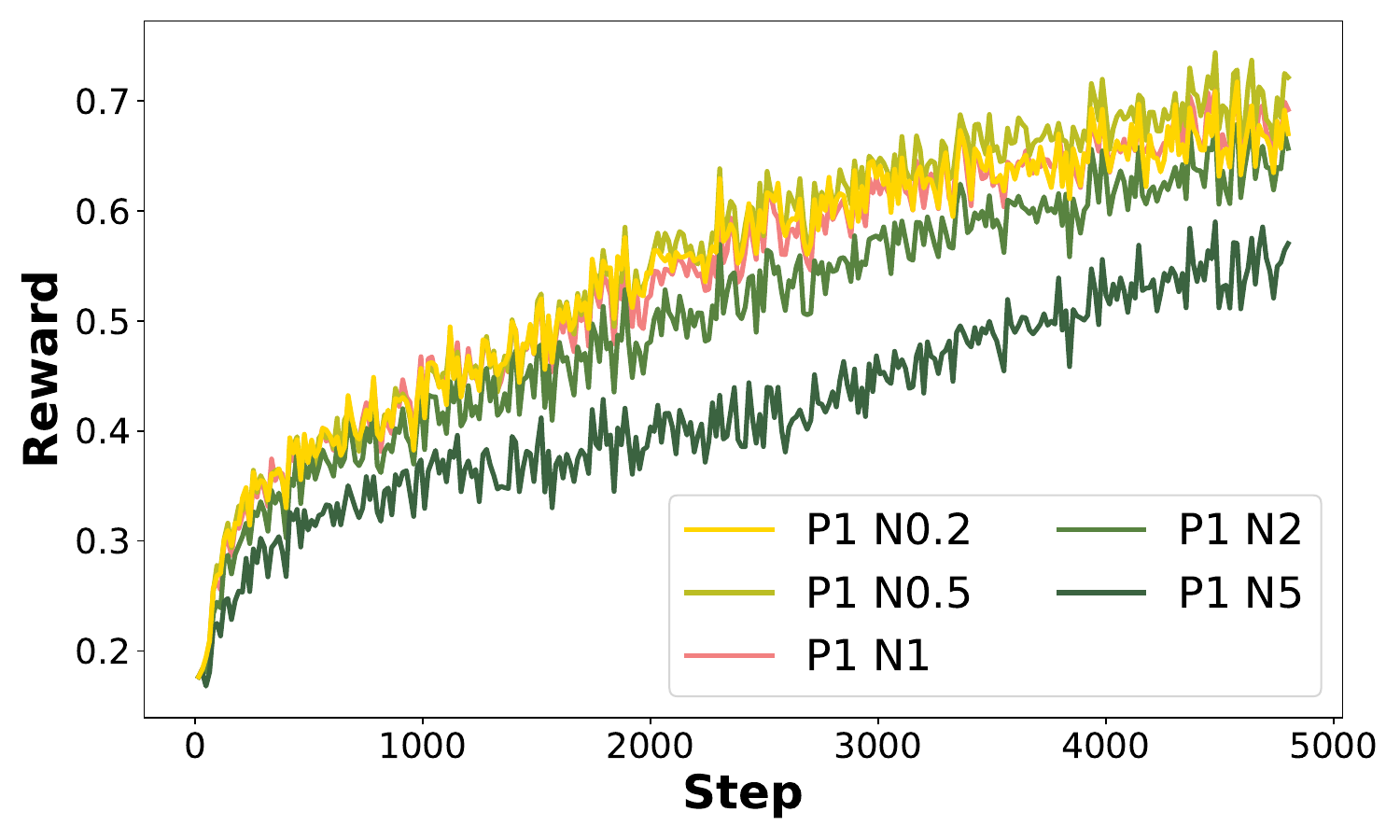}
        \caption{Reward}
    \end{subfigure}
    \begin{subfigure}[b]{0.32\linewidth}
        \centering
        \includegraphics[width=\linewidth]{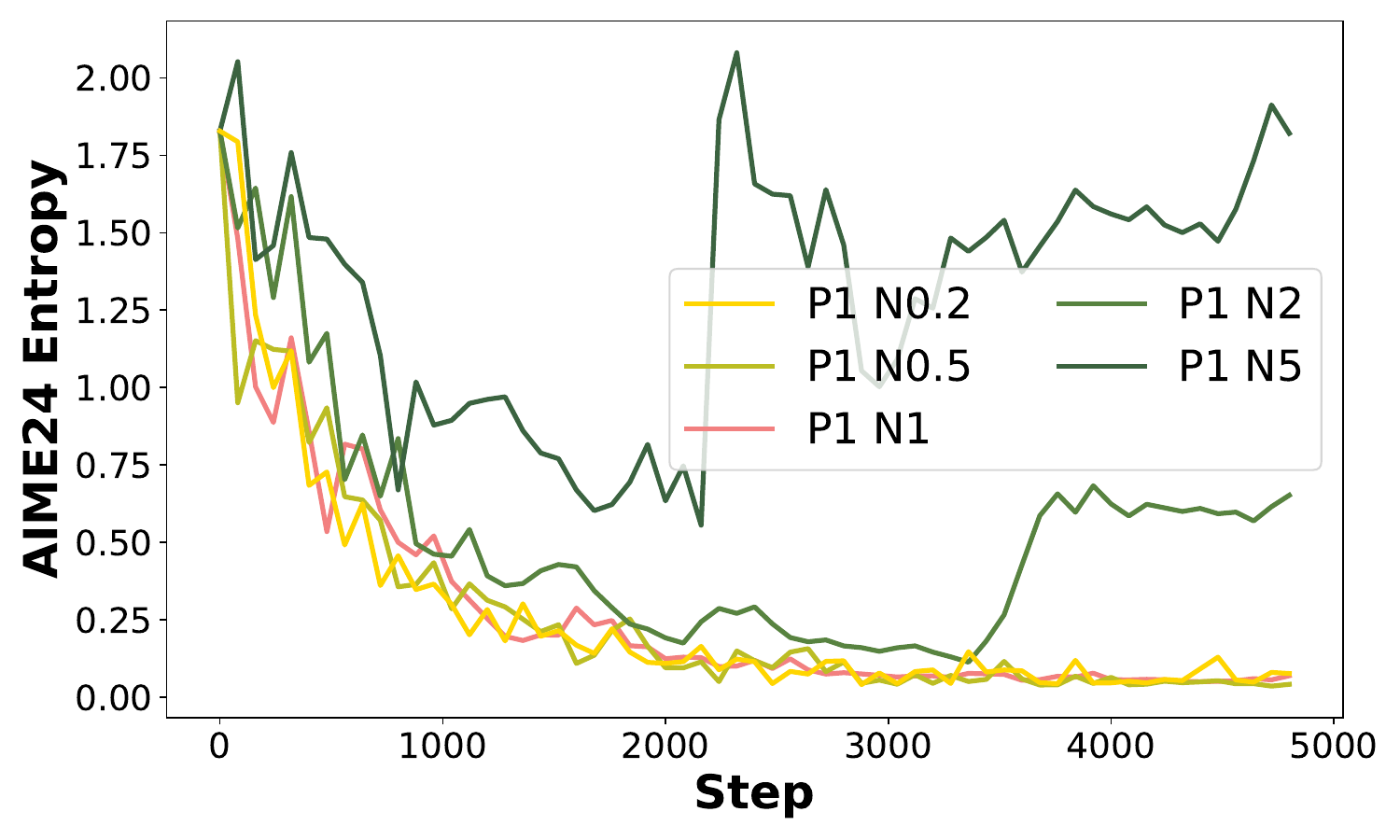}
        \caption{AIME24 Entropy}
    \end{subfigure}
    \begin{subfigure}[b]{0.32\linewidth}
        \centering
        \includegraphics[width=\linewidth]{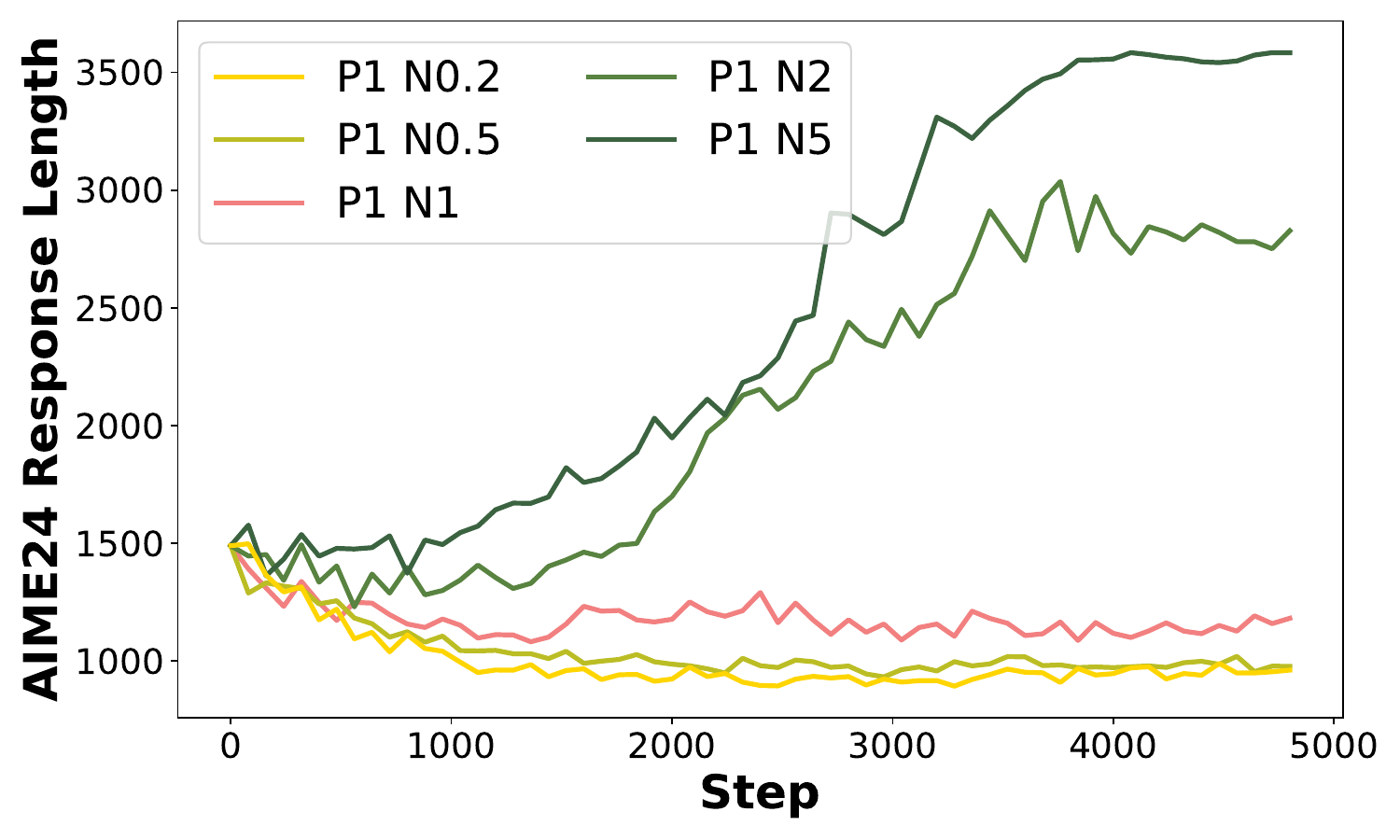}
        \caption{AIME24 Length}
    \end{subfigure}
    \begin{subfigure}[b]{0.32\linewidth}
        \centering
        \includegraphics[width=\linewidth]{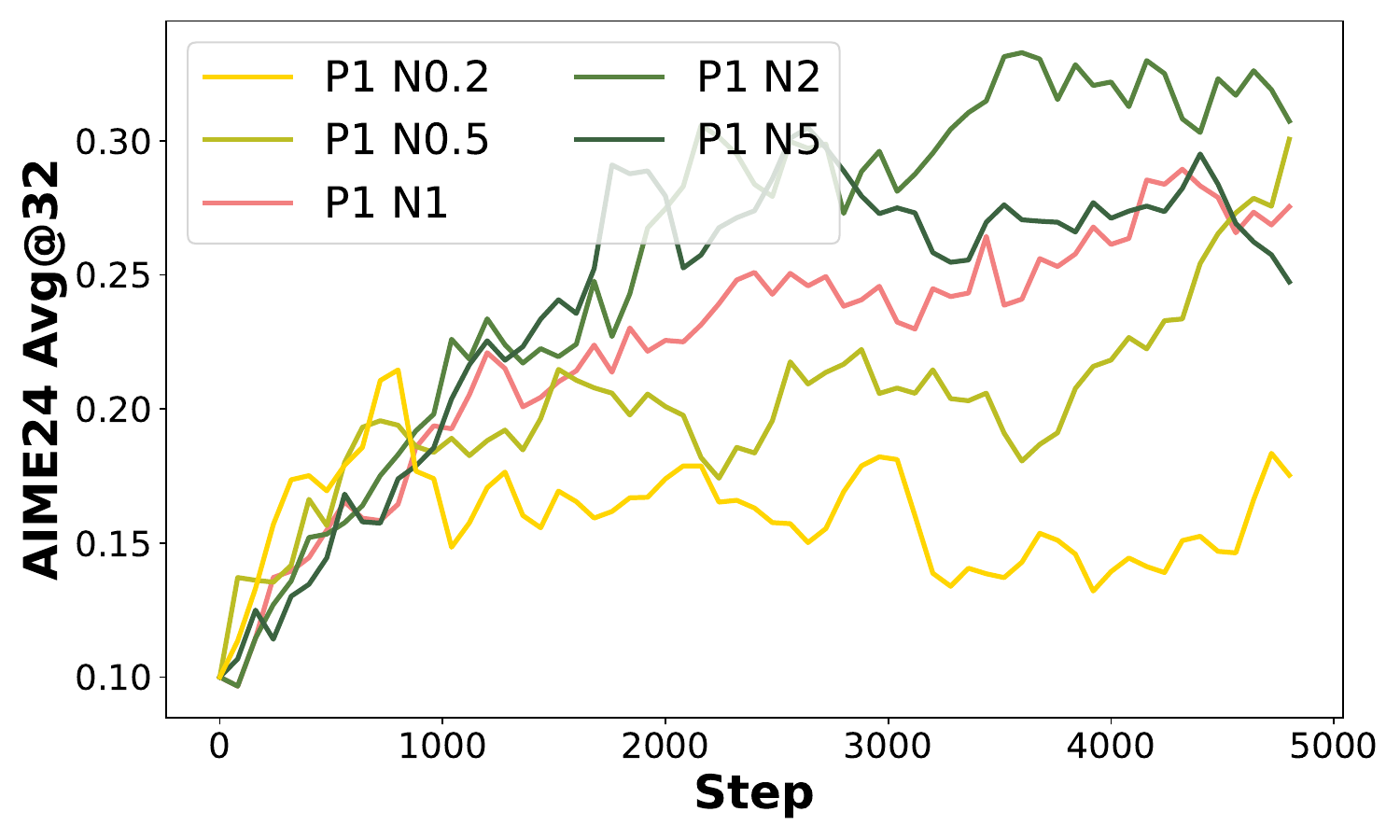}
        \caption{AIME24 Avg@32}
    \end{subfigure}
    \begin{subfigure}[b]{0.32\linewidth}
        \centering
        \includegraphics[width=\linewidth]{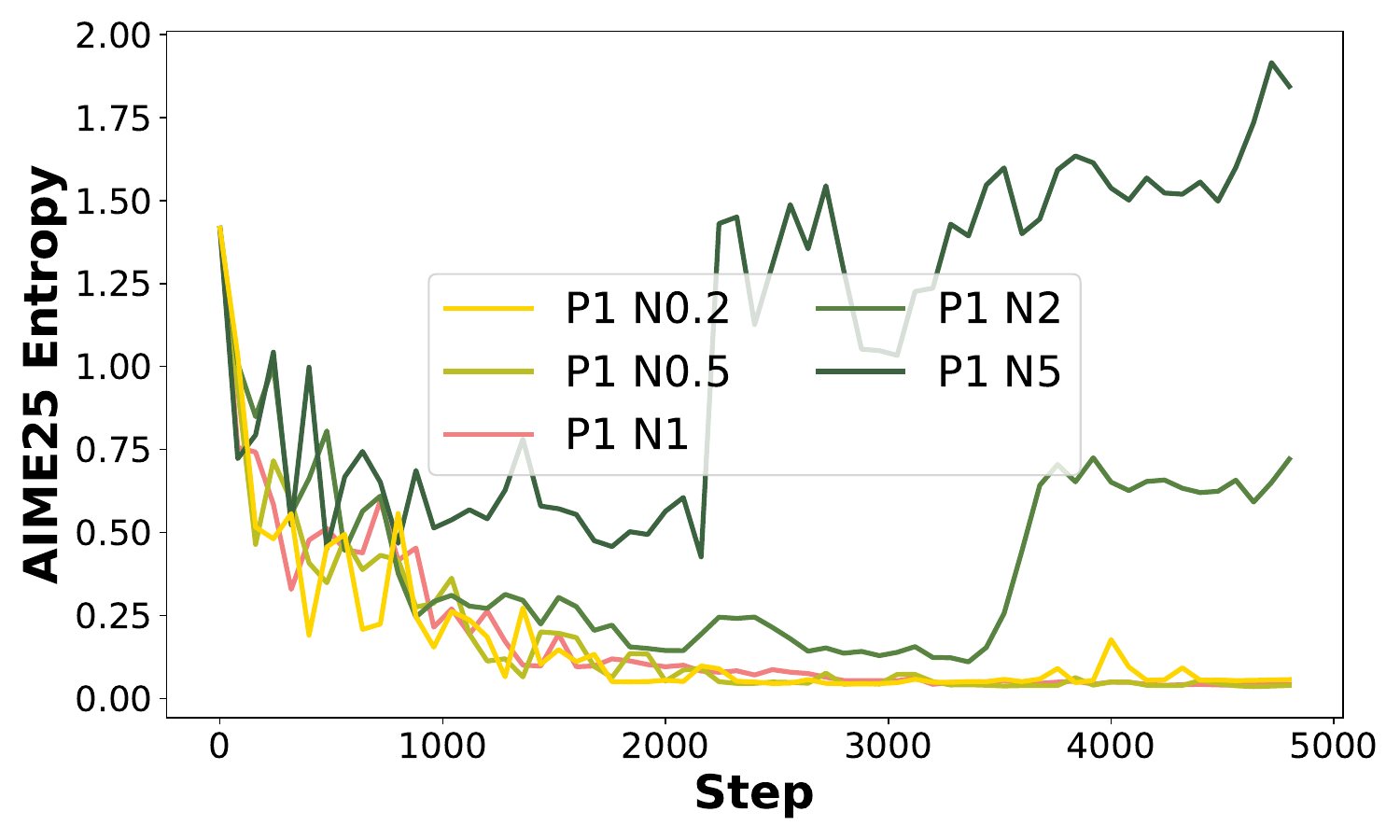}
        \caption{AIME25 Entropy}
    \end{subfigure}
    \begin{subfigure}[b]{0.32\linewidth}
        \centering
        \includegraphics[width=\linewidth]{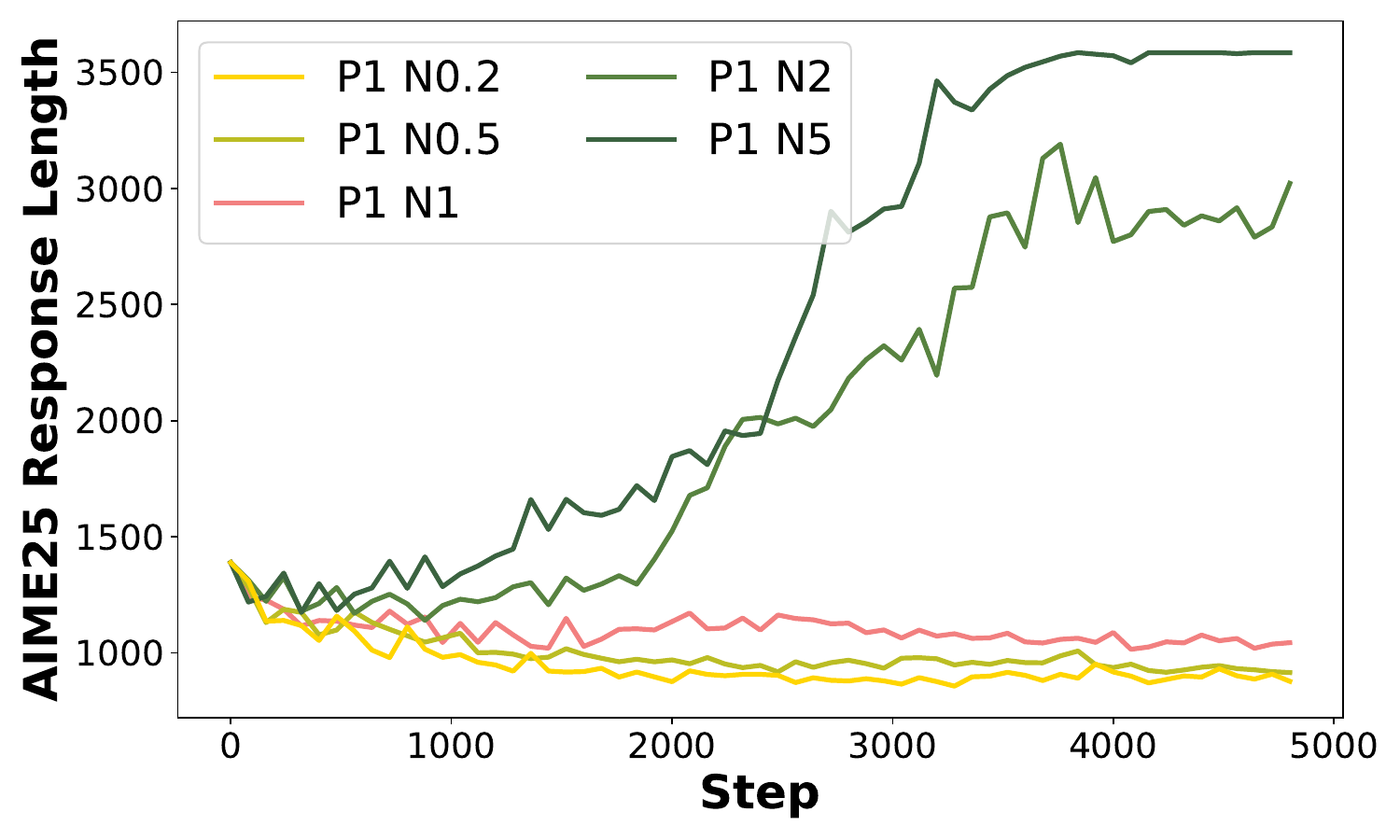}
        \caption{AIME25 Length}
    \end{subfigure}
    \begin{subfigure}[b]{0.32\linewidth}
        \centering
        \includegraphics[width=\linewidth]{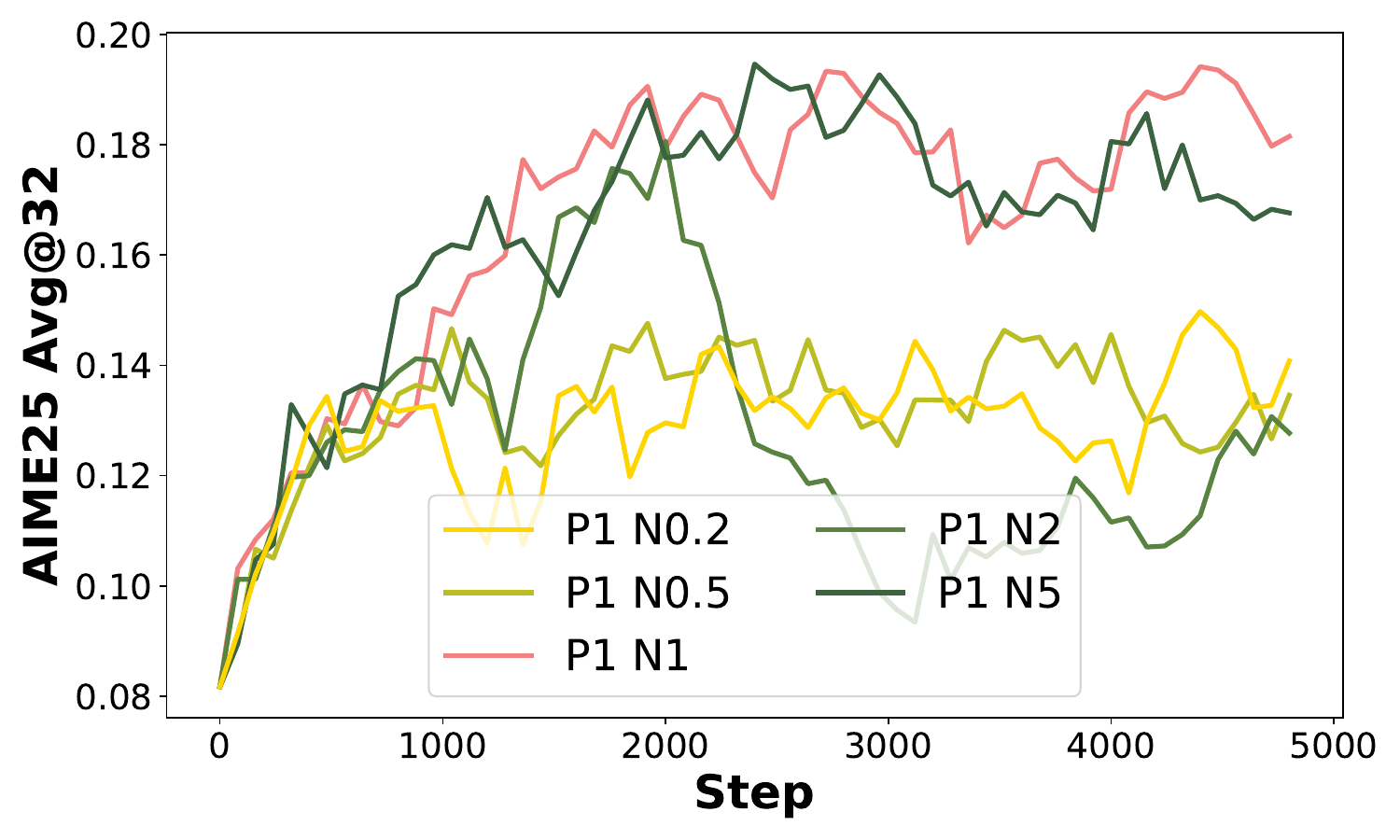}
        \caption{AIME25 Avg@32}
    \end{subfigure}
    \caption{Different training dynamics of polarity-level negative sample advantage shaping on Qwen2.5-7B-Math.}
\label{fig:polarity-nas-training_dynamic}
\end{figure*}

\begin{figure*}[t]
    \centering
    \begin{subfigure}[b]{0.32\linewidth}
        \centering
        \includegraphics[width=\linewidth]{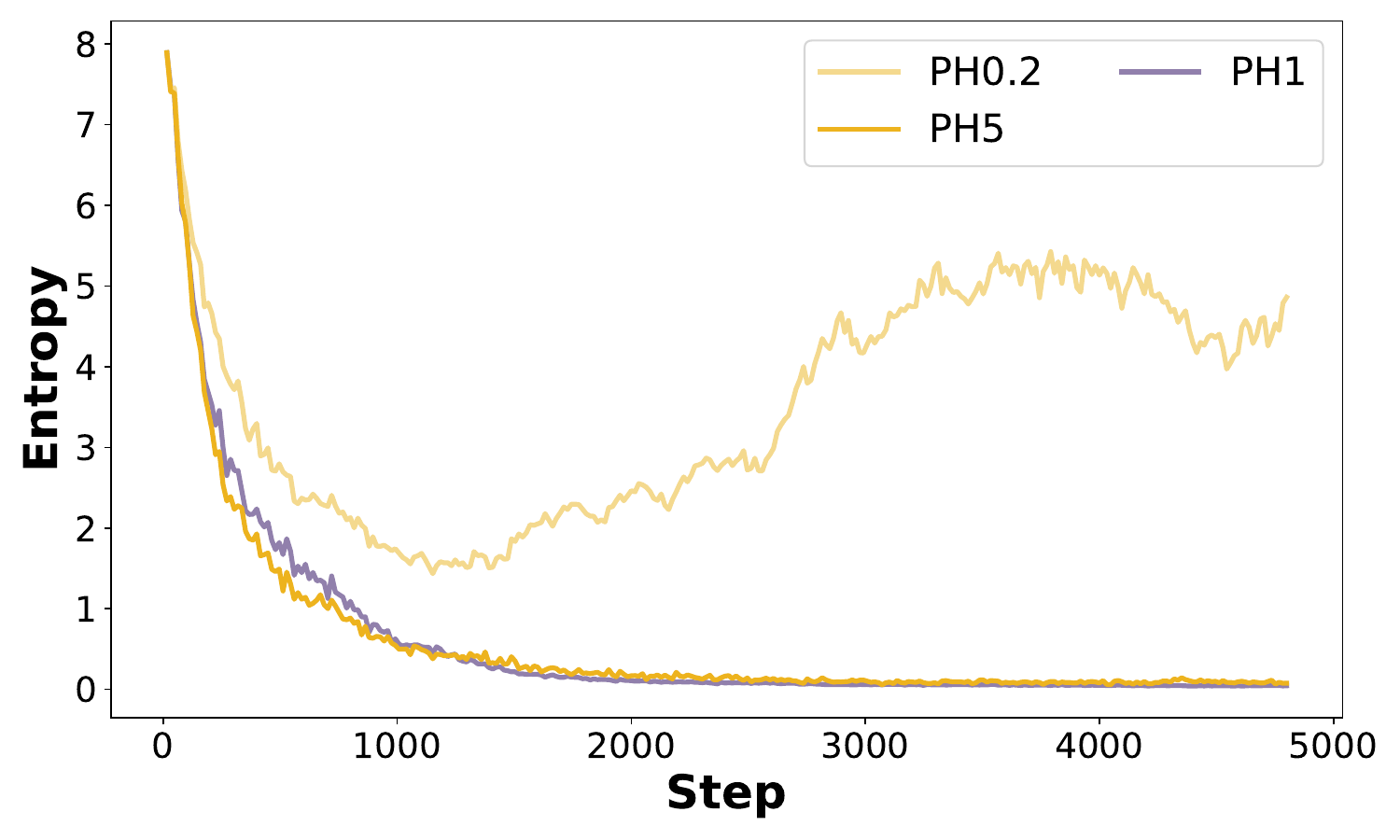}
        \caption{Entropy}
    \end{subfigure}
    \begin{subfigure}[b]{0.32\linewidth}
        \centering
        \includegraphics[width=\linewidth]{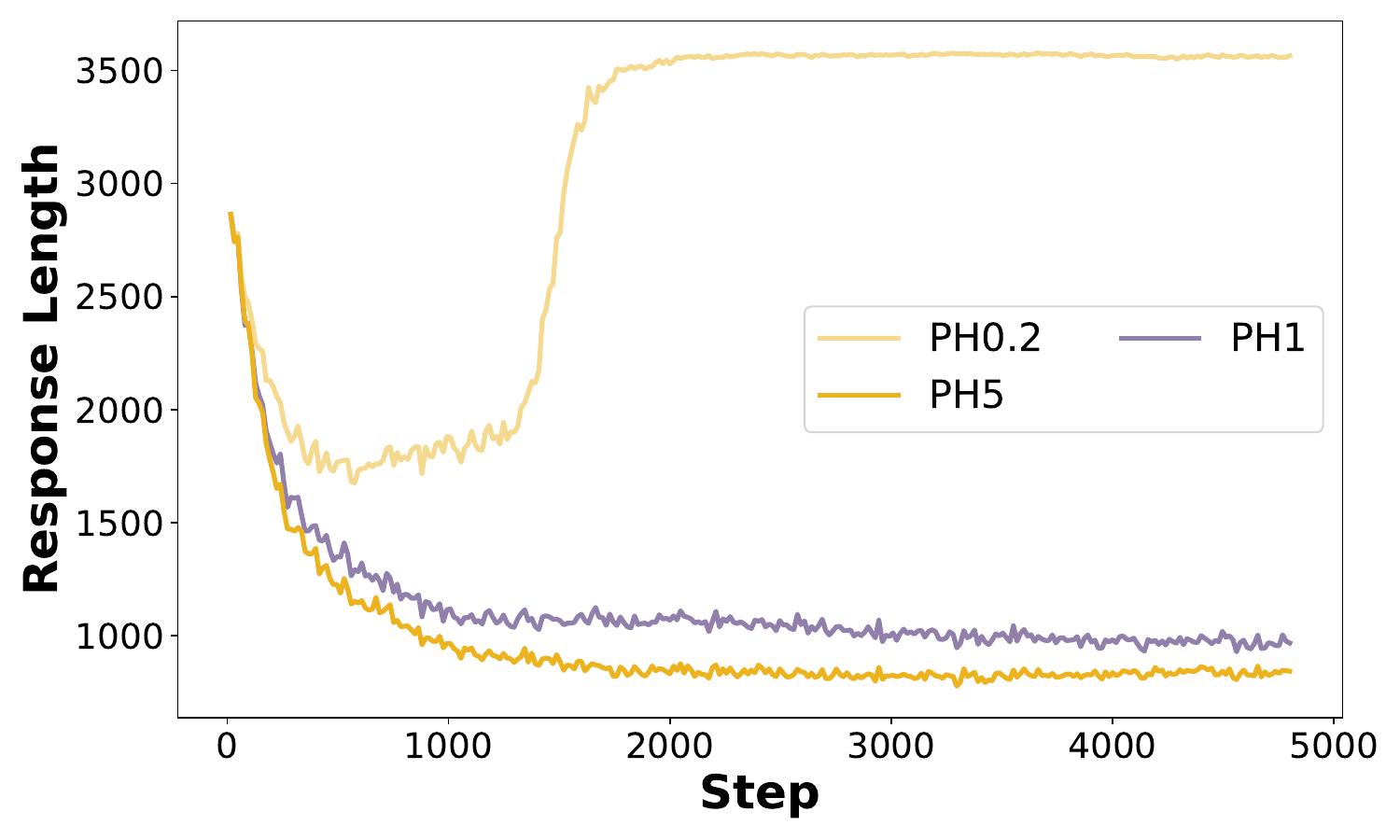}
        \caption{Length}
    \end{subfigure}
    \begin{subfigure}[b]{0.32\linewidth}
        \centering
        \includegraphics[width=\linewidth]{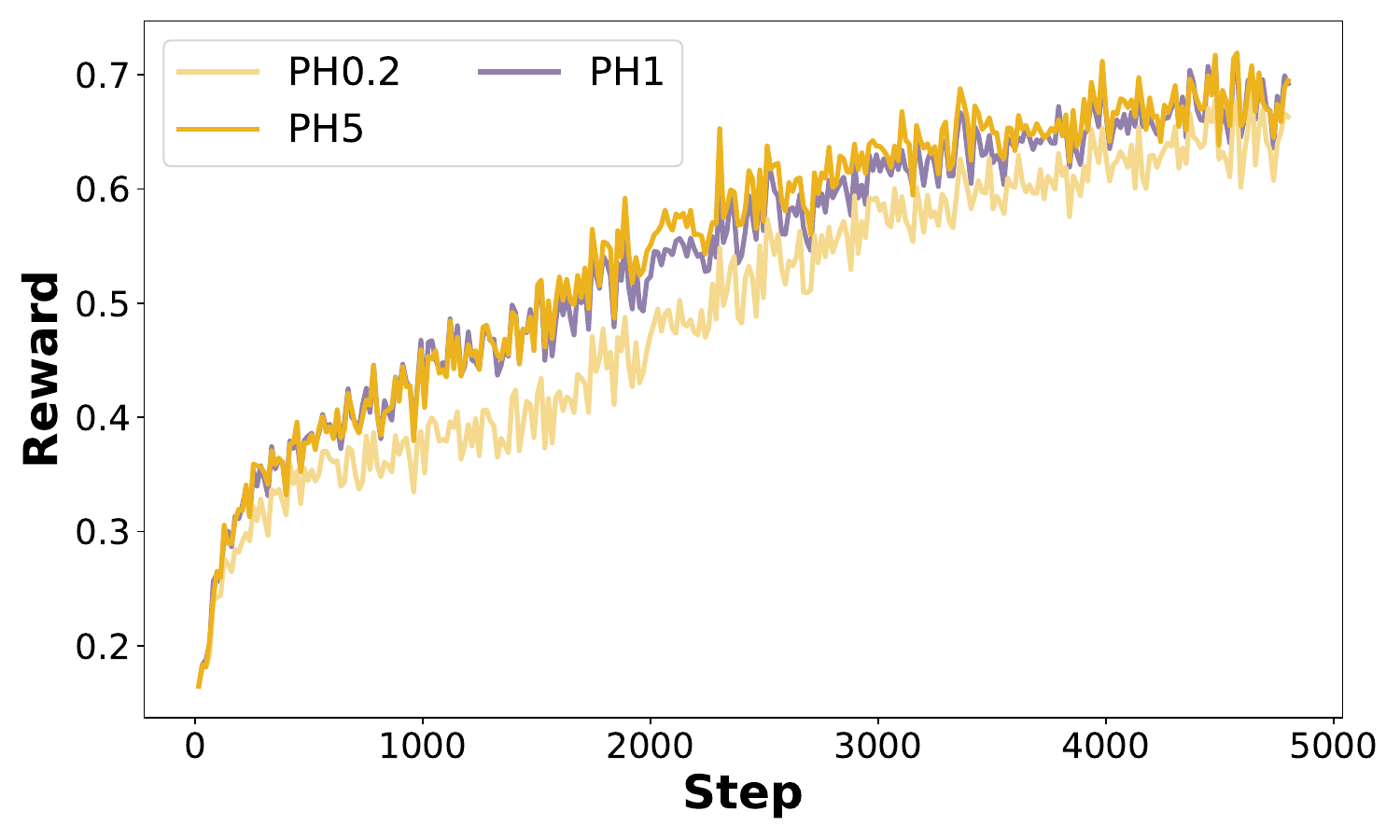}
        \caption{Reward}
    \end{subfigure}
    \begin{subfigure}[b]{0.32\linewidth}
        \centering
        \includegraphics[width=\linewidth]{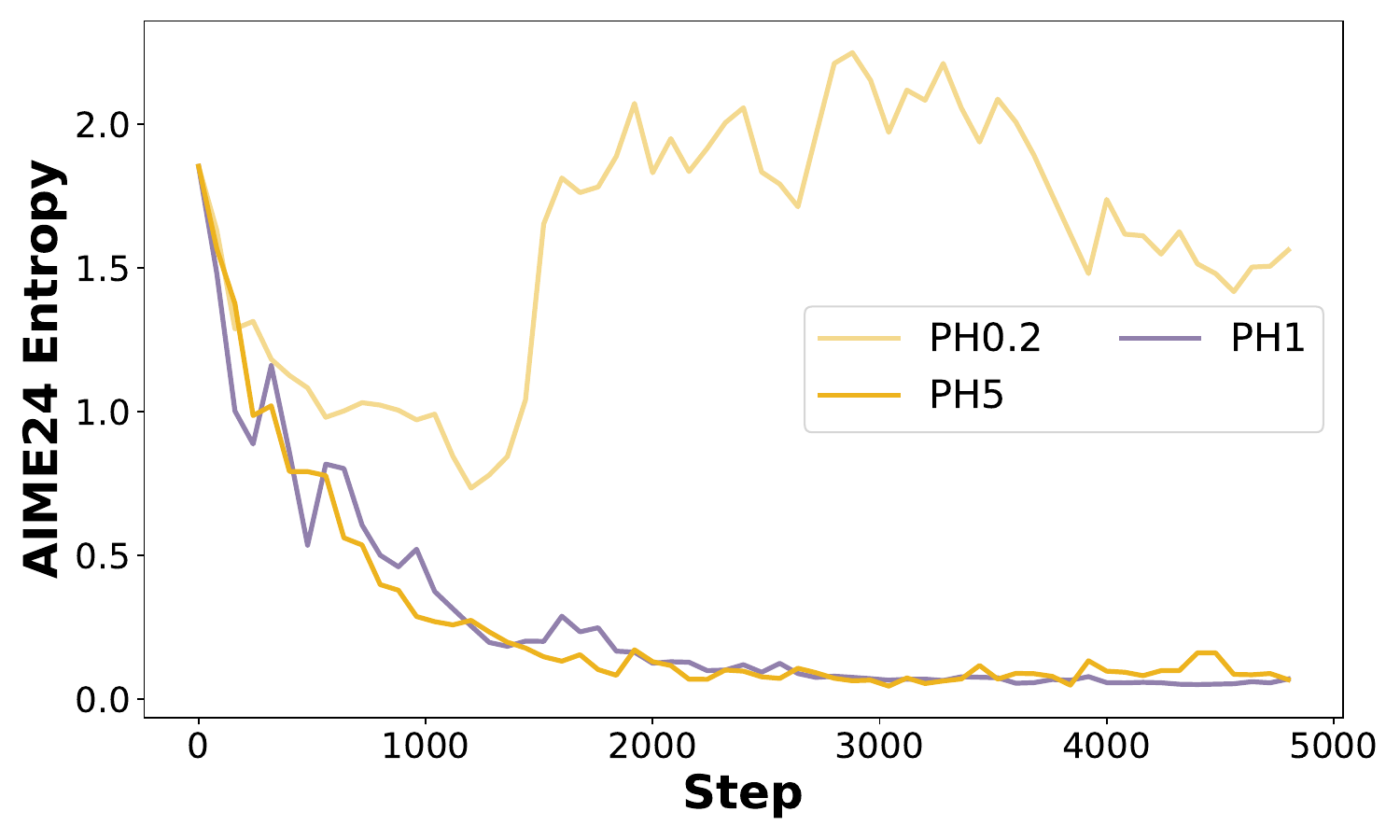}
        \caption{AIME24 Entropy}
    \end{subfigure}
    \begin{subfigure}[b]{0.32\linewidth}
        \centering
        \includegraphics[width=\linewidth]{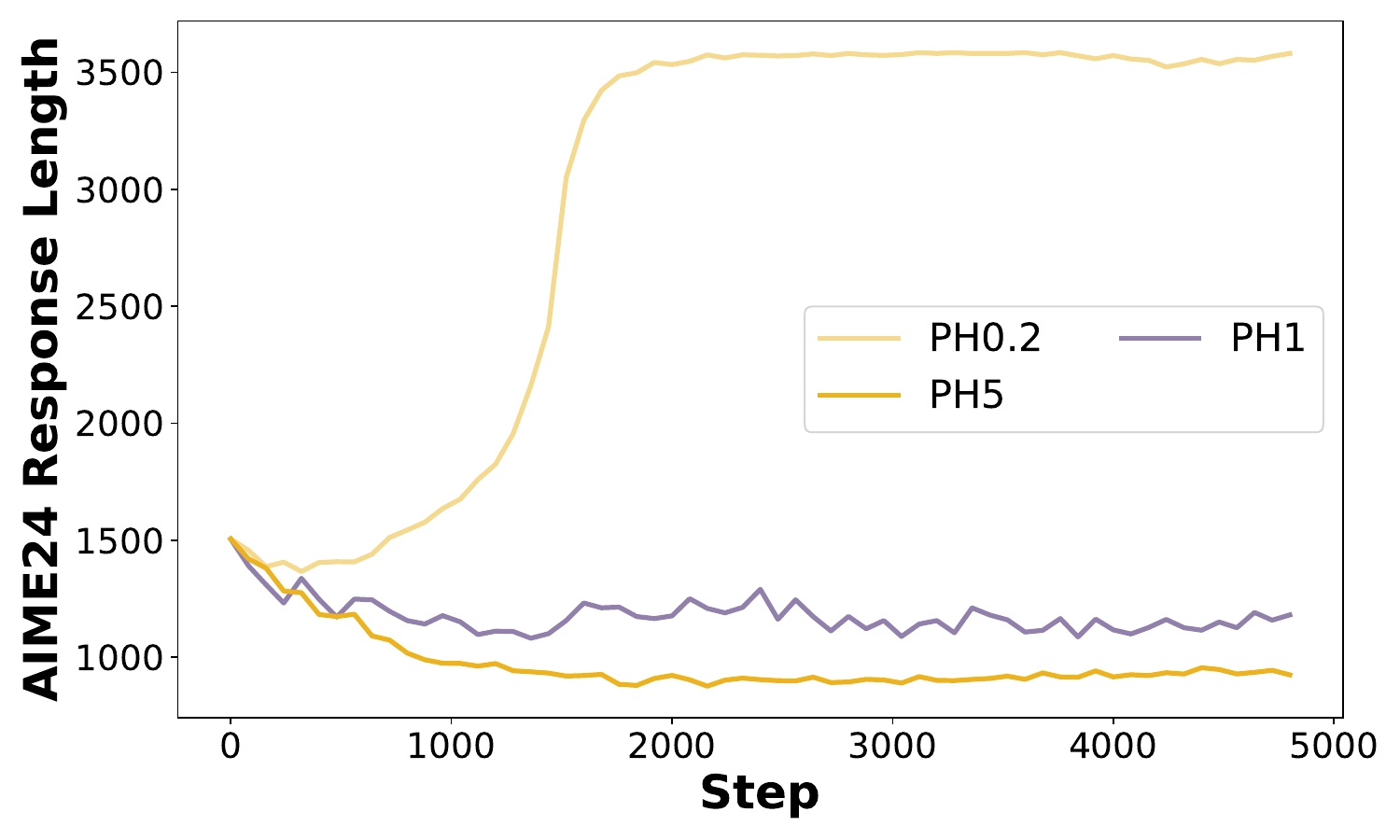}
        \caption{AIME24 Length}
    \end{subfigure}
    \begin{subfigure}[b]{0.32\linewidth}
        \centering
        \includegraphics[width=\linewidth]{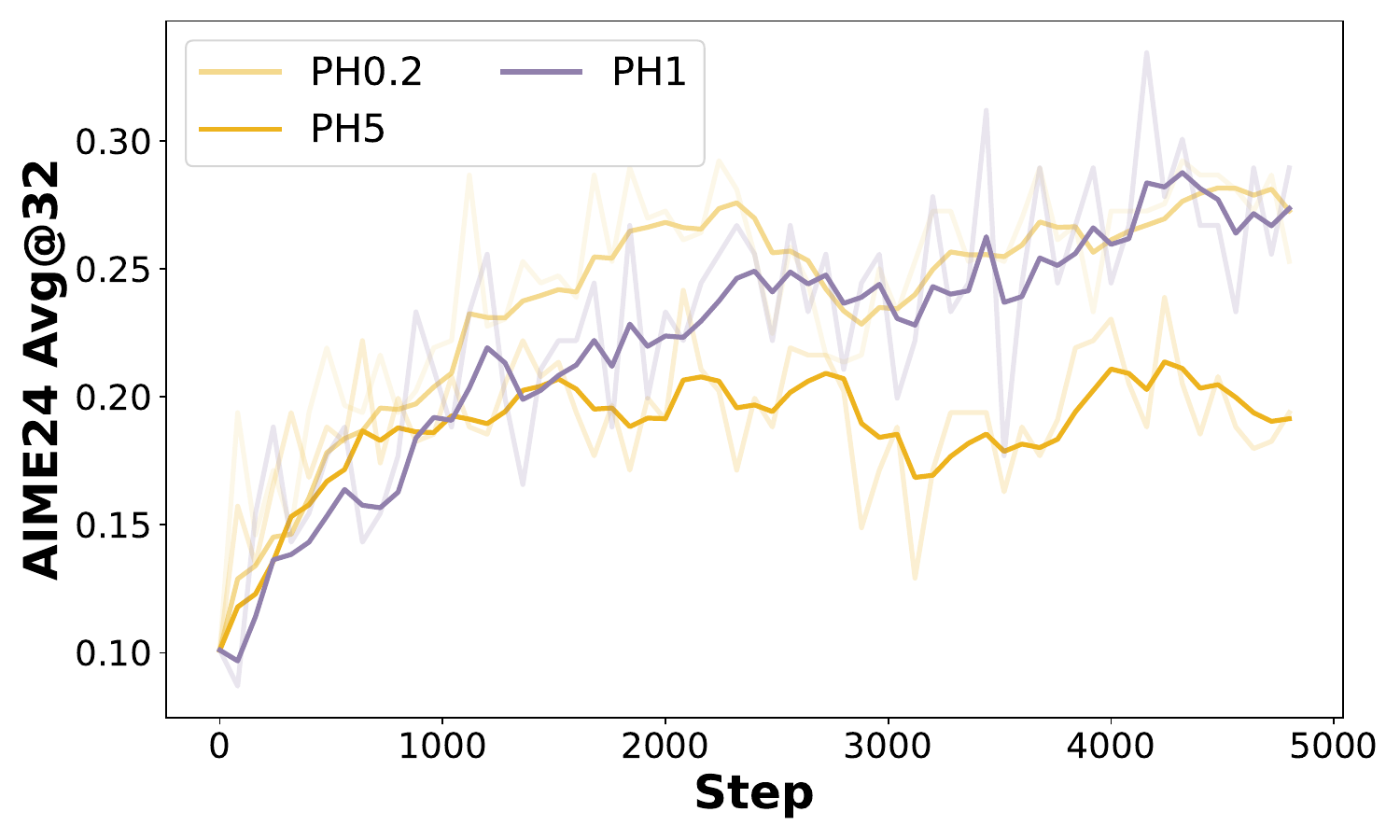}
        \caption{AIME24 Avg@32}
    \end{subfigure}
    \begin{subfigure}[b]{0.32\linewidth}
        \centering
        \includegraphics[width=\linewidth]{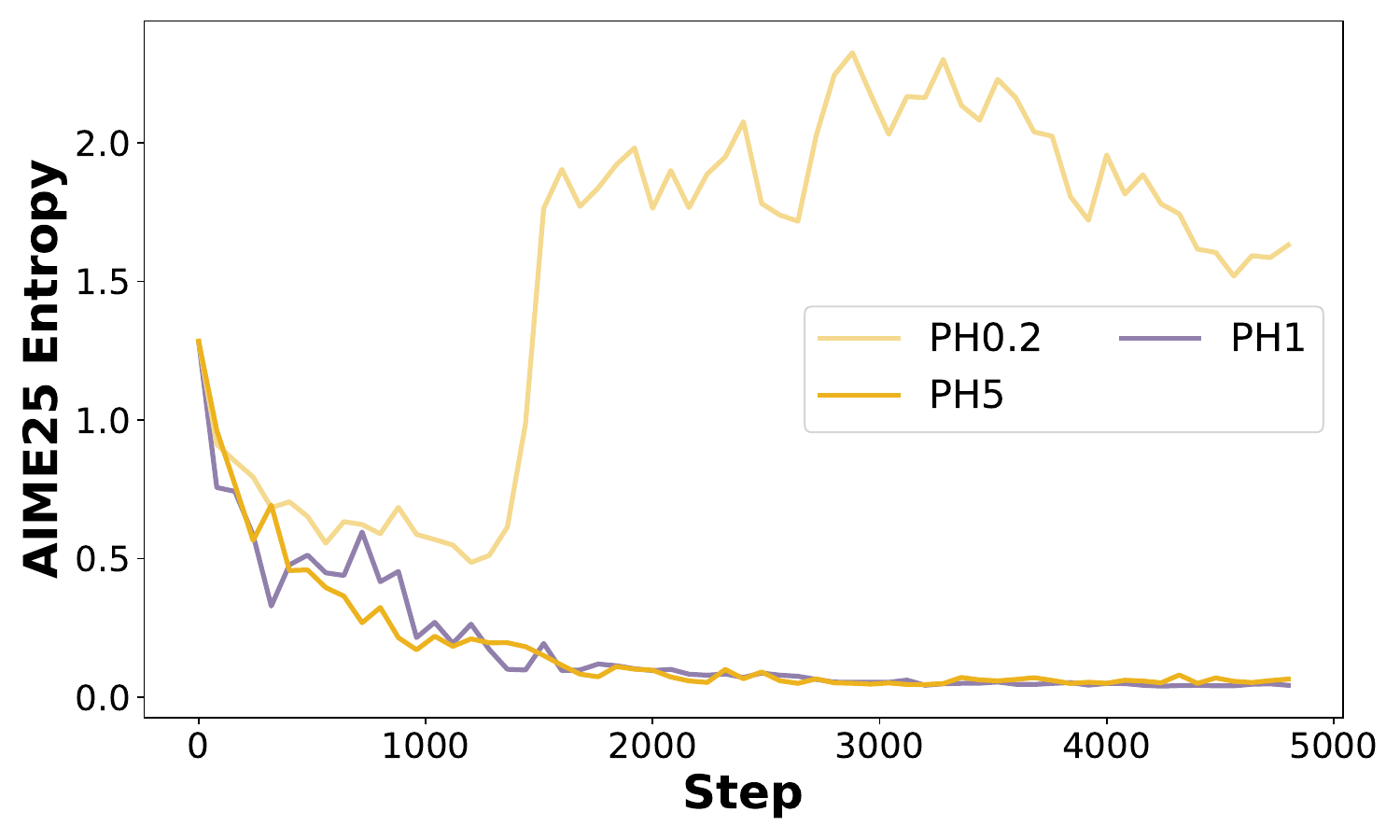}
        \caption{AIME25 Entropy}
    \end{subfigure}
    \begin{subfigure}[b]{0.32\linewidth}
        \centering
        \includegraphics[width=\linewidth]{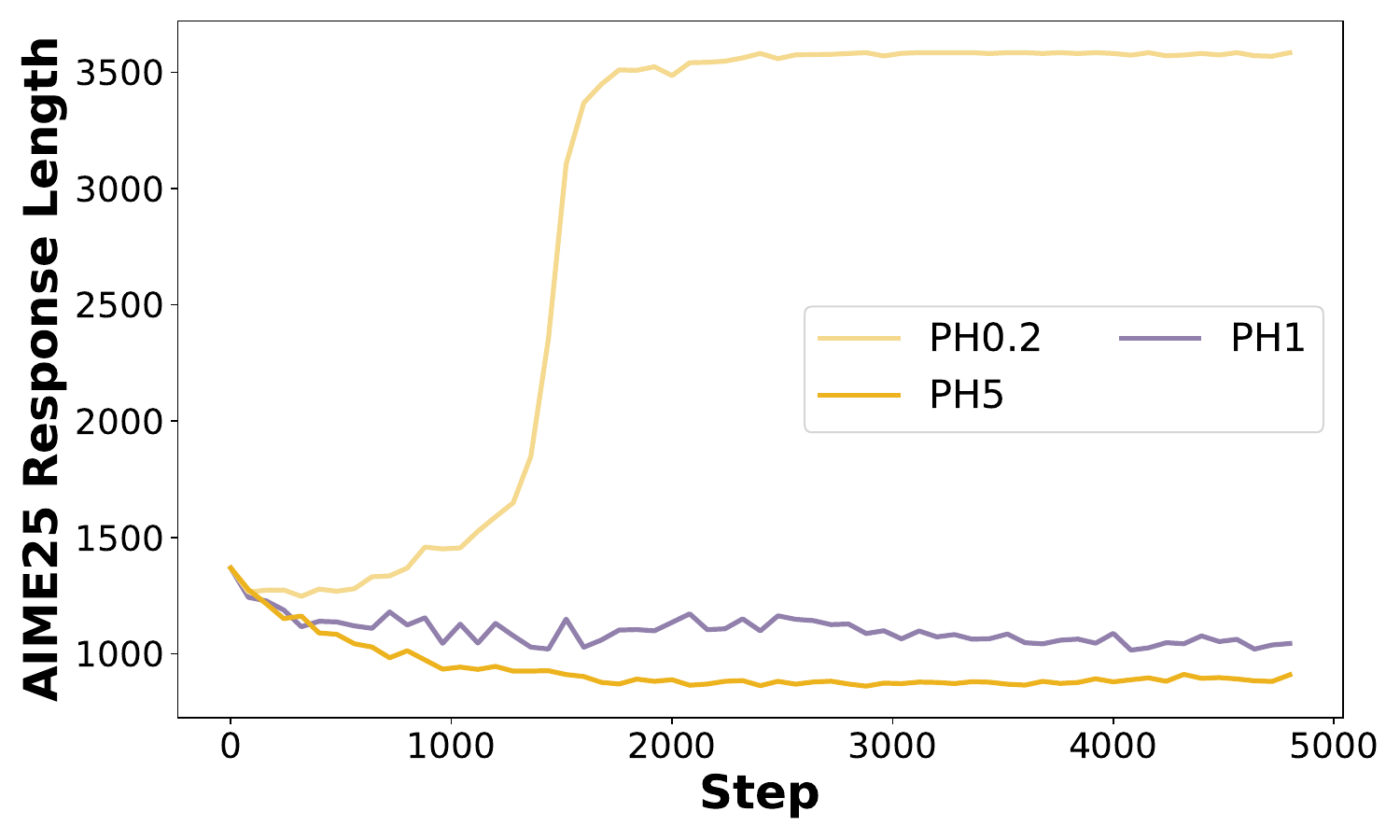}
        \caption{AIME25 Length}
    \end{subfigure}
    \begin{subfigure}[b]{0.32\linewidth}
        \centering
        \includegraphics[width=\linewidth]{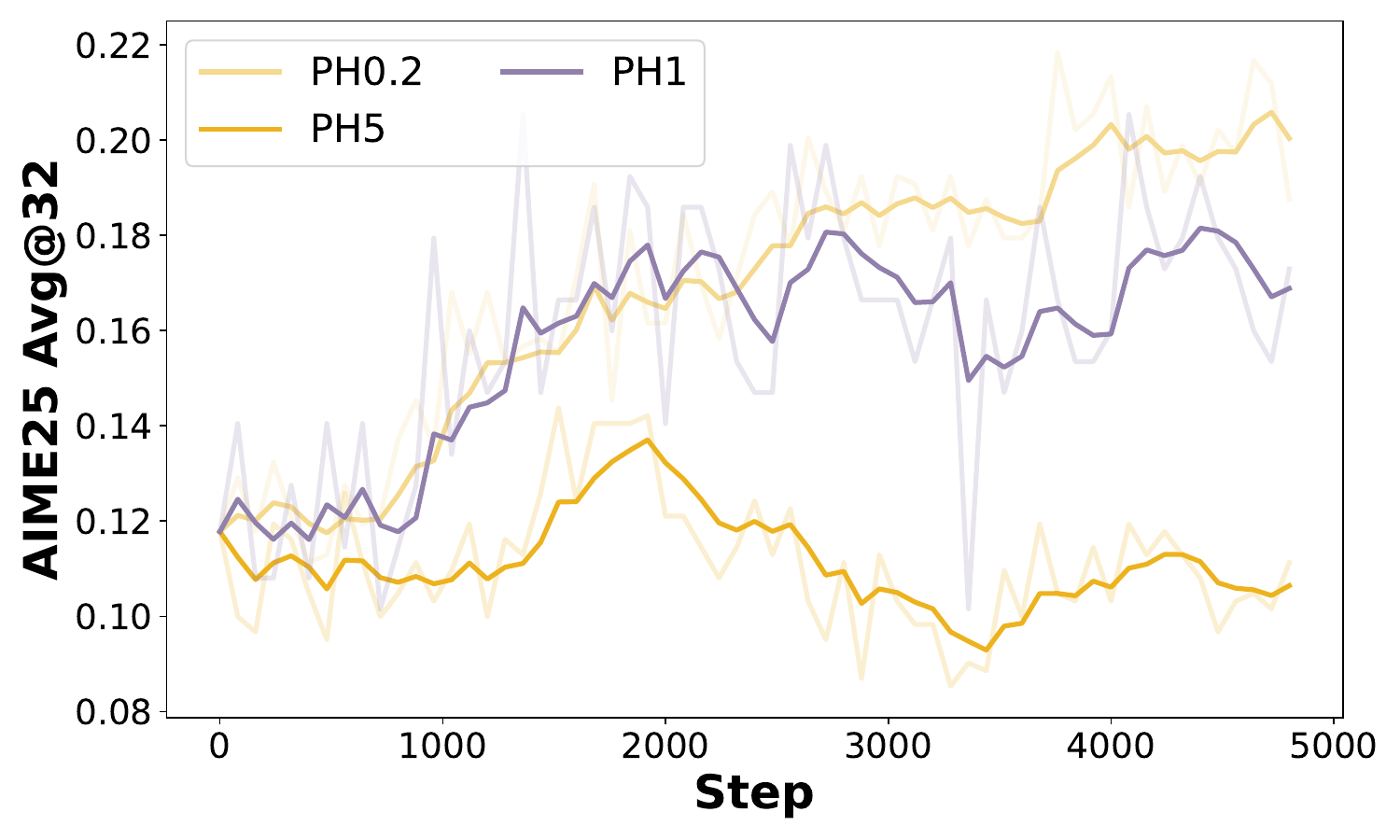}
        \caption{AIME25 Avg@32}
    \end{subfigure}
    \caption{RLVR training dynamics on positive high entropy token advantage shaping.}
\label{fig:token-entropy-ph-training_dynamic}
\end{figure*}

\begin{figure*}[t]
    \centering
    \begin{subfigure}[b]{0.32\linewidth}
        \centering
        \includegraphics[width=\linewidth]{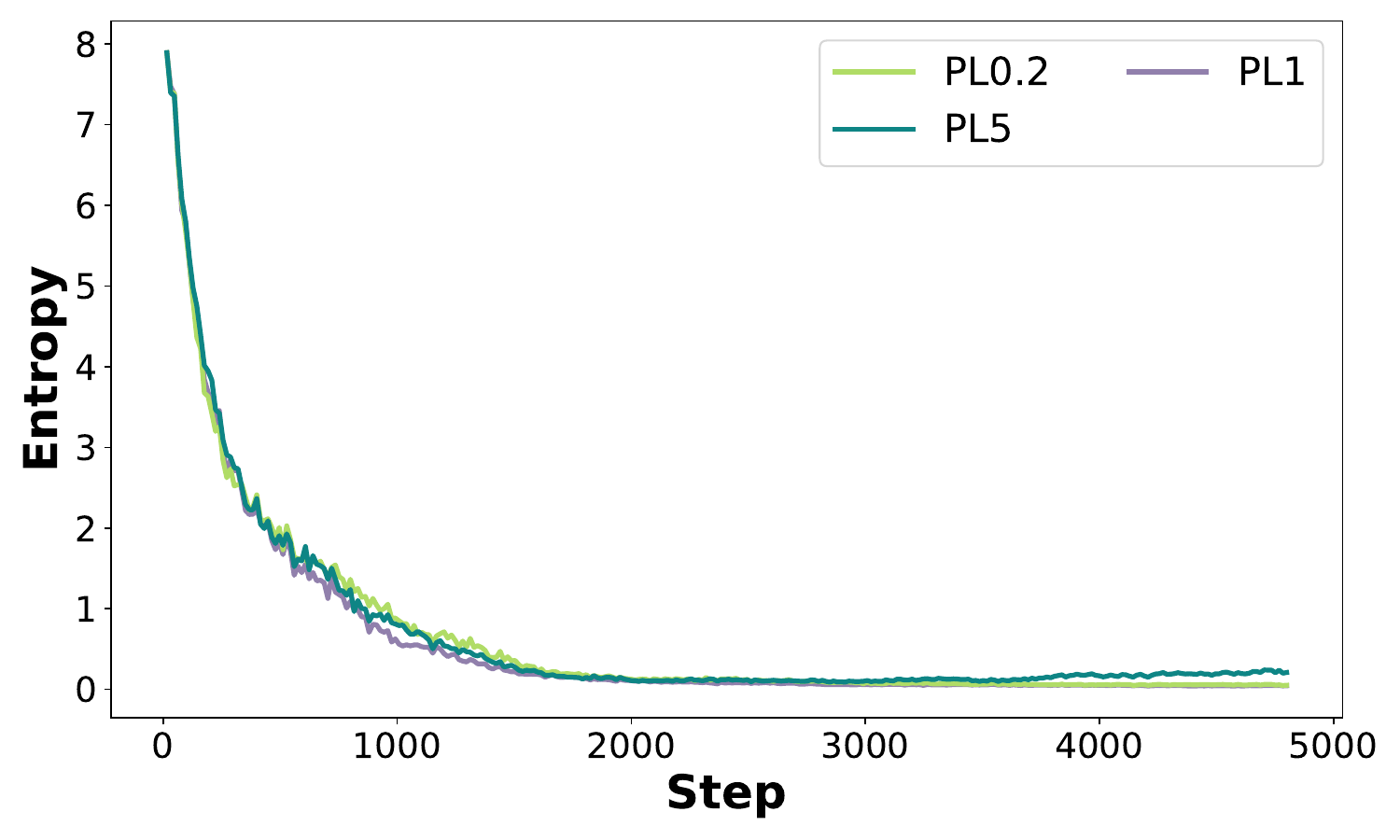}
        \caption{Entropy}
    \end{subfigure}
    \begin{subfigure}[b]{0.32\linewidth}
        \centering
        \includegraphics[width=\linewidth]{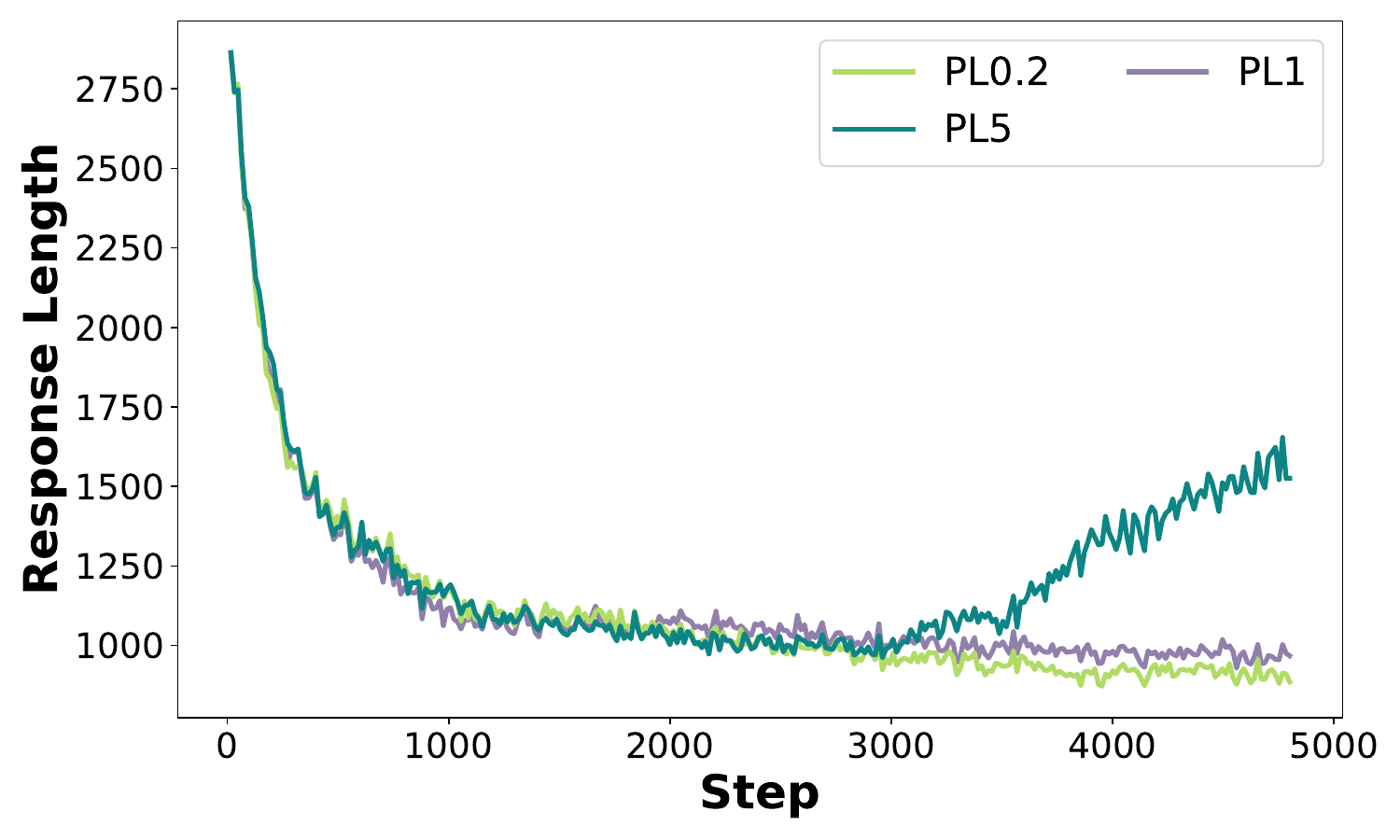}
        \caption{Length}
    \end{subfigure}
    \begin{subfigure}[b]{0.32\linewidth}
        \centering
        \includegraphics[width=\linewidth]{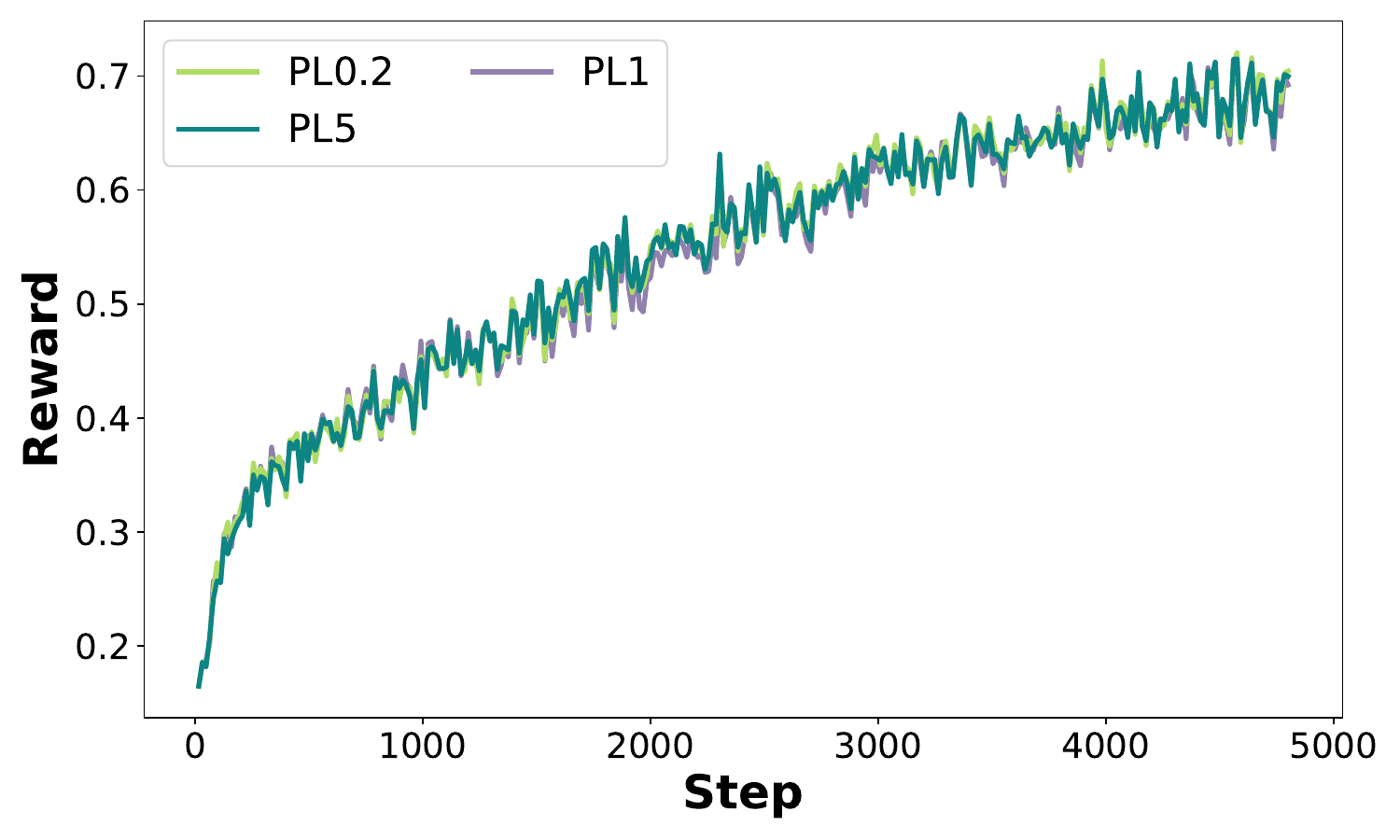}
        \caption{Reward}
    \end{subfigure}
    \begin{subfigure}[b]{0.32\linewidth}
        \centering
        \includegraphics[width=\linewidth]{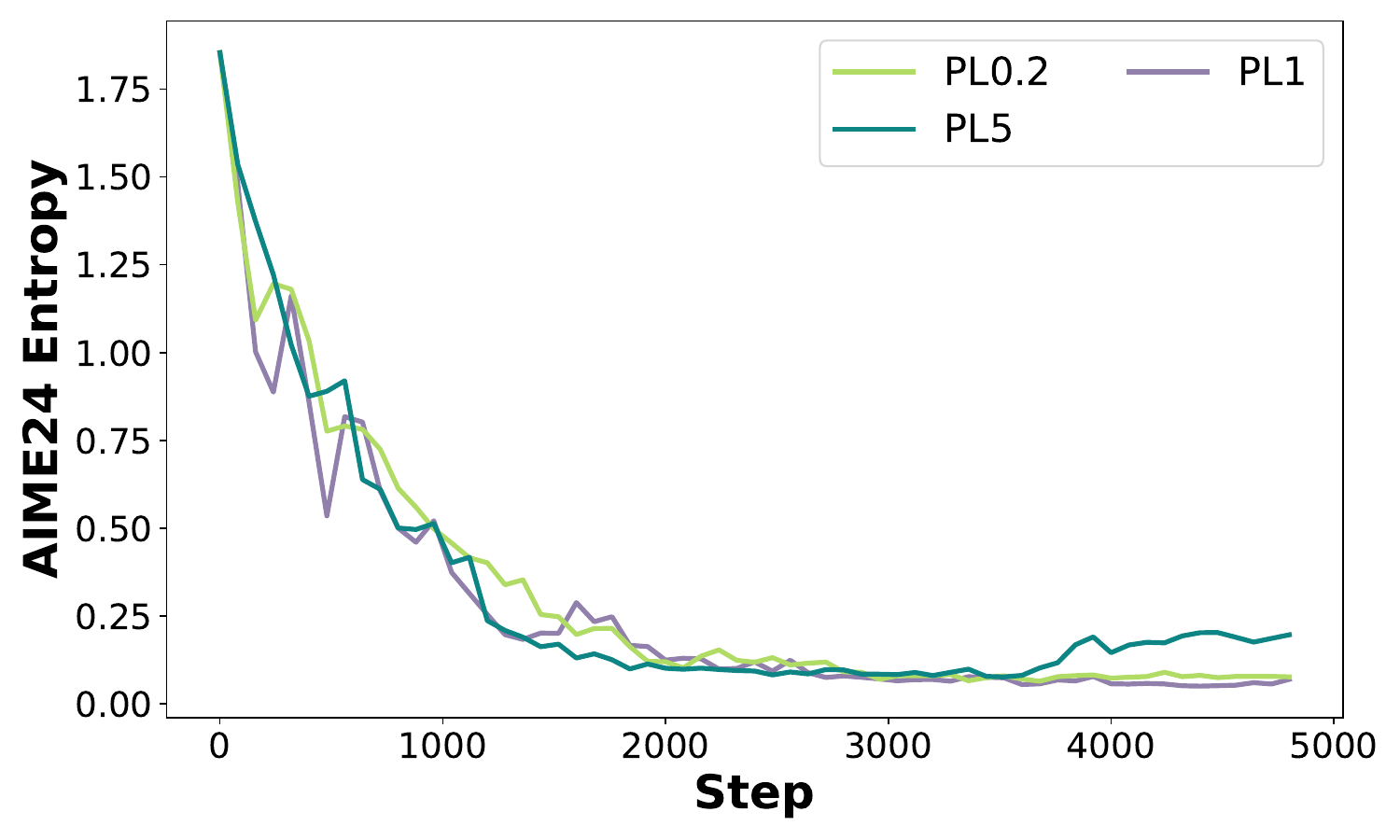}
        \caption{AIME24 Entropy}
    \end{subfigure}
    \begin{subfigure}[b]{0.32\linewidth}
        \centering
        \includegraphics[width=\linewidth]{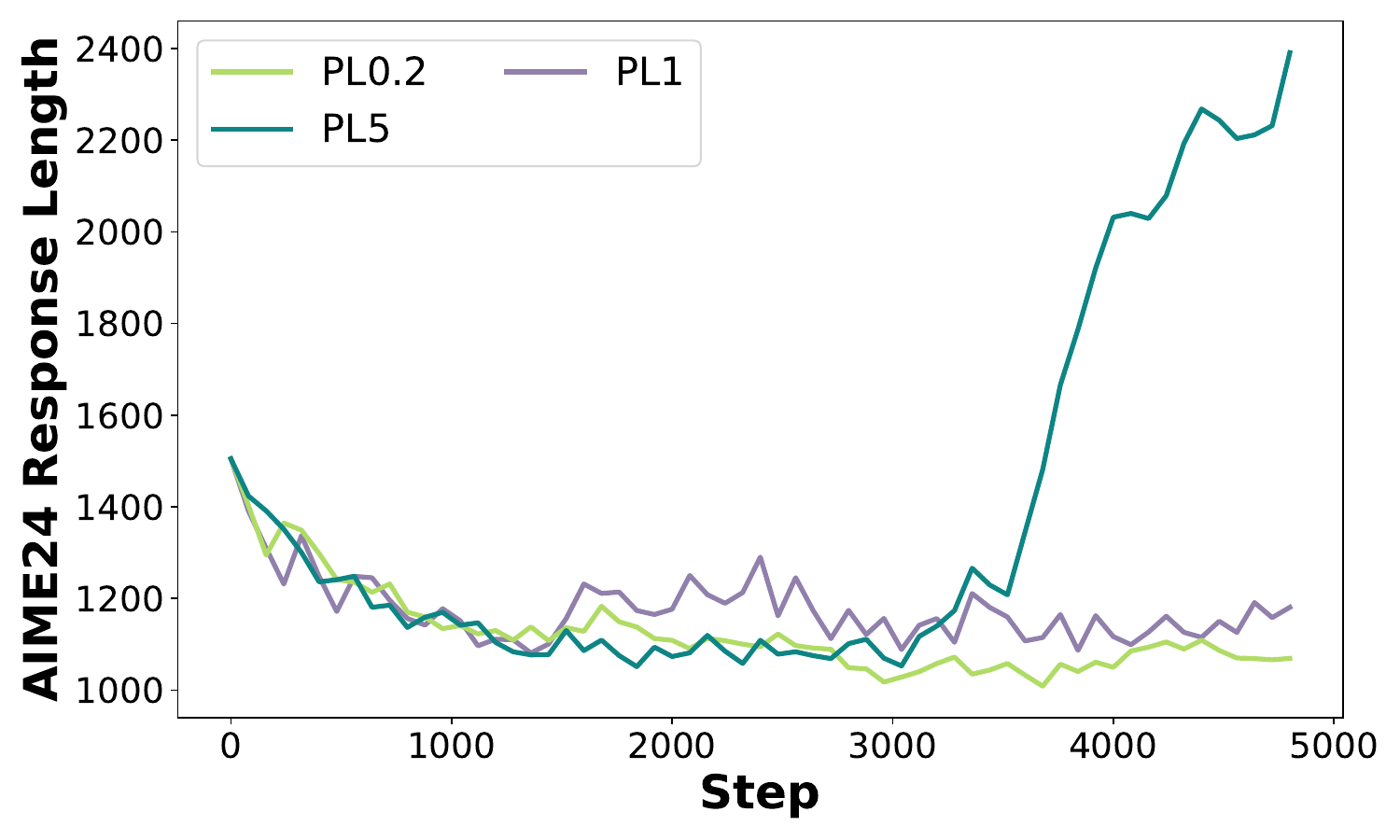}
        \caption{AIME24 Length}
    \end{subfigure}
    \begin{subfigure}[b]{0.32\linewidth}
        \centering
        \includegraphics[width=\linewidth]{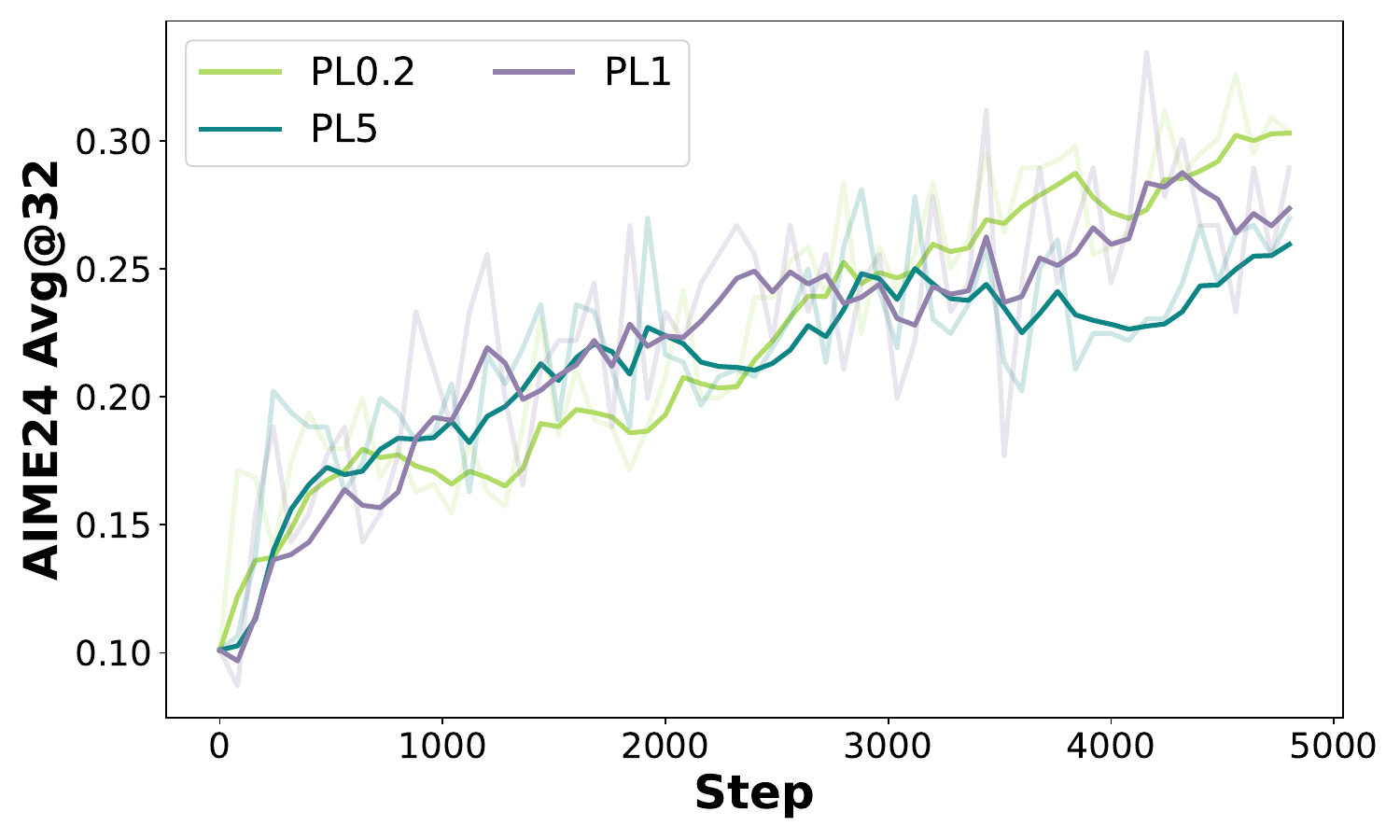}
        \caption{AIME24 Avg@32}
    \end{subfigure}
    \begin{subfigure}[b]{0.32\linewidth}
        \centering
        \includegraphics[width=\linewidth]{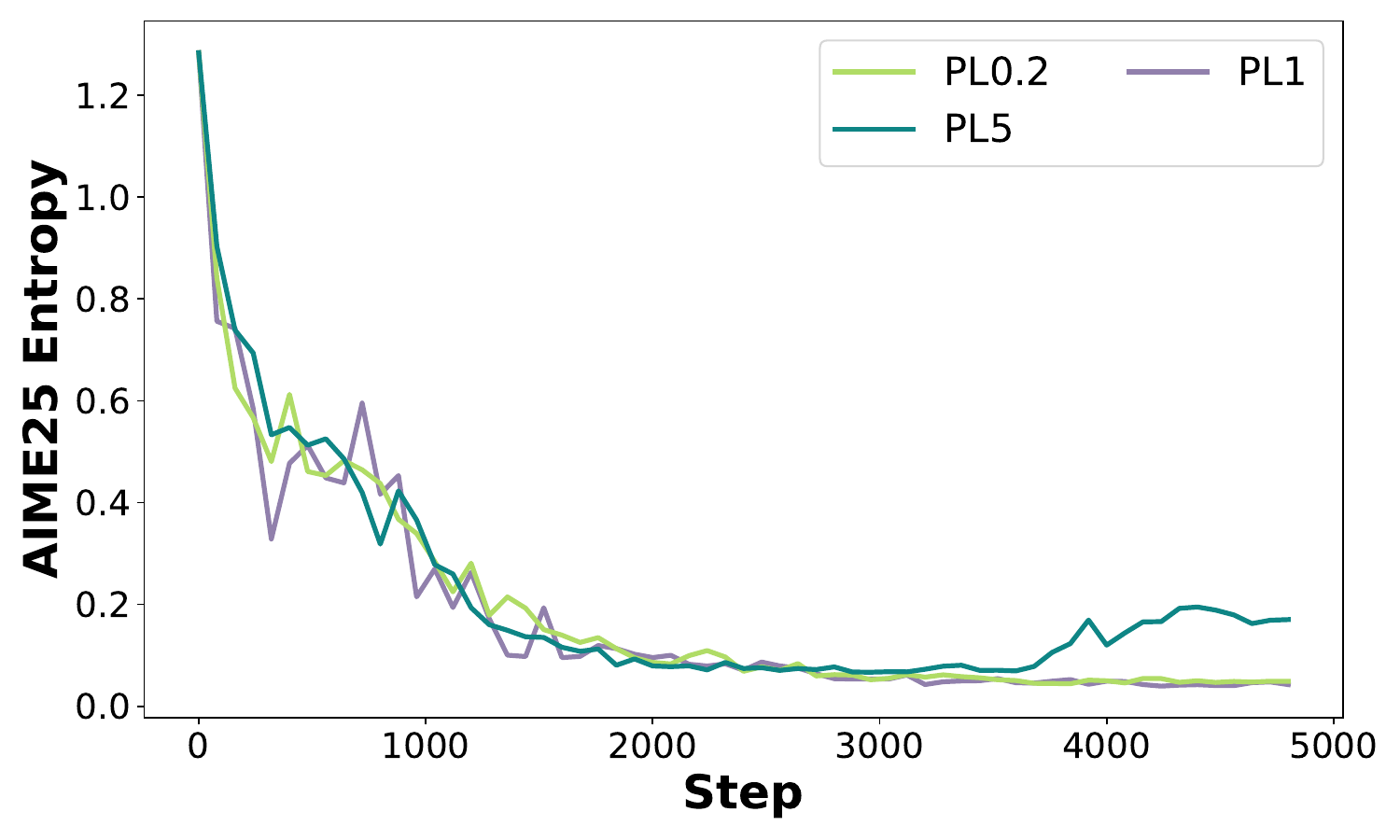}
        \caption{AIME25 Entropy}
    \end{subfigure}
    \begin{subfigure}[b]{0.32\linewidth}
        \centering
        \includegraphics[width=\linewidth]{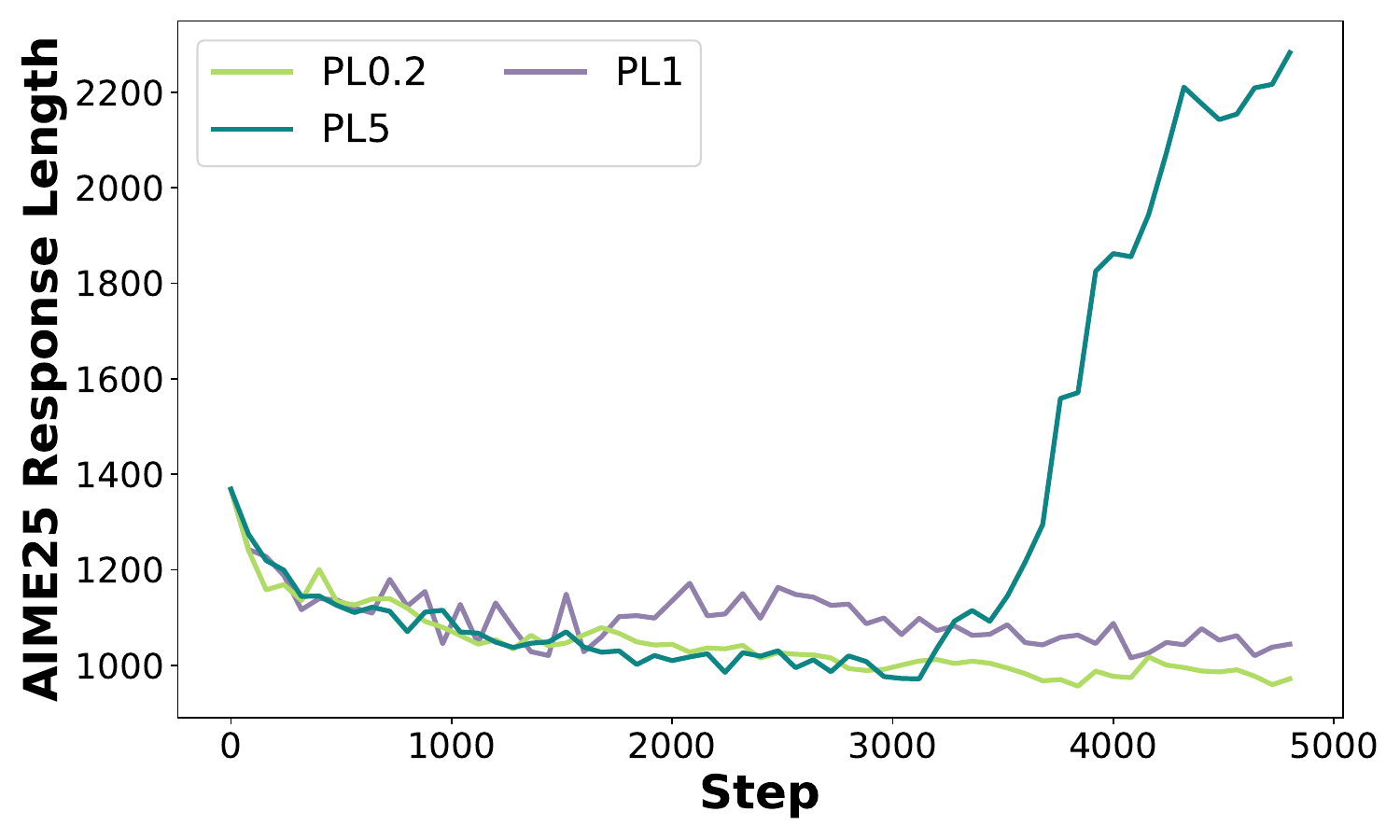}
        \caption{AIME25 Length}
    \end{subfigure}
    \begin{subfigure}[b]{0.32\linewidth}
        \centering
        \includegraphics[width=\linewidth]{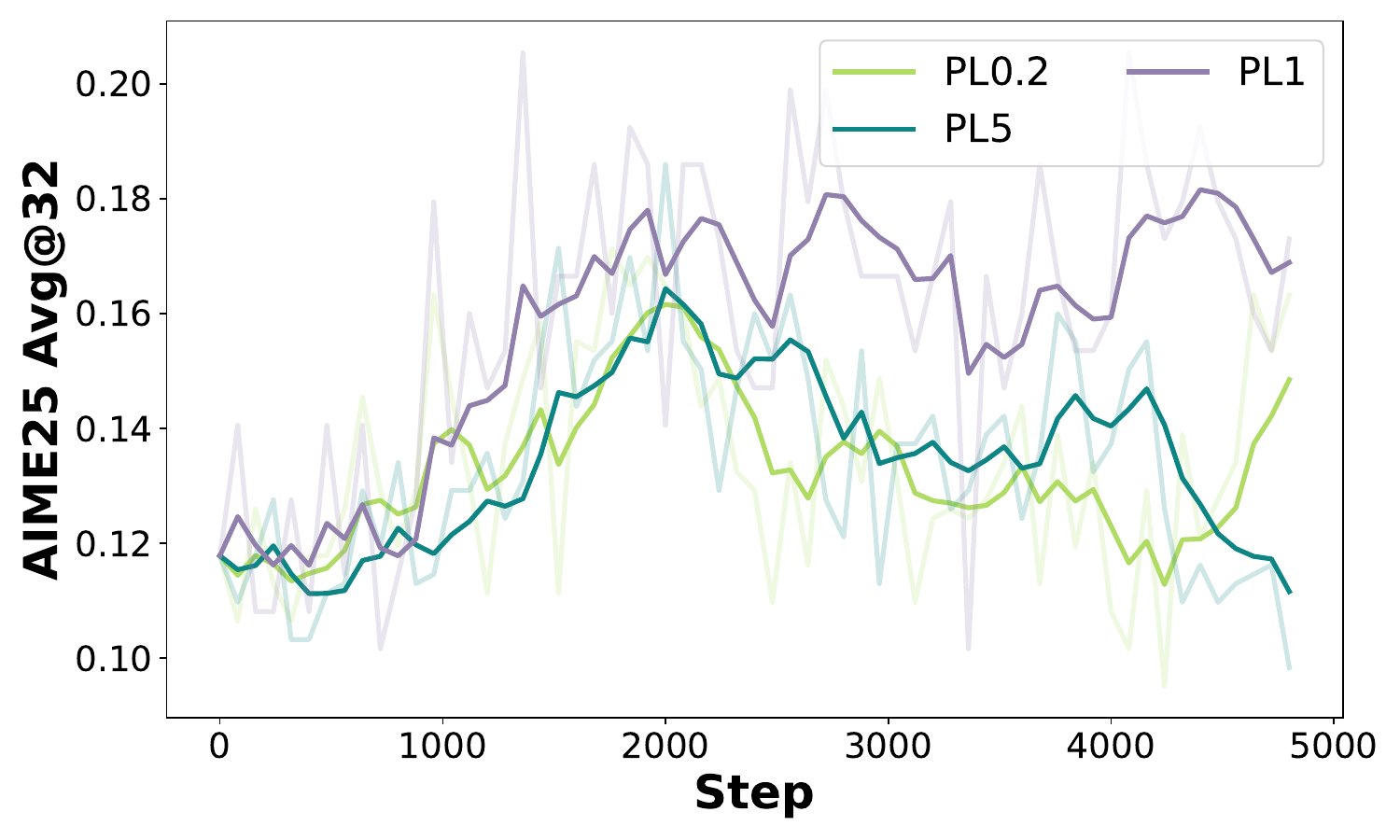}
        \caption{AIME25 Avg@32}
    \end{subfigure}
    \caption{RLVR training dynamics on positive low entropy token advantage shaping.}
\label{fig:token-entropy-pl-training_dynamic}
\end{figure*}

\begin{figure*}[t]
    \centering
    \begin{subfigure}[b]{0.32\linewidth}
        \centering
        \includegraphics[width=\linewidth]{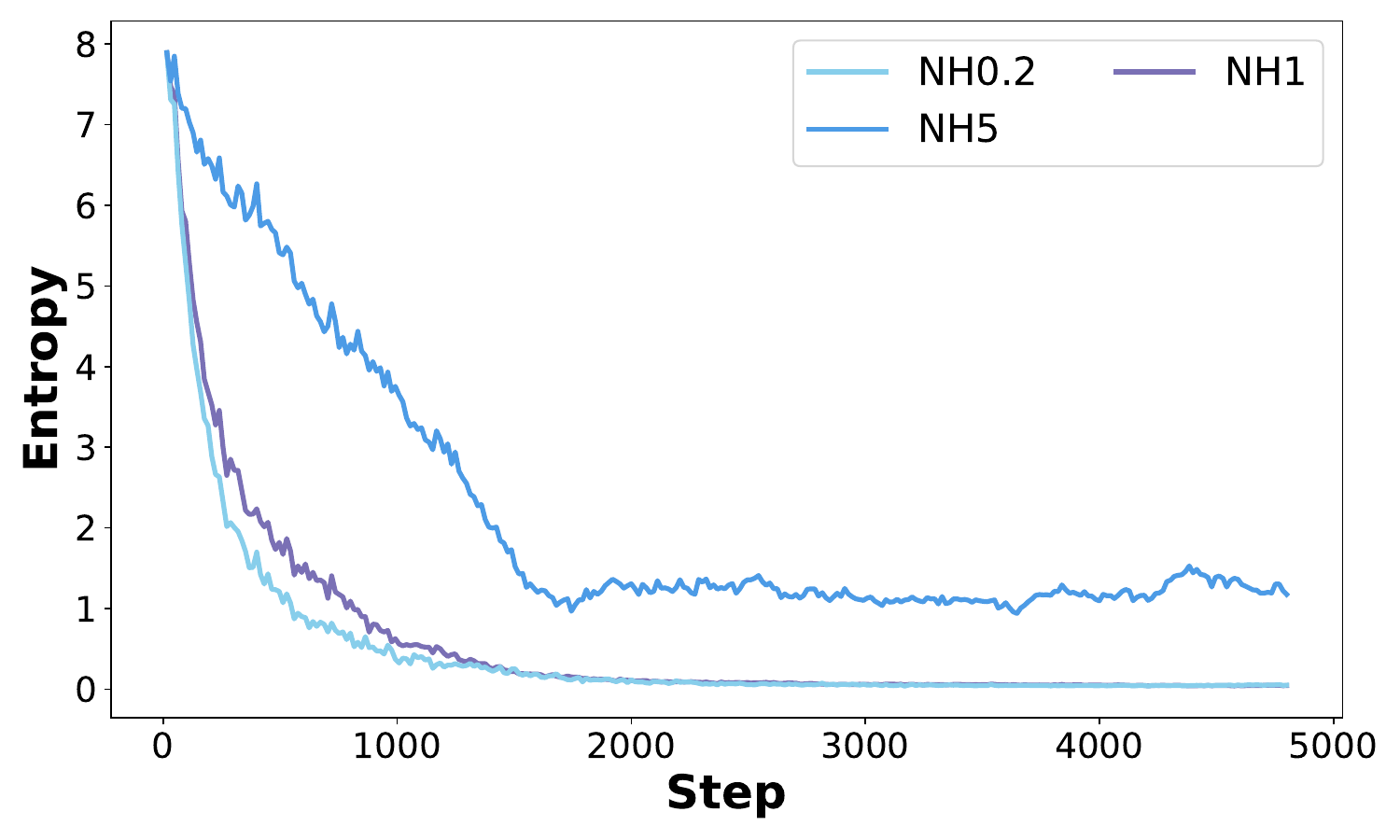}
        \caption{Entropy}
    \end{subfigure}
    \begin{subfigure}[b]{0.32\linewidth}
        \centering
        \includegraphics[width=\linewidth]{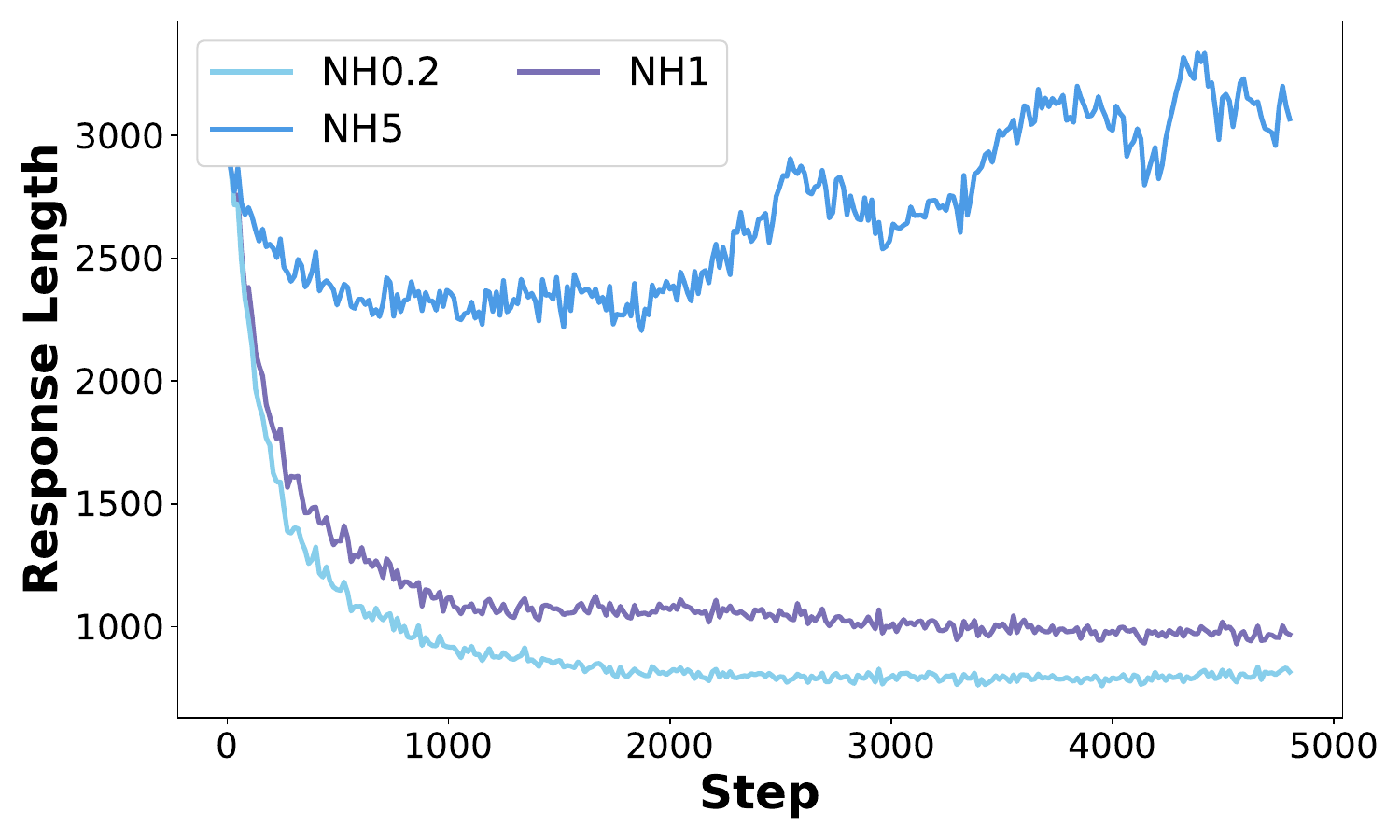}
        \caption{Length}
    \end{subfigure}
    \begin{subfigure}[b]{0.32\linewidth}
        \centering
        \includegraphics[width=\linewidth]{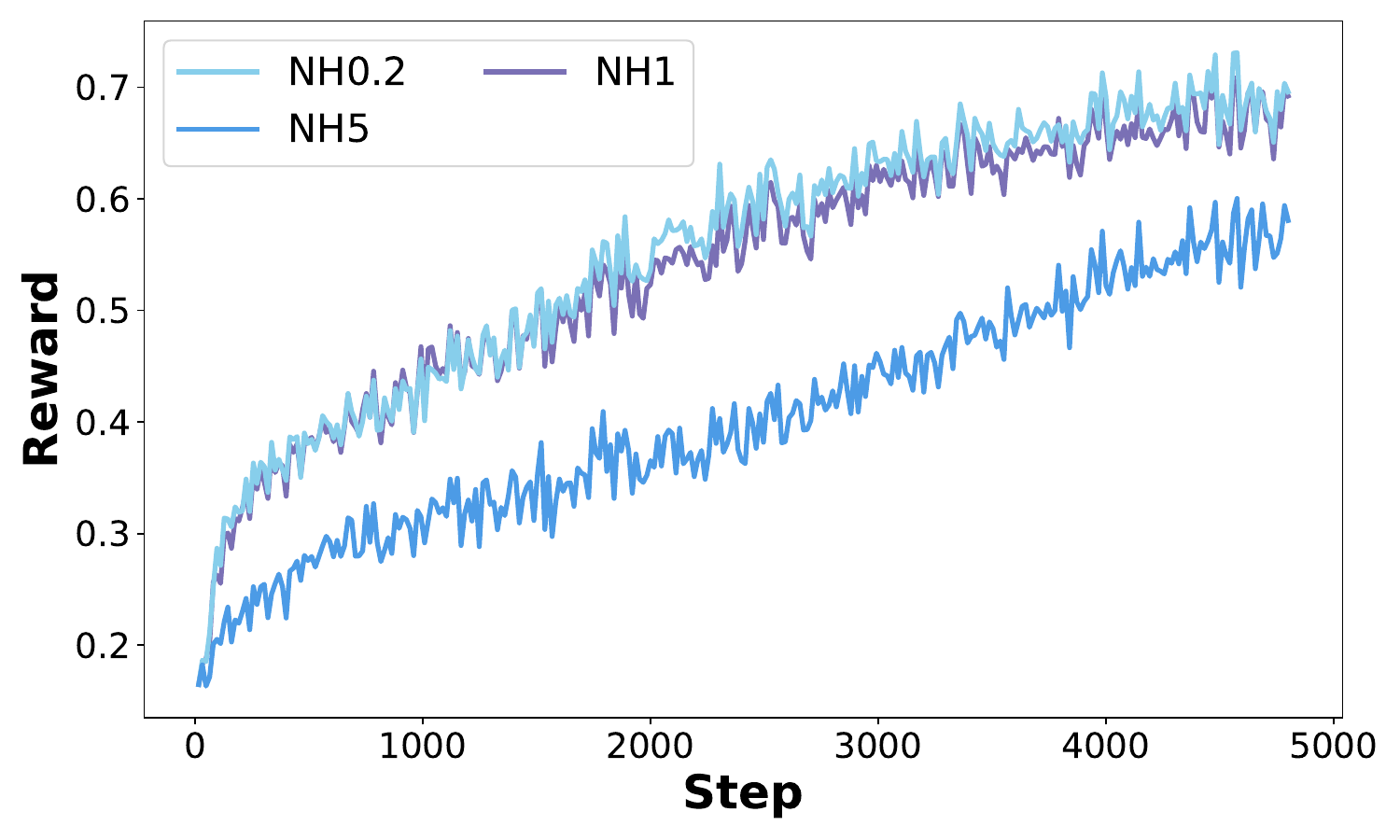}
        \caption{Reward}
    \end{subfigure}
    \begin{subfigure}[b]{0.32\linewidth}
        \centering
        \includegraphics[width=\linewidth]{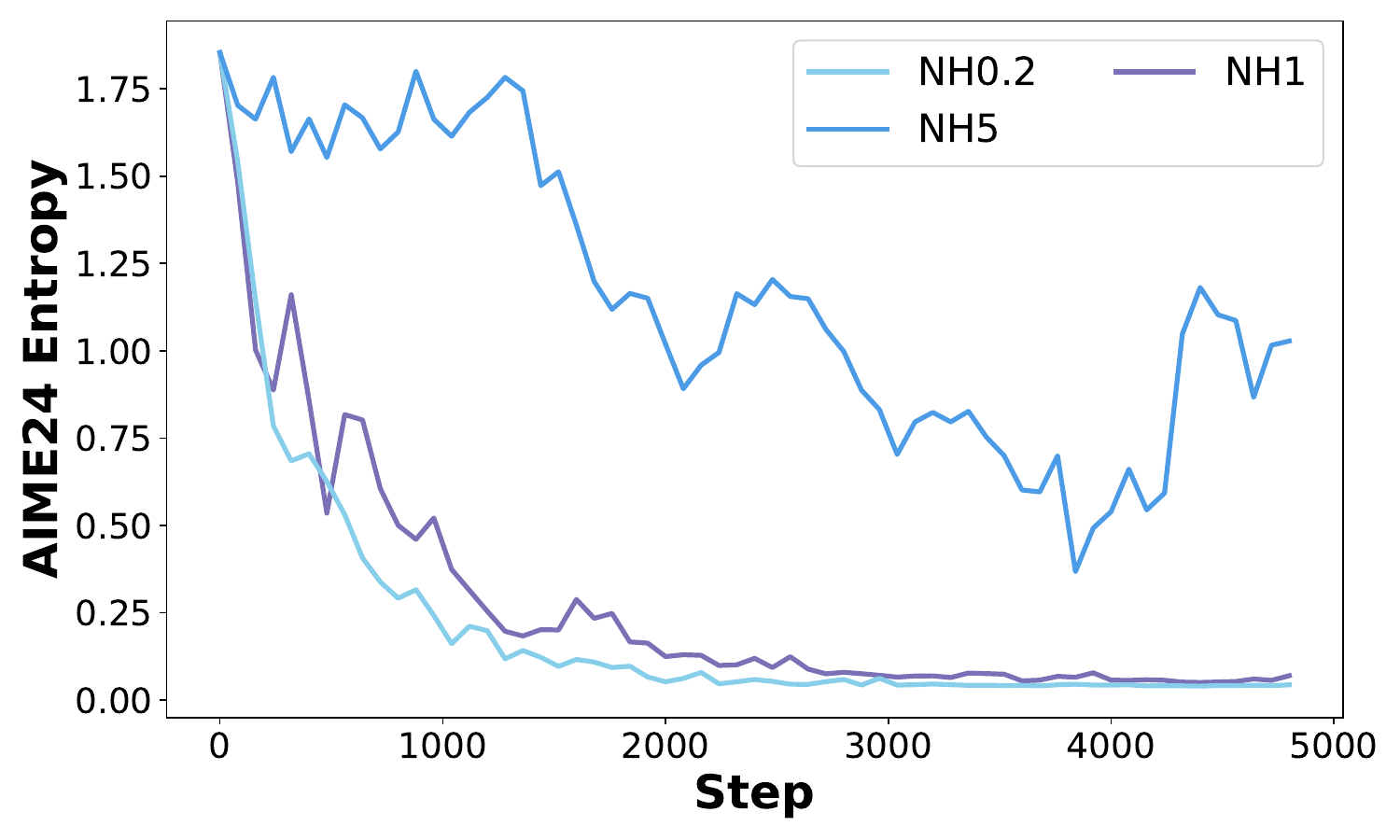}
        \caption{AIME24 Entropy}
    \end{subfigure}
    \begin{subfigure}[b]{0.32\linewidth}
        \centering
        \includegraphics[width=\linewidth]{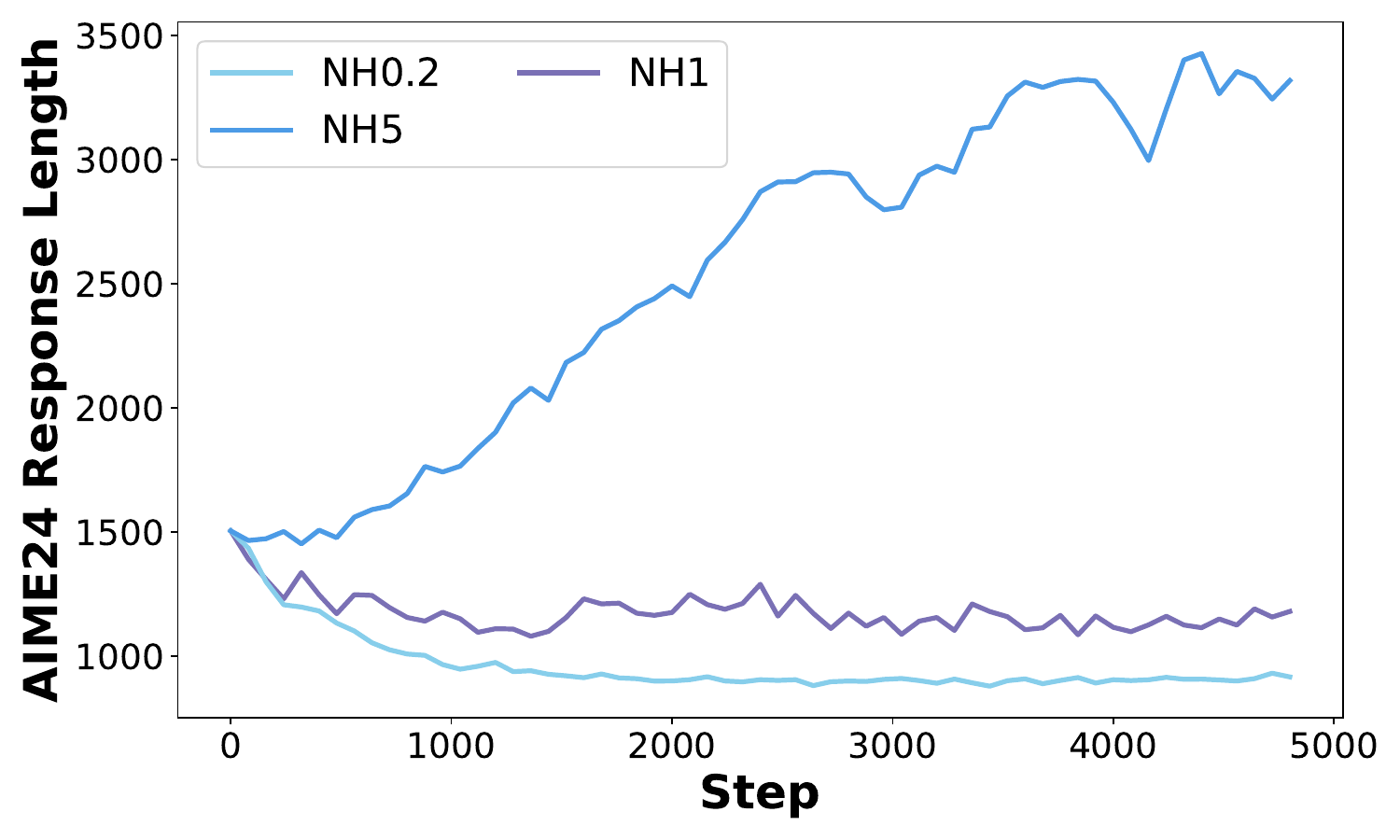}
        \caption{AIME24 Length}
    \end{subfigure}
    \begin{subfigure}[b]{0.32\linewidth}
        \centering
        \includegraphics[width=\linewidth]{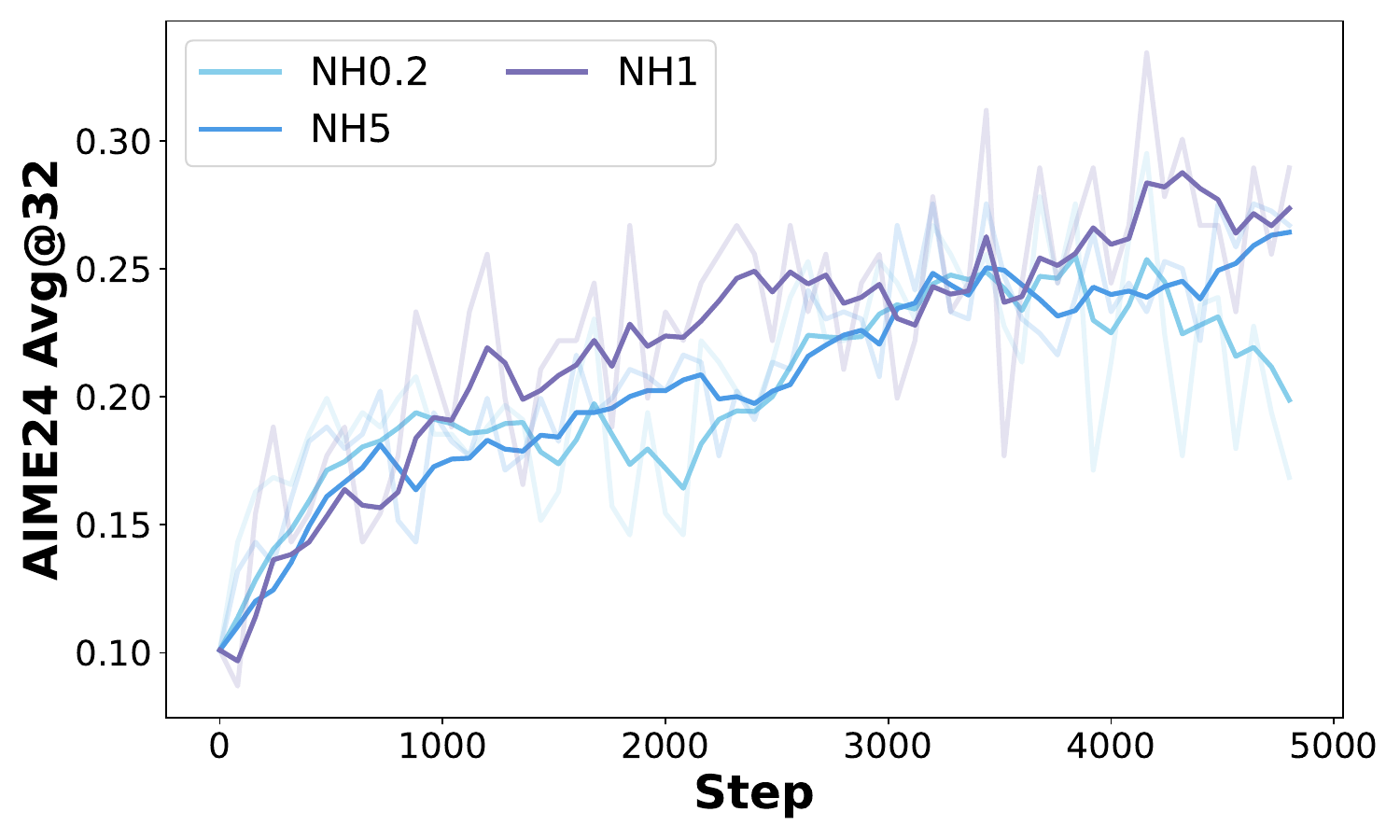}
        \caption{AIME24 Avg@32}
    \end{subfigure}
    \begin{subfigure}[b]{0.32\linewidth}
        \centering
        \includegraphics[width=\linewidth]{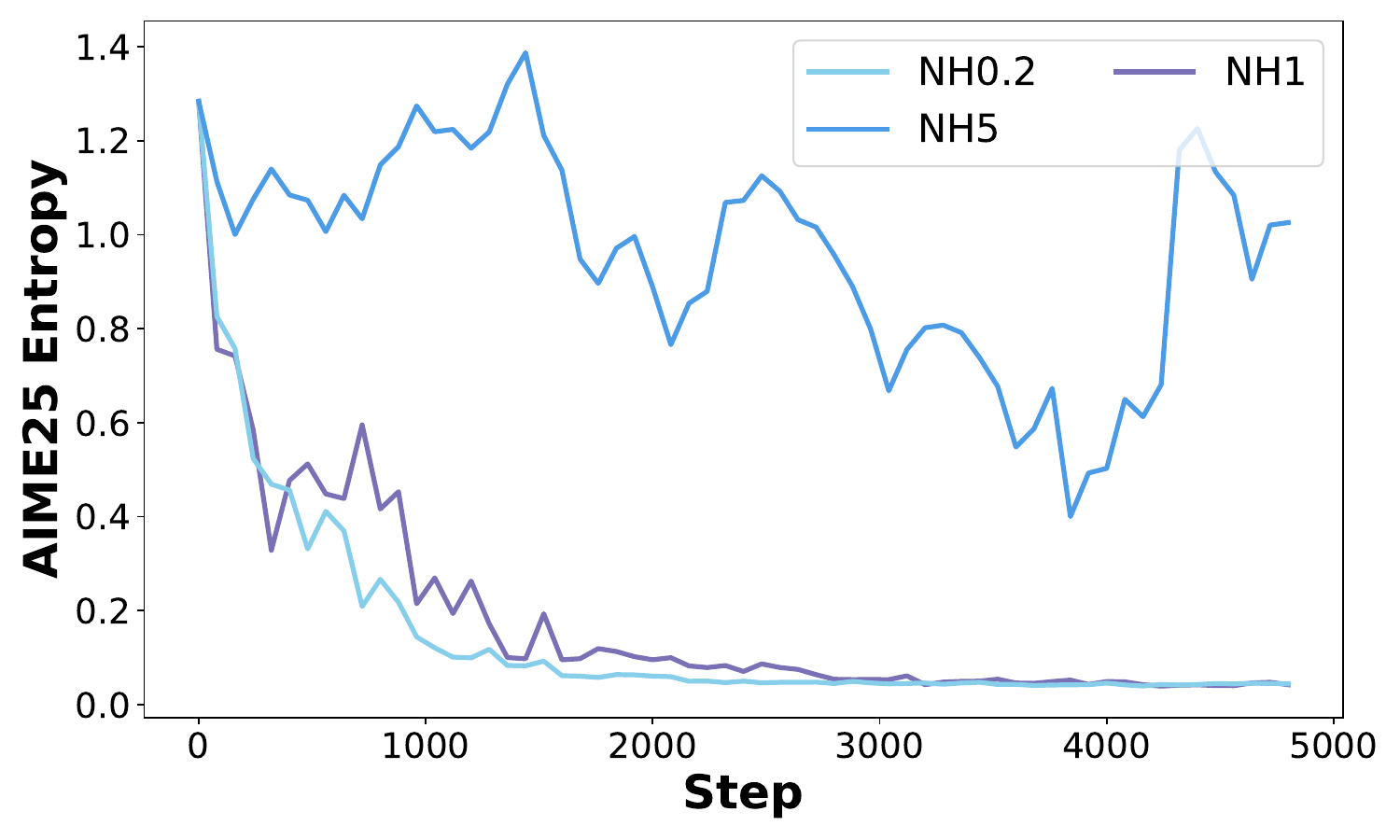}
        \caption{AIME25 Entropy}
    \end{subfigure}
    \begin{subfigure}[b]{0.32\linewidth}
        \centering
        \includegraphics[width=\linewidth]{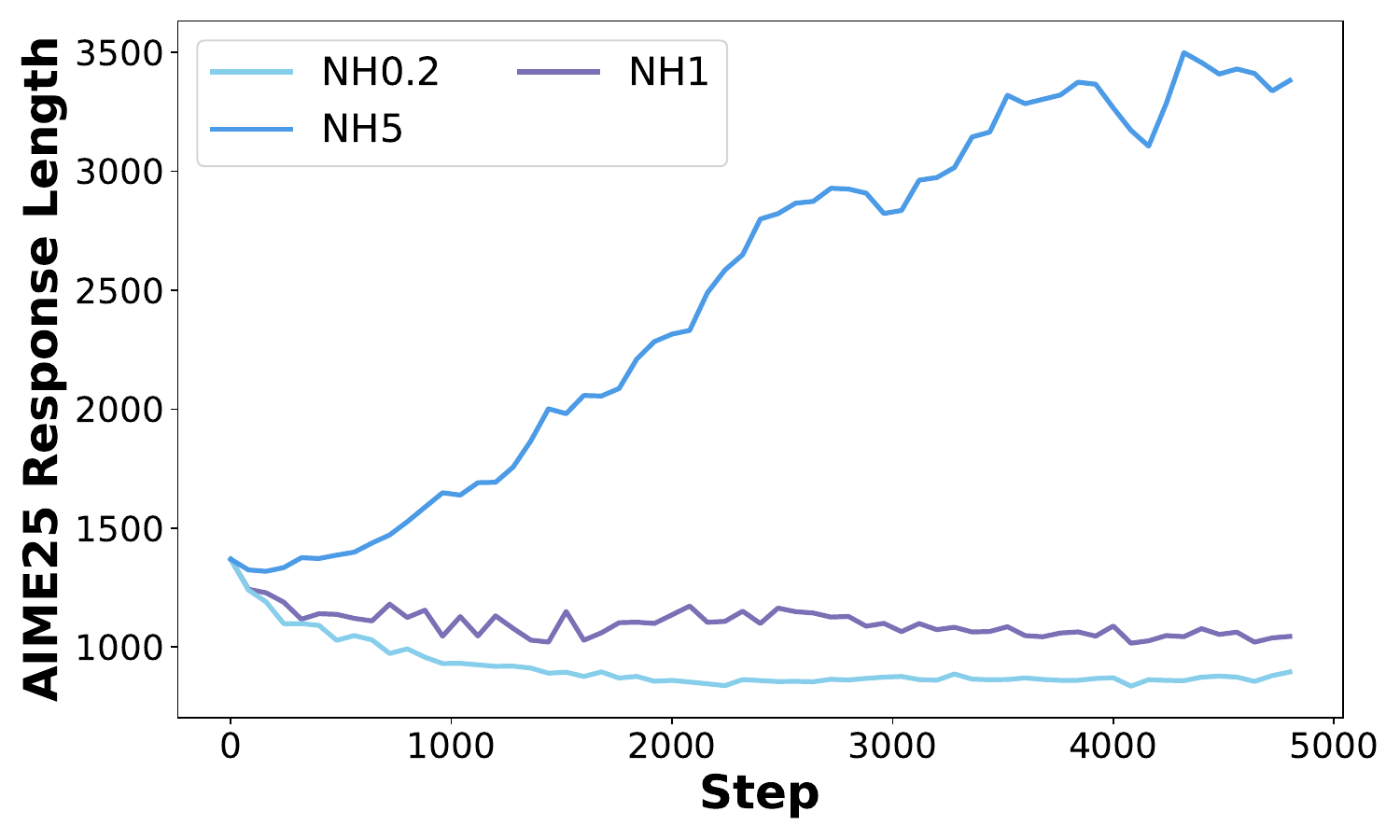}
        \caption{AIME25 Length}
    \end{subfigure}
    \begin{subfigure}[b]{0.32\linewidth}
        \centering
        \includegraphics[width=\linewidth]{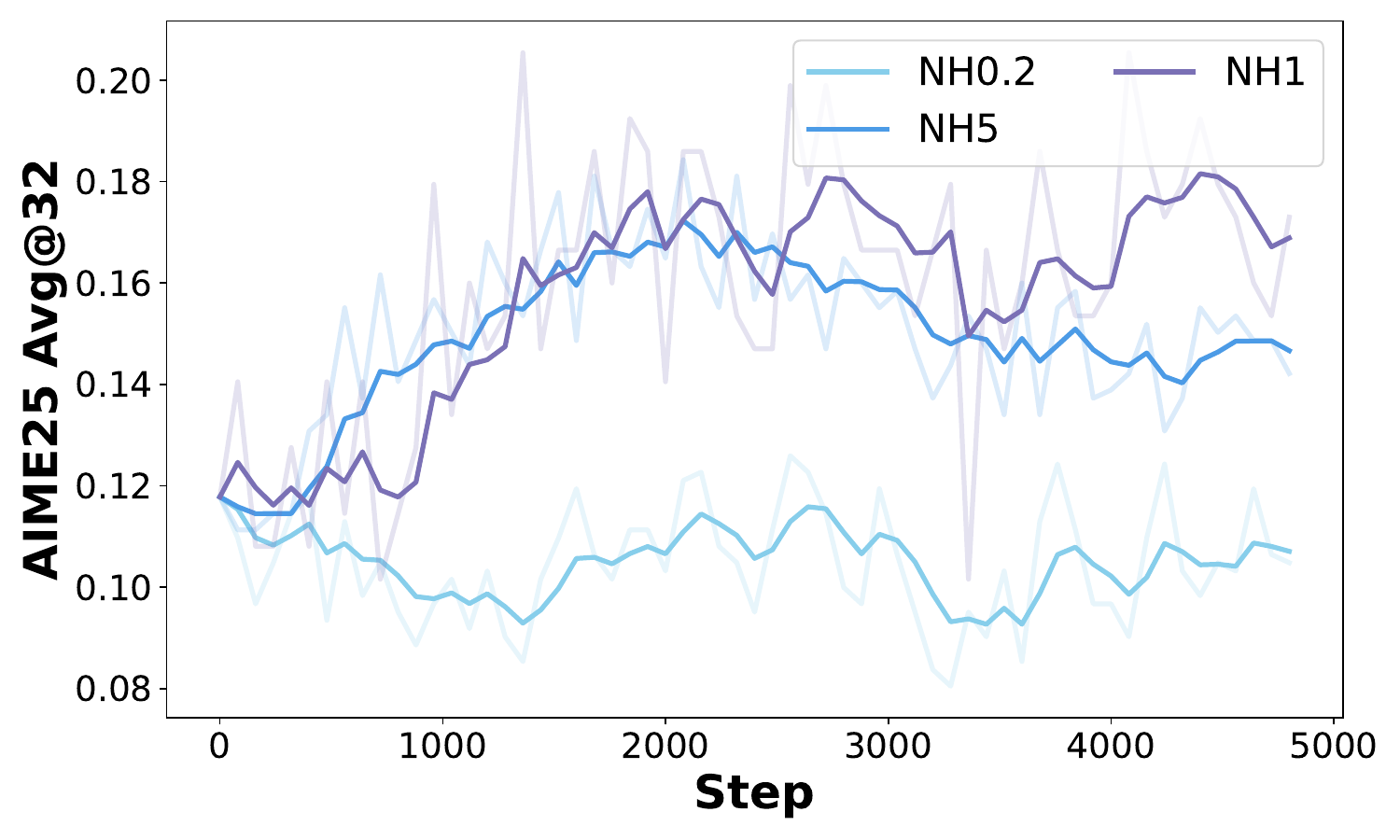}
        \caption{AIME25 Avg@32}
    \end{subfigure}
    \caption{RLVR training dynamics on negative high entropy token advantage shaping.}
\label{fig:token-entropy-nh-training_dynamic}
\end{figure*}

\begin{figure*}[t]
    \centering
    \begin{subfigure}[b]{0.32\linewidth}
        \centering
        \includegraphics[width=\linewidth]{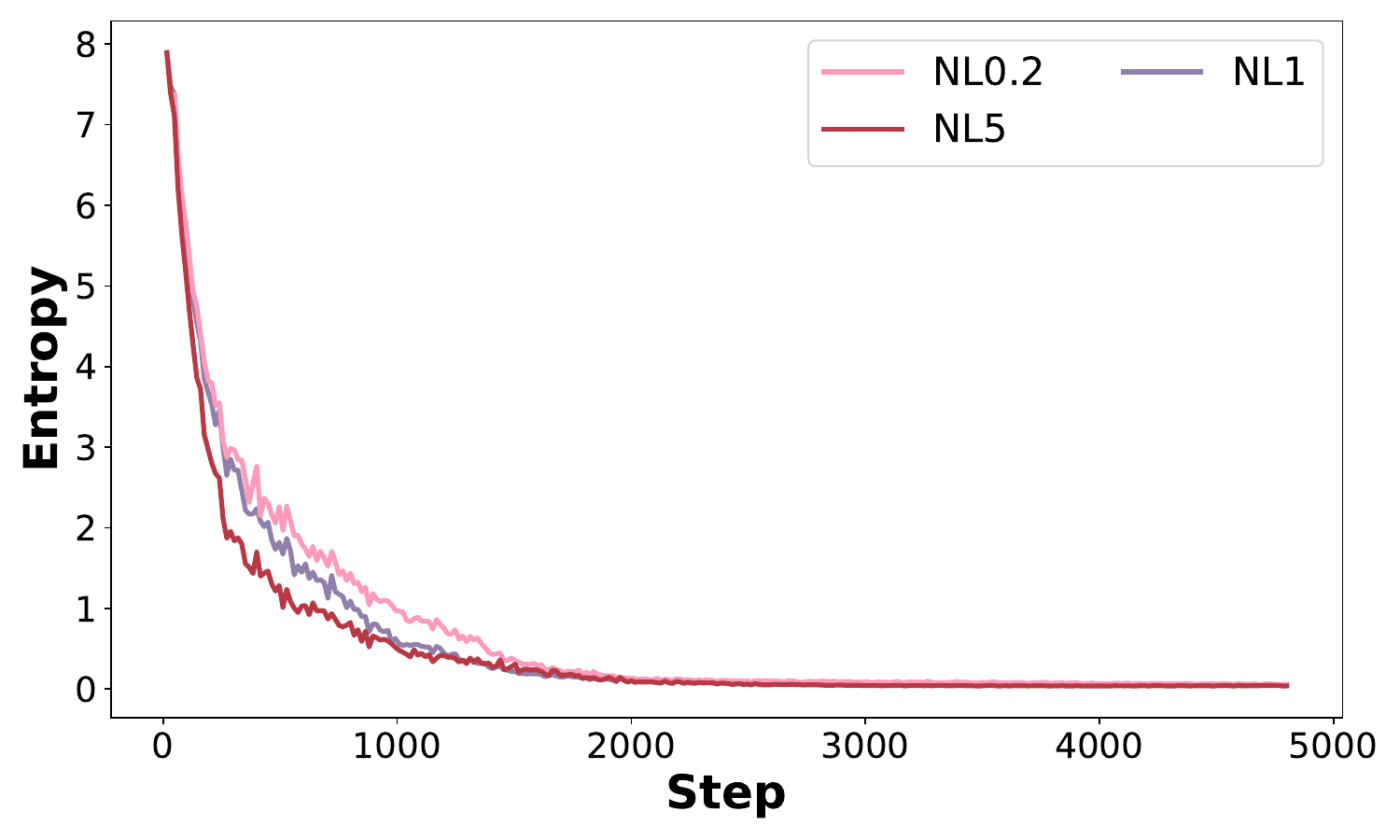}
        \caption{Entropy}
    \end{subfigure}
    \begin{subfigure}[b]{0.32\linewidth}
        \centering
        \includegraphics[width=\linewidth]{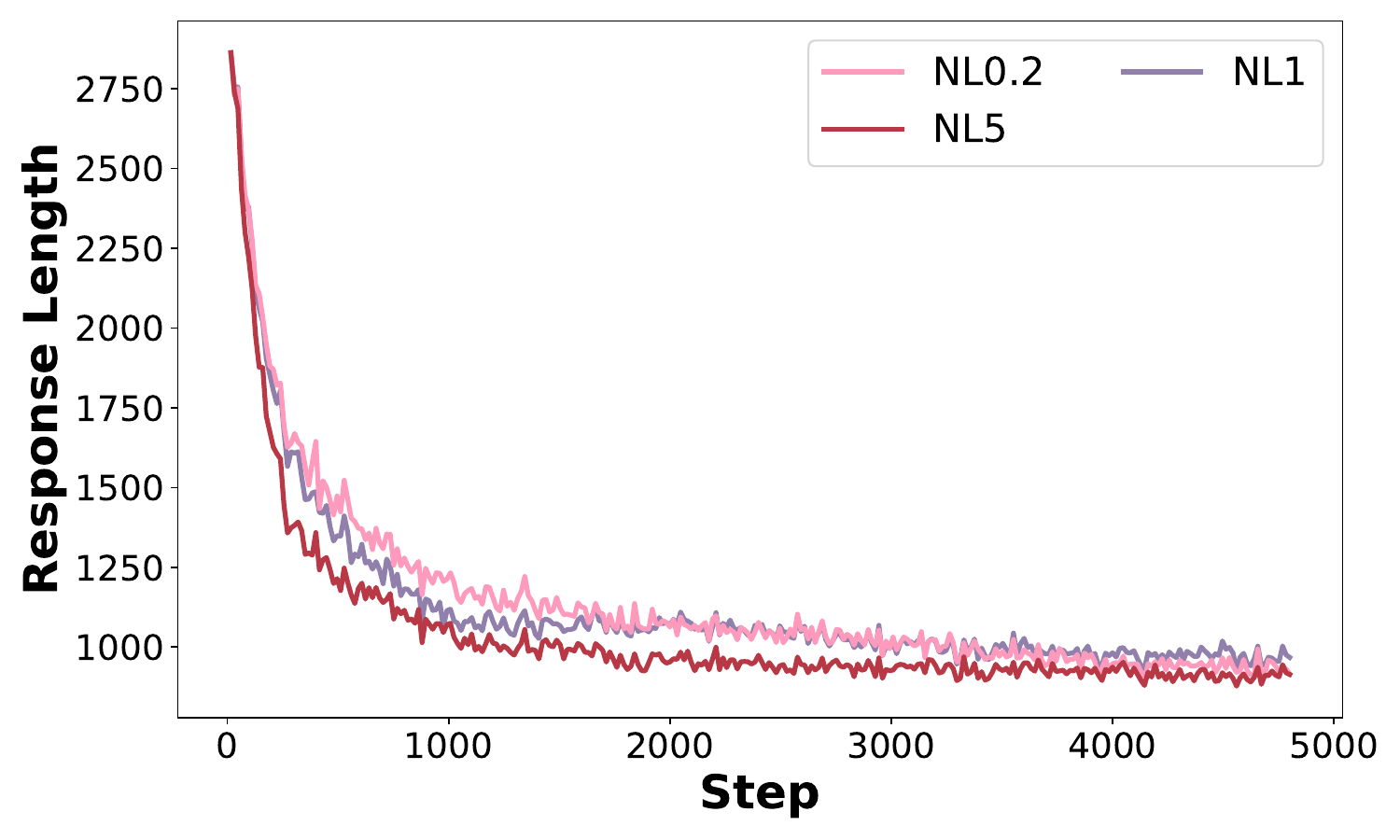}
        \caption{Length}
    \end{subfigure}
    \begin{subfigure}[b]{0.32\linewidth}
        \centering
        \includegraphics[width=\linewidth]{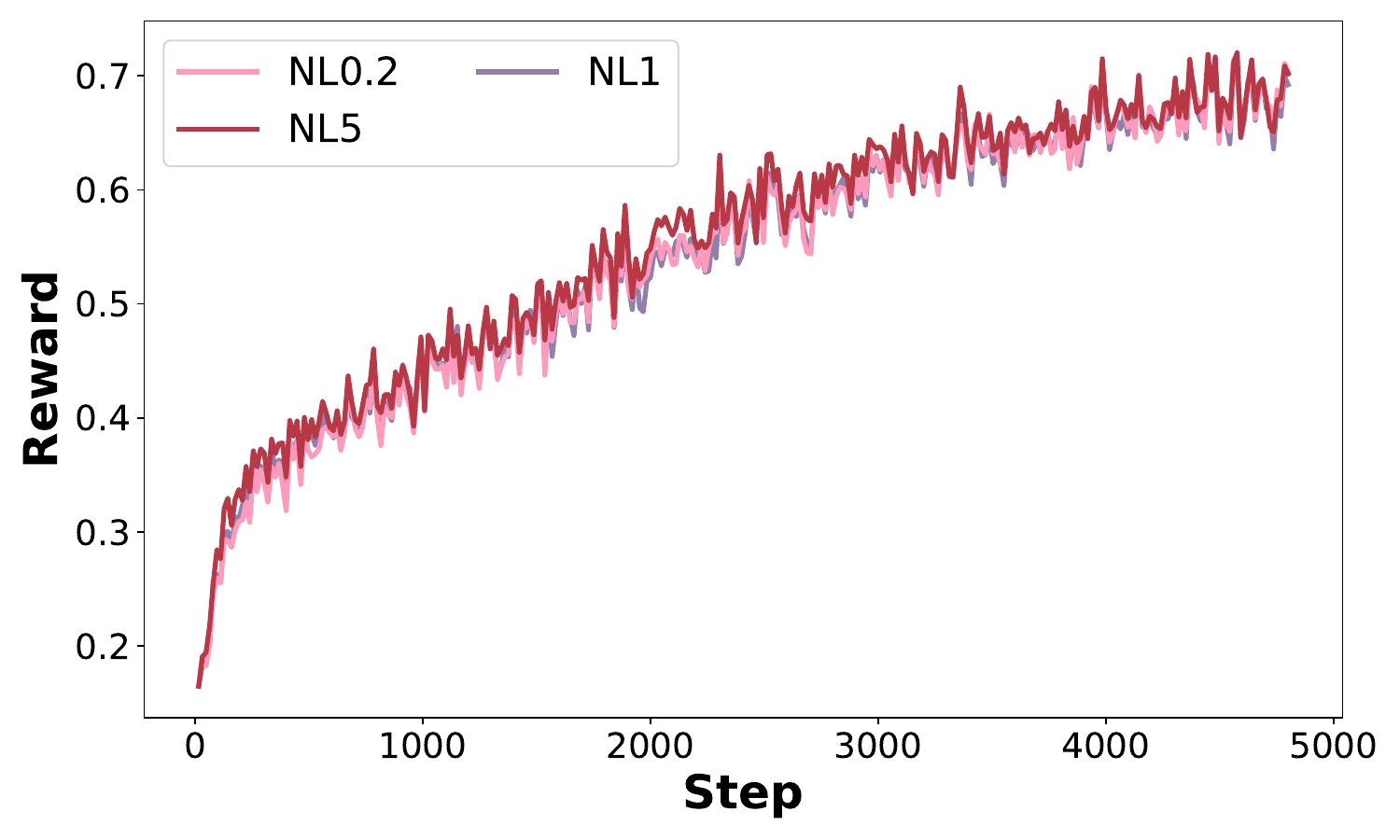}
        \caption{Reward}
    \end{subfigure}
    \begin{subfigure}[b]{0.32\linewidth}
        \centering
        \includegraphics[width=\linewidth]{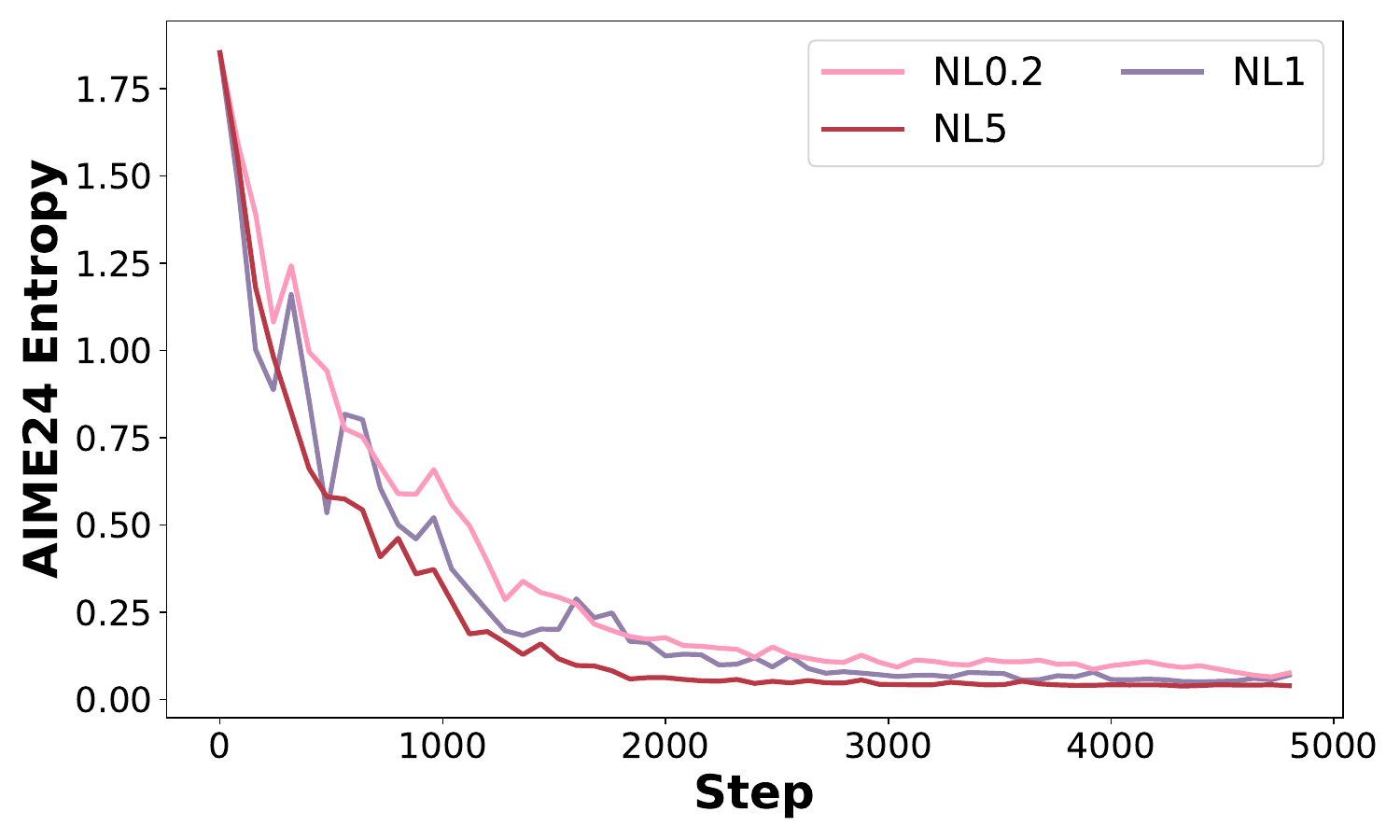}
        \caption{AIME24 Entropy}
    \end{subfigure}
    \begin{subfigure}[b]{0.32\linewidth}
        \centering
        \includegraphics[width=\linewidth]{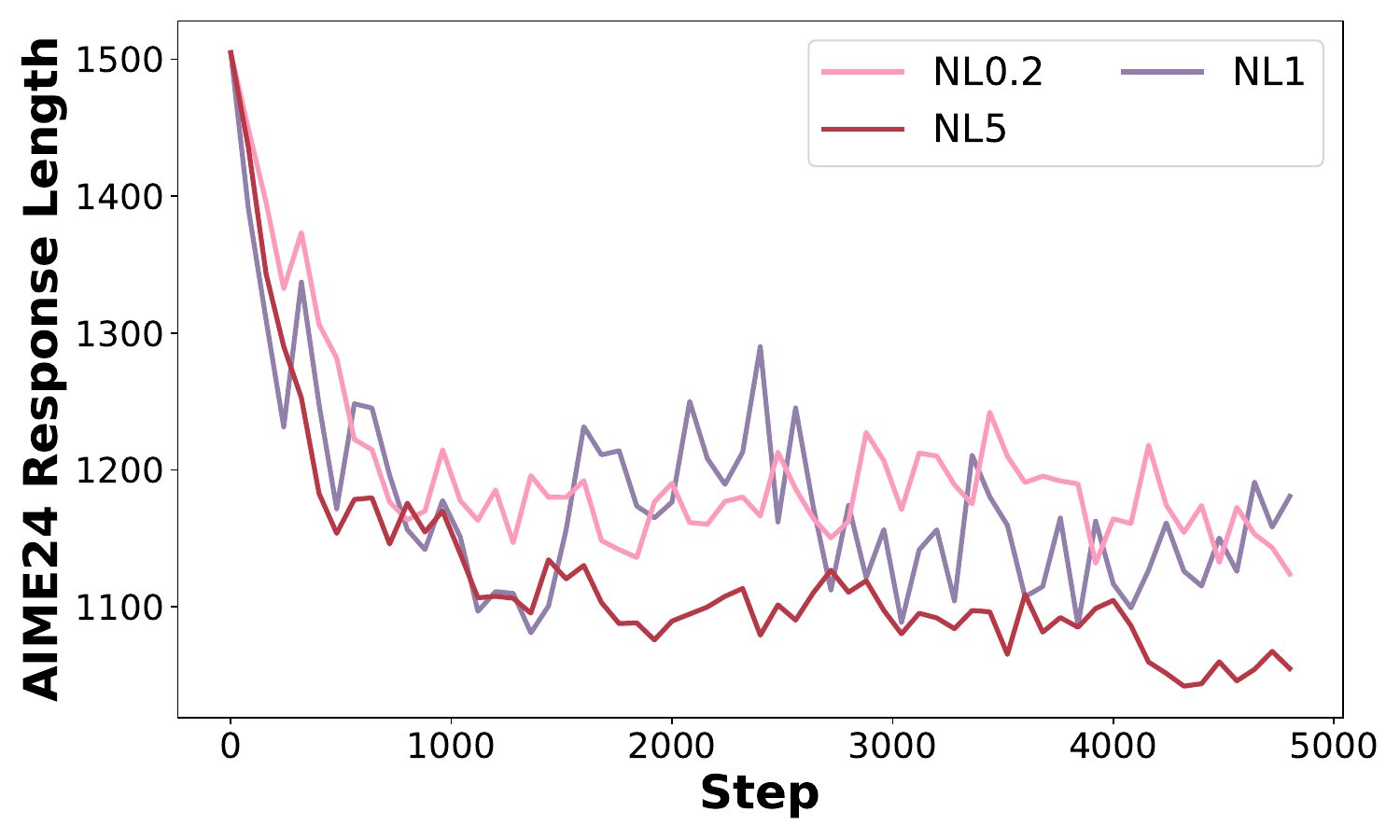}
        \caption{AIME24 Length}
    \end{subfigure}
    \begin{subfigure}[b]{0.32\linewidth}
        \centering
        \includegraphics[width=\linewidth]{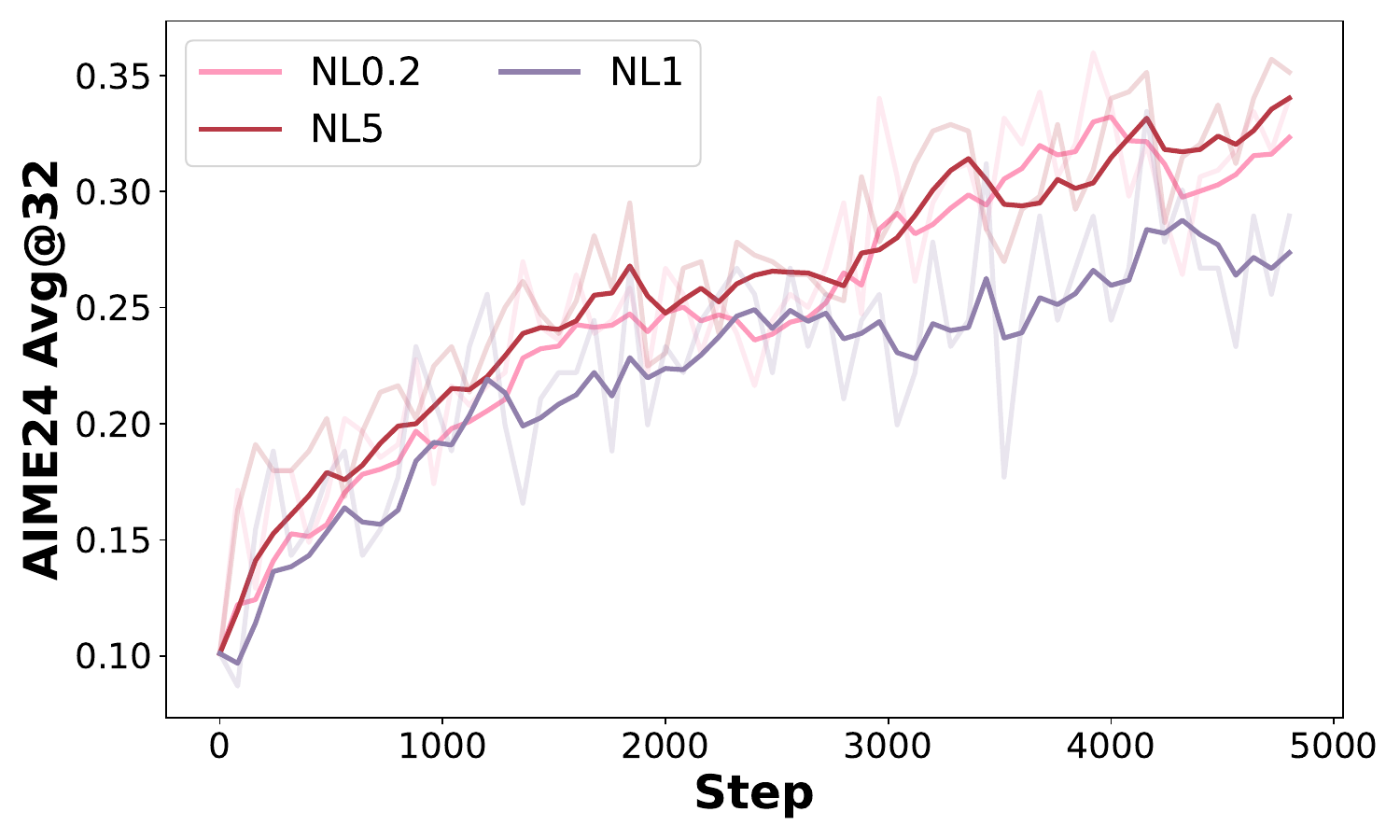}
        \caption{AIME24 Avg@32}
    \end{subfigure}
    \begin{subfigure}[b]{0.32\linewidth}
        \centering
        \includegraphics[width=\linewidth]{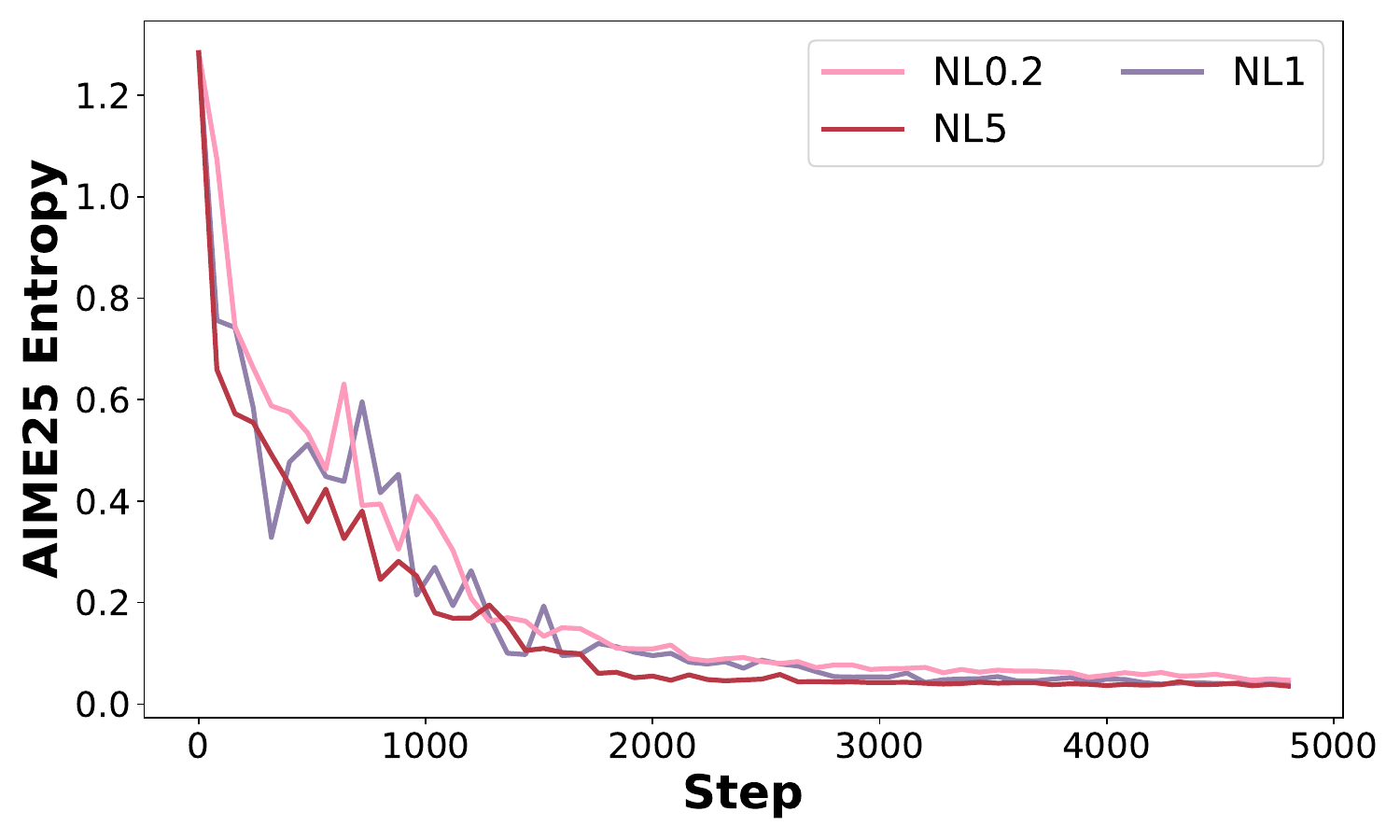}
        \caption{AIME25 Entropy}
    \end{subfigure}
    \begin{subfigure}[b]{0.32\linewidth}
        \centering
        \includegraphics[width=\linewidth]{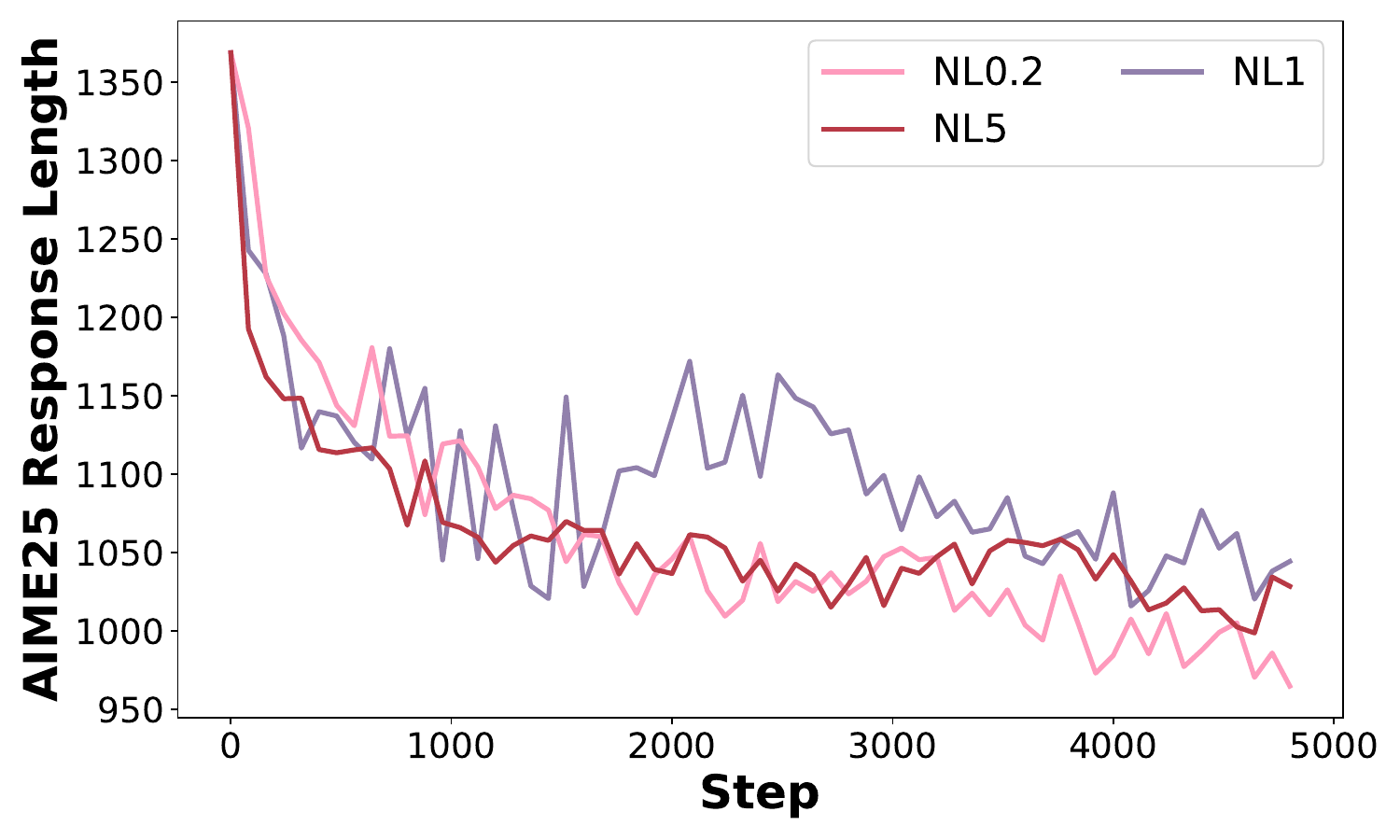}
        \caption{AIME25 Length}
    \end{subfigure}
    \begin{subfigure}[b]{0.32\linewidth}
        \centering
        \includegraphics[width=\linewidth]{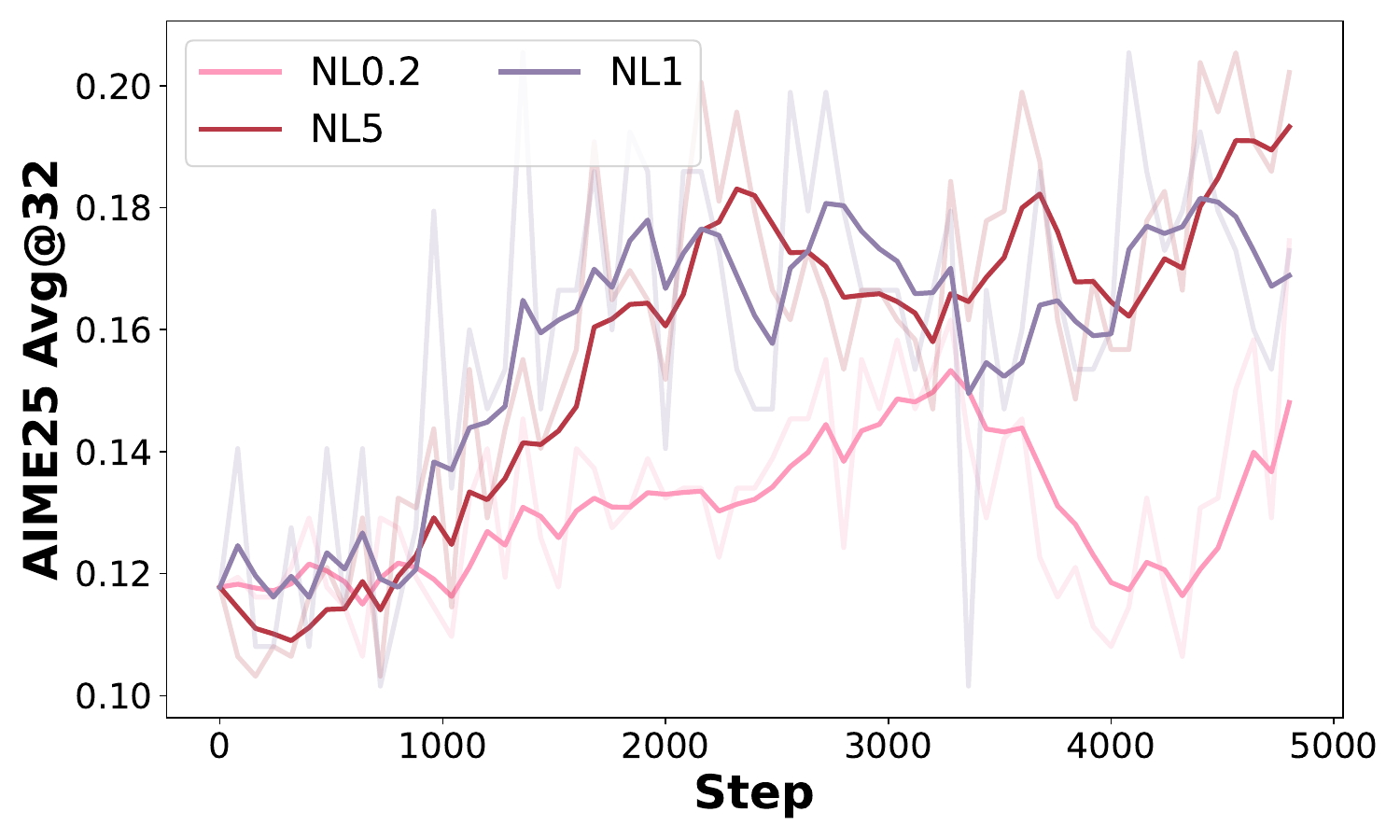}
        \caption{AIME25 Avg@32}
    \end{subfigure}
    \caption{RLVR training dynamics on negative low entropy token advantage shaping.}
\label{fig:token-entropy-nl-training_dynamic}
\end{figure*}

\begin{figure*}[t]
    \centering
    \begin{subfigure}[b]{0.32\linewidth}
        \centering
        \includegraphics[width=\linewidth]{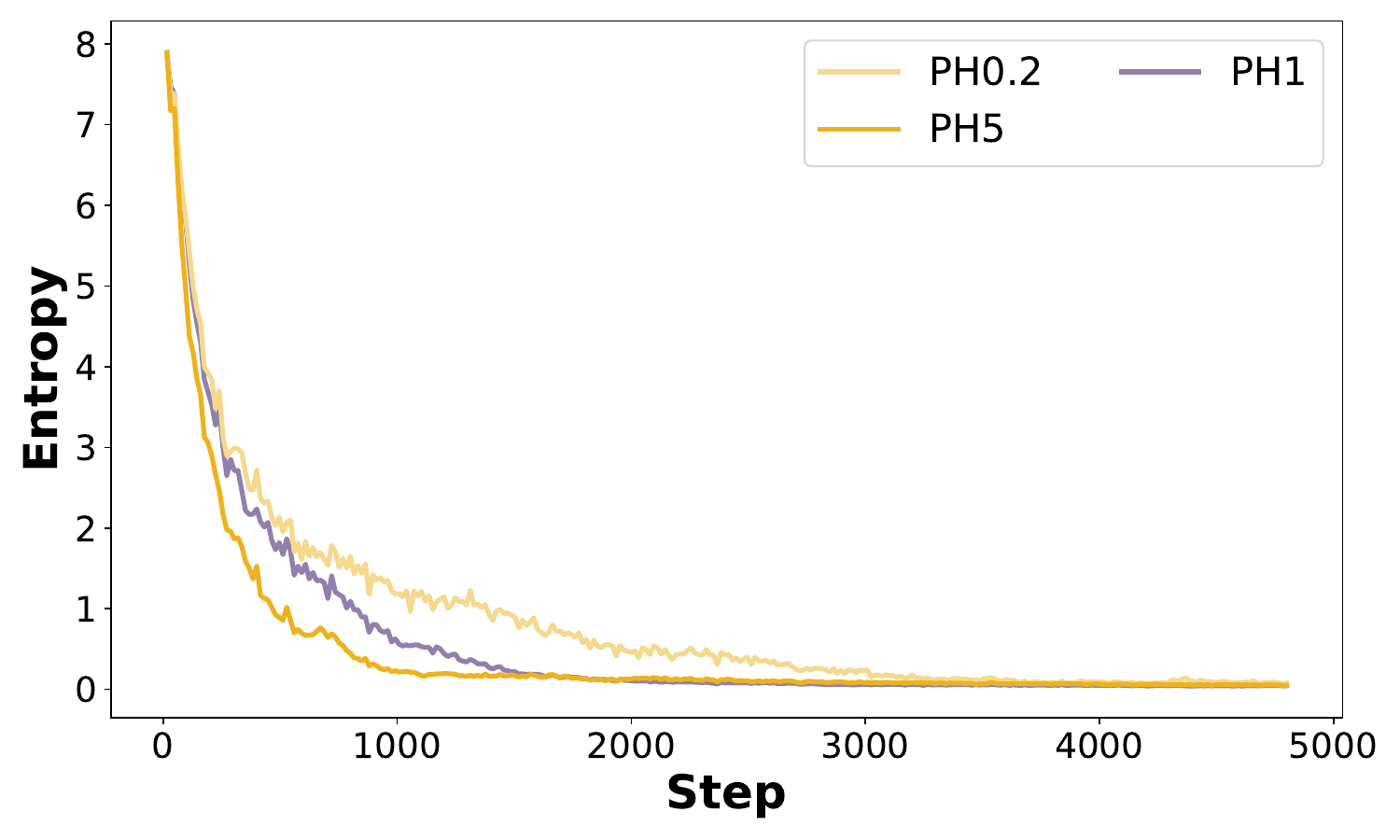}
        \caption{Entropy}
    \end{subfigure}
    \begin{subfigure}[b]{0.32\linewidth}
        \centering
        \includegraphics[width=\linewidth]{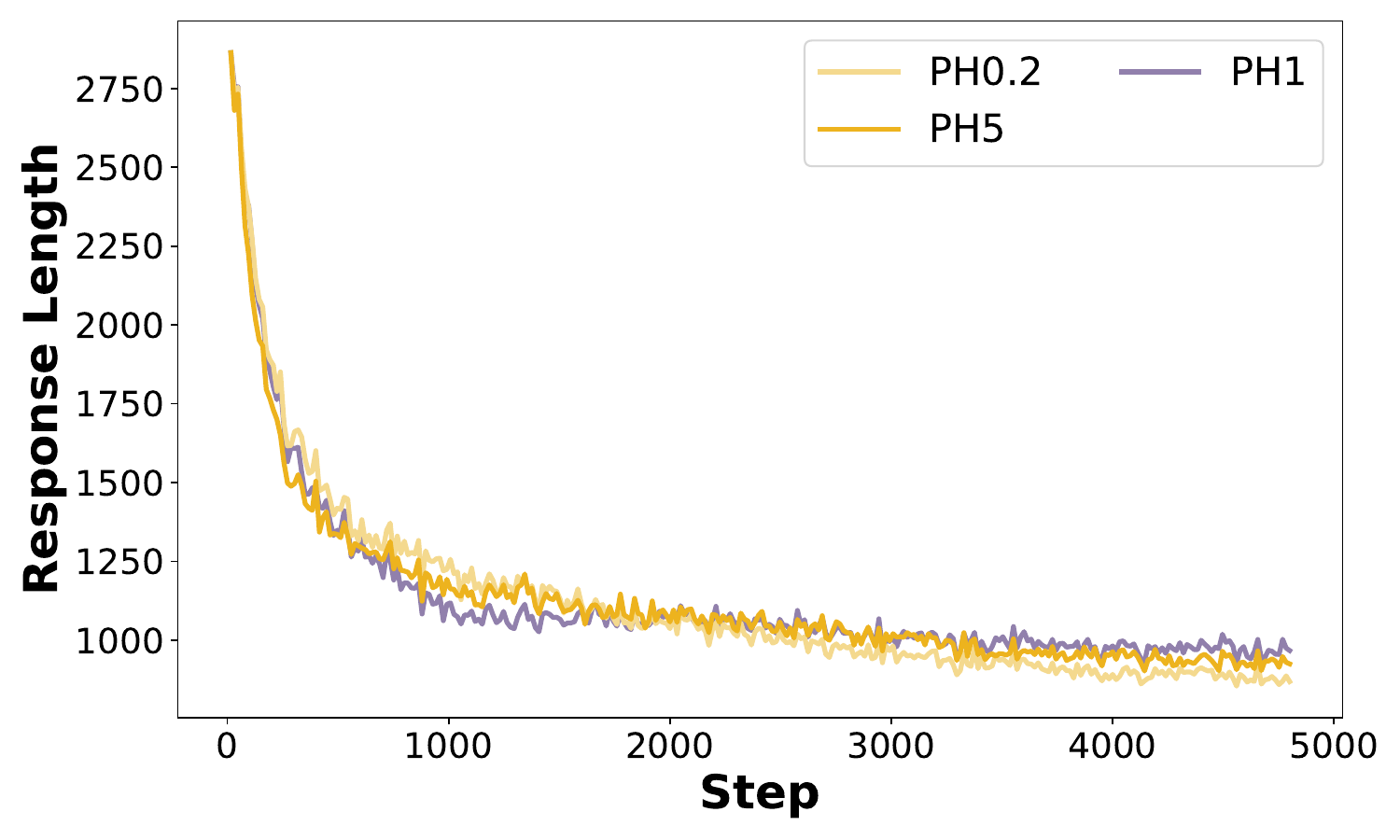}
        \caption{Length}
    \end{subfigure}
    \begin{subfigure}[b]{0.32\linewidth}
        \centering
        \includegraphics[width=\linewidth]{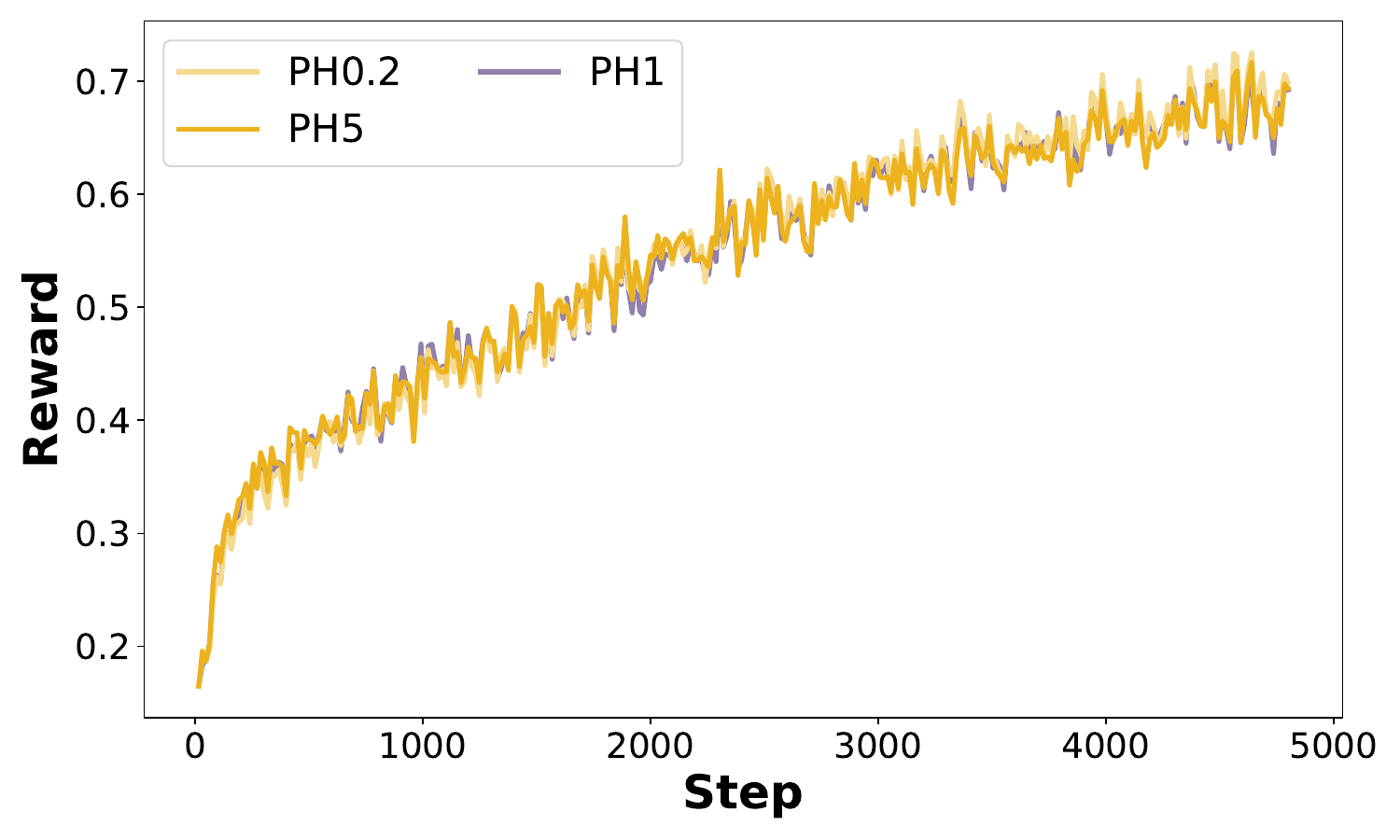}
        \caption{Reward}
    \end{subfigure}
    \begin{subfigure}[b]{0.32\linewidth}
        \centering
        \includegraphics[width=\linewidth]{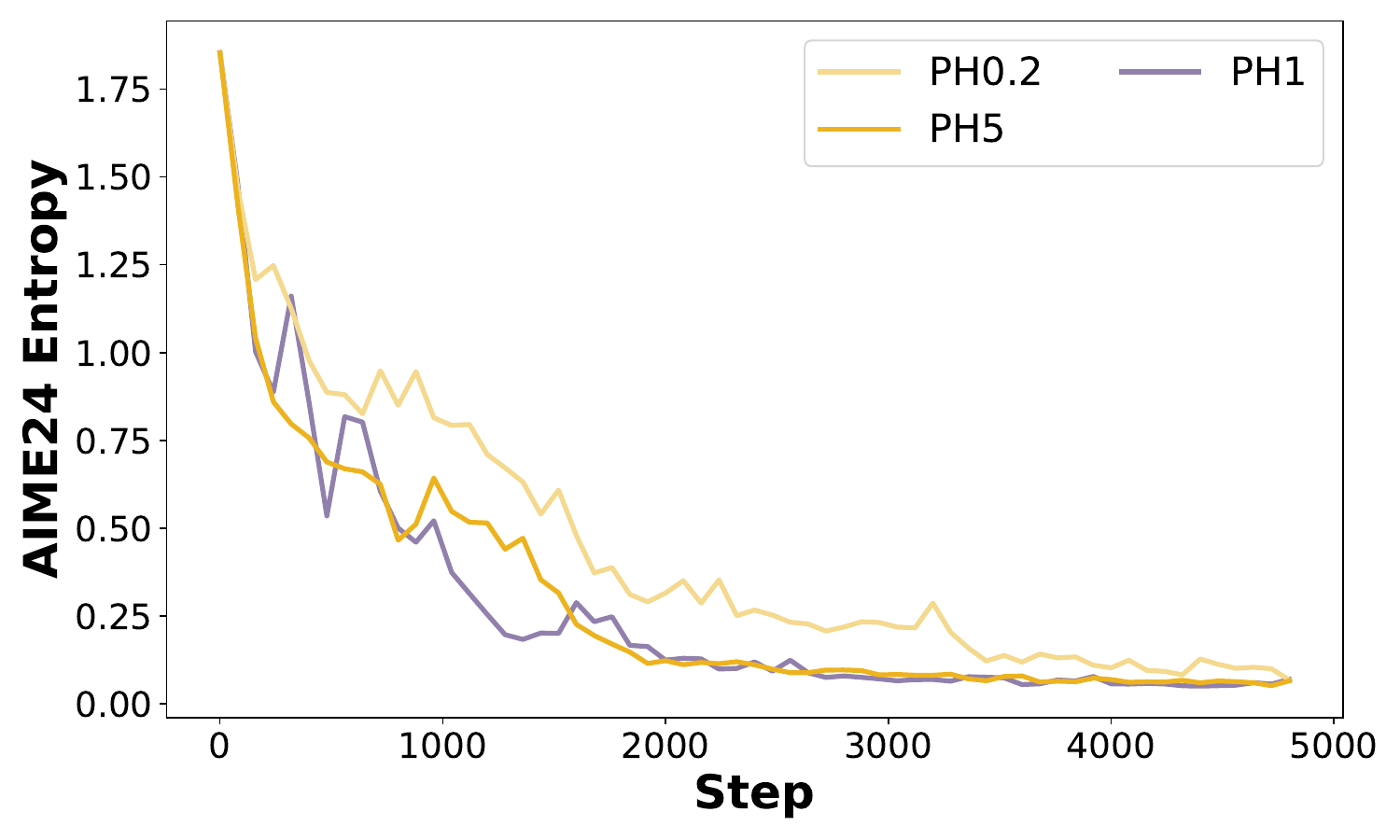}
        \caption{AIME24 Entropy}
    \end{subfigure}
    \begin{subfigure}[b]{0.32\linewidth}
        \centering
        \includegraphics[width=\linewidth]{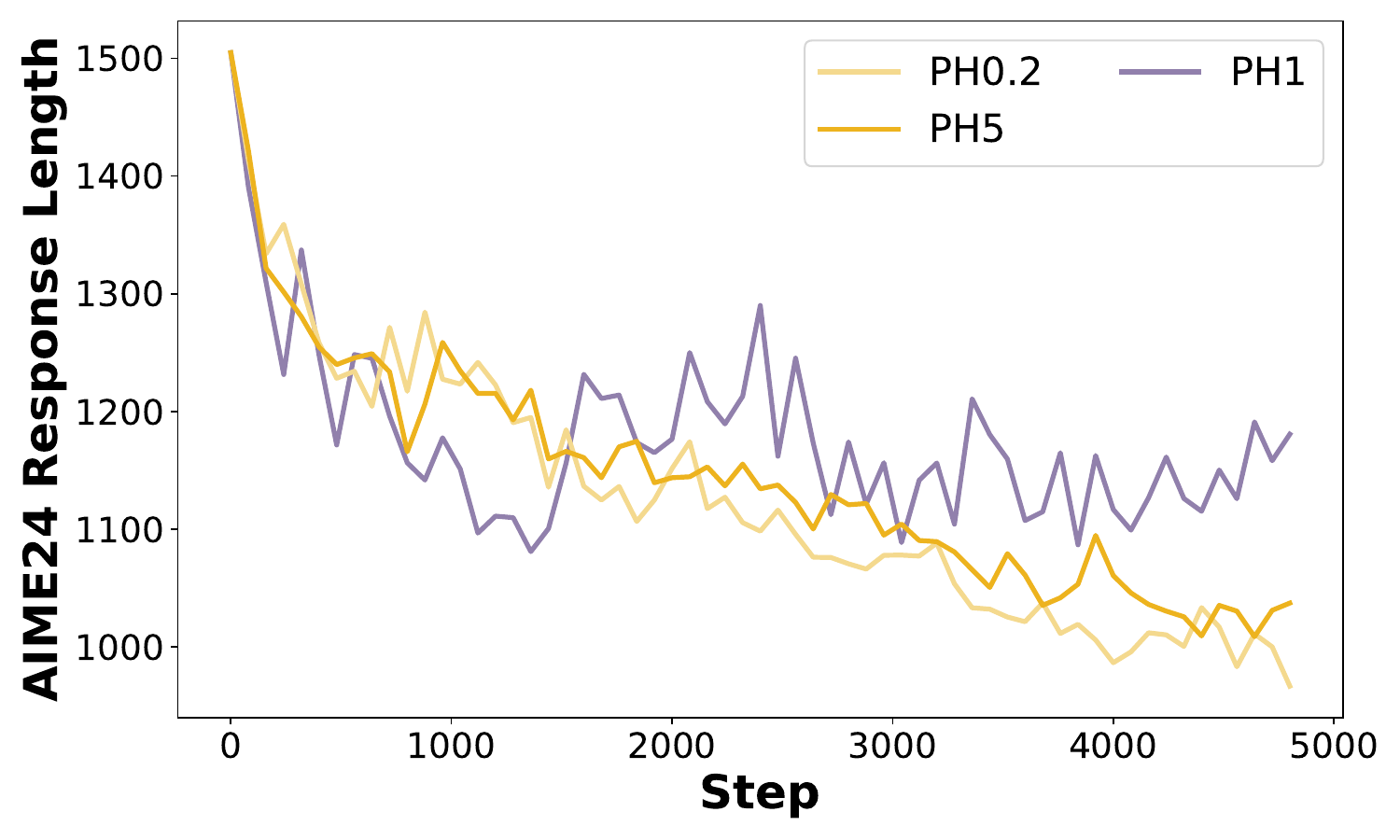}
        \caption{AIME24 Length}
    \end{subfigure}
    \begin{subfigure}[b]{0.32\linewidth}
        \centering
        \includegraphics[width=\linewidth]{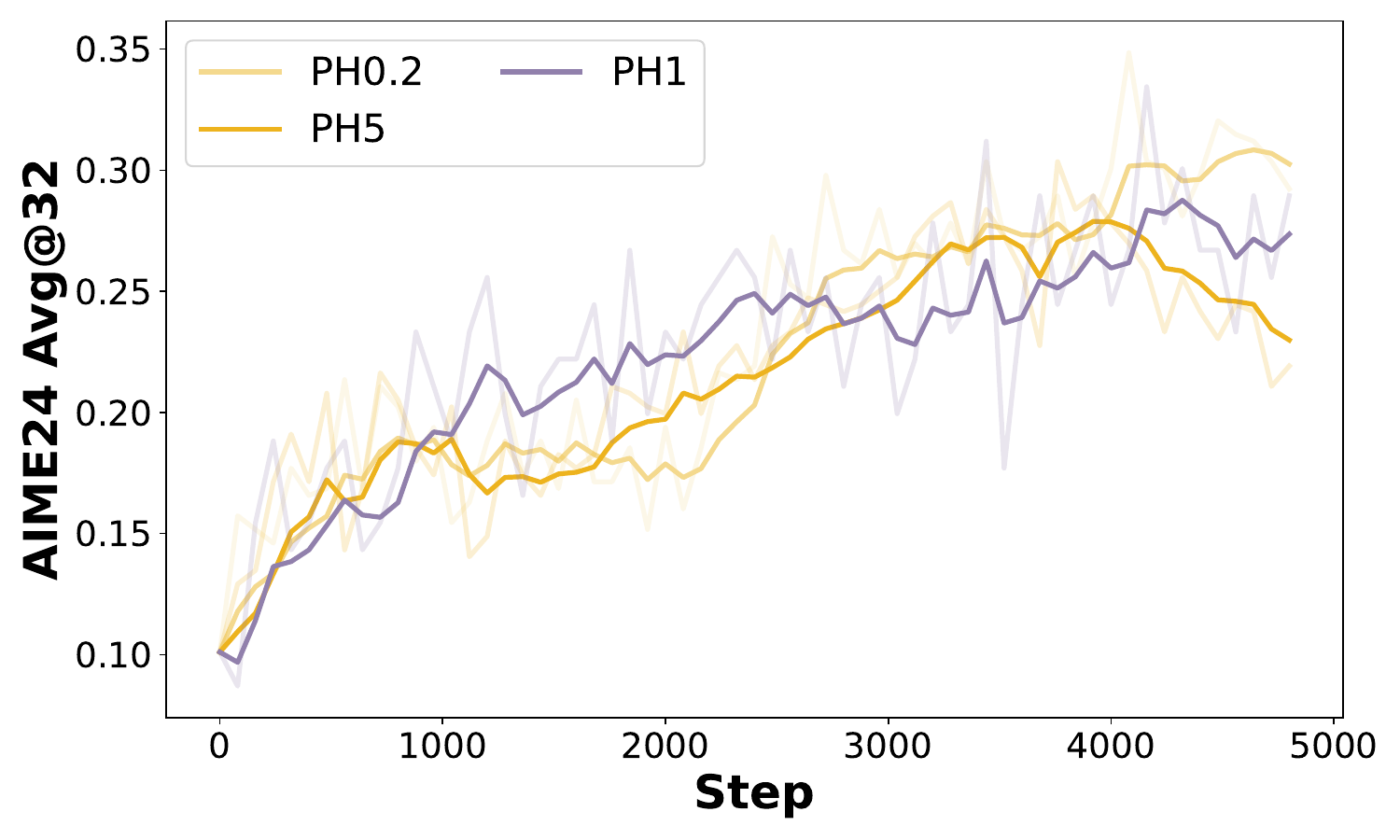}
        \caption{AIME24 Avg@32}
    \end{subfigure}
    \begin{subfigure}[b]{0.32\linewidth}
        \centering
        \includegraphics[width=\linewidth]{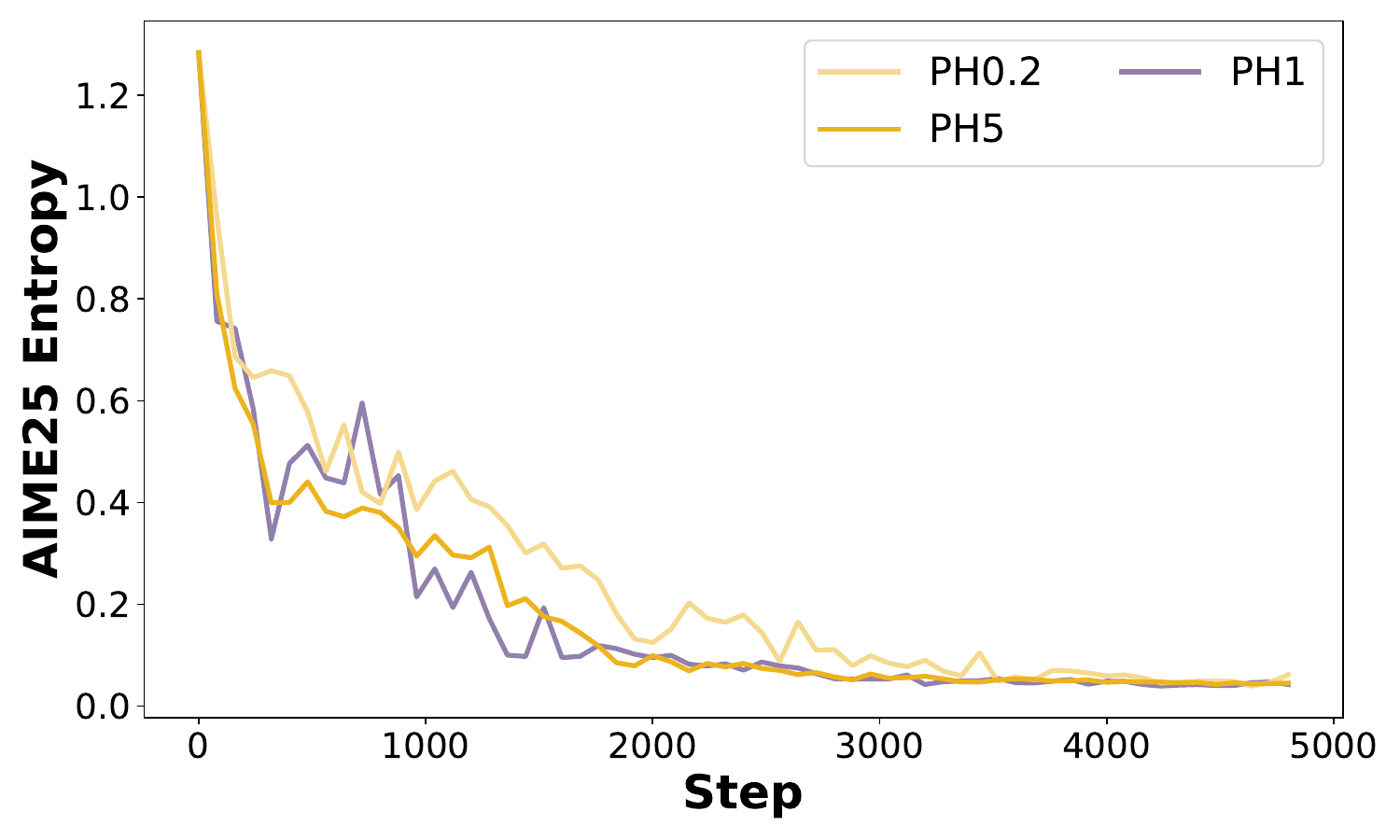}
        \caption{AIME25 Entropy}
    \end{subfigure}
    \begin{subfigure}[b]{0.32\linewidth}
        \centering
        \includegraphics[width=\linewidth]{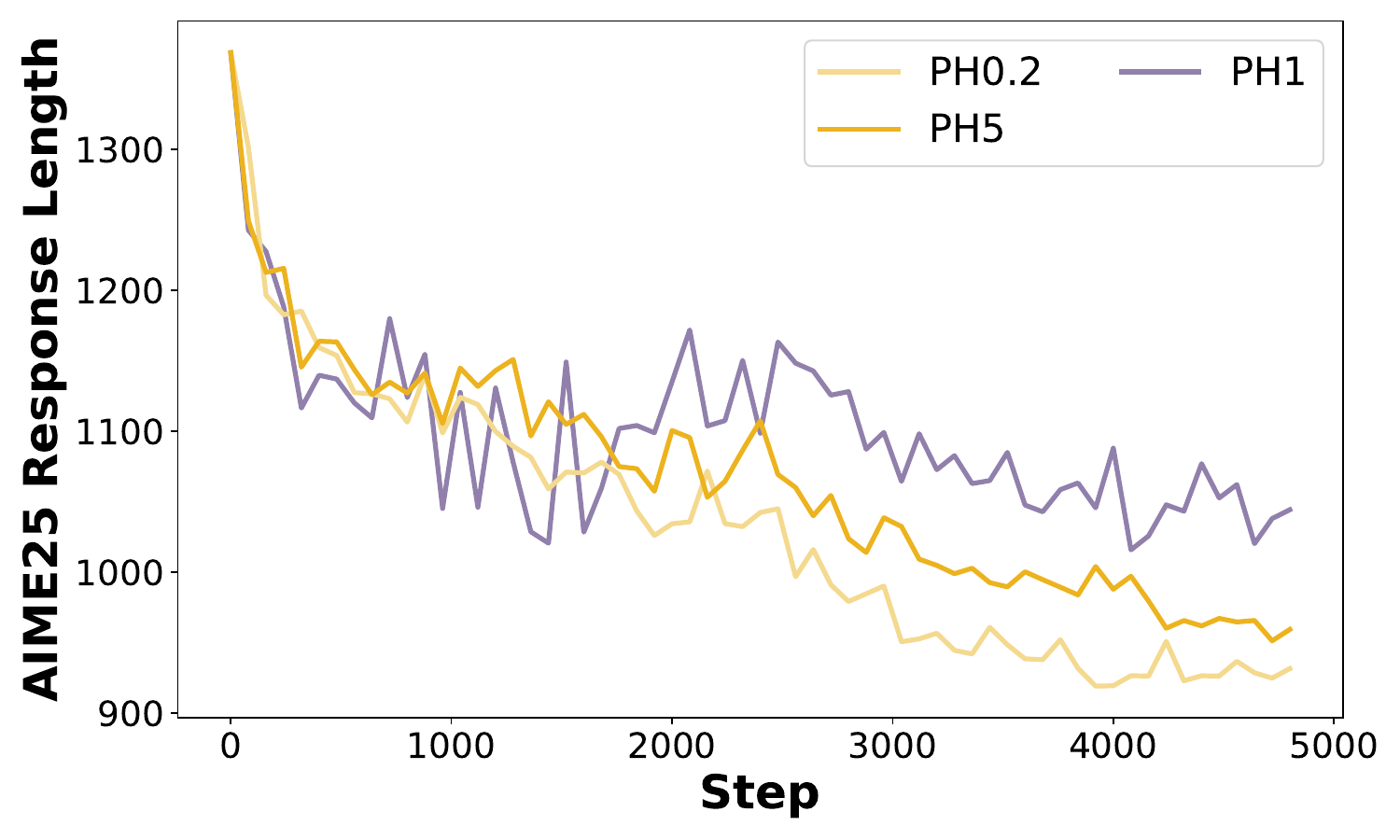}
        \caption{AIME25 Length}
    \end{subfigure}
    \begin{subfigure}[b]{0.32\linewidth}
        \centering
        \includegraphics[width=\linewidth]{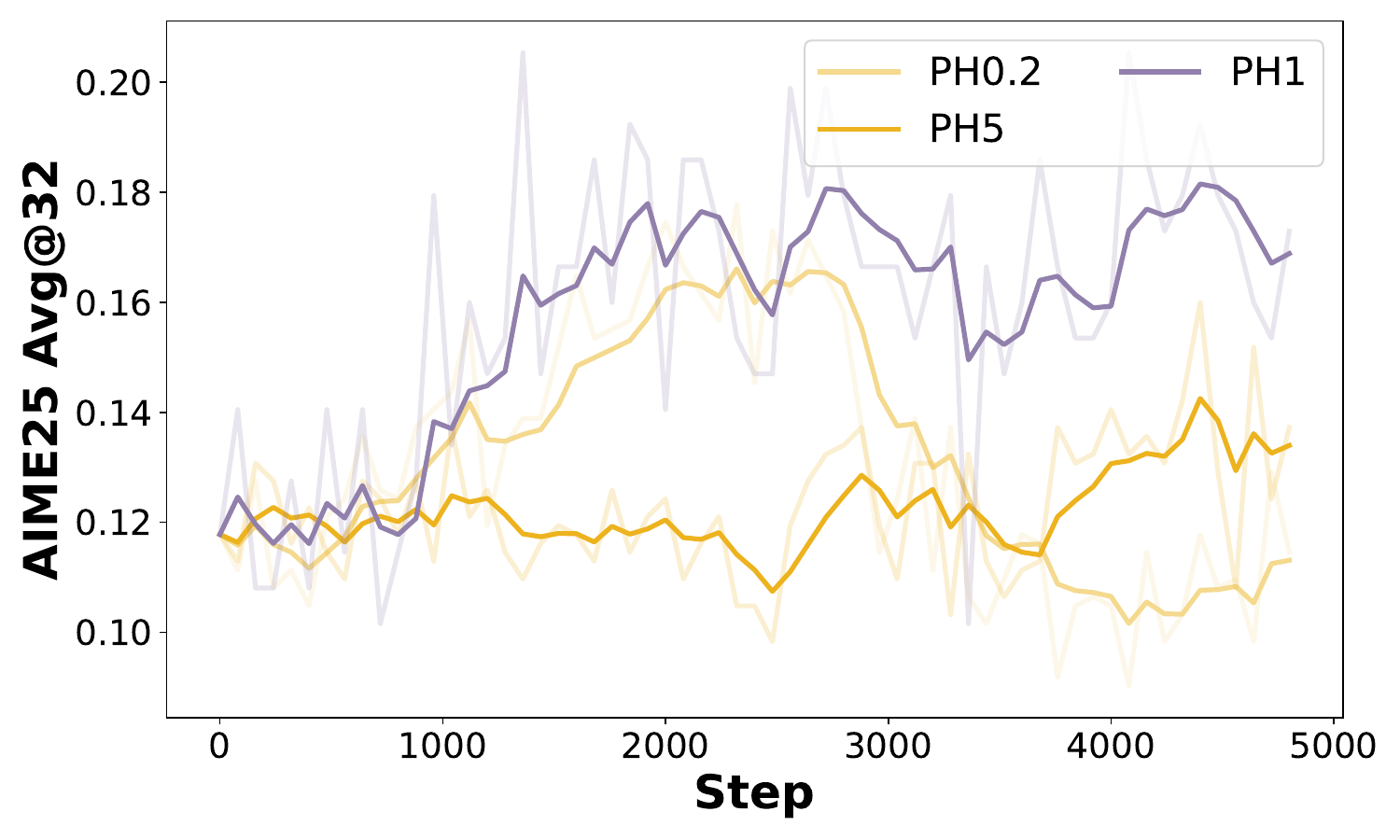}
        \caption{AIME25 Avg@32}
    \end{subfigure}
    \caption{RLVR training dynamics on positive high probability token advantage shaping.}
\label{fig:token-prob-ph-training_dynamic}
\end{figure*}

\begin{figure*}[t]
    \centering
    \begin{subfigure}[b]{0.32\linewidth}
        \centering
        \includegraphics[width=\linewidth]{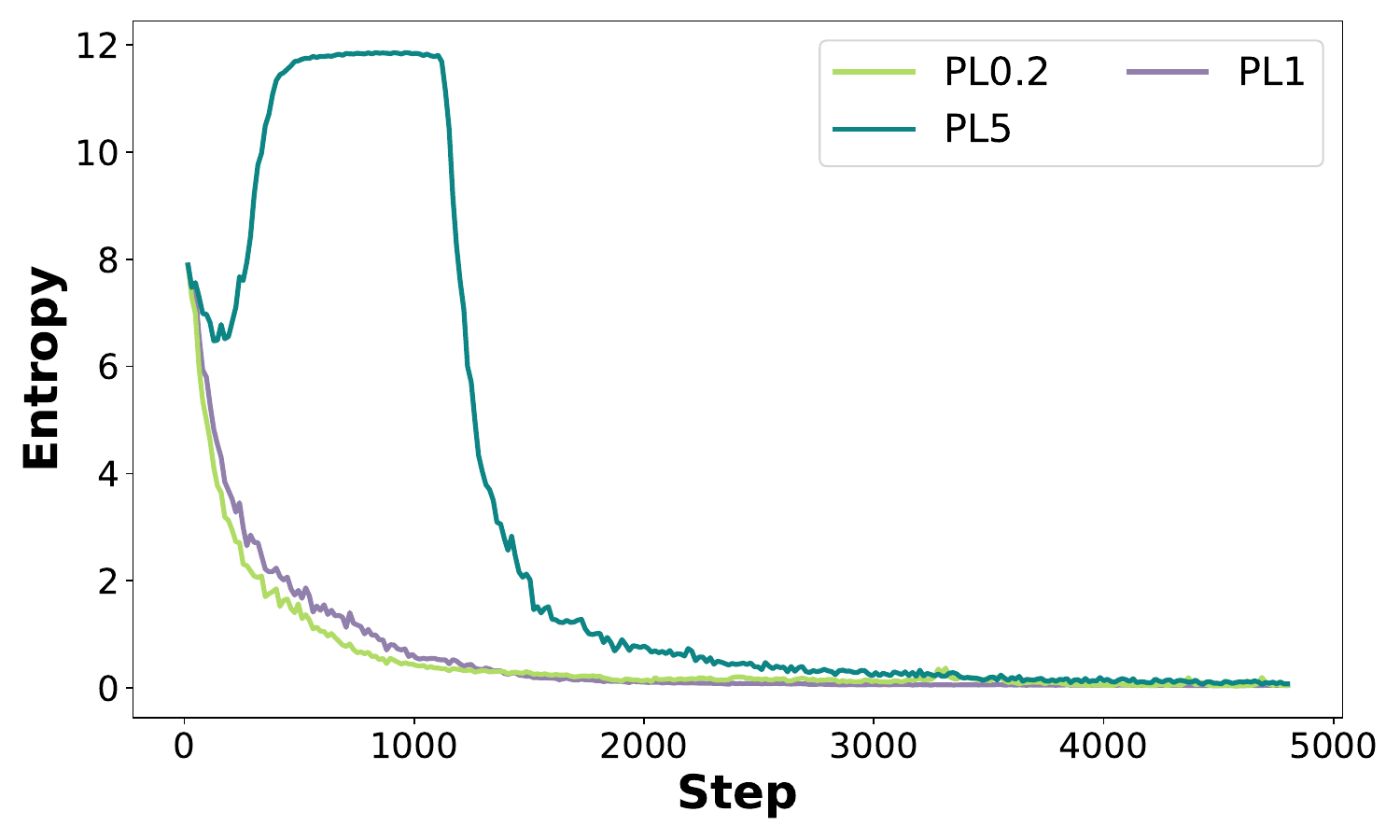}
        \caption{Entropy}
    \end{subfigure}
    \begin{subfigure}[b]{0.32\linewidth}
        \centering
        \includegraphics[width=\linewidth]{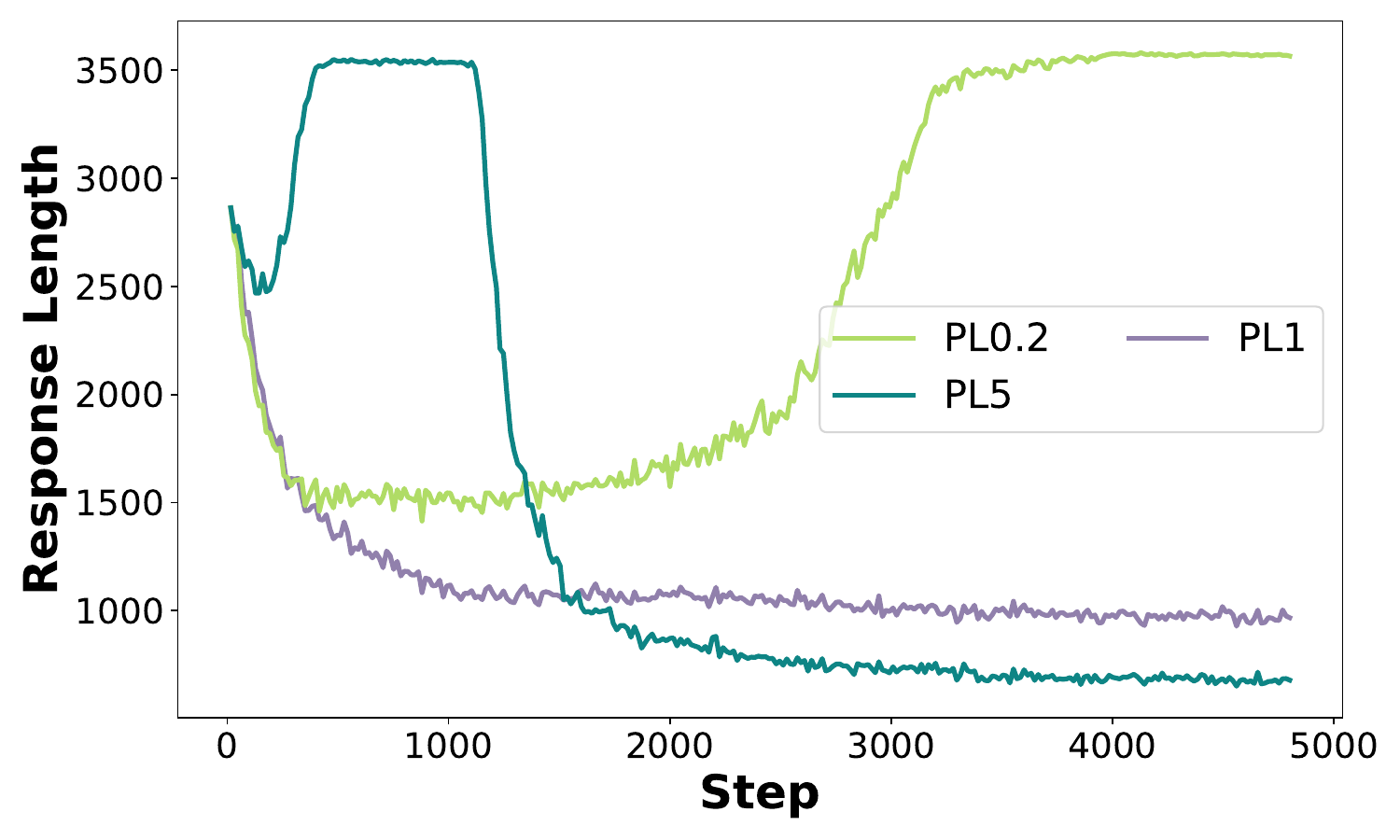}
        \caption{Length}
    \end{subfigure}
    \begin{subfigure}[b]{0.32\linewidth}
        \centering
        \includegraphics[width=\linewidth]{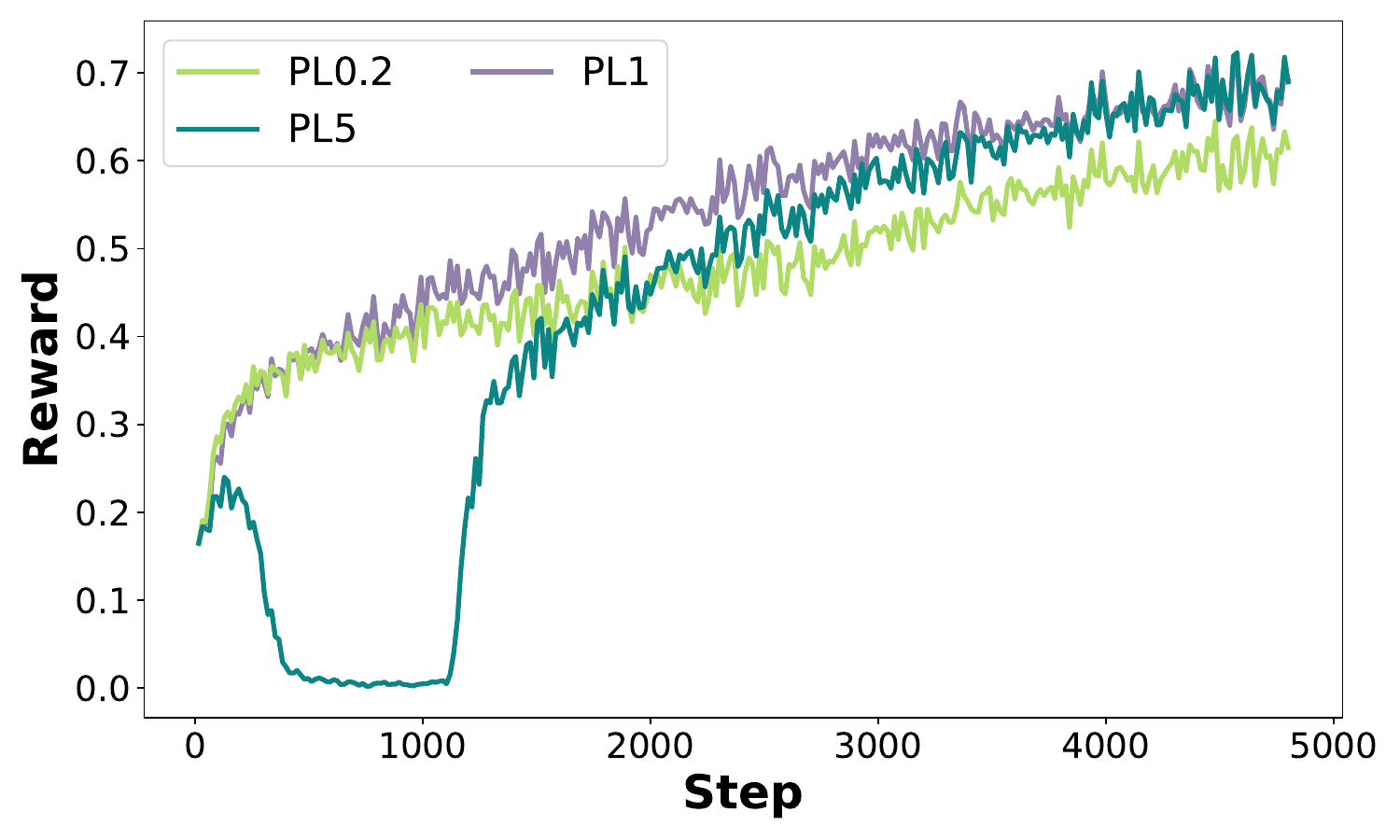}
        \caption{Reward}
    \end{subfigure}
    \begin{subfigure}[b]{0.32\linewidth}
        \centering
        \includegraphics[width=\linewidth]{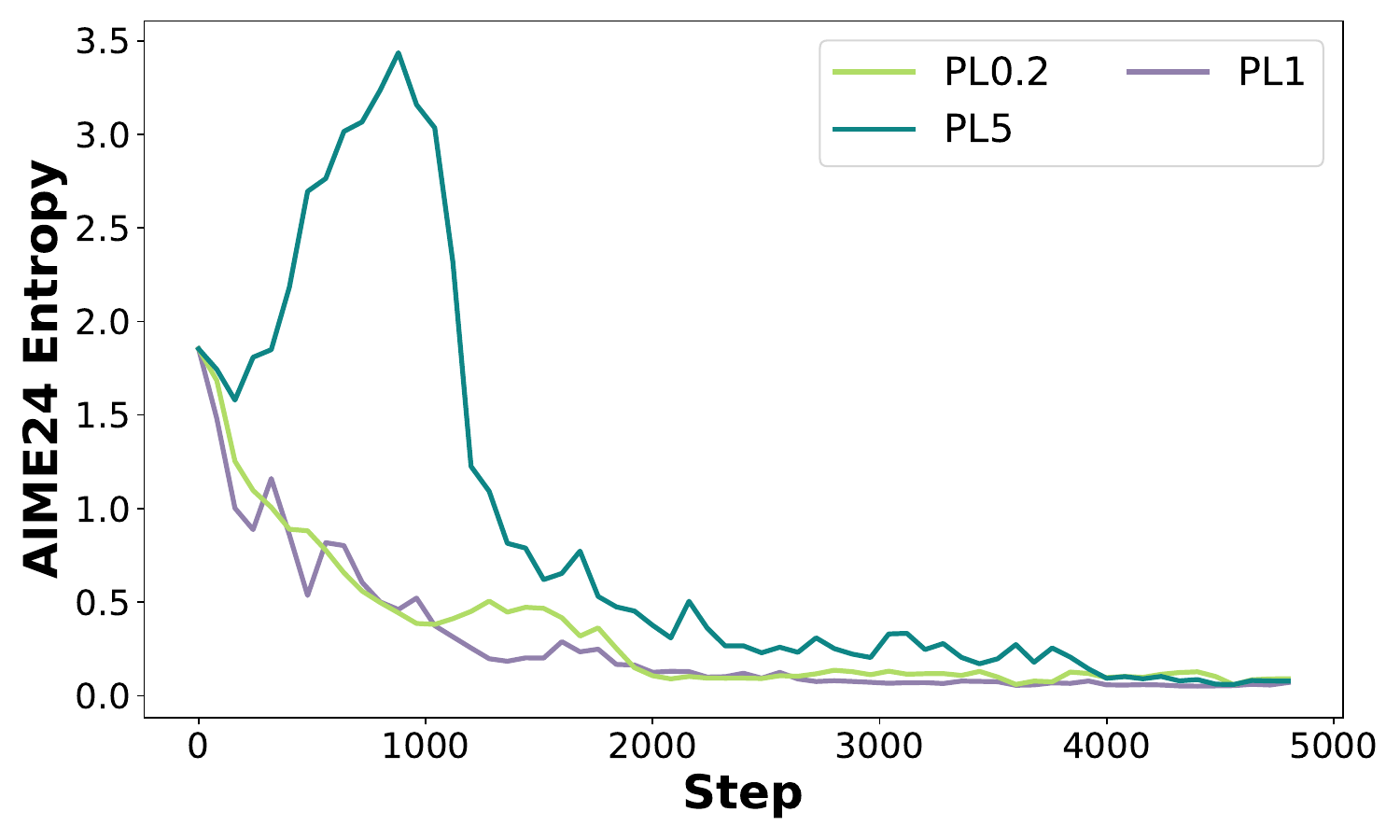}
        \caption{AIME24 Entropy}
    \end{subfigure}
    \begin{subfigure}[b]{0.32\linewidth}
        \centering
        \includegraphics[width=\linewidth]{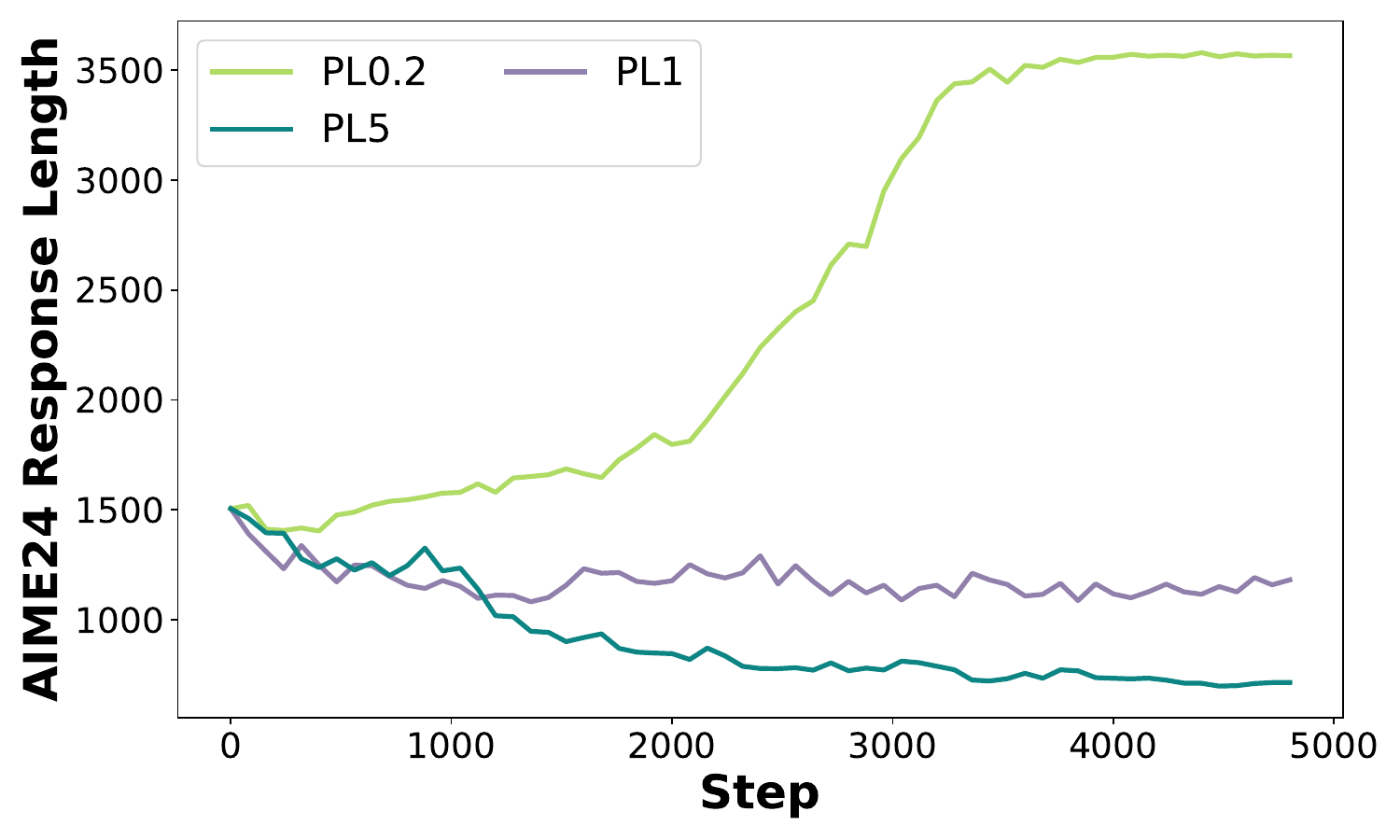}
        \caption{AIME24 Length}
    \end{subfigure}
    \begin{subfigure}[b]{0.32\linewidth}
        \centering
        \includegraphics[width=\linewidth]{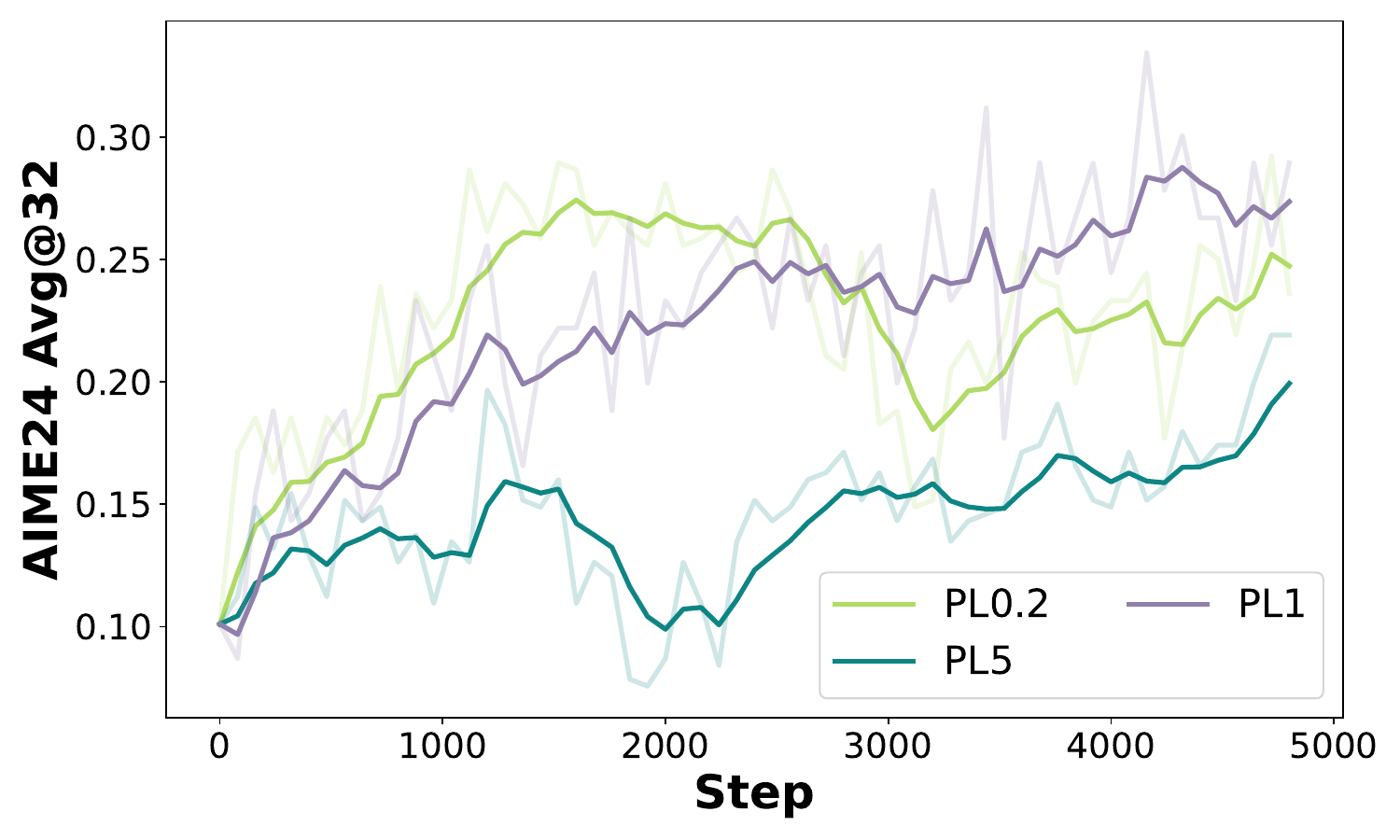}
        \caption{AIME24 Avg@32}
    \end{subfigure}
    \begin{subfigure}[b]{0.32\linewidth}
        \centering
        \includegraphics[width=\linewidth]{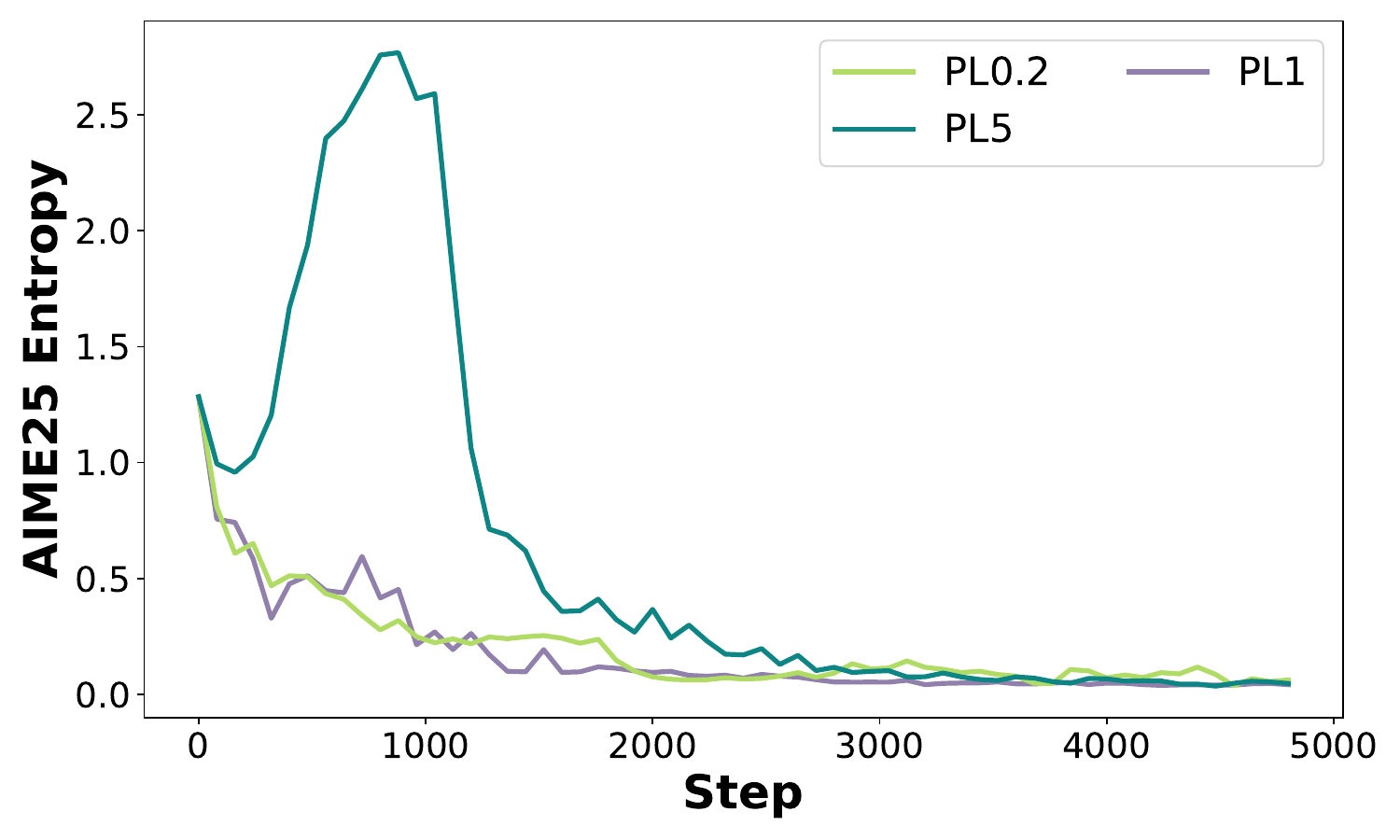}
        \caption{AIME25 Entropy}
    \end{subfigure}
    \begin{subfigure}[b]{0.32\linewidth}
        \centering
        \includegraphics[width=\linewidth]{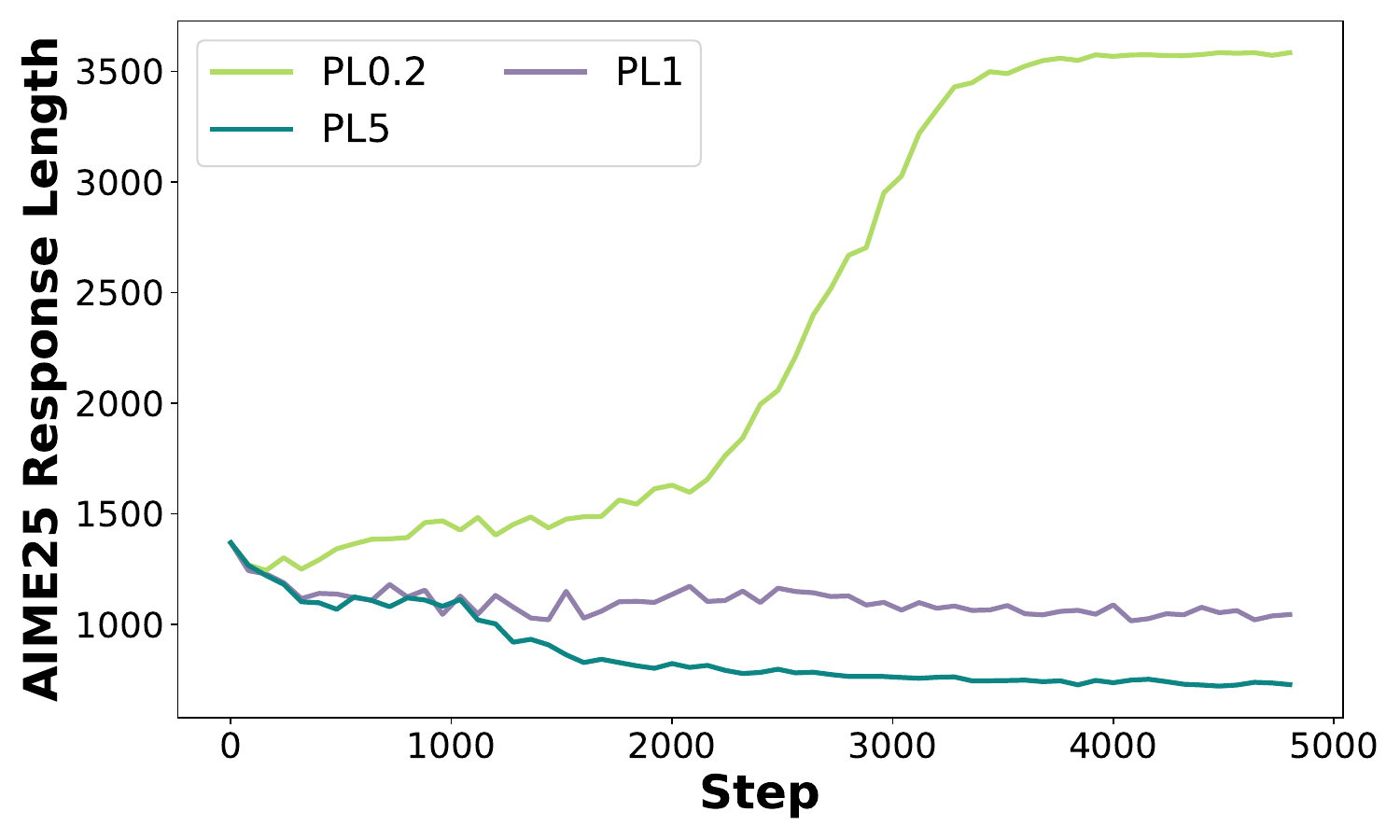}
        \caption{AIME25 Length}
    \end{subfigure}
    \begin{subfigure}[b]{0.32\linewidth}
        \centering
        \includegraphics[width=\linewidth]{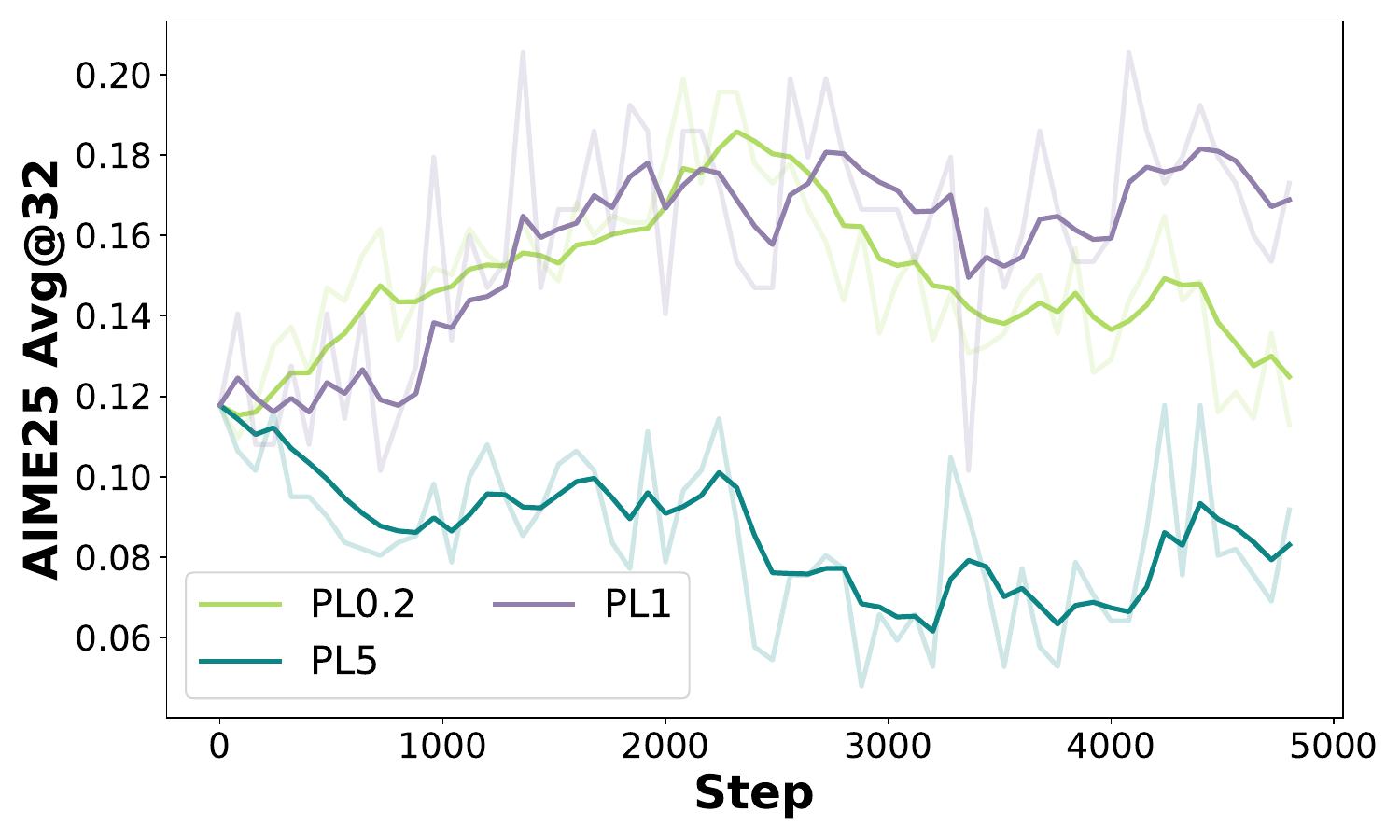}
        \caption{AIME25 Avg@32}
    \end{subfigure}
    \caption{RLVR training dynamics on positive low probability token advantage shaping.}
\label{fig:token-prob-pl-training_dynamic}
\end{figure*}

\begin{figure*}[t]
    \centering
    \begin{subfigure}[b]{0.32\linewidth}
        \centering
        \includegraphics[width=\linewidth]{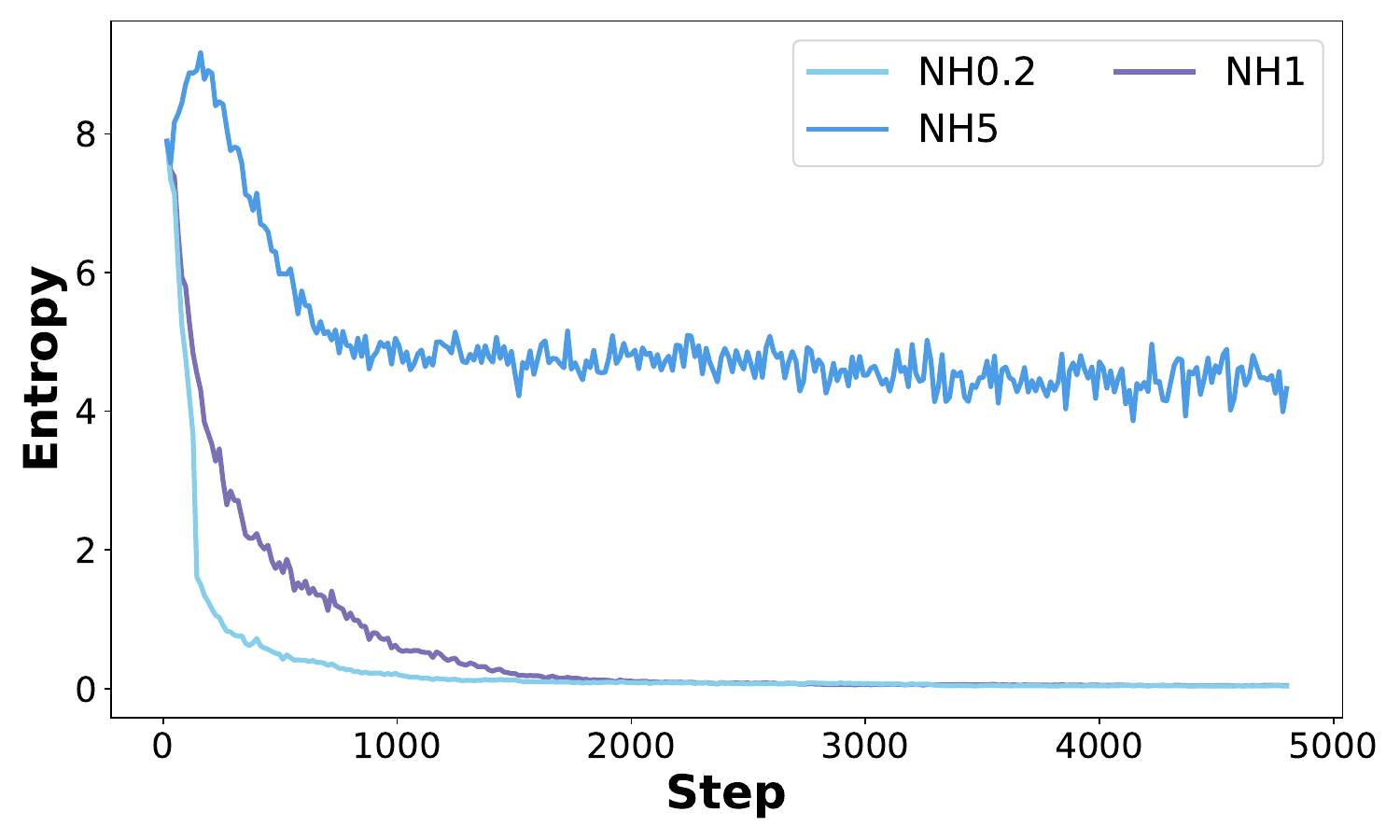}
        \caption{Entropy}
    \end{subfigure}
    \begin{subfigure}[b]{0.32\linewidth}
        \centering
        \includegraphics[width=\linewidth]{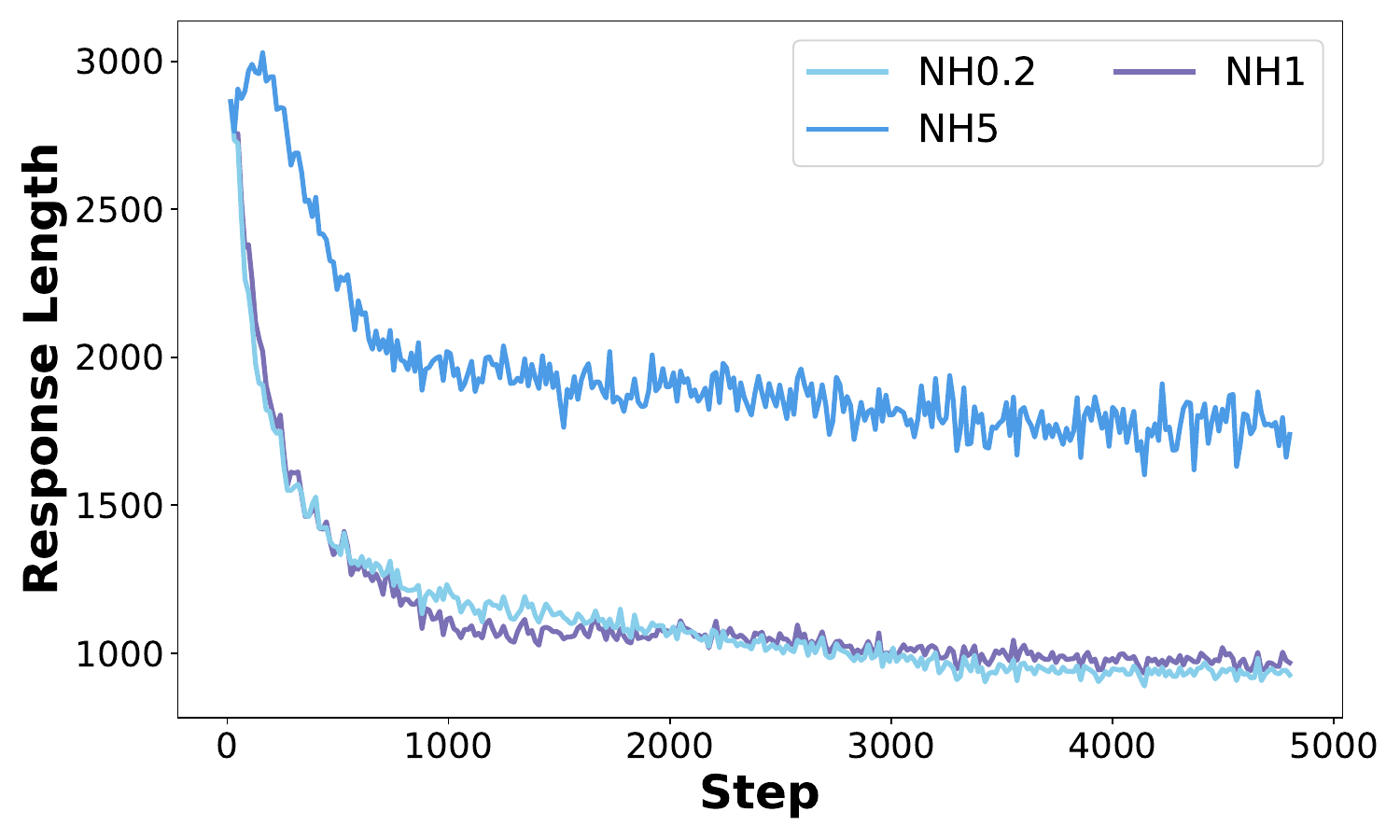}
        \caption{Length}
    \end{subfigure}
    \begin{subfigure}[b]{0.32\linewidth}
        \centering
        \includegraphics[width=\linewidth]{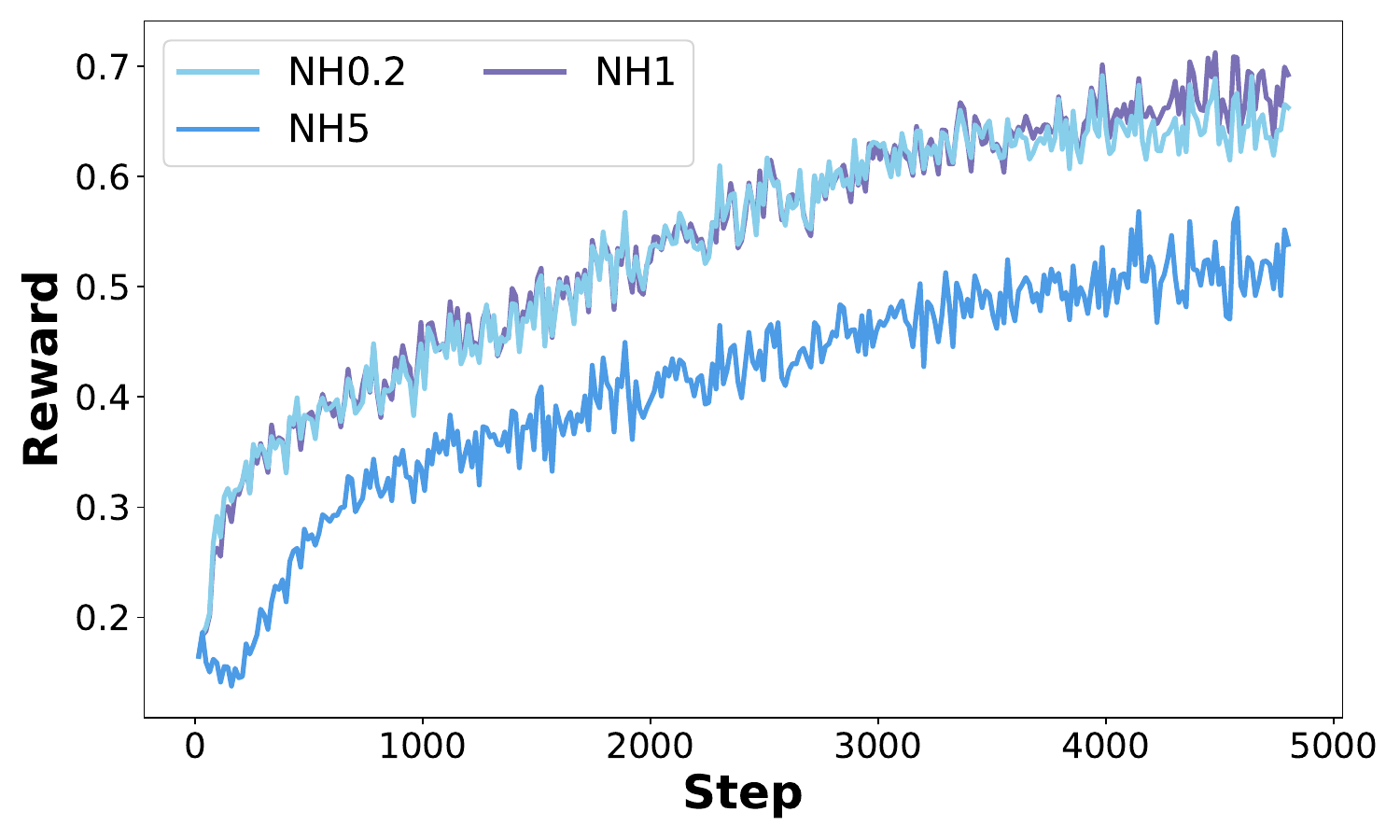}
        \caption{Reward}
    \end{subfigure}
    \begin{subfigure}[b]{0.32\linewidth}
        \centering
        \includegraphics[width=\linewidth]{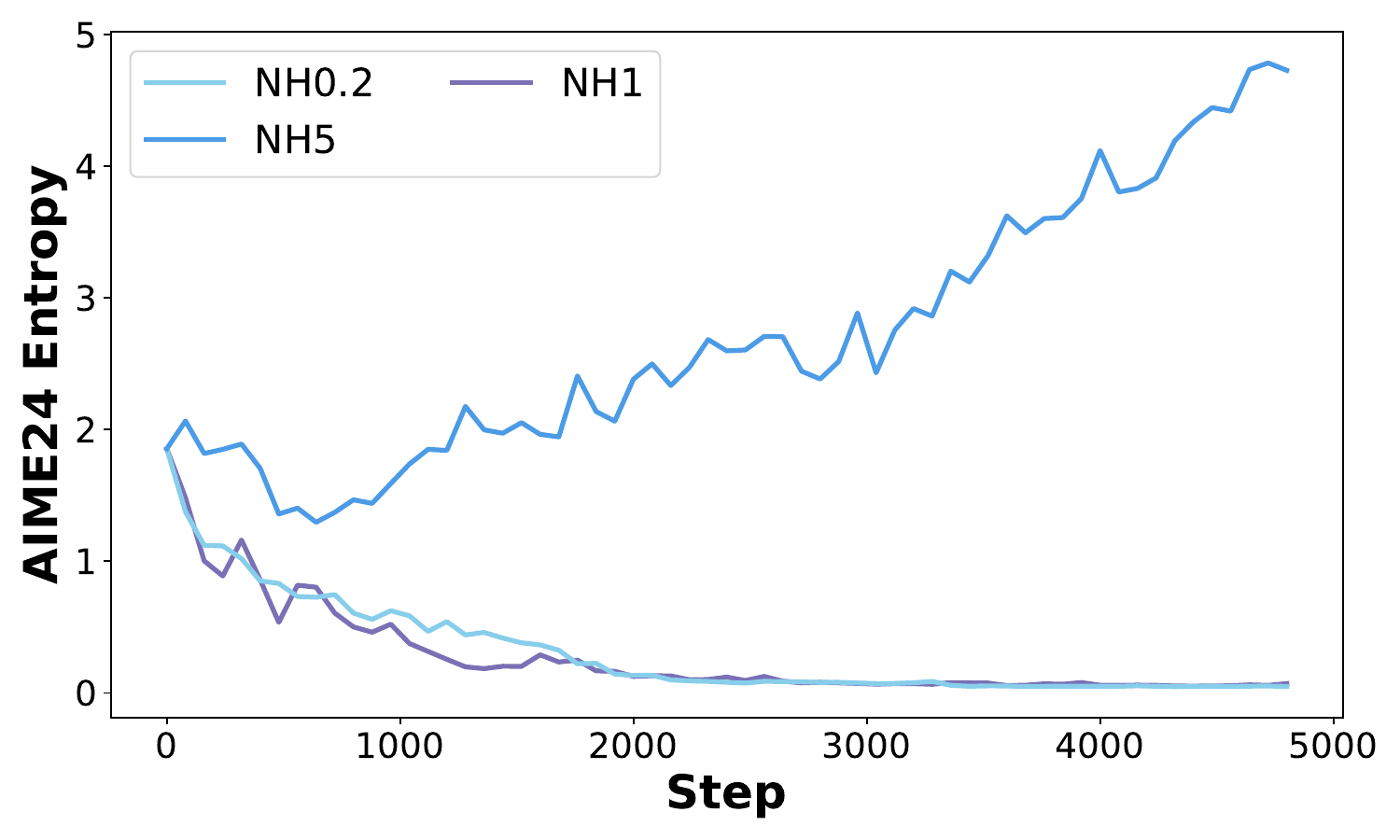}
        \caption{AIME24 Entropy}
    \end{subfigure}
    \begin{subfigure}[b]{0.32\linewidth}
        \centering
        \includegraphics[width=\linewidth]{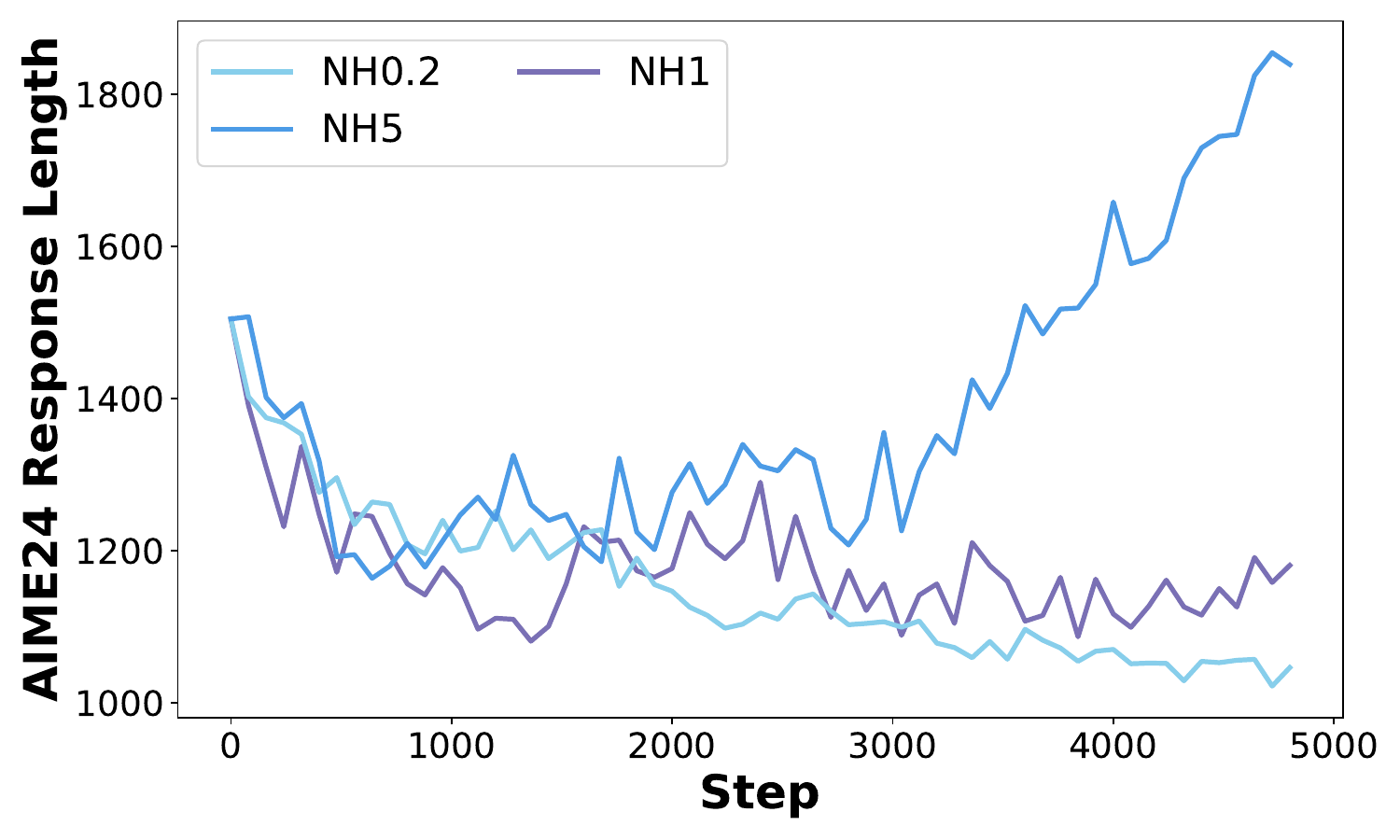}
        \caption{AIME24 Length}
    \end{subfigure}
    \begin{subfigure}[b]{0.32\linewidth}
        \centering
        \includegraphics[width=\linewidth]{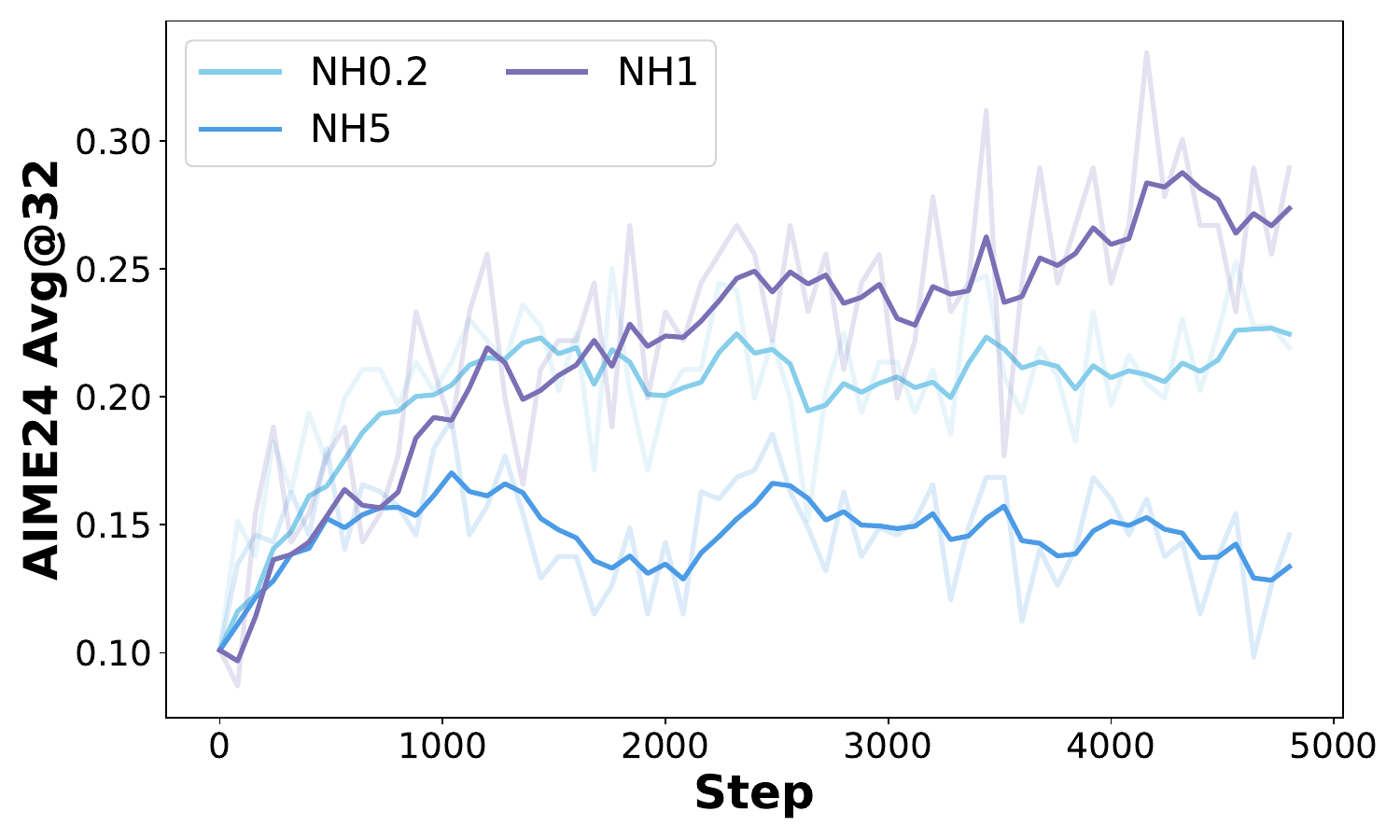}
        \caption{AIME24 Avg@32}
    \end{subfigure}
    \begin{subfigure}[b]{0.32\linewidth}
        \centering
        \includegraphics[width=\linewidth]{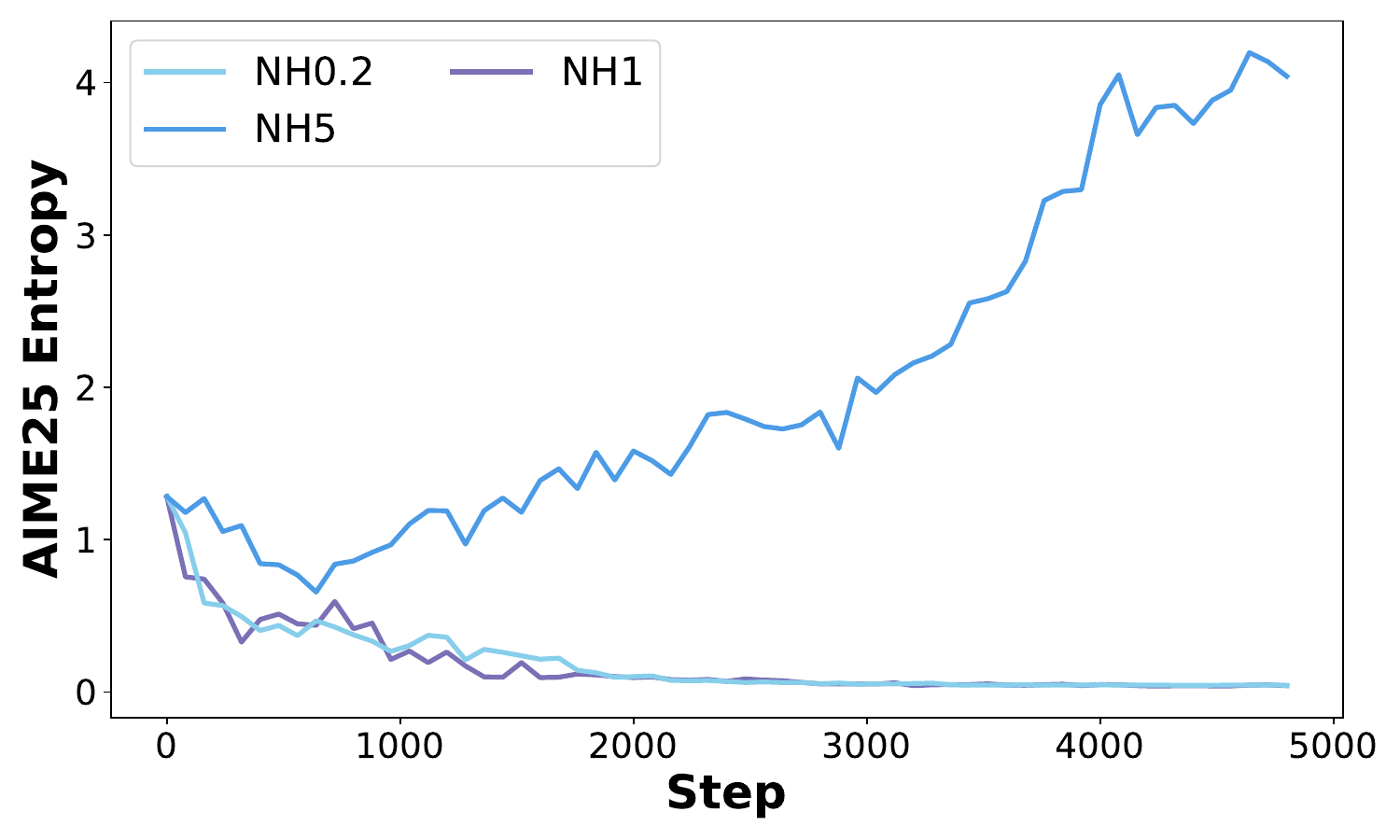}
        \caption{AIME25 Entropy}
    \end{subfigure}
    \begin{subfigure}[b]{0.32\linewidth}
        \centering
        \includegraphics[width=\linewidth]{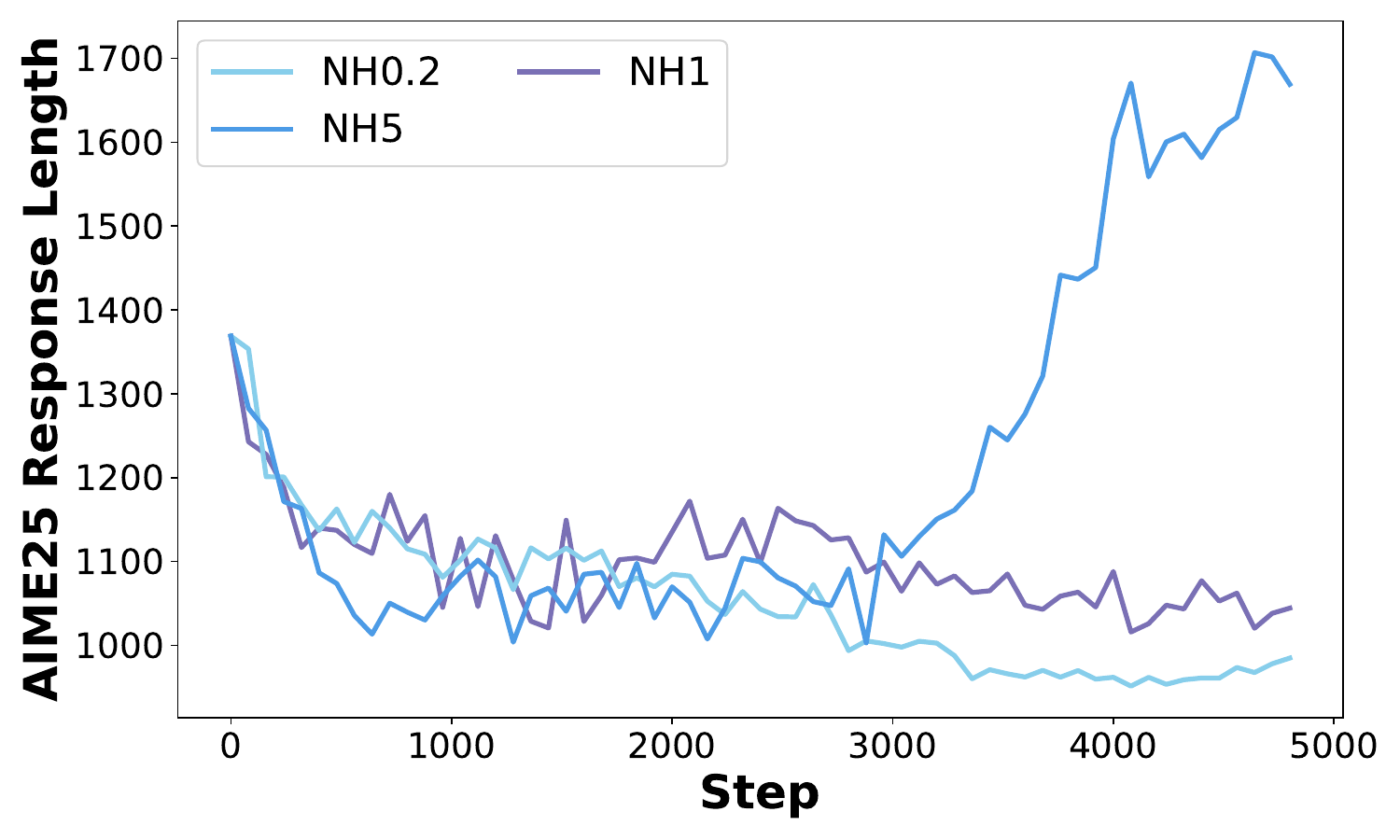}
        \caption{AIME25 Length}
    \end{subfigure}
    \begin{subfigure}[b]{0.32\linewidth}
        \centering
        \includegraphics[width=\linewidth]{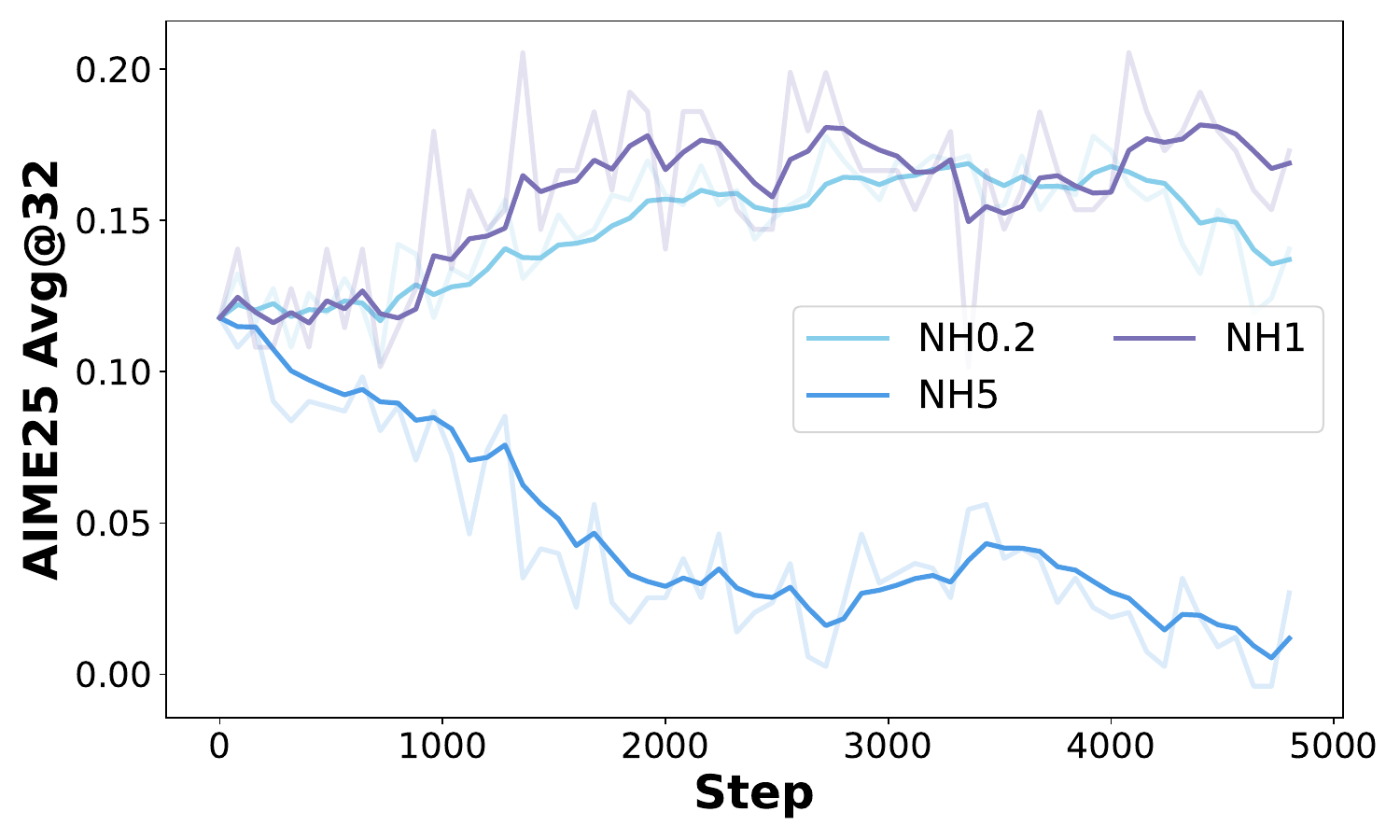}
        \caption{AIME25 Avg@32}
    \end{subfigure}
    \caption{RLVR training dynamics on negative high probability token advantage shaping.}
\label{fig:token-prob-nh-training_dynamic}
\end{figure*}

\begin{figure*}[t]
    \centering
    \begin{subfigure}[b]{0.32\linewidth}
        \centering
        \includegraphics[width=\linewidth]{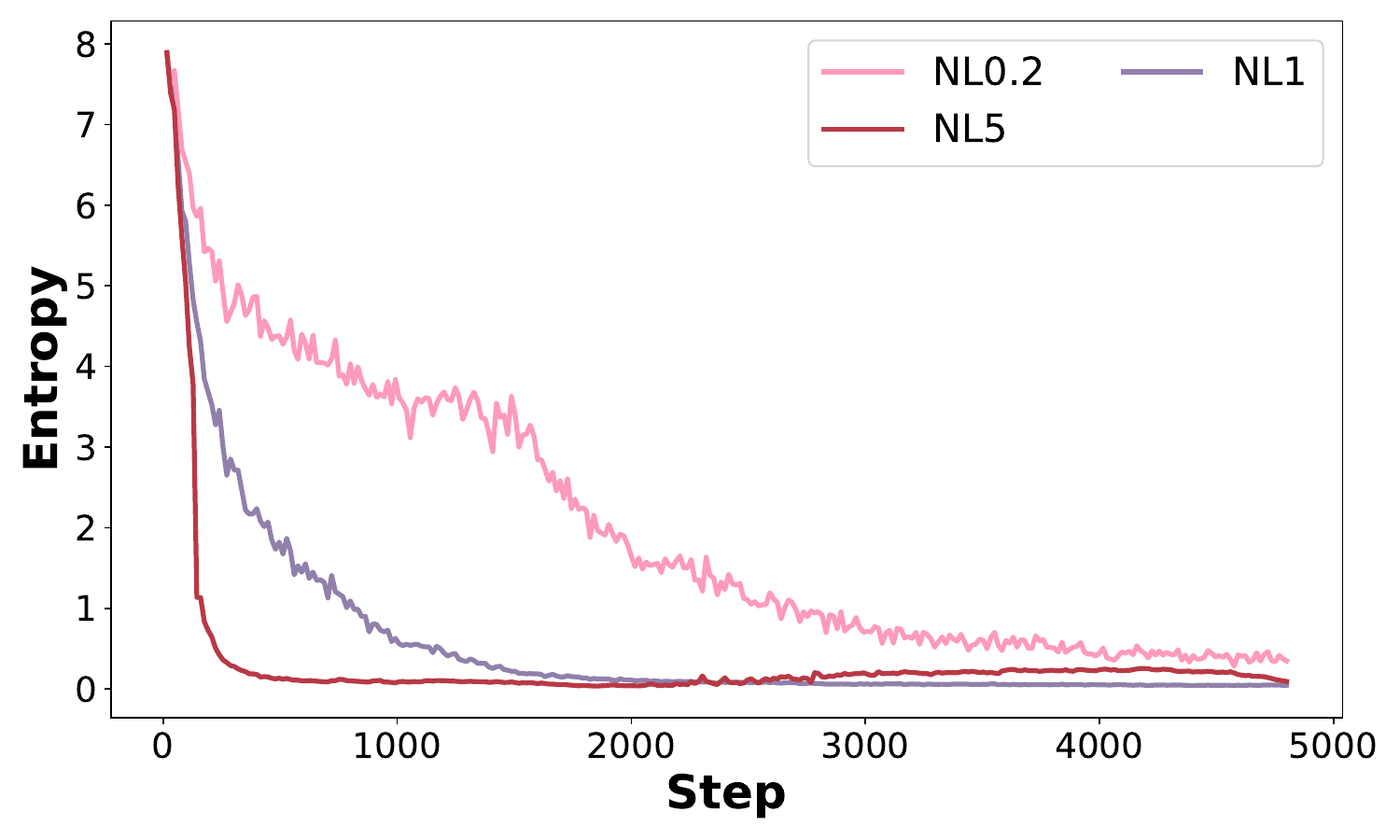}
        \caption{Entropy}
    \end{subfigure}
    \begin{subfigure}[b]{0.32\linewidth}
        \centering
        \includegraphics[width=\linewidth]{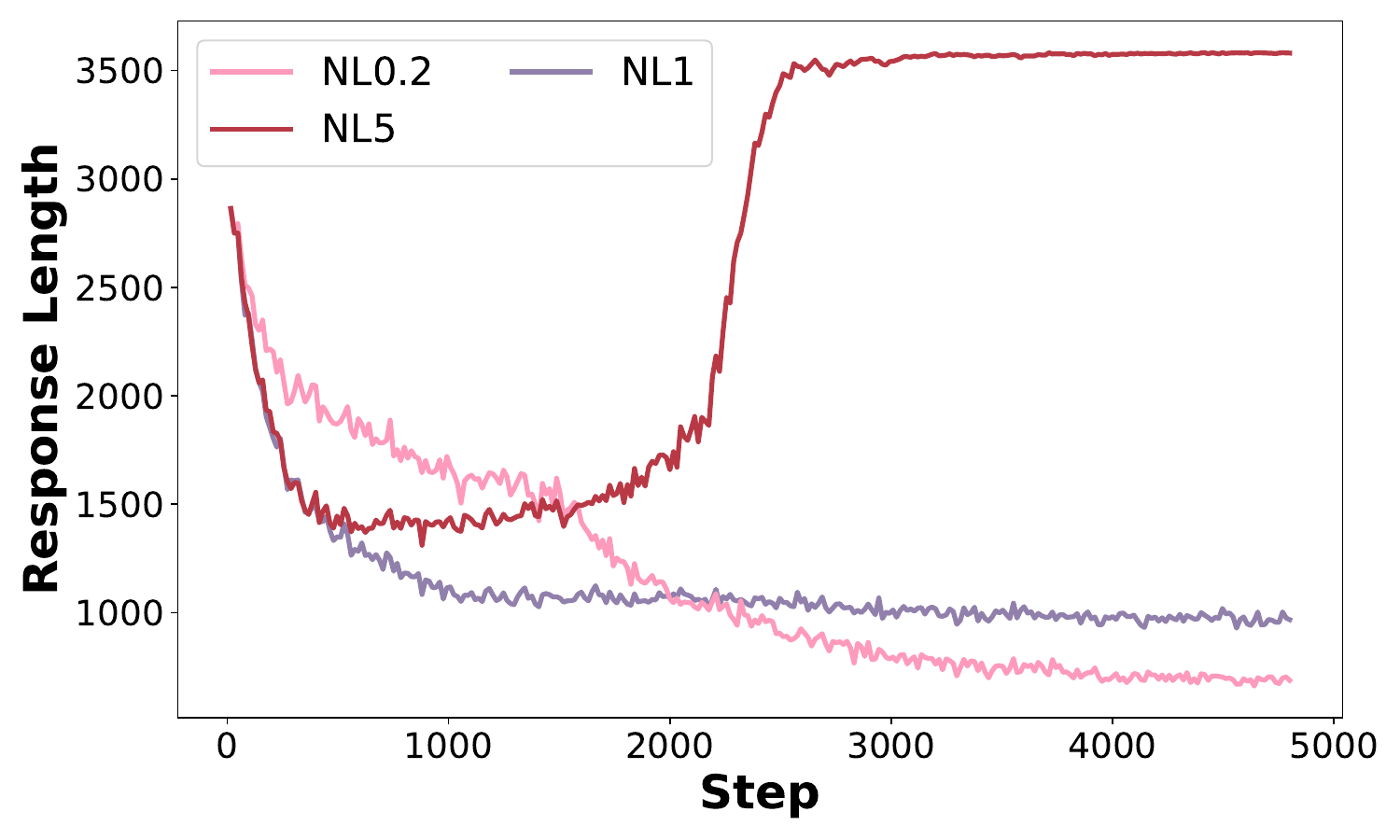}
        \caption{Length}
    \end{subfigure}
    \begin{subfigure}[b]{0.32\linewidth}
        \centering
        \includegraphics[width=\linewidth]{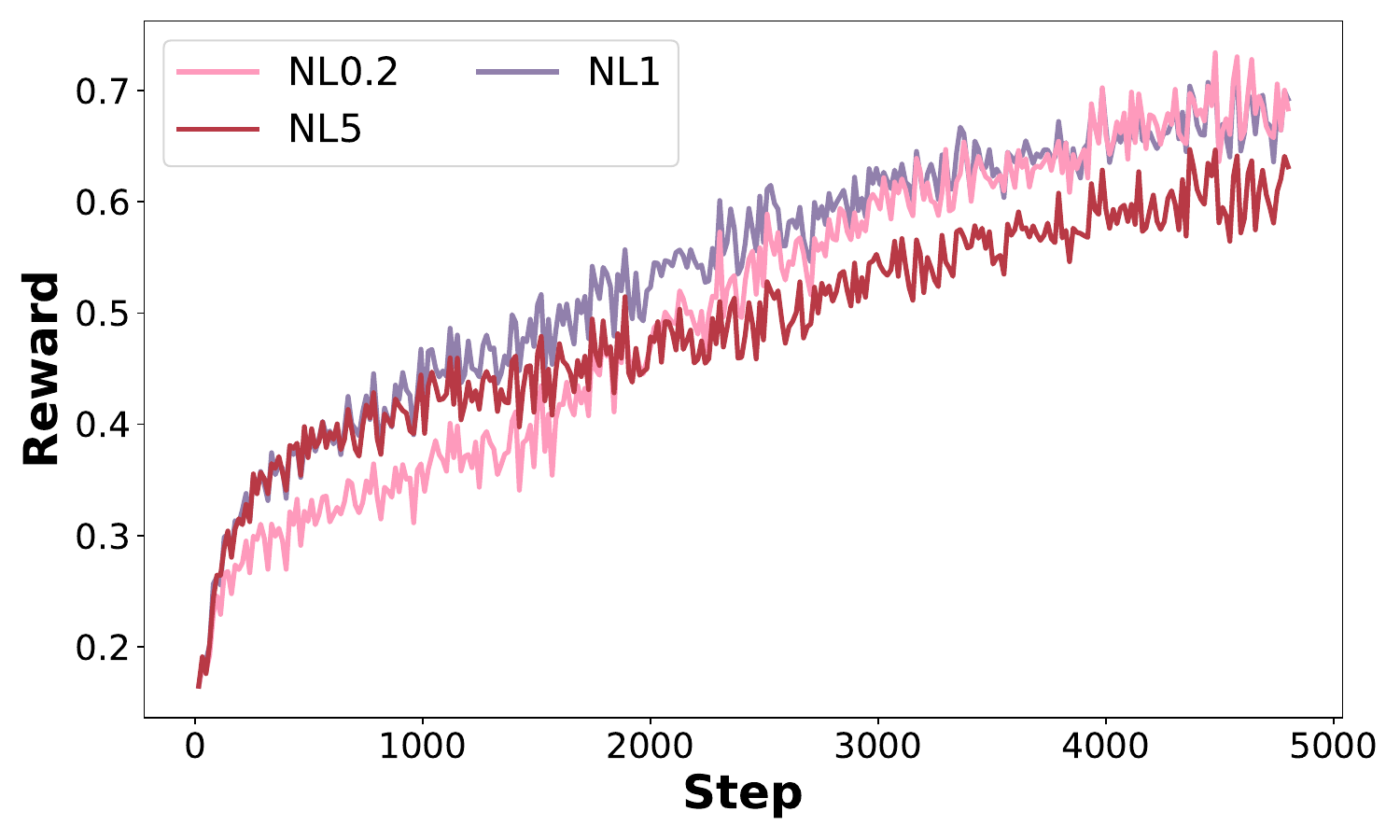}
        \caption{Reward}
    \end{subfigure}
    \begin{subfigure}[b]{0.32\linewidth}
        \centering
        \includegraphics[width=\linewidth]{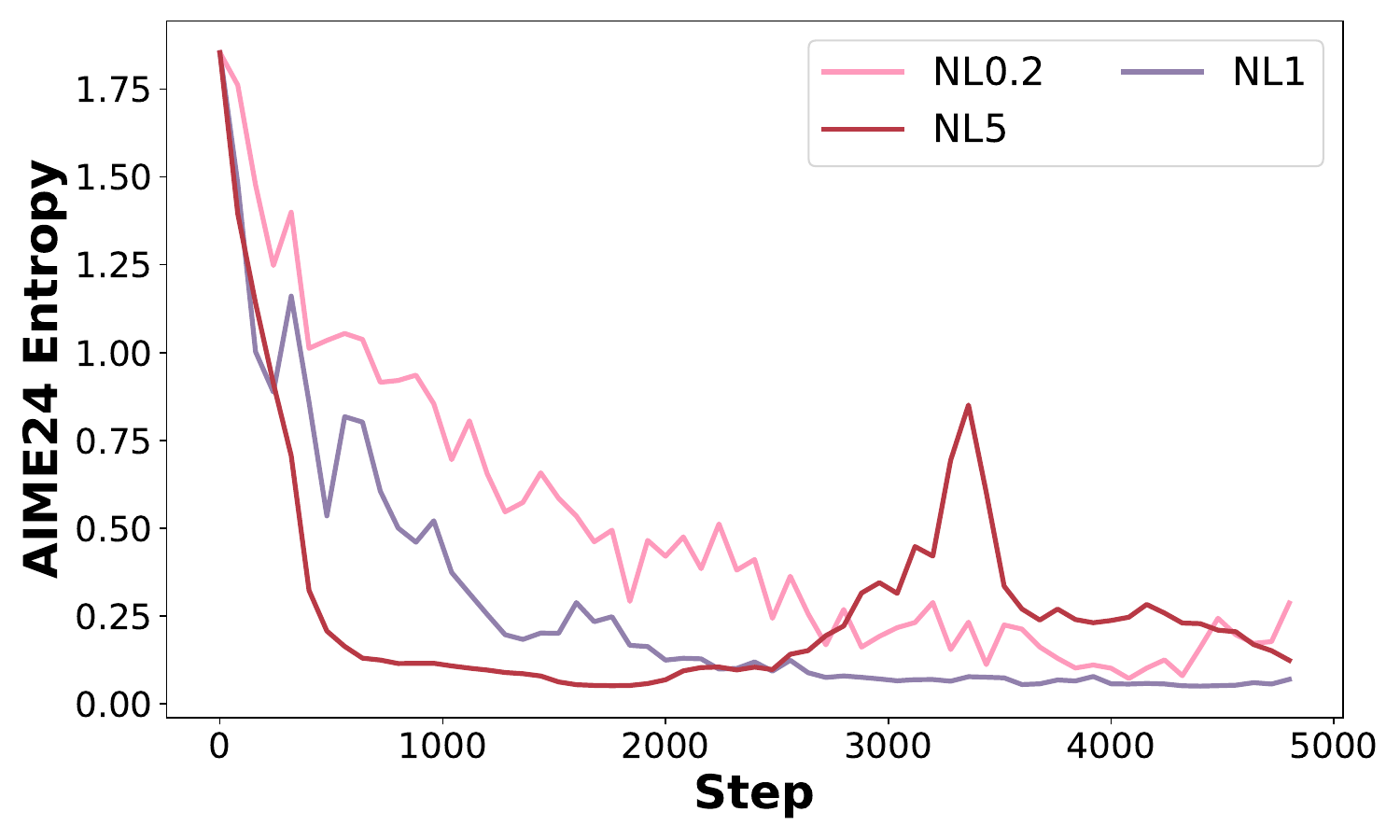}
        \caption{AIME24 Entropy}
    \end{subfigure}
    \begin{subfigure}[b]{0.32\linewidth}
        \centering
        \includegraphics[width=\linewidth]{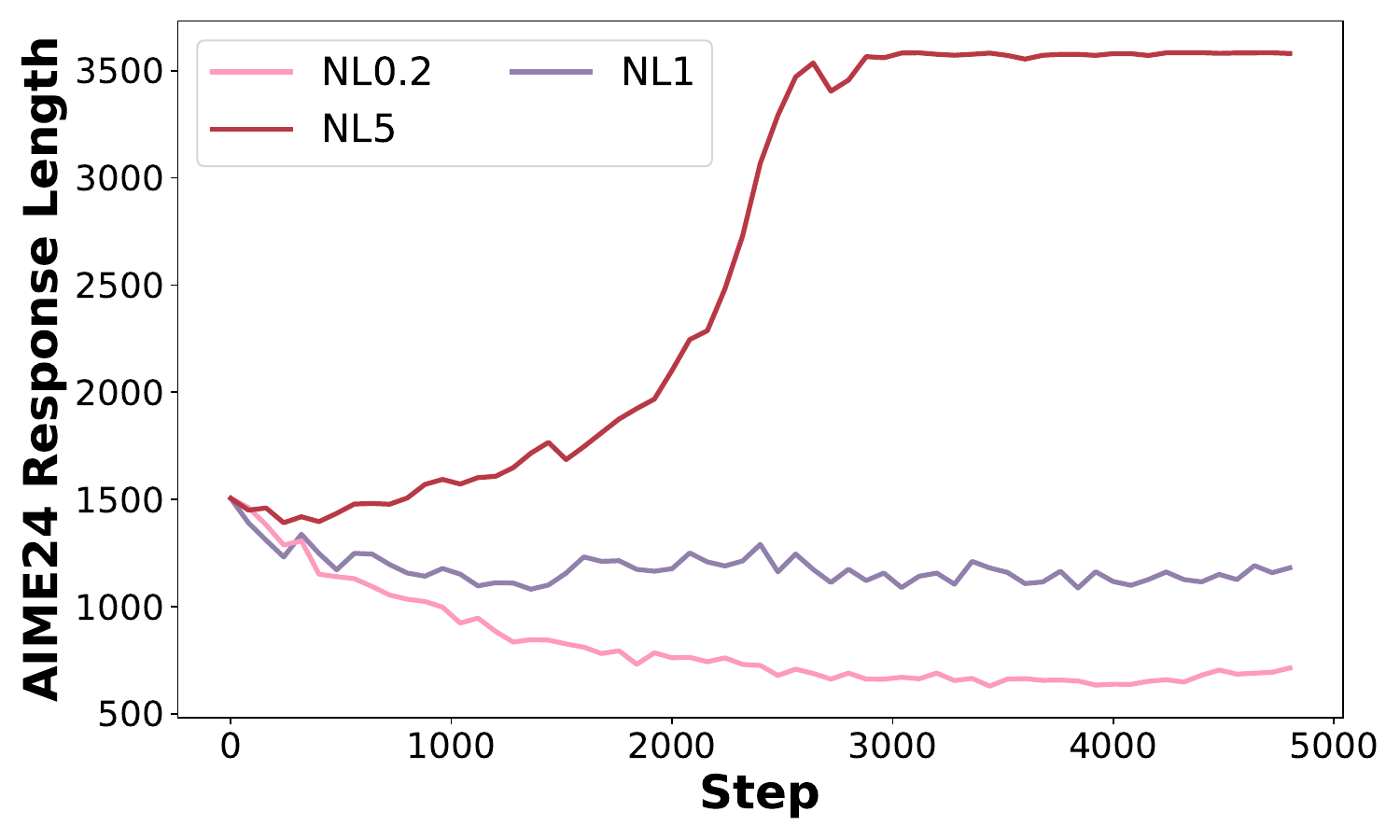}
        \caption{AIME24 Length}
    \end{subfigure}
    \begin{subfigure}[b]{0.32\linewidth}
        \centering
        \includegraphics[width=\linewidth]{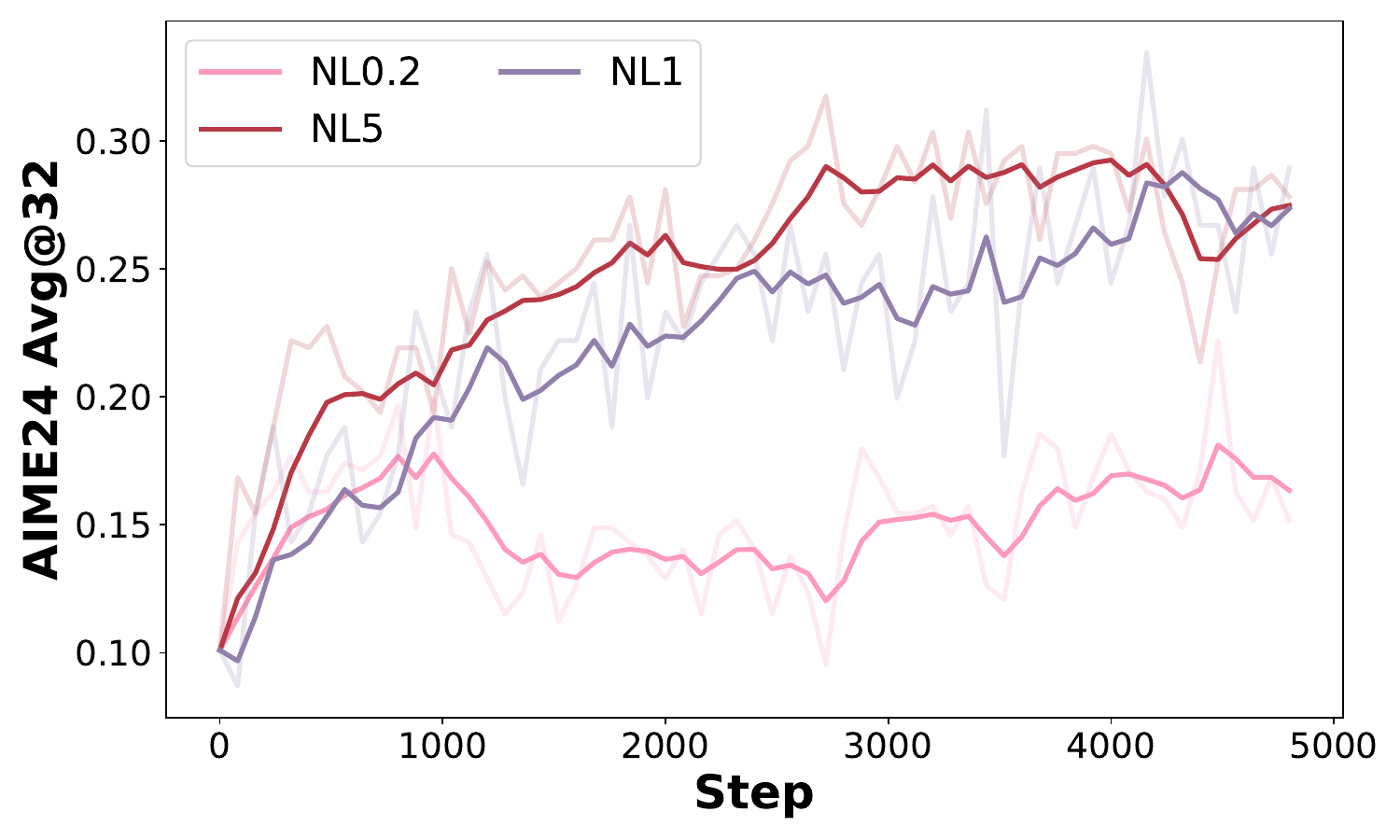}
        \caption{AIME24 Avg@32}
    \end{subfigure}
    \begin{subfigure}[b]{0.32\linewidth}
        \centering
        \includegraphics[width=\linewidth]{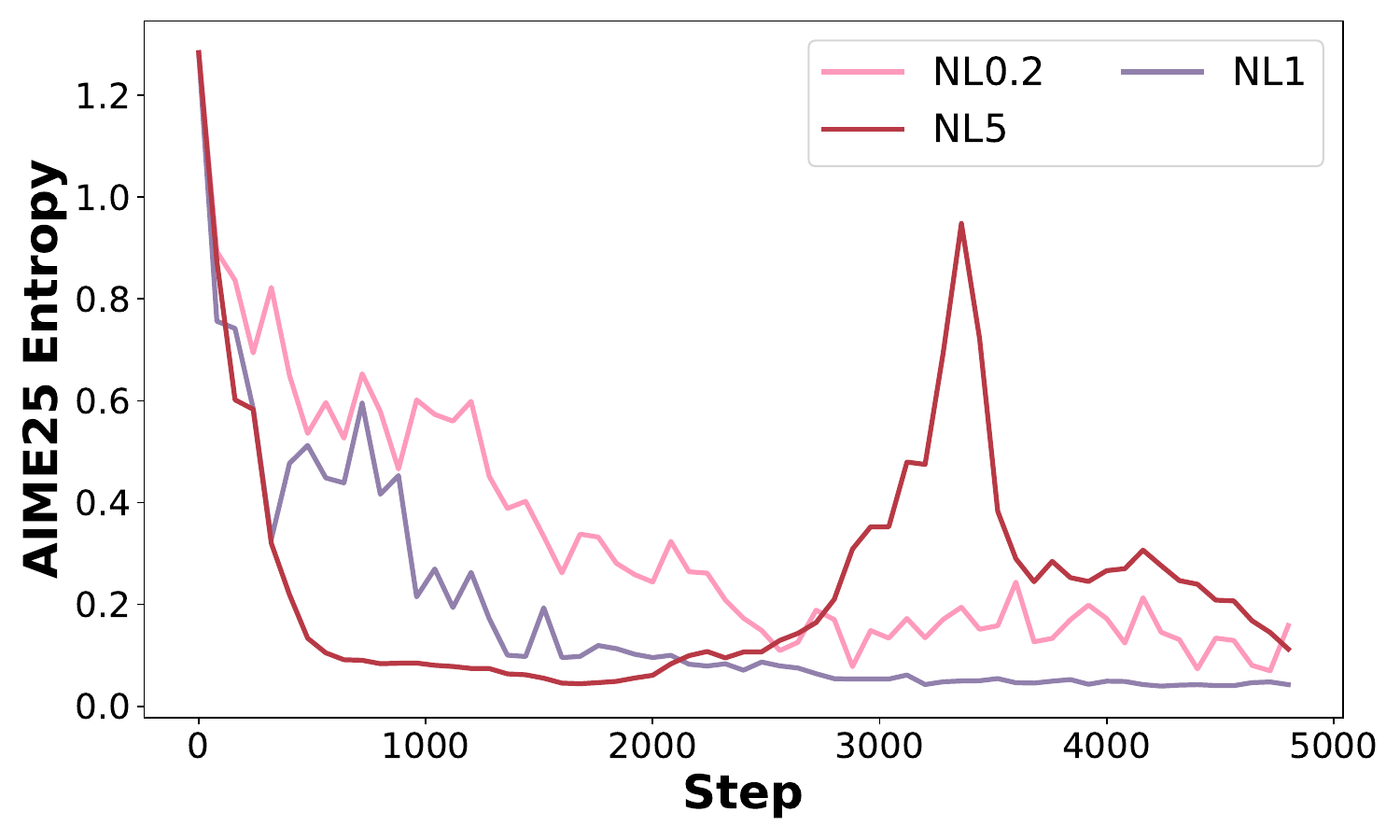}
        \caption{AIME25 Entropy}
    \end{subfigure}
    \begin{subfigure}[b]{0.32\linewidth}
        \centering
        \includegraphics[width=\linewidth]{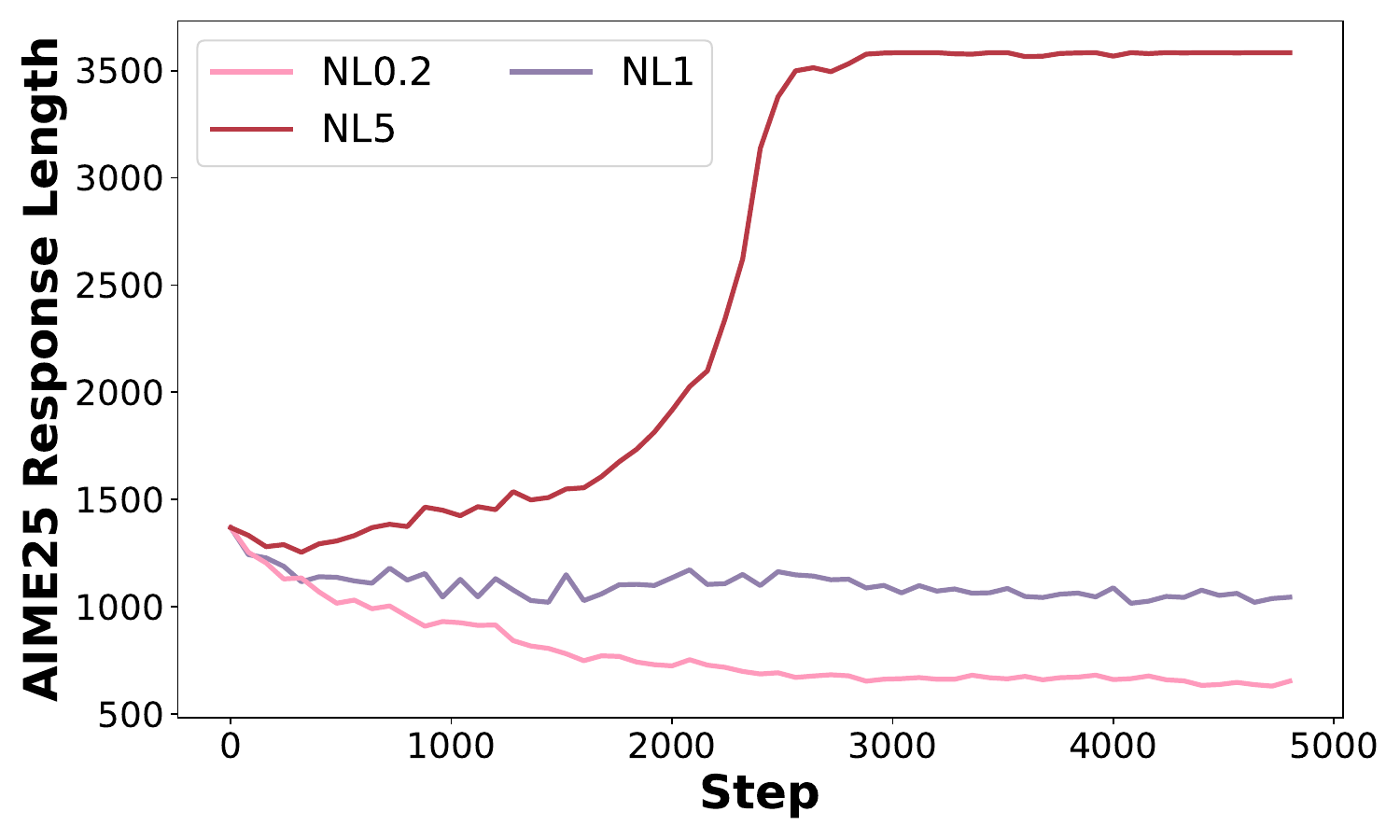}
        \caption{AIME25 Length}
    \end{subfigure}
    \begin{subfigure}[b]{0.32\linewidth}
        \centering
        \includegraphics[width=\linewidth]{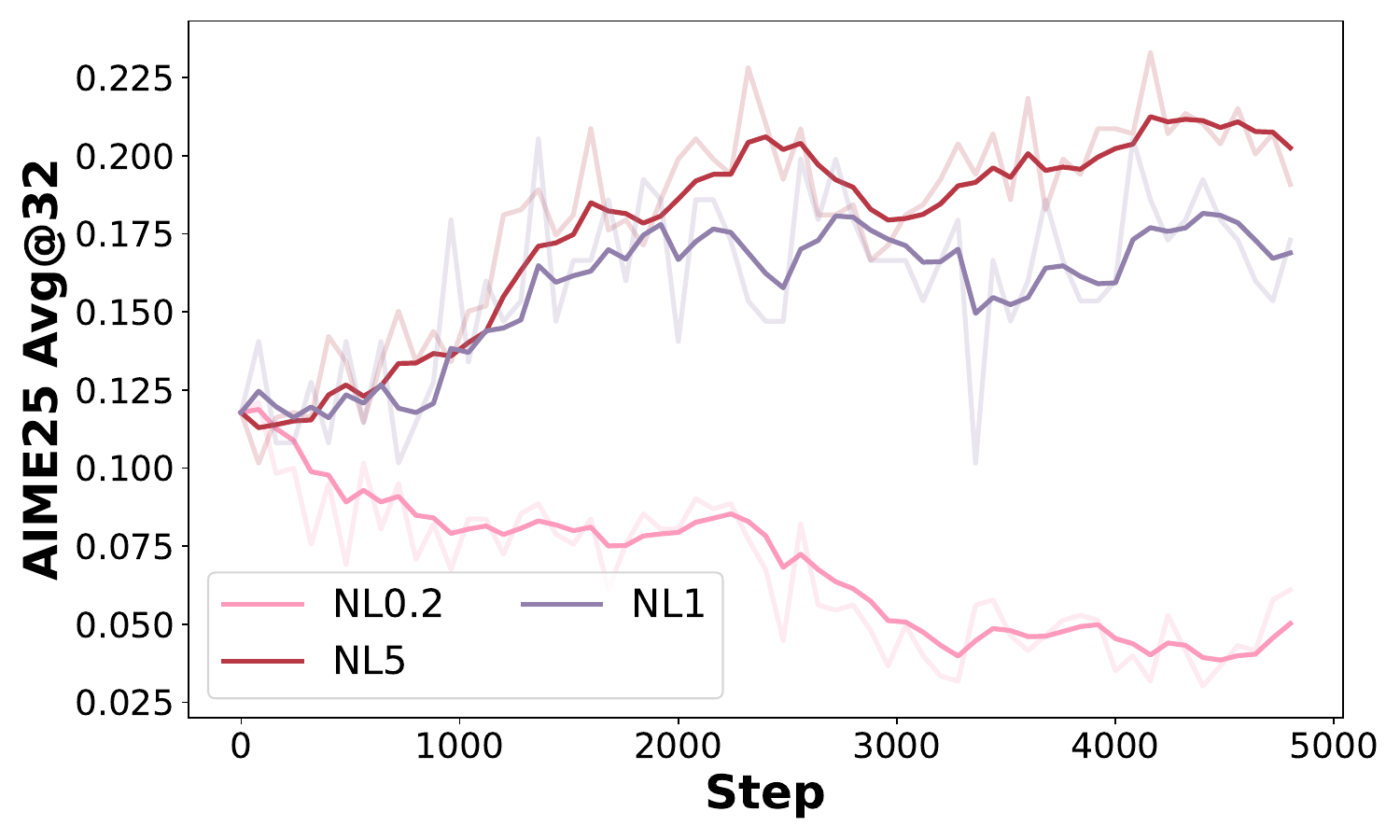}
        \caption{AIME25 Avg@32}
    \end{subfigure}
    \caption{RLVR training dynamics on negative low probability token advantage shaping.}
\label{fig:token-prob-nl-training_dynamic}
\end{figure*}

\begin{figure*}[t]
    \centering
    \begin{subfigure}[b]{0.32\linewidth}
        \centering
        \includegraphics[width=\linewidth]{figures/token_pl_ratio/PL_Ratio_Group_0-2/Entropy.pdf}
        \caption{Entropy}
    \end{subfigure}
    \begin{subfigure}[b]{0.32\linewidth}
        \centering
        \includegraphics[width=\linewidth]{figures/token_pl_ratio/PL_Ratio_Group_0-2/Response_Length.pdf}
        \caption{Length}
    \end{subfigure}
    \begin{subfigure}[b]{0.32\linewidth}
        \centering
        \includegraphics[width=\linewidth]{figures/token_pl_ratio/PL_Ratio_Group_0-2/Reward.pdf}
        \caption{Reward}
    \end{subfigure}
    \begin{subfigure}[b]{0.32\linewidth}
        \centering
        \includegraphics[width=\linewidth]{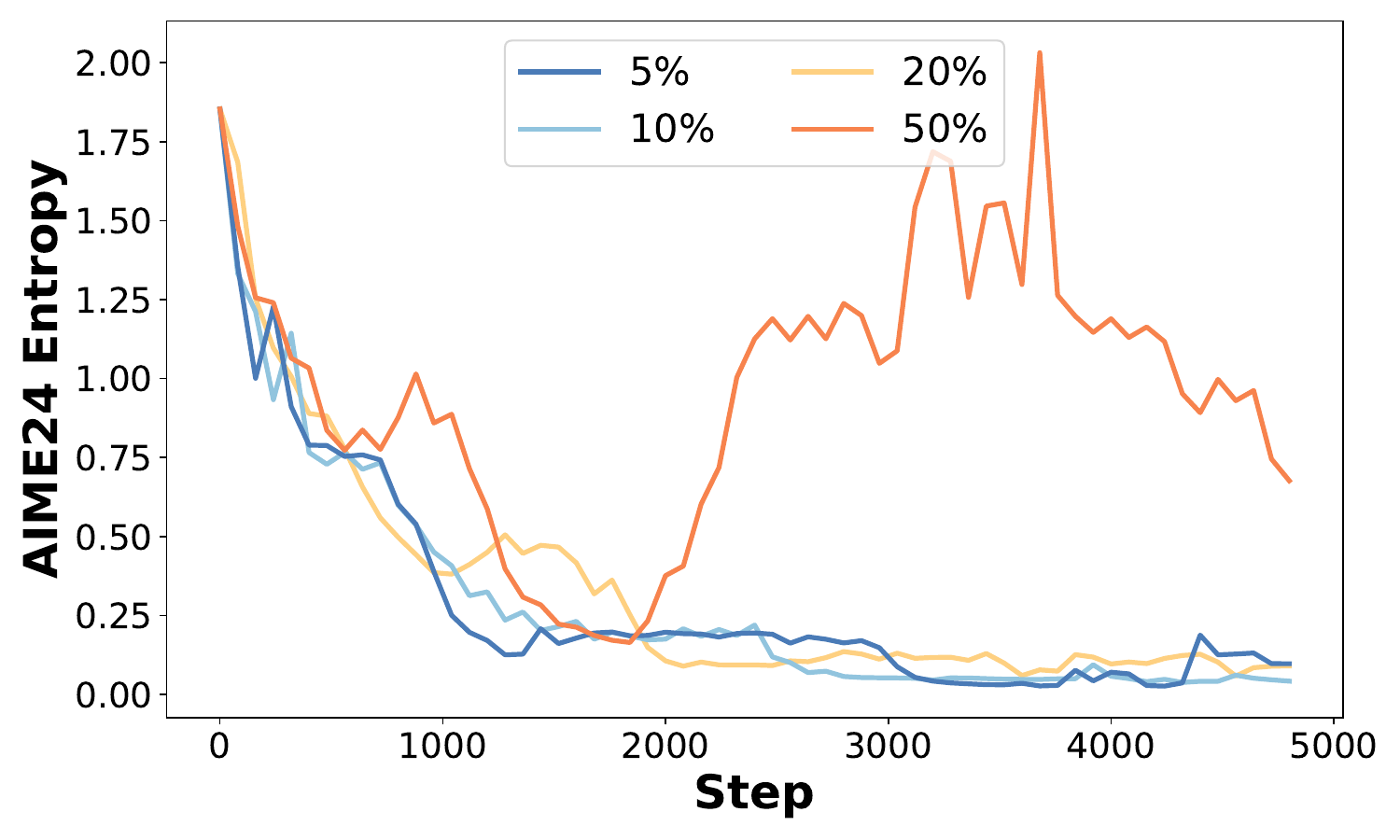}
        \caption{AIME24 Entropy}
    \end{subfigure}
    \begin{subfigure}[b]{0.32\linewidth}
        \centering
        \includegraphics[width=\linewidth]{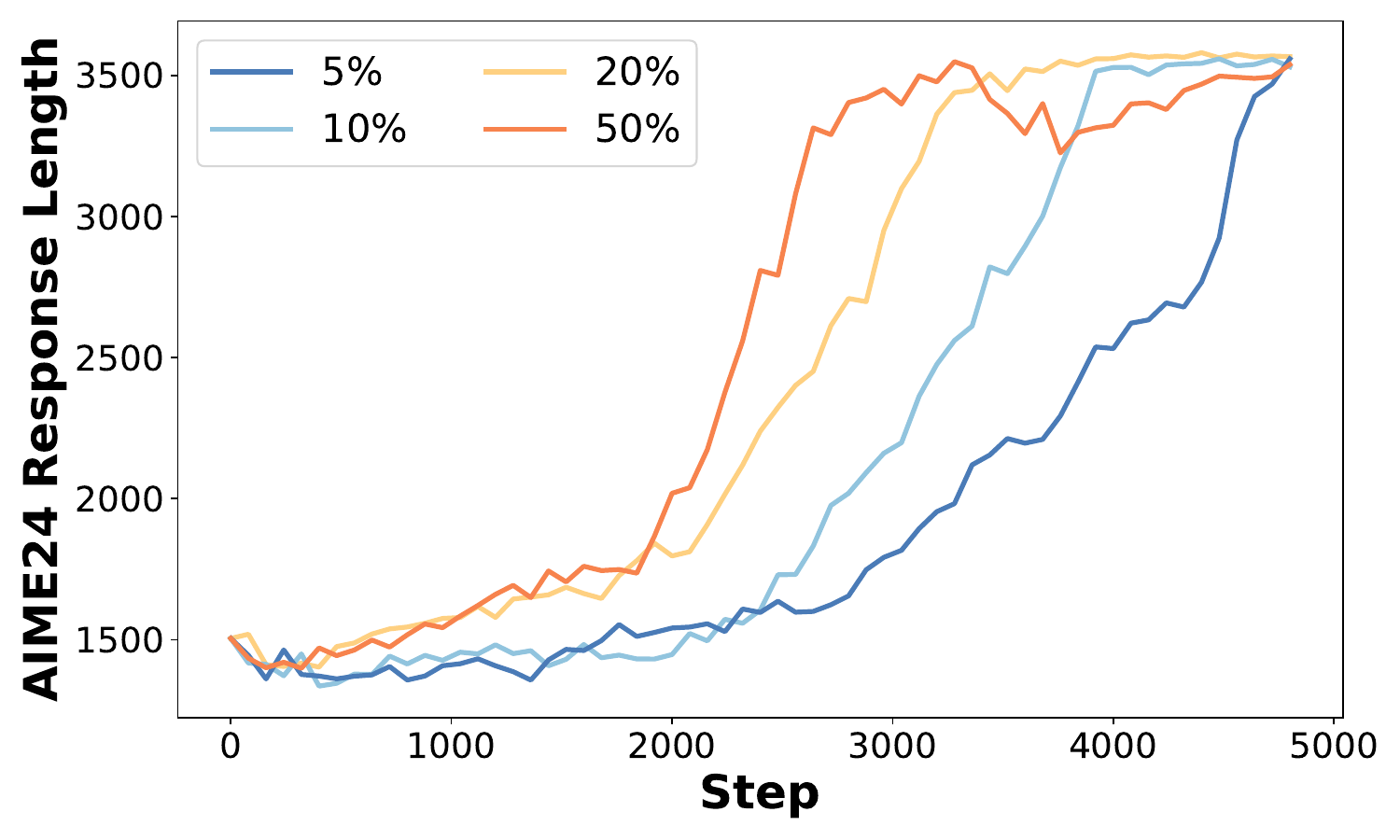}
        \caption{AIME24 Length}
    \end{subfigure}
    \begin{subfigure}[b]{0.32\linewidth}
        \centering
        \includegraphics[width=\linewidth]{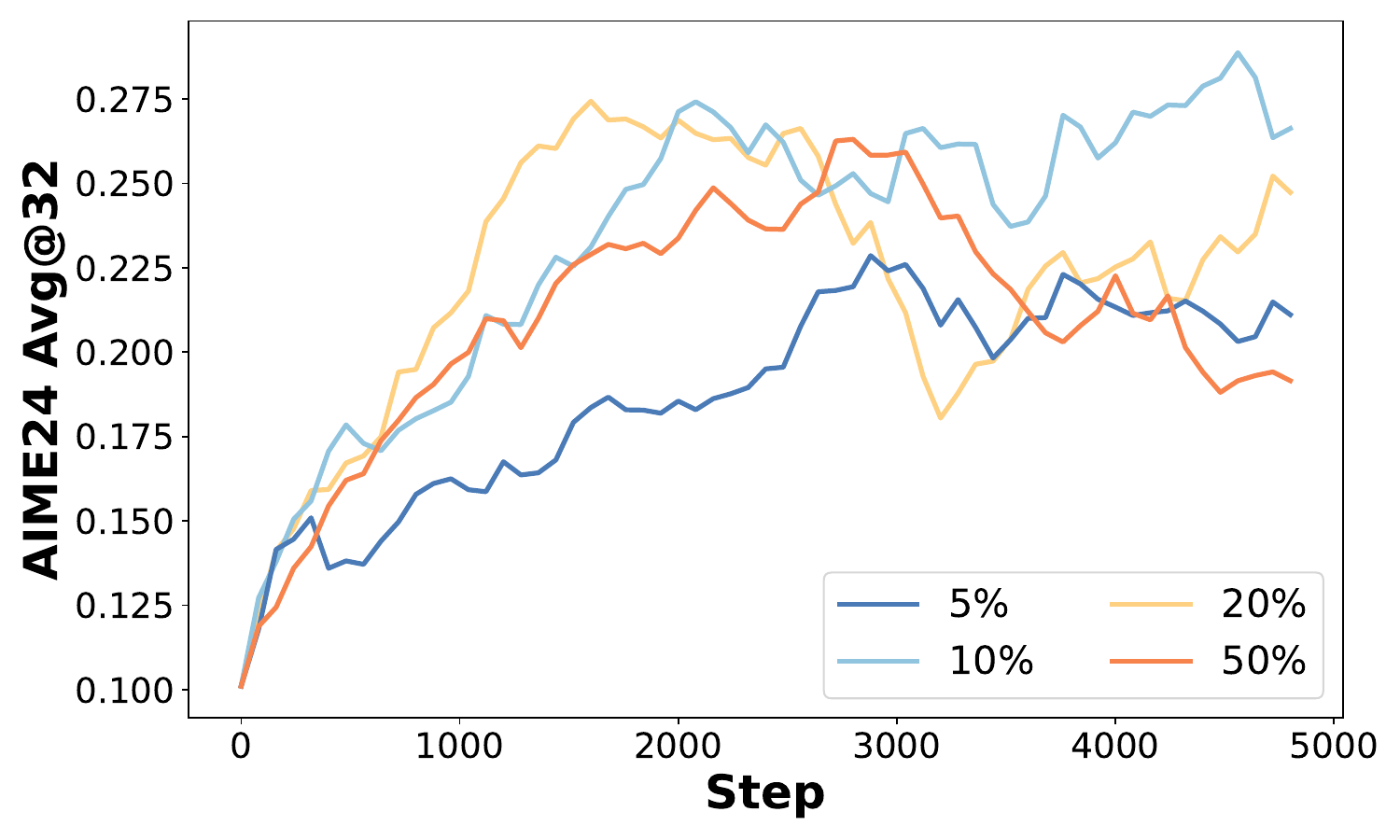}
        \caption{AIME24 Avg@32}
    \end{subfigure}
    \begin{subfigure}[b]{0.32\linewidth}
        \centering
        \includegraphics[width=\linewidth]{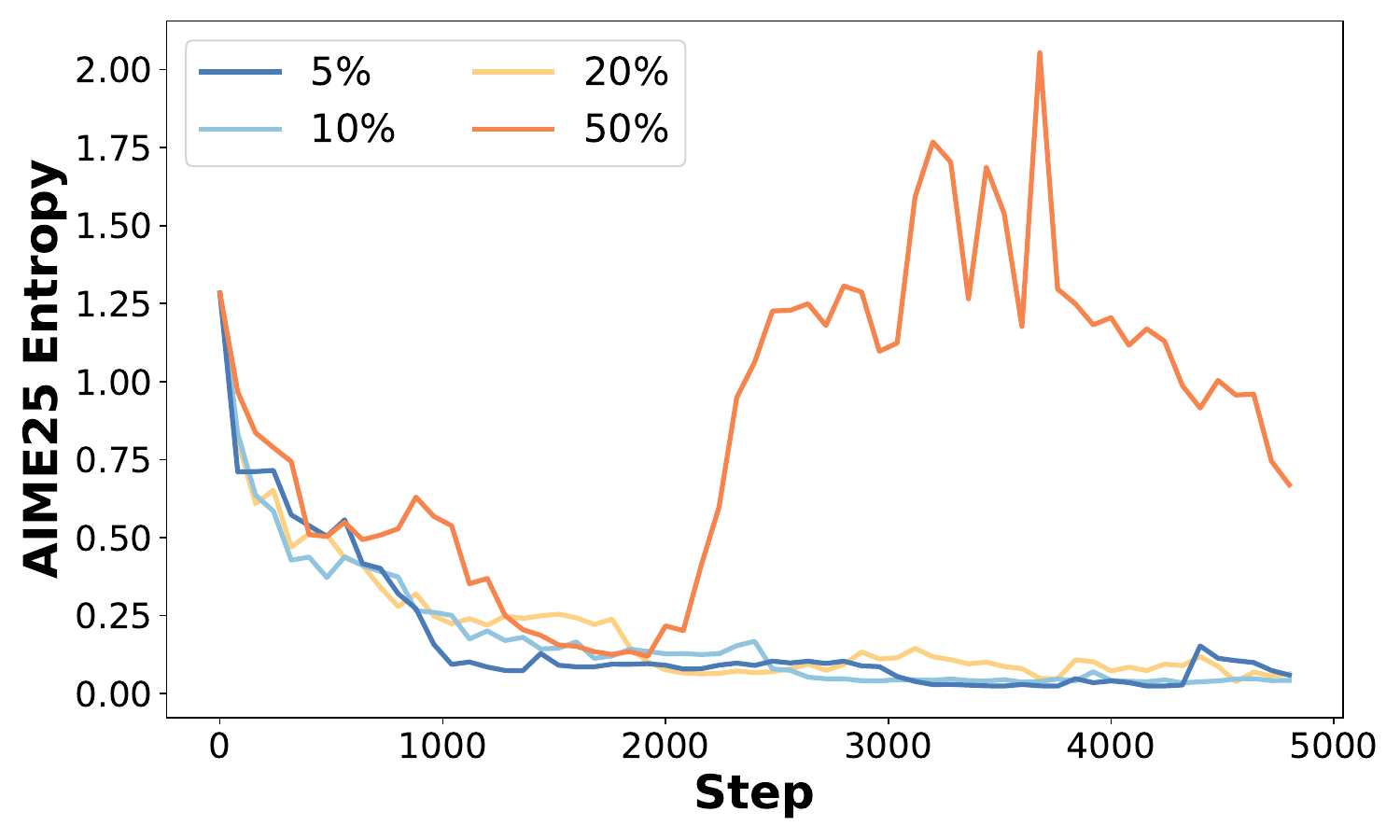}
        \caption{AIME25 Entropy}
    \end{subfigure}
    \begin{subfigure}[b]{0.32\linewidth}
        \centering
        \includegraphics[width=\linewidth]{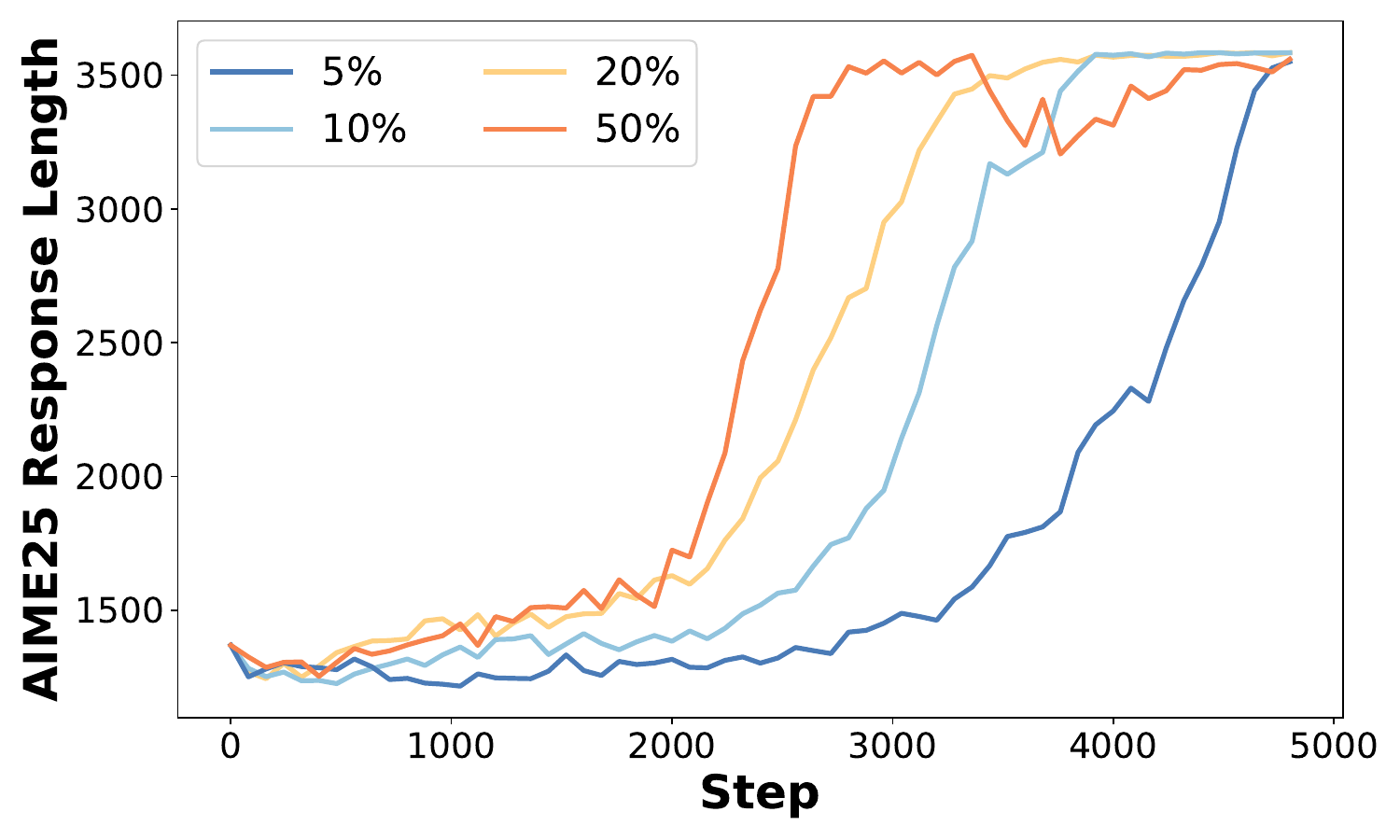}
        \caption{AIME25 Length}
    \end{subfigure}
    \begin{subfigure}[b]{0.32\linewidth}
        \centering
        \includegraphics[width=\linewidth]{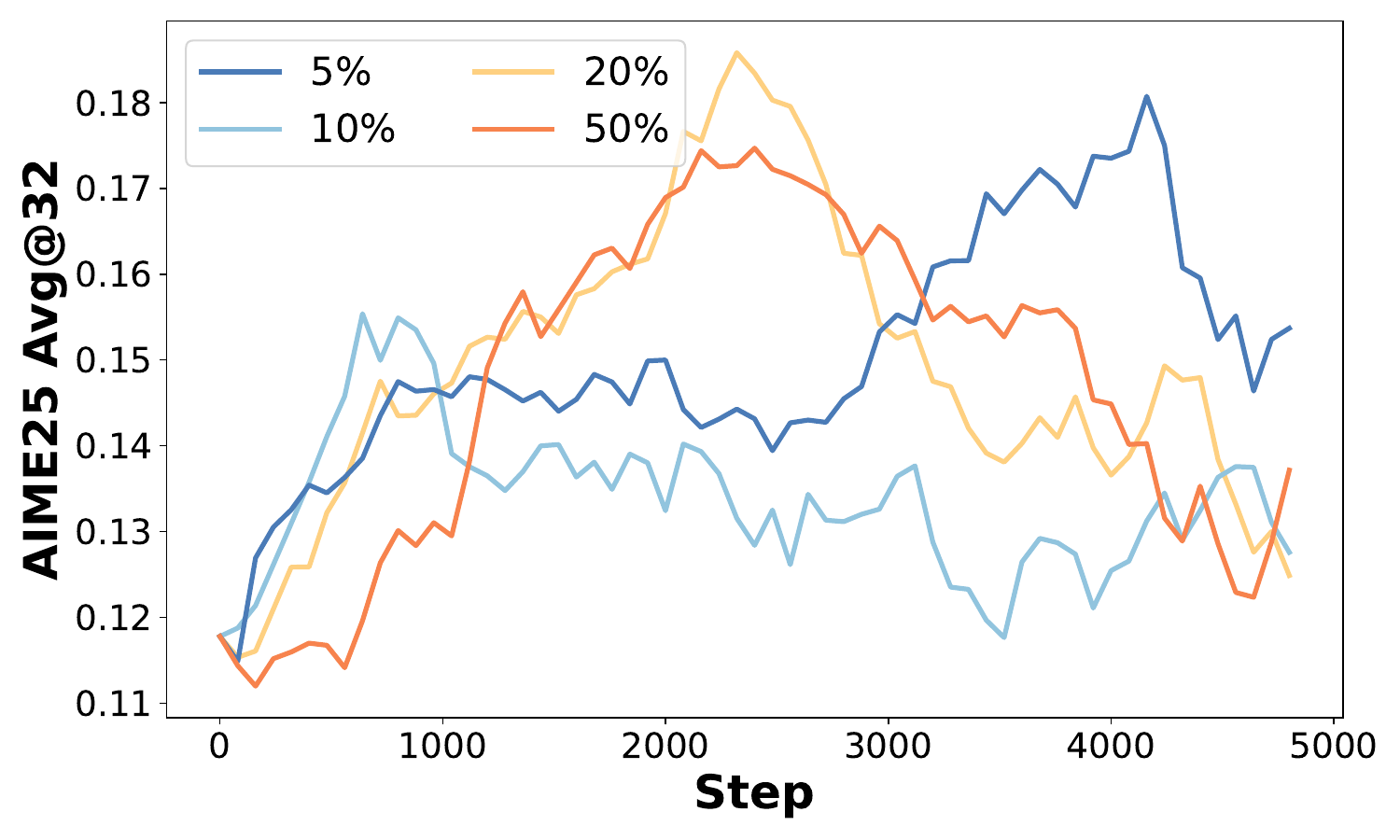}
        \caption{AIME25 Avg@32}
    \end{subfigure}
    \caption{RLVR training dynamics under different token-shaped ratios for low-probability positive tokens (scaled by 0.2).}
\label{fig:token-ratio-pl-training_dynamic}
\end{figure*}

In this part, we provide detailed training dynamics in our experiments.
Figure~\ref{fig:qwen2.5-7b-math-training_dynamic}, ~\ref{fig:qwen3-8b-base-training_dynamic}, ~\ref{fig:ds-qwen-7b-training_dynamic} present training dynamics of positive sample reinforcement, negative sample reinforcement, and DAPO across three different LLMs.
Figure~\ref{fig:sharpen-discovery3} and Figure~\ref{fig:sharpen-discovery4} present training behaviors (\ie sharpen and discovery) on Qwen2.5-7B-Math and Deepseek-R1-Distilled-Qwen-7B.
Figure~\ref{fig:polarity-pas-training_dynamic} and Figure~\ref{fig:polarity-nas-training_dynamic} show the results of polarity-level advantage shaping.
Figure~\ref{fig:token-entropy-ph-training_dynamic}, Figure~\ref{fig:token-entropy-pl-training_dynamic}, Figure~\ref{fig:token-entropy-nh-training_dynamic} and Figure~\ref{fig:token-entropy-nl-training_dynamic} illustrate the entropy-based token-level advantage shaping.
Figure~\ref{fig:token-prob-ph-training_dynamic}, Figure~\ref{fig:token-prob-pl-training_dynamic}, Figure~\ref{fig:token-prob-nh-training_dynamic} and Figure~\ref{fig:token-prob-nl-training_dynamic} illustrate the probability-based token-level advantage shaping.
Figure~\ref{fig:token-ratio-pl-training_dynamic} shows the results of different shaped token ratios in token-level advantage shaping.





\end{document}